\documentclass[lettersize,journal]{IEEEtran}
\usepackage{amsmath,amsfonts}
\usepackage{algorithmic}
\usepackage{algorithm}
\usepackage{graphicx} %
\usepackage{pgffor} %
\usepackage{longtable}
\usepackage[table,xcdraw]{xcolor} %
\usepackage[caption=false,font=normalsize,labelfont=sf,textfont=sf]{subfig} 
\usepackage{array}
\usepackage{subfiles} %
\usepackage{titletoc} %
\usepackage{multirow} %
\usepackage{textcomp}
\usepackage{stfloats}
\usepackage{url}
\usepackage{verbatim}
\usepackage{cite}
\usepackage{amsmath}
\usepackage{hyperref}

\begin{document}


\title{LLM4CMO: Large Language Model-aided Algorithm Design for Constrained Multiobjective Optimization}

\author{Zhen-Song Chen,~\IEEEmembership{Senior Member,~IEEE,} Hong-Wei Ding, Xian-Jia Wang, Witold Pedrycz,~\IEEEmembership{Life Fellow,~IEEE,}
        
\thanks{This work was supported in part by the National Natural Science Foundation of China under Grants 72571203, 72171182 and 72031009. (\textit{Corresponding authors: Zhen-Song Chen and Hong-Wei Ding}) } 
\thanks{Z.-S.~Chen is with the School of Civil Engineering, Wuhan University, Wuhan 430072, China and Faculty of Business, City University of Macau, 999078, Macao Special Administrative Region of China (e-mail: zschen@whu.edu.cn).}
\thanks{H. W. Ding is with the School of Civil Engineering, Wuhan University, Wuhan 430072, China (e-mail: hwding606@whu.edu.cn).}
\thanks{X. J. Wang is with the Economics and Management School, Wuhan University, Wuhan 430072, China (e-mail: wangxj@whu.edu.cn).}
\thanks{W. Pedrycz is with the Department of Electrical and Computer Engineering, University of Alberta, Edmonton, AB T6G 2R3, Edmonton, Canada. (e-mail address: wpedrycz@ualberta.ca)}
}



\maketitle








\begin{abstract}
Constrained multi-objective optimization problems (CMOPs) frequently arise in real-world applications where multiple conflicting objectives must be optimized under complex constraints. Existing dual-population two-stage algorithms have shown promise by leveraging infeasible solutions to improve solution quality. However, designing high-performing constrained multi-objective evolutionary algorithms (CMOEAs) remains a challenging task due to the intricacy of algorithmic components. Meanwhile, large language models (LLMs) offer new opportunities for assisting with algorithm design; however, their effective integration into such tasks remains underexplored. To address this gap, we propose LLM4CMO, a novel CMOEA based on a dual-population, two-stage framework. In Stage 1, the algorithm identifies both the constrained Pareto front (CPF) and the unconstrained Pareto front (UPF). In Stage 2, it performs targeted optimization using a combination of hybrid operators (HOps), an epsilon-based constraint-handling method, and a classification-based UPF-CPF relationship strategy, along with a dynamic resource allocation (DRA) mechanism. To reduce design complexity, the core modules, including HOps, epsilon decay function, and DRA, are decoupled and designed through prompt template engineering and LLM-human interaction. Experimental results on six benchmark test suites and ten real-world CMOPs demonstrate that LLM4CMO outperforms eleven state-of-the-art baseline algorithms. Ablation studies further validate the effectiveness of the LLM-aided modular design. These findings offer preliminary evidence that LLMs can serve as efficient co-designers in the development of complex evolutionary optimization algorithms. The code associated with this article is available
at~\url{https://anonymous.4open.science/r/LLM4CMO971}
\end{abstract}
\footnote{This work has been submitted to the IEEE for possible publication. Copyright may be transferred without notice, after which this version may no longer be accessible.}




    


\section{Introduction}
Constrained multi-objective optimization problems (CMOPs) are among the most common types of optimization challenges encountered in real-world applications, frequently arising in fields such as transportation~\cite{zuo2014vehicle}, power system design~\cite{mishra2019butterfly} and mechanical design~\cite{saravanan2009evolutionary,kannan1994augmented}. Solving CMOPs typically involves formulating real-world problems as multiple objective functions subject to both equality and inequality constraints. A solution is considered feasible only if it satisfies all specified constraints. Mathematically, a CMOP is generally formulated as follows:
    \begin{equation}
        \begin{aligned}
        \text{minimize} \quad & f(\mathbf{x}) = \left[ f_1(\mathbf{x}), f_2(\mathbf{x}), \ldots, f_m(\mathbf{x})\right]^T \\
        \text{s.t.} \quad & g_j(\mathbf{x}) \leq 0 \quad (j = 1, 2, \ldots, p) \\
        & h_k(\mathbf{x}) = 0 \quad (k = 1, 2, \ldots, q) \\
        & \mathbf{x} \in (x_1, x_2, \ldots, x_D) \subseteq \mathbb{R}^D
        \end{aligned}
    \end{equation}
where $f(\mathbf{x})$ is the objective functions; $g(\mathbf{x})$ represents the inequality constraints and $h(\mathbf{x})$ represents the equality constraints; $\mathbf{x}$ is a $D$-dimensional vector, which expresses a solution; $\mathbb{R}^D$ is a $D$-dimensional decision space. For every CMOP, constraints should be expressed as $p$ inequality constraints and $q$ equality constraints. When a $D$-dimensional vector satisfies all constraints (inequality and equality) within the decision space \( \mathbb{R}^D \), it is termed a feasible solution; otherwise, it is considered infeasible. The degree of constraint violation can be quantified using the Constraint Violation (CV) metric. A higher CV value indicates a greater deviation from feasibility. The CV~\cite{tian2020coevolutionary} is typically calculated as follows:  
    \begin{equation}
    CV(\mathbf{x}) = \sum\nolimits_{j=1}^{p} \max\left(0, \, g_j(\mathbf{x})\right) + \sum\nolimits_{k=1}^{q} (\left| h_k(\mathbf{x} ) \right| - \delta)
    \end{equation}
where $CV(\mathbf{x})$ quantifies the degree to which the current solution $\mathbf{x}$ deviates from the feasible region by measuring its satisfaction of the constraint conditions. When the value of $CV(\mathbf{x})$ is zero, $\mathbf{x}$ is considered feasible. $\delta$ is minute positive value for equality constraint to ease the strictness of the equality constraints.

In CMOPs, single-population algorithms such as C-NSGA-II~\cite{deb2002fast} and MOEA/D~\cite{zhang2007moea} enhance traditional Multi-Objective Evolutionary Algorithms (MOEAs) by incorporating Constraint Handling Techniques (CHTs), including the Constraint Dominance Principle (CDP), decomposition strategies, and the epsilon method~\cite{takahama2005constrained}. However, these algorithms have reached performance limits when addressing increasingly complex CMOP tasks. To overcome this limitation, researchers developed dual-population optimization methods~\cite{tian2020coevolutionary} that deploy auxiliary populations to explore unknown solution space boundaries and employ weak collaboration mechanisms for evolutionary information exchange between main and auxiliary populations, and many new algorithms have been derived~\cite{xu2025handling,ming2023constraint,li2024competitive}. While this approach broadens the search scope, it suffers from inefficient auxiliary population information utilization. Recent dual-population algorithms focus on efficiently utilizing auxiliary population information and dynamically balancing population exploitation and exploration capabilities~\cite{zou2021dual,ming2021dual}. Some algorithms employ two-stage control strategies that divide dual-population optimization into learning and exploration phases~\cite{liang2022utilizing,liu2024coevolutionary}. The learning phase focuses on understanding relationships between the Unconstrained Pareto Front (UPF) and Constrained Pareto Front (CPF), along with population characteristics. The exploration phase formulates decision-making schemes based on learning outcomes, strengthening the population's search capabilities and significantly improving algorithmic convergence. Algorithms like URCMO~\cite{liang2022utilizing} effectively enhance population diversity by combining UPF-CPF relationships with hybrid operators. However, existing UPF-CPF learning mechanisms are limited by problem decision space scale and relationship clarity. In some CMOP instances, accurately identifying CPF-UPF relationships proves difficult, and algorithms may encounter deceptive constraint interference, resulting in poor convergence, feasibility, and diversity performance. Additionally, distance metric-based methods for evaluating UPF and CPF convergence face slow stage 1 convergence challenges. Therefore, balancing algorithmic feasibility, convergence, and diversity during CMOP solution processes remains a core research challenge~\cite{liang2022survey}.

Large Language Model (LLM) has developed rapidly in recent years, with applications expanding across dialogue interaction, mathematical calculation, engineering design and algorithm design~\cite{zhao2023survey,liu2024systematic}. The multi-faceted capabilities of LLMs have given rise to the concept of LLM4SR~\cite{luo2025llm4sr}. Optimization problems have attracted numerous scholars exploring LLM application potential. Researchers have leveraged the knowledge of LLM via Chain of Thought (CoT) techniques to model and solve optimization problems through prompt template design~\cite{wei2022chain}. In algorithm design, heuristic search algorithms including Funsearch~\cite{romera2024mathematical} and EOH~\cite{liu2024evolution} have achieved remarkable results in bin packing (BP) and Traveling Salesman Problem (TSP) optimization through evolutionary prompting. In multi-objective optimization, some studies have explored using LLMs as population generators for MOPs or CMOPs~\cite{liu2025large,wang2024large}, but these investigations remain preliminary. Performance improvements compared to mainstream algorithms are modest. Therefore, investigating LLM effectiveness in CMOP algorithm design represents a promising research direction.

Looking at existing algorithms, Constrained Multi-Objective Evolutionary Algorithms (CMOEAs) have evolved from single-stage to multi-stage processes, from single-operator to multi-operator approaches, while also developing various complex mechanisms, making their design quite challenging. Currently, high-performance algorithms aim to achieve a balance between exploration and exploitation to enhance comprehensive performance in terms of diversity, feasibility, and convergence. Although LLMs have demonstrated strong performance in areas such as automated design of heuristics, their application in CMOEA improvement and automated design paradigms remains limited by the inherent complexity of the algorithms themselves. These challenges highlight a core research gap: despite the potential of LLMs in optimization algorithm design, their application in current CMOEAs remains underexplored. The complexity of mainstream CMOEAs, with the obstacles posed by their multi-stage processes and intricate mechanisms, also presents opportunities. This prompts us to investigate: Can LLMs aid algorithm designers in developing more effective CMOEA?

Based on this motivation, we propose applying LLMs to key modules within basic CMOP algorithm frameworks to assist in the design of core components.
We constructed a dual-population two-stage algorithm framework where the Stage 1 employs dynamic learning strategies, and the Stage 2 controls exploration and exploitation through epsilon methods and UPF-CPF relationships. The Stage 2 comprises three core modules: a population offspring generation strategy using HOps combinations, an epsilon equation, and a Dynamic Resource Allocation (DRA) mechanism. We used LLM as an auxiliary tool, designing a set of initial prompts for each design task and interacting with it to improve and enhance the existing algorithm framework. Experimental results show that the algorithm obtained through LLM-aided interactive design has significantly improved the performance of the initial algorithm framework. We comprehensively evaluated the algorithm on standard test suites and real-world optimization problems, comparing it with mainstream algorithms to verify effectiveness and superiority.


The main contributions of this paper are summarized as follows:

\begin{itemize}
    \item We propose a novel algorithm, LLM4CMO, for solving CMOPs. In the first stage, a learning and dynamic relaxation mechanism is introduced to balance the convergence of the UPF and the CPF, as well as to control learning duration. In the second stage, UPF-CPF relationship types and an epsilon decay function are employed for phase control. HOps perform targeted optimization based on the relationship type, while the DRA mechanism adaptively balances computational resources between the main and auxiliary populations.
    \item We use LLM as a tool to assist algorithm design. For the three core modules (HOps, epsilon decay function, and DRA) in the basic CMOEA algorithm framework, we designed initial prompts and conducted interactive design with LLM, completing the design and optimization of these three core modules. We preliminarily explored the feasibility of applying interactive methods with LLM as an auxiliary tool in complex algorithm design.
    \item We conduct comprehensive experiments on six benchmark test suites and ten real-world CMOP instances, comparing LLM4CMO against eleven state-of-the-art algorithms. The results show that LLM4CMO achieves superior performance across standard evaluation metrics. Additionally, ablation studies on the three core modules confirm their individual contributions and validate the effectiveness of the LLM-aided modular design strategy.
\end{itemize}

This article is organized as follows. Section \ref{sec2} reviews related work, including CHTs, mainstream  CMOEAs, the type of UPF-CPF relationship, mating pools and operators, LLMs for algorithm design and the motivation of this study. Section \ref{sec3} provides the base algorithm framework and the LLM-aided design methodology. Section \ref{sec4} tests the proposed algorithm, LLM4CMO, compares its performance against 11 baseline algorithms and ablation studies, and analyzes the experimental results. Finally, Section~\ref{sec5} and Section~\ref{sec6} present the conclusions and limitations.


\section{Related works} \label{sec2}

CMOPs are widely encountered in real-world scenarios that require simultaneous optimization of multiple conflicting objectives under complex constraint conditions. Over the years, numerous algorithmic frameworks, CHTs, and evolutionary strategies have been developed to address the challenges of feasibility, convergence, and diversity. This section reviews key developments in five major areas: (i) CHTs, (ii) mainstream CMOEAs, (iii) the relationship between the UPF and CPF, (iv) mating pool and operator design, and (v) emerging approaches that leverage LLMs for algorithm design.

\subsection{Constraint Handling Techniques}

Among CHTs, CDP is one of the most widely used approaches for solving CMOPs. CDP selects elite solutions based on the comparison of constraint violation values ($\mathbf{CV}$). Its standard rules include: a) Feasible solutions dominate infeasible ones; b) Among infeasible solutions, those with smaller $\mathbf{CV}$ dominate those with larger values; c) Among feasible solutions, Pareto dominance is applied. Several MOEAs are extended to CMOEAs by incorporating CDP modules. For example, C-NSGA-II~\cite{deb2002fast} and C-MOEA/D~\cite{jain2013evolutionary}. However, CDP's strict discriminative rules often hinder exploration, limiting performance in highly constrained or complex landscapes. To address this, Takahama and Sakai~\cite{takahama2005constrained} proposed the epsilon constraint-handling method, which relaxes constraint enforcement via a positive tolerance factor. Later studies~\cite{fan2019push, fan2019improved} introduced self-adaptive and dynamically adaptive epsilon functions to further enhance exploration in CMOEAs. Liu et al.~\cite{liu2025constraint} integrated reference vectors and angle-based subspace division~\cite{fan2016angle, fan2019moea} to improve diversity, guiding solution distribution along vector directions and enhancing outward exploration based on minimal objective values. Qiao et al.~\cite{qiao2024cooperative} proposed a strategy that generates new populations by moving individuals away from current population centers, thus improving spatial coverage. 


\subsection{Mainstream CMOEAs}

Mainstream CMOEAs typically employ dual-population and multi-stage strategies to solve CMOPs. Tian et al.~\cite{tian2020coevolutionary} proposed the CCMO to separately handle the UPF and CPF. Using a dual-population structure with weak collaborative relationships, CCMO respectively explores feasibility and diversity, thereby improving solution quality under complex constraints. DDCMOEA proposed by Ming et al.~\cite{ming2021novel} employs a dual-population framework where the main population consistently incorporates constraints, while the auxiliary population only considers constraints after converging to the UPF.  Zou et al.~\cite{zou2021dual} introduced a two-stage algorithm known as CAEAD. In the second stage, it employs dynamic interaction and stage switching, using the epsilon method to guide the auxiliary population toward gradual convergence with the main population. Two-stage algorithms divided by the PPS method~\cite{fan2019push}, such as BiCo~\cite{liu2021handling} and cDPEA~\cite{ming2021dual}, focus on intermediate solutions between populations and stage-based strategies for UPF and CPF optimization, respectively. These approaches aim to enhance algorithmic exploration capability in the second stage.  From a multi-stage constraint increment perspective, Ma et al.~\cite{ma2021multi} proposed MSCMO to accelerate convergence by gradually expanding from single to multiple constraints across stages. CMOEA-MS, established by Tian et al.~\cite{tian2021balancing}, adaptively switches stages based on feasibility rates, dynamically balancing diversity exploration and local optimization needs by prioritizing constraints over objectives.
The CMEGL proposed by Qiao et al.~\cite{qiao2023evolutionary} and the CMOEMT proposed by Ming et al.~\cite{ming2022constrained} introduced the concept of multi-task evolution, leveraging knowledge transfer between two auxiliary tasks and the main task. Sun et al.~\cite{sun2022multistage} proposed the C3M, which adopts a three-stage framework with sorting strategies, adding constraints incrementally while prioritizing primary constraints to reduce invalid searches. The CMOES  by Ming et al.~\cite{ming2024even} advocated for uniform search strategies in promising regions to accelerate non-dominated solution escape from local optima while relaxing feasible solutions for broader feasible region exploration. Both URCMO proposed by Liang et al.~\cite{liang2022utilizing} and the CMOE-DAS proposed by Liu et al.~\cite{liu2024coevolutionary} first learn UPF-CPF relationships in stage one, then optimize solutions in stage two through different operator combinations and environmental selection strategies. The diversity archive in CMOEACD proposed by Liu et al.~\cite{liu2025constraint} employs angle vector-based guidance for solution generation to enhance solution diversity. 

\subsection{Types of UPF-CPF Relationships}
 Ma and Wang~\cite{ma2019evolutionary} introduced the concepts of the UPF and CPF and defined a set of benchmark functions for analyzing their relationships. They categorized the relationship between UPF and CPF into four types: (i) completely coincident, (ii) UPF partially containing CPF, (iii) partially separated, and (iv) completely separated. They demonstrated that tailoring CDP strategies to specific UPF-CPF types improves overall algorithmic performance. Building on this, URCMO ~\cite{liang2022utilizing} proposed a parameter-controlled CDP classification method, further subdividing the relationships into six subcategories and summarizing them into three general solution strategies. More recently, COME-DAS ~\cite{liu2024coevolutionary} enhanced this classification scheme by incorporating multiple critical metrics for dynamic supervision. These efforts focus on refining the categorization of UPF-CPF types and tailoring constraint-handling methods accordingly.
   
  \subsection{Mating Pools and Operators}

    The core operation of evolutionary algorithms typically involves three steps: mating pool generation, offspring creation via variation operators, and environmental selection by combining offspring with the parent population.
    Mating pools are generated from parent populations, and there are mainly two mainstream generation strategies. One is tournament sorting based on parental fitness, which involves polynomial selection to filter out higher-quality parents; the other is full random selection, where parents are randomly selected to form the mating pool, introducing randomness. Operators are a crucial component of algorithms, determining the solution quality, diversity, and convergence speed of the algorithm. Genetic Algorithm (GA)~\cite{holland1992genetic} and Differential Evolution (DE)~\cite{storn1997differential} operators are the most commonly used ones. Three different operators evolved from DE operators are also widely applied in operator combinations. Zhang et al.~\cite{zhang2009jade} designed DE/current-to-pbest/1 for exploring poor solutions to provide promising progress directions by historical information. The DE/current-to-rand/1 mutation strategy generates new candidate solutions by combining the current individual with the scaled difference vector between two randomly selected individuals and a third randomly selected individual. and DE/transfer~\cite{liang2022utilizing} for enabling information exchange between two populations. The most important function of HOps is to provide diversity.
    
    \subsection{LLM-aided Algorithm Design}
    Large language models (LLMs) have demonstrated strong capabilities and broad applicability in algorithm design, with several recent studies exploring their role in optimization. To address robust optimization and adaptive robust optimization problems, Dimitris and Margaritis~\cite{bertsimas2024robust} employed prompt engineering strategies to guide ChatGPT-4 in formulating mathematical optimization models and generating corresponding solution code. Wang et al.~\cite{wang2024large} and Liu et al.~\cite{liu2025large} utilized LLMs as mutation operators to generate partial solutions for multi-objective optimization problems (MOPs) and constrained multi-objective optimization problems (CMOPs), respectively. Other related studies have focused on heuristic algorithm design by refining prompts to improve the performance of LLM-generated heuristics in downstream optimization tasks~\cite{liu2024evolution, romera2024mathematical, yang2023large}. These works primarily target classic combinatorial problems such as Boolean Programming (BP)~\cite{garey1981approximation} and the Traveling Salesman Problem (TSP)~\cite{reinelt1992fast}. In addition, Niki and Thomas~\cite{van2024llamea} explored meta-heuristic algorithm discovery using ChatGPT-3.5, applying it to the 5-dimensional black-box optimization benchmark suite. This line of research highlights the growing potential of LLMs to participate in the automated discovery and refinement of optimization strategies. However, the use of LLMs for designing and optimizing more complex CMOEA algorithms remains unexplored.

    \subsection{Motivation}
    At the algorithmic level, current mainstream approaches adopt dual-population and two-stage frameworks or even multi-population and multi-stage architectures to establish search populations under different auxiliary tasks. These frameworks leverage information from infeasible regions through weak exchange strategies, exploring broader plausible regions by mining insights from infeasible solutions. The core objective is to balance exploration and exploitation. Algorithm performance is typically evaluated across three key dimensions: feasibility, convergence, and diversity. Dual-population algorithms exemplified by CCMO have spawned numerous derivatives. However, multi-population and multi-stage frameworks alone cannot fully optimize performance on downstream tasks, and the resulting CMOEAs often exhibit complex structures. 
    Although LLMs demonstrate strong capabilities in comprehension, reasoning, and modeling, and perform excellently in operator fusion, heuristic search, and code generation, there remains a question in designing a complete dual-population two-stage CMOEA: \textit{is it feasible for LLMs to design and optimize complex CMOEAs?} Therefore, we have conducted a preliminary exploration on LLMs as tools for assisting algorithm design.

    A high-performing CMOEA must excel across multiple metrics, requiring both adaptability and stability across diverse task types. According to the \textit{No Free Lunch} theorem~\cite{wolpert1997no}, no single algorithm performs optimally across all problem classes. Both URCMO and COEA-DAS demonstrate promising performance with their HOps based on UPF-CPF relationship. These HOps are designed manually and rely on precise determination of the UPF-CPF relationship type. Therefore, We have weakened the classification, dividing it into four types: complete overlap (Type-1), partial overlap (Type-2 or 3), complete separation (Type-4), and unclear.

    HOps presents a combinatorial recommendation problem that aligns perfectly with interaction of LLMs. By structuring prompt templates to contextualize tasks, leveraging LLMs to recommend effective HOps combinations based on task-specific backgrounds is inherently feasible. Designing task-specific epsilon decay functions is also a critical module for CMOEAs, and LLMs appear to have high feasibility and optimization potential for designing such functions. From an operational efficiency perspective, balancing resource allocation between popMain and popAux in dual-population frameworks is critical to minimizing waste. The DRA mechanism optimizes computational efficiency by generating offspring of varying intensities through coefficient-based allocation strategies.

    In summary, our research objectives are divided into two aspects. On one hand, we extract modular tasks from the original complex CMOEA algorithm for LLMs to design and optimize, aiming to obtain a CMOEA algorithm. On the other hand, we explore the feasibility of LLMs as auxiliary tools for complex algorithm design.
\section{Methodology} \label{sec3}
    The LLM4CMO proposed in this paper adopts a basic framework of dual-population and two-stage, where the two stages are divided into learning and exploration. In addition, we focus on LLM-aided interactive improvement of the core modules: HOps, epsilon decay function and DRA. In this section, we will focus on the description of the algorithm design, compare the differences and connections with the existing dual-population and two-stage CMOEAs, and introduce the base framework and LLM-aided methods in detail.
\subsection{Base Framework Design} 

\begin{figure}[htbp!]
        \centering
        \includegraphics[width=0.99\linewidth]{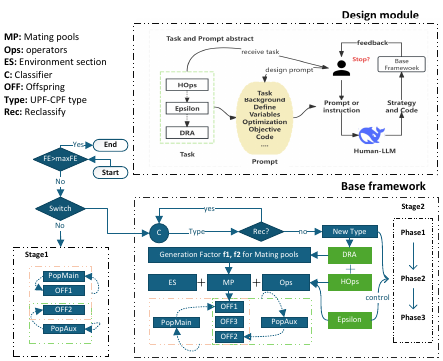}
        \caption{The proposed base dual-population two-stage CMOEA framework (down) and the LLM–human interactive module design flowchart (up). Here, $C$ denotes the classification strategy, following the approach used in URCMO. The Rec updates the UPF-CPF type during Phase 1 of Stage 2. $Off_3$ represents offspring generated via the opposition-based mechanism. The Type classification includes: (1) completely overlap, (2) UPF and CPF partially overlap, (3) completely separated, and (4) unclear. The Design module sequentially guides the LLM-aided development of the HOps, epsilon decay function, and DRA mechanism.}
        \label{fig:Base framework}
    \end{figure}

Our framework extends existing dual-population, two-stage CMOEAs (See Fig.~\ref{fig:Base framework}). During Stage 1, the two populations co-evolve: popMain approximates the CPF, while popAux targets the UPF. For certain problems, rapidly approaching the CPF and UPF proves more valuable than optimizing solution quality in Stage 1. To accelerate exploration, we, therefore, disconnect offspring exchange between popAux and popMain, allowing popAux to focus exclusively on UPF discovery. We adopt a distance metric (Eq. \eqref{eq:rs}) and a dynamic discrimination mechanism (Eq. \eqref{eq:rsc}) to determine CPF and UPF convergence, thereby identifying Stage 1 completion. We employ the distance metrics as used in the DDCMOEA \cite{ming2021novel}:
    \begin{align}
        \label{eq:rs}
        r_s =
        \begin{cases}
        r_z(g)=\max_{i = 1,2,\ldots,M}\left\{\frac{\vert z^g_i - z^{g - l}_i\vert}{\max\{\vert z^{g - l}_i\vert,\Delta\}}\right\}\\
        r_n(g)=\max_{i = 1,2,\ldots,M}\left\{\frac{\vert n^g_i - n^{g - l}_i\vert}{\max\{\vert n^{g - l}_i\vert,\Delta\}}\right\}\\
        r_a(g)=\max_{i = 1,2,\ldots,M}\left\{\frac{\vert a^g_i - a^{g - l}_i\vert}{\max\{\vert a^{g - l}_i\vert,\Delta\}}\right\} \\
        r_s(g) = \max(r_s(g),r_a(g),r_n(g))
        \end{cases}
    \end{align}
where $z^g=[z^g_1,\ldots,z^g_M]$, $n^g=[n^g_1,\ldots,n^g_M]$, and $a^g=[a^g_1,\ldots,a^g_M]$ are the ideal, nadir, and average points in the $g$-th generation of popAux, respectively. $l$ is the generation gap, and $\Delta$ is a value small enough to ensure that the denominator terms in Eq. \eqref{eq:rs} are not zero (e.g., $10^{-7}$). $S_t$ serves as a stage transition identifier, determining whether the algorithm should switch to Stage 2 based on the discriminant condition of the dynamic relaxation metric $r_s$, as defined in Eq.~\eqref{eq:rsc}.
    \begin{align}
        \label{eq:rsc}
        S_t = \begin{cases}
            1 & \text{if } r_s < 0.001 \land g > 10 \\
            1 & \text{if } r_s < 0.02 \land g > 100 \\
            1 & \text{if } r_s < 0.05 \land g > 150 \\
            1 & \text{if } g > 250 \\
            0 & \text{otherwise}
        \end{cases}
    \end{align}
The smaller the decision threshold of $r_s$, the stricter the judgment, which leads to longer learning time and affects the optimization in the second stage. Therefore, we adopt a dynamic approach to adapt to different CMOPs, with parameters set as coarse condition relaxation. We verified its effectiveness in the ablation study section.

The pseudocode for Stage 1 of LLM4CMO  is summarized in Algorithm~\ref{algo:stage1}.

Stage 2 employs epsilon controls, type controls, and a DRA mechanism to balance exploration and exploitation. This mechanism guides offspring generation through three optimization phases, each utilizing specific epsilon values. The algorithm applies different operators and mating pools according to problem type, with a dynamic transition discriminant determining when to switch between stages.

Upon entering Stage 2, the first phase continuously updates problem types to refine the accuracy achieved in Stage 1. The three-phase optimization ensures robust algorithm performance across various CMOPs by employing HOps and environmental selection strategies tailored to different optimization problems and UPF-CPF types.

A critical parameter in this framework is $N_s$, which controls the number of popAux generated, thereby reducing redundancy and improving algorithmic efficiency. This parameter is determined in the final step of algorithm design, with its calculation defined as follows:
\begin{equation}
    N_s = \max(25,(1-fr_2)\cdot N).
\end{equation}
Here, $fr_2$ represents the feasible ratio of the popAux, and $N$ denotes the population size, which is set to 100 for all test problems.

\begin{algorithm}
\footnotesize
\caption{\footnotesize Procedure of Stage 1: Dual-Population Generation and Selection}
\label{algo:stage1}
\begin{algorithmic}[1]
\REQUIRE Population size $N$, current populations $Pop_{main}^g$, $Pop_{aux}^g$, selection pressure $\epsilon$
\ENSURE Updated populations $Pop_{main}^{g+1}$, $Pop_{aux}^{g+1}$

\STATE Generate offspring $Off_1$ by randomly mating parents from $Pop_{main}^g$ using GA operators
\STATE Generate offspring $Off_2$ by randomly mating parents from $Pop_{aux}^g$ using GA operators
\STATE Construct candidate population $P_1 \leftarrow Pop_{main}^g \cup Off_1 \cup Off_2$
\STATE Construct candidate population $P_2 \leftarrow Pop_{aux}^g \cup Off_2$
\STATE Select $Pop_{main}^{g+1} \leftarrow \text{CDP}(P_1, \epsilon=0)$ \hfill \COMMENT{CDP}
\STATE Select $Pop_{aux}^{g+1} \leftarrow \text{CDP}(P_2, \epsilon=\infty)$ \hfill \COMMENT{${\text{CDP}_{\epsilon}}$}

\RETURN $Pop_{main}^{g+1}$, $Pop_{aux}^{g+1}$
\end{algorithmic}
\end{algorithm}
 
\begin{algorithm}
 \footnotesize
\caption{ \footnotesize Procedure of Stage 2: Evolution of core modules for epsilon and type control with LLM-aided design}
\label{algo:stage2}
\begin{algorithmic}[1]
\REQUIRE Population size $N$, current populations $Pop_{main}^g$, $Pop_{aux}^g$, $\epsilon$, solution type $type$, counter $cnt$
\ENSURE $Pop_{main}^{g+1}$, $Pop_{aux}^{g+1}$, $cnt$, $type$

\STATE $N_s \leftarrow 40$
\STATE Compute feasible ratios: $fr_{main}, fr_{aux}$
\IF{$(fr_{main} = 1 \lor fr_{main} = 0) \land fr_{aux} = 0 \land \epsilon > 0.0005$}
    \STATE Compute $Off_3$ using Eq.~\eqref{eq:opposite} \COMMENT{Opposition Offspring}
\ENDIF

\STATE Compute $f_1, f_2$ using Algorithm~\ref{algo:DRA} \COMMENT{DRA}
\STATE Generate offspring $Off_1, Off_2$ using Algorithm~\ref{algo:HOps} \COMMENT{HOps}
\STATE $Off \leftarrow Off_1 \cup Off_2 \cup Off_3$

\STATE Select $Pop_{main}^{g+1} \leftarrow \text{CDP}(Pop_{main}^g \cup Off, \epsilon = 0)$

\IF{$\epsilon \geq 0.195$} 
    \STATE Generate $N_s$ candidates: $Pop_{aux}^{g+1} \leftarrow \text{CDP}(Pop_{aux}^g \cup Off, \epsilon = \infty)$  \\ \COMMENT{Phase-1: Exploration}
    \STATE $ntype \leftarrow \text{Reclassify}(Pop_{aux}^{g+1})$ 
    \IF{$ntype \neq type$}
        \STATE $cnt \leftarrow cnt + 1$
        \IF{$cnt > 3$}
            \STATE $type \leftarrow ntype$ \COMMENT{Update UPF-CPF type}
        \ENDIF
    \ENDIF
    
\ELSIF{$\epsilon \leq 0.005 \lor type = 1 \lor type = 2$}
    \IF{$type = 1$}
        \STATE Generate $N_s$ candidates: $Pop_{aux}^{g+1} \leftarrow \text{CDP}(Pop_{aux}^g \cup Off, \epsilon = 0)$ \COMMENT{Phase-3: Exploitation}
    \ELSE
        \STATE Generate $N_s$ candidates: $Pop_{aux}^{g+1} \leftarrow \text{CDP}(Pop_{aux}^g \cup Off, \epsilon)$
    \ENDIF

\ELSE
    \STATE $Pop_{aux}^{g+1} \leftarrow$ Algorithm~\ref{algo:VAENV}  \COMMENT{Phase-2: Angle-based Selection}
\ENDIF

\RETURN $Pop_{main}^{g+1}$, $Pop_{aux}^{g+1}$, $cnt$, $type$
\end{algorithmic}
\end{algorithm}

In addition, the opposite population mechanism was adapted from \cite{qiao2024cooperative}  to generate offspring to prevent algorithm degradation on problems with deceptive constraints. The generative opposite population mechanism produces these offspring according to the following procedure:
\begin{align}    \label{eq:opposite}
    \begin{cases} 
        tc = \tanh(\log(popsize) * 0.8) \\
        off_i = (lb_i + up_i) \cdot tc - decs_i\\
        Off_{3} = (off_1,off_2,...,off_{popsize})
    \end{cases}
\end{align}
Here, $Off_3$ represents the opposite generations of popAux, $lb_i$ and $up_i$ are the lower and upper bounds of the current CMOP decision variables, respectively. $off_i$ denotes a single solution, $decs_i$ represents the decision variables of the $i$-th solution in popAux, and $tc$ serves as a conversion coefficient. The pseudocode for Stage 2 of LLM4CMO is summarized in Algorithm~\ref{algo:stage2}.

In summary, this work focuses on improving three key algorithmic modules:  a) HOps, b) Epsilon decay function and c) DRA mechanism. The pseudocode of LLM4CMO is summarized in Algorithm \ref{algo:base}.

\begin{algorithm}
 \footnotesize
    \begin{algorithmic}[1]
        \caption{ \footnotesize LLM4CMO algorithm Base Framework}
        \label{algo:base}
        \REQUIRE Problem (CMOPs), $MaxFE$ (Evaluation counts), $N$ (Population size).
        \ENSURE $Pop_{main}^{final}$
        \STATE $Pop_{main}^{0} \leftarrow RandomInitialization(N)$ 
        \STATE $Pop_{aux}^{0} \leftarrow RandomInitialization(N)$
        \STATE $g \leftarrow 1$, $flag \leftarrow 0$, $R_s \leftarrow 1$, $S_t \leftarrow 0$ \COMMENT{Initialization}
        \WHILE{$FE < MaxFE$}
            \IF{$flag = 0$}
                \STATE Update $r_s^{g}$ by Eq. \eqref{eq:rs} \COMMENT{Distance discriminant}
                \STATE Update $S_t$ by Eq. \eqref{eq:rsc} \COMMENT{Switch condition}
                
                \IF{$S_t = 1$}
                    \STATE $flag \leftarrow 1$
                    \STATE $Switch \leftarrow Problem.FE$
                \ENDIF
                
                \STATE Update $Pop_{main}^{g},Pop_{aux}^{g}$ by Algorithm \ref{algo:stage1} \COMMENT{Stage-1}
            \ELSE
                \STATE Update $Pop_{main}^{g},Pop_{aux}^{g}$ by Algorithm \ref{algo:stage2} \COMMENT{Stage-2}
            \ENDIF
            \STATE $g \leftarrow g + 1$
        \ENDWHILE
        \STATE \textbf{return} $Pop_{main}^{final}$
    \end{algorithmic}
\end{algorithm}
    
\subsection{LLM-aided Design of Core Modules}

The flowchart of the design modules is shown in Fig. \ref{fig:Base framework}. We employed prompt template design, feedback mechanisms, and interactive processes, structuring them in the sequence of HOps, epsilon decay function, and DRA. Since operators are fundamental to algorithmic performance, we ensure that the entire design process is robust and employs iterative optimization to achieve optimal conditions. For LLM selection, we use DeepSeek-R1~\cite{guo2025deepseek} to complete the design of all modules. (The training data cutoff date for DeepSeek-R1 is July 2024.).The parameter information of DeepSeek-R1 is set by default from the official website. We design task-specific prompt templates based on individual modules and incorporate observational information gathered during actual testing.


\subsubsection{HOps}
Operators constitute the fundamental components of CMOEAs, determining both population diversity and evolutionary direction. Leveraging the UPF-CPF relationship, various HOps have already achieved satisfactory performance. We define combination sets encompassing mating pools, operators, and environmental selection mechanisms. Table~\ref{tab:MOE} summarizes all HOps task configurations, while Algorithm~\ref{algo:HOps} presents the optimized HOps design generated through LLM-aided methodology. The designer's provision of suggestions or attempts based on context also constitutes a vital part of the design process. Unlike automatic search methods, such as grid search, the integration of LLMs allows humans to utilize LLMs as a tool for algorithm design in a more intuitive way, while also achieving a certain degree of interpretability.

    \begin{table}[H]
    \centering
    \caption{Mating Pools, Operators and Environment Select methods are used in HOps LLM-aided design task.}
    \label{tab:MOE}
    \resizebox{\linewidth}{!}{%
    \begin{tabular}{ccc}
    \hline
    \textbf{}                   & \textbf{PopMain}                          & \textbf{PopAux}                         \\ \hline
    \textbf{Mating Pools}       & \multicolumn{2}{c}{{[}tournament,tournament{]} or {[}tournament,random{]}}          \\
    \textbf{Operators}      & \multicolumn{2}{c}{{[}DE,GA,DE-current-to-rand,DE-current-to-pbest,DE-transfer{]}}  \\
    \textbf{Environment Select} & \multicolumn{2}{c}{{[}CDP, epsilon-CDP, epsilon with angle vector and subregion{]}} \\ \hline
    \end{tabular}%
    }
    \end{table}
\subsubsection{Epsilon Decay Function}
The epsilon decay function controls Stage 2 operations for popAux. In Stage 1, popAux employs an epsilon method with epsilon set to infinity, but in Stage 2, we require popAux to converge toward popMain. This convergence aims to ensure adequate feasible solution counts and promote convergence of both popAux and popMain to similar feasible regions.

We implement an initial epsilon decay function defined by the following equation from:
    \begin{equation}
        \label{eq:epsilon init}
        \epsilon = \epsilon_0 \cdot \exp^{-20*(FE-Switch)/(maxFE-Switch)}
    \end{equation}
where $\epsilon_0$ is the initialization value, which is set to 0.2; $\epsilon$ is the current epsilon value; $FE$ is the current evaluation count; $maxFE$ is the maximum number of function evaluations; and $Switch$ is the transition point where the algorithm switches from Stage 1 to Stage 2.

For the base framework, we use epsilon and UPF-CPF type controls to regulate the entire Stage 2 process, as shown in Algorithm~\ref{algo:stage2}.

\subsubsection{DRA Mechanism}
The DRA mechanism controls the strength of offspring generation by assigning factors $f_1$ and $f_2$ to popMain and popAux, respectively. Without DRA, $f_1$ and $f_2$ are computed as follows:
    \begin{align}
            f_1 \cdot count_1 + f_2 \cdot count_2 = 2
    \end{align}
where $f_1$ and $f_2$ represent the generation factors for individual operators in popMain and popAux, respectively, and $count_1$ and $count_2$ denote the number of operators in each population. Through interaction with the LLM, we obtained an improved DRA mechanism. The decision function is shown in Eq. \eqref{eq:DDRA} and the DRA is shown in Algorithm \ref{algo:DRA}.

\begin{footnotesize}
    \small
    \begin{align}
    \label{eq:DDRA}
    f_1,f_2 = 
    \begin{cases} 
    \left(\frac{S-3f_2}{2},\ 0.25{+}r\cdot\frac{S-1.25}{3}\right), & \text{type=3},\, \text{cnt$>$3} \\
    \left(S-3f_2,\ 0.25{+}r\cdot\frac{S-1}{3}\right), & \text{type=3} \\
    \left(0.25{+}r_1\cdot\frac{S-1}{2},\ 0.25{+}r_2\cdot\frac{S-1}{2}\right), & \text{otherwise}
    \end{cases}
    \end{align}
    \vspace{-1em}
\end{footnotesize} 

\noindent where $S$ is determined by the parameter $ll$ and $denom$ as shown in Algorithm~\ref{algo:DRA}. The ratios $r_1$ and $r_2$ are computed as $r_1 = fr_1/denom$ and $r_2 = fr_2/denom$, respectively.

\subsubsection{Design Process}
We employed the prompt templates illustrated in supplementary materials (Sec.~\ref{sec:A8}) as foundational design frameworks, developing them sequentially: HOps, epsilon decay function, and DRA mechanism. Our interactive LLM-based design methodology uses comprehensive downstream task performance as the termination criterion. The design process incorporates human guidance through flexible natural language prompts, which fall into three categories: structural modifications, local parameter adjustments, and performance feedback. Interaction continues until satisfactory performance is achieved, at which point optimal algorithmic configurations are selected. Each module addresses distinct optimization objectives, necessitating specialized design strategies.
\begin{algorithm}
 \footnotesize
\caption{ \footnotesize Environment Selection Based on Vector and Angle Subregion}
\label{algo:VAENV}
\begin{algorithmic}[1]
\REQUIRE $Pop_{aux}^g$, $Off_1$, $Off_2$, $Off_3$, $N_s$, $\epsilon$
\ENSURE $Pop_{aux}^{g+1}$

\STATE $Off \leftarrow Off_1 \cup Off_2 \cup Off_3$
\STATE $A \leftarrow Pop_{aux}^g \cup Off$
\STATE Compute dominance relationship in $A$ to get set $S$
\STATE $[Objs, Cons, CV] \leftarrow \text{Extract from } S$
\STATE Generate reference vectors $W$ with $N_s$ directions in $m$-objective space
\STATE $z_{\max}, z_{\min} \leftarrow \text{Compute from } CV$
\STATE Normalize objectives: $ObjNorm \leftarrow \text{Normalize}(Objs, z_{\max}, z_{\min})$
\STATE $h \leftarrow \text{Minimum angular 2-norm distance between } ObjNorm \text{ and } W$

\FOR{$i = 1$ to $N_s$}
    \STATE $A_1 \leftarrow \{ x \in A \mid \text{AngularDistance}(x, W_i) < h \}$
    \IF{$A_1 = \emptyset$}
        \STATE $A_1 \leftarrow \{ x \in A \mid \text{AngularDistance}(x, W_i) = \min \}$
        \STATE \textbf{continue}
    \ENDIF

    \STATE Compute total score $T = CV(x) + \text{AngularDistance}(x, W_i)$ for $x \in A_1$
    \STATE $FA \leftarrow \{ x \in A_1 \mid T(x) \leq 0 \}$
    \IF{$FA = \emptyset$}
        \STATE Select $x^* \in A_1$ with minimum angular distance
    \ELSE
        \STATE Select $x^* \in FA$ with minimum $T$
    \ENDIF
\ENDFOR

\STATE $P_1 \leftarrow$ selected solutions from $A$
\STATE $P_2 \leftarrow$ remaining solutions in $A$
\STATE $N_p \leftarrow \text{length}(Pop_{aux}^{g})$

\IF{$N_p < 25$}
    \STATE $P \leftarrow P_1 \cup P_2[1:25-N_p]$
\ELSE
    \STATE $P \leftarrow P_1$
\ENDIF

\STATE Compute fitness using $\epsilon$: $Fitness \leftarrow \text{ComputeFitness}(P, \epsilon)$
\STATE $Pop_{aux}^{g+1} \leftarrow$ Top-$N_s$ ranked solutions in $P$ by $Fitness$
\RETURN $Pop_{aux}^{g+1}$
\end{algorithmic}
\end{algorithm}

\begin{algorithm}
 \footnotesize
    \caption{ \footnotesize DRA designed by LLM-aided}
    \label{algo:DRA}
    \begin{algorithmic}[1]
        \REQUIRE $type$, $ll$, $fr_1$, $fr_2$, $cnt$
        \ENSURE $f_1$, $f_2$
        \STATE $\text{denom} \gets f_1 + f_2 + 1$ \COMMENT{Total}
        \STATE $postive_{ll} \leftarrow max(0,ll)$ \COMMENT{Fix}
        \STATE $corr \leftarrow (fr_2 / donom) \times postive_{ll}$
        \STATE $S \leftarrow 2 + corr$
        \STATE get $f_1$,$f_2$ by Eq. (\ref{eq:DDRA}). \COMMENT{Allocation}
        \STATE $f_1 f_2,\leftarrow max(f_\cdot,0.25+1e-6)$ 
        \RETURN $f_1$,$f_2$
    \end{algorithmic}
\end{algorithm}

\begin{algorithm}
 \footnotesize
    \begin{algorithmic}[1]
        \caption{\footnotesize Generate Offspring by HOps}
        \label{algo:HOps}
        \REQUIRE $N$, $Pop_{main}^g$, $Pop_{aux}^g$, $type$, $f_1$, $f_2$, $cnt$
        \ENSURE  $Off_1$,$Off_2$
        \STATE Get Mating Pools methods and Operators by $type$
        \STATE Fix Mating pools counts by $f_1$ and $f_2$
        \STATE Prepare Mating pools $M1_1,M1_2,M2_1,M2_2$
        \STATE Generation Offspring $Off1_1,Off1_2,Off2_1,..$ from the everyone operator
        \STATE $Off_1,Off_2 \leftarrow$ Combination Offspring
    \RETURN $Off_1$, $Off_2$
    \end{algorithmic}
\end{algorithm}
For HOps, we employed the LLM to determine appropriate mating pool selections and operator combinations for different UPF-CPF types, while environmental selection strategies were relatively straightforward. We tested operator performance on downstream tasks using given schemes to guide the interaction process. By informing the LLM of observed phenomena and providing descriptions of operator performance along with fine-tuning suggestions for specific settings, we obtained the final strategy through multiple rounds of interaction, as shown in Table~\ref{tab:final HOps}.

\begin{table}[!]
\centering
\caption{HOps is designed by LLM-aided. T represents tournament selection and R represents random selection to get mating pools.}
\label{tab:final HOps}
\resizebox{0.99\linewidth}{!}{%
\begin{tabular}{ccccc}
\hline
\multirow{2}{*}{Type} & \multicolumn{2}{c}{Operators} & \multicolumn{2}{c}{Mating pools} \\
 & PopMain & PopAux & PopMain & PopAux \\ \hline
Type-1 & $DE,GA$ &\texttt{$DE_{trans},DE_{pbest}$} & T\&T & T\&T \\
Type-2 & $GA,DE$ &$DE_{trans},DE_{pbest}$ & T\&R & T\&R \\
Type-3 & \text{$DE_{rand}$} & $DE_{rand},DE_{pbest},DE$ & T\&R & T\&R \\
Type-4 & $GA,DE$ & $DE_{trans},DE_{rand},DE_{pbest}$ & T\&R & T\&R \\ \hline
\end{tabular}%
}
\end{table}

\begin{table}[H]
    \centering
    \caption{The process of HV charge during we get HOps by LLM interaction.}
    \label{tab:HOps}
    \resizebox{0.99\linewidth}{!}{%
    \begin{tabular}{ccccc}
    \hline
    \textbf{Problem} & \textbf{HOps1} & \textbf{HOps2} & \textbf{HOps3} & \textbf{HOps4} \\ \hline
    \textbf{CF1} & 5.6323e-1 (6.41e-4) = & 5.6282e-1 (5.67e-4) - & {\color[HTML]{3333E9} \textbf{5.6349e-1 (5.70e-4) =}} & 5.6349e-1 (4.06e-4) \\
    \textbf{CF2} & 6.7739e-1 (7.43e-4) = & 6.7617e-1 (1.35e-3) - & 6.7724e-1 (7.02e-4) = & {\color[HTML]{3333E9} \textbf{6.7747e-1 (7.76e-4)}} \\
    \textbf{CF3} & 2.4577e-1 (4.33e-2) = & 2.4294e-1 (4.60e-2) = & 2.3671e-1 (4.47e-2) = & {\color[HTML]{3333E9} \textbf{2.5257e-1 (4.45e-2)}} \\
    \textbf{DASCMOP1} & 1.9651e-1 (4.85e-2) - & 2.1242e-1 (2.96e-4) = & {\color[HTML]{3333E9} \textbf{2.1252e-1 (2.74e-4) =}} & 2.1241e-1 (4.52e-4) \\
    \textbf{DASCMOP2} & {\color[HTML]{3333E9} \textbf{3.5547e-1 (1.71e-4) +}} & 3.5509e-1 (1.00e-4) = & 3.5510e-1 (9.06e-5) = & 3.5512e-1 (8.63e-5) \\
    \textbf{DASCMOP3} & 2.2657e-1 (3.81e-2) - & 3.0379e-1 (2.24e-2) = & {\color[HTML]{3333E9} \textbf{3.0654e-1 (1.92e-2) =}} & 3.0055e-1 (2.97e-2) \\
    \textbf{LIRCMOP1} & 2.1144e-1 (1.39e-2) - & 2.3648e-1 (1.29e-3) = & {\color[HTML]{3333E9} \textbf{2.3663e-1 (6.87e-4) =}} & 2.3642e-1 (1.02e-3) \\
    \textbf{LIRCMOP5} & 2.8183e-1 (5.32e-2) = & 2.8169e-1 (5.32e-2) = & {\color[HTML]{3333E9} \textbf{2.9151e-1 (2.61e-4) =}} & 2.9146e-1 (2.62e-4) \\
    \textbf{LIRCMOP9} & 4.5704e-1 (7.30e-2) = & {\color[HTML]{3333E9} \textbf{5.1183e-1 (7.14e-2) =}} & 4.9702e-1 (7.96e-2) = & 4.7458e-1 (7.36e-2) \\
    \textbf{LIRCMOP11} & 6.6989e-1 (2.36e-2) = & 6.7037e-1 (4.35e-2) - & 6.7262e-1 (3.11e-2) = & {\color[HTML]{3333E9} \textbf{6.7887e-1 (2.66e-2)}} \\
    \textbf{MW3} & 5.4442e-1 (4.73e-4) = & 5.4358e-1 (5.36e-4) - & 5.4447e-1 (5.74e-4) = & {\color[HTML]{3333E9} \textbf{5.4457e-1 (4.00e-4)}} \\
    \textbf{MW4} & 8.4146e-1 (4.04e-4) = & 8.3228e-1 (4.53e-3) - & {\color[HTML]{3333E9} \textbf{8.4147e-1 (3.54e-4) =}} & 8.4133e-1 (4.70e-4) \\
    \textbf{MW5} & 3.2438e-1 (2.21e-4) = & 3.2430e-1 (1.68e-4) - & 3.2432e-1 (3.73e-4) = & {\color[HTML]{3333E9} \textbf{3.2444e-1 (1.60e-4)}} \\
    \textbf{MW9} & 3.9672e-1 (2.55e-3) + & 3.9482e-1 (2.92e-3) = & 3.9463e-1 (2.27e-3) = & 3.9387e-1 (4.03e-3) \\ \hline
    \textbf{+/-/=} & 2/3/9 & 0/6/8 & 0/0/14 &  \\
    \textbf{Best} & 1 & 1 & 6 & 5 \\ \hline
    \end{tabular}%
    }
\end{table}
We adopted a diverse set of CMOPs with different UPF-CPF relationship types from various benchmark test suites to optimize three core modules. During the optimization process, each of the three modules achieved good performance despite varying difficulty levels. For HOps, our objective was to identify combinations with minimal shortcomings, indicating relatively good performance across different UPF-CPF type CMOPs. Given the enormous difficulty and resource consumption required for finding optimal combinations in such combinatorial optimization problems, our initial HOps design used URCMO as the baseline reference (achieving superior performance in 4 cases and comparable performance in 7 cases). We iteratively revised and generated three additional HOps recommendations through LLM interaction, comparing their performance changes as shown in Table~\ref{tab:HOps}. For the HOps task, we employed lightweight LLM interventions to derive near-optimal combinations. Upon obtaining the final HOps design, we proceeded to the epsilon decay function design task, which followed a workflow analogous to the HOps task. The results are presented in Table~\ref{tab:EDF}. To validate effectiveness, we selected a subset of promising interaction results for testing, noting that this task exhibited greater complexity than the HOps task. For DRA, minimal LLM interaction was required, confined to template prompting, which yielded appropriately calibrated DRA functions.

We observed that when results from HOps and epsilon decay function designs exceeded the performance of previous modules, we continued interacting with the LLM to achieve further improvements. We conducted sensitivity tests and analyses on the epsilon parameter for the first generated HOps1 and the last generated HOps4, respectively, to rule out the possibility that the performance improvement is caused by hyperparameter sensitivity rather than design. Result in supplementary material Sec.~\ref{sec:SA}.

\begin{table*}[htb!]
\centering
\caption{The process of HV charge during we get epsilon decay functions by LLM interaction.}
\label{tab:EDF}
\resizebox{\linewidth}{!}{%
\begin{tabular}{ccccccccc}
\hline
\textbf{Problem} & \textbf{HOps4} & \textbf{Eps2} & \textbf{Eps3} & \textbf{Eps4} & \textbf{Eps5} & \textbf{Eps6} & \textbf{Eps7} & \textbf{Eps8} \\ \hline
\textbf{CF1} & 5.6349e-1 (4.06e-4) = & 5.6379e-1 (8.10e-4) + & 5.6410e-1 (7.35e-4) + & {\color[HTML]{3333E9} 5.6416e-1 (4.80e-4) +} & 5.6234e-1 (6.93e-4) - & 5.6277e-1 (8.27e-4) - & 5.6351e-1 (7.12e-4) = & 5.6336e-1 (8.72e-4) \\
\textbf{CF2} & {\color[HTML]{3333E9} 6.7747e-1 (7.76e-4) =} & 6.7698e-1 (1.44e-3) = & 6.7719e-1 (1.05e-3) = & 6.7706e-1 (1.39e-3) = & 6.7730e-1 (7.92e-4) = & 6.7715e-1 (9.76e-4) = & 6.7703e-1 (1.02e-3) = & 6.7704e-1 (1.13e-3) \\
\textbf{CF3} & 2.5257e-1 (4.45e-2) = & 2.4684e-1 (3.36e-2) = & 2.5328e-1 (3.65e-2) = & 2.4885e-1 (4.49e-2) = & 2.2139e-1 (4.56e-2) - & 2.4796e-1 (3.27e-2) = & {\color[HTML]{3333E9} 2.5663e-1 (3.56e-2) =} & 2.4963e-1 (3.95e-2) \\
\textbf{DASCMOP1} & 2.1241e-1 (4.52e-4) - & 2.1217e-1 (3.86e-4) - & 2.1216e-1 (4.04e-4) - & {\color[HTML]{3333E9} 2.1272e-1 (3.59e-4) =} & 2.1264e-1 (3.04e-4) = & 2.1250e-1 (2.92e-4) - & 2.1249e-1 (3.26e-4) - & 2.1270e-1 (3.28e-4) \\
\textbf{DASCMOP2} & 3.5512e-1 (8.63e-5) - & 3.5505e-1 (7.58e-5) - & 3.5503e-1 (9.63e-5) - & 3.5510e-1 (9.29e-5) - & 3.5512e-1 (8.64e-5) - & 3.5512e-1 (1.01e-4) - & 3.5510e-1 (8.86e-5) - & {\color[HTML]{3333E9} 3.5518e-1 (8.34e-5)} \\
\textbf{DASCMOP3} & 3.0055e-1 (2.97e-2) - & 3.1221e-1 (1.98e-4) - & 3.1216e-1 (1.86e-4) - & 3.0689e-1 (1.78e-2) = & 3.1234e-1 (1.15e-4) = & 3.0676e-1 (1.99e-2) - & 3.1229e-1 (2.08e-4) - & {\color[HTML]{3333E9} 3.1239e-1 (1.01e-4)} \\
\textbf{LIRCMOP1} & 2.3642e-1 (1.02e-3) - & 2.3648e-1 (2.43e-4) - & 2.3649e-1 (3.24e-4) - & 2.3690e-1 (3.28e-4) = & 2.3695e-1 (3.41e-4) = & {\color[HTML]{3333E9} 2.3700e-1 (5.16e-4) =} & 2.3682e-1 (3.89e-4) = & 2.3692e-1 (4.61e-4) \\
\textbf{LIRCMOP5} & 2.9146e-1 (2.62e-4) - & 2.9150e-1 (5.39e-4) = & 2.9158e-1 (2.15e-4) = & 2.9158e-1 (2.91e-4) = & 2.9158e-1 (4.96e-4) = & 2.9159e-1 (3.07e-4) = & 2.9146e-1 (5.34e-4) = & {\color[HTML]{3333E9} 2.9162e-1 (2.99e-4)} \\
\textbf{LIRCMOP9} & 4.7458e-1 (7.36e-2) - & 4.8011e-1 (7.71e-2) - & 4.9906e-1 (7.91e-2) = & 5.0676e-1 (6.92e-2) = & {\color[HTML]{3333E9} 5.4720e-1 (3.48e-2) =} & 5.3385e-1 (5.16e-2) = & 5.3920e-1 (4.18e-2) = & 5.3093e-1 (5.17e-2) \\
\textbf{LIRCMOP11} & 6.7887e-1 (2.66e-2) = & 6.5507e-1 (9.13e-2) = & 6.8519e-1 (2.26e-2) = & 6.8065e-1 (5.48e-2) = & 6.8749e-1 (1.92e-2) = & {\color[HTML]{3333E9} 6.9282e-1 (6.14e-3) =} & 6.9051e-1 (1.02e-2) = & 6.8834e-1 (1.27e-2) \\
\textbf{MW3} & 5.4457e-1 (4.00e-4) + & 5.4420e-1 (9.04e-4) = & 5.4442e-1 (5.68e-4) = & 5.4460e-1 (3.91e-4) + & 5.4457e-1 (3.53e-4) + & 5.4447e-1 (4.33e-4) = & {\color[HTML]{3333E9} 5.4465e-1 (2.90e-4) +} & 5.4431e-1 (4.97e-4) \\
\textbf{MW4} & 8.4133e-1 (4.70e-4) = & 8.4138e-1 (4.59e-4) = & 8.4134e-1 (4.09e-4) = & 8.4139e-1 (5.92e-4) = & 8.4140e-1 (5.73e-4) = & 8.4138e-1 (6.17e-4) = & {\color[HTML]{3333E9} 8.4158e-1 (3.41e-4) =} & 8.4152e-1 (3.29e-4) \\
\textbf{MW5} & 3.2444e-1 (1.60e-4) = & 3.2441e-1 (2.91e-4) = & 3.2426e-1 (7.23e-4) = & 3.2424e-1 (5.01e-4) = & {\color[HTML]{3333E9} 3.2447e-1 (8.67e-5) =} & 3.2434e-1 (4.27e-4) = & 3.2437e-1 (2.68e-4) = & 3.2430e-1 (4.26e-4) \\
\textbf{MW9} & 3.9387e-1 (4.03e-3) = & 3.9387e-1 (2.92e-3) = & 3.9419e-1 (2.34e-3) = & 3.9446e-1 (2.13e-3) = & 3.9495e-1 (2.17e-3) = & 3.9462e-1 (2.25e-3) = & {\color[HTML]{3333E9} 3.9557e-1 (2.16e-3) =} & 3.9474e-1 (3.33e-3) \\ \hline
\textbf{+/-/=} & 1/6/7 & 1/5/8 & 1/4/9 & 2/1/11 & 1/3/10 & 0/4/10 & 1/3/10 &  \\
\textbf{best} & 1 & 0 & 0 & 2 & 2 & 2 & 4 & 3 \\ \hline
\end{tabular}%
}
\end{table*}

\section{Experimental Study} \label{sec4}
\subsection{Experiment Settings}
\subsubsection{Metrics and Benchmark Test Suites}
For all CMOEAs, we evaluate their performance by comparing Hyper-Volume (HV)~\cite{zitzler2002multiobjective} values and Inverse Generation Distance (IGD)~\cite{bosman2003balance}. A lower IGD value and a higher HV value indicate better performance of the corresponding algorithm.
In this study, we select six benchmark test suites: CF~\cite{mashwani2016multiobjective}, DASCMOP~\cite{fan2020difficulty}, LIRCMOP~\cite{fan2019improved}, MW~\cite{ma2019evolutionary}, DOC~\cite{liu2019handling}, and FCP~\cite{yuan2021indicator}. Additionally, we test the proposed LLM4CMO on a series of real-world CMOPs~\cite{kumar2021benchmark}. 

These benchmark datasets are commonly used for evaluating CMOEA performance and cover most CMOP types. CF generally involves test problems with simple constraints that facilitate feasible region exploration, though achieving complete Pareto front (PF) diversity remains challenging. DASCMOP and MW employ complex combinatorial constraints that partition feasible regions into irregular shapes, with DASCMOP converging to the PF more slowly than MW. Both problem sets present exploration challenges but exhibit clearer UPF-CPF relationships compared to CF. LIRCMOP features extremely strict constraints and narrow decision spaces, making feasibility and diversity maintenance difficult and potentially causing algorithms to become trapped in local optima during search. DOC incorporates constraints in both decision variables and objectives, creating large distances between the UPF and CPF. FCP focuses on indicator-based constraint handling with complex constraint structures. We comprehensively evaluated our algorithm using these six benchmark test suites.

\subsubsection{Baseline Algorithms}

We compared our algorithm with 11 baseline algorithms: CCMO~\cite{tian2020coevolutionary}, CAEAD~\cite{zou2021dual}, cDEAP~\cite{ming2021dual}, MSCMO~\cite{ma2021multi}, CMOEAMS~\cite{tian2021balancing}, BiCO~\cite{liu2021handling}, URCMO~\cite{liang2022utilizing}, CMEGL~\cite{qiao2023evolutionary}, C3M~\cite{sun2022multistage}, CMOES~\cite{ming2024even}, and CMOEMT~\cite{ming2022constrained}. CCMO represents the first dual-population CMOEA that effectively utilizes UPF and CPF information. CAEAD introduces dynamic switching for Stage 2 to improve upon CCMO by enhancing utilization efficiency and represents the Progressive Preference Strategy (PPS) method. Bico use a archieve population to save high quality feasible solutions, coevoling with popMain. CMOEAMS proposes adaptive switching methods in Stage 2. cDEAP based on PPS methodology, balance UPF and CPF information to improve convergence and feasibility ratios. CMOES employs non-dominated solutions and distance functions to control diversity and explore promising regions. Both CMOEMT and CMEGL utilize auxiliary tasks and knowledge transfer, optimizing main tasks by leveraging information explored in auxiliary tasks. MSCMO and C3M are multi-stage CMOEAs that incorporate constraints through stage transitions. URCMO leverages UPF-CPF relationships and employs operator combinations to address different relationship types.

All algorithms were implemented in PlatEMO v4.12~\cite{tian2017platemo} and executed using MATLAB R2023a on Intel(R) Xeon(R) Gold 6248 CPU @2.50GHz. Each algorithm was independently run 30 times to obtain mean and standard deviation values for HV and IGD metrics. For different CMOPs, we set varying maximum numbers of fitness evaluations (maxFEs): CF (200,000), DASCMOP (150,000), MW and FCP (100,000), while LIRCMOP, DOC, and real-world CMOPs were all set to 300,000. Additionally, the Wilcoxon rank-sum test was performed at a significance level of 0.05 to assess statistical significance of performance differences between LLM4CMO and each baseline algorithm. The symbols +/-/= indicate that the baseline algorithm performs superior to, inferior to, or comparable with LLM4CMO, respectively.
The Multiproblem Wilcoxon’s signed-rank test is employed in algorithm comparison and ablation experiments. A significance level of 0.05 is used for the $p$-value, with a larger $R^+$ and a smaller $R^-$ being more desirable.
\begin{table}[htp!]
    \centering
    \caption{Wilcoxon Rank Sum Test Result and Multi-Problem Wilcoxon Signed Rank Test Results with Respect to HV and IGD on All 61 Functions}
    \label{tab:HVIGD}
    \resizebox{\linewidth}{!}{%
    \begin{tabular}{cccccc}
    \hline
    LLM4CMO vs. & HV(+/-/=) & $R^+$ & $R^-$ & $p$-value & level=0.05 \\ \hline
    CCMO & 48/7/6 & 1782.00 & 109.00 & 1.90E-09 & YES \\
    CAEAD & 54/4/3 & 1679.00 & 212.00 & 1.38E-07 & YES \\
    cDPEA & 46/7/8 & 1723.00 & 168.00 & 2.70E-09 & YES \\
    MSCMO & 52/5/4 & 1840.00 & 51.00 & 1.00E-10 & YES \\
    CMOEAMS & 51/5/5 & 1823.00 & 68.00 & 3.00E-10 & YES \\
    BiCo & 52/3/6 & 1830.00 & 61.00 & 2.00E-10 & YES \\
    CMEGL & 46/7/8 & 1744.00 & 147.00 & 9.70E-09 & YES \\
    C3M & 53/5/3 & 1674.00 & 217.00 & 1.67E-07 & YES \\
    CMOES & 48/5/8 & 1793.00 & 98.00 & 1.10E-09 & YES \\
    URCMO & 44/9/8 & 1589.00 & 302.00 & 3.80E-06 & YES \\
    CMOEMT & 46/5/10 & 1758.00 & 133.00 & 5.30E-09 & YES \\ \hline
    LLM4CMO vs. & IGD(+/-/=) & $R^+$ & $R^-$ & $p-$value & level=0.05 \\ \hline
    CCMO & 45/10/6 & 1755.00 & 136.00 & 6.10E-09 & YES \\
    CAEAD & 50/6/5 & 1607.00 & 284.00 & 2.02E-06 & YES \\
    cDPEA & 46/8/7 & 1778.00 & 113.00 & 2.20E-09 & YES \\
    MSCMO & 46/7/8 & 1830.00 & 61.00 & 2.00E-10 & YES \\
    CMOEAMS & 49/5/7 & 1813.00 & 78.00 & 5.00E-10 & YES \\
    BiCo & 52/6/3 & 1807.00 & 84.00 & 6.00E-10 & YES \\
    CMEGL & 43/11/7 & 1709.00 & 182.00 & 4.16E-08 & YES \\
    C3M & 49/5/7 & 1611.00 & 280.00 & 1.75E-06 & YES \\
    CMOES & 44/8/9 & 1748.00 & 143.00 & 8.20E-09 & YES \\
    URCMO & 36/10/15 & 1443.00 & 448.00 & 3.52E-04 & YES \\
    CMOEMT & 42/7/12 & 1668.00 & 223.00 & 2.11E-07 & YES\\
    \hline
    \end{tabular}%
    }
\end{table}

\subsection{Results Description and Analysis}
The detailed test results on the six test suites and real-world problems can be found in the supplementary materials (Sec.~\ref{sec:A4}).
\subsubsection{CF}
The CF test suite comprises conventional CMOPs with relatively lenient constraints, serving as fundamental test problems. However, for two-stage dual-population algorithms, maintaining solution diversity in such problems is challenging. Populations tend to evolve continuously in a specific direction, generating exclusively feasible solutions, which impedes algorithm convergence to the optimal PF. We compared LLM4CMO against 11 baseline algorithms, and results show that LLM4CMO achieves the best performance in terms of the HV metric. Specifically, LLM4CMO performs optimally on six problems (CF3-7 and CF9), while underperforming optimal baselines on the remaining four problems. Notably, our algorithm employs fuzzy processing for UPF-CPF types, and the operator combinations identified by the LLM may overly preserve diversity, making it difficult to further optimize problems with easily accessible feasible regions and ambiguous UPF-CPF relationships.
\subsubsection{MW and DASCMOP}
MW and DASCMOP are both irregular CMOPs. We compared algorithm performance on the MW test suite, where our algorithm achieved 7 optimal results in the HV metric, while BiCO and URCMO each achieved 2, and cDEAP achieved 1. To more intuitively compare the convergence of algorithm combinations found by the LLM when handling different UPF-CPF types, we analyzed performance across relationship categories. For types where UPF-CPF are completely overlapping (MW2, 4, 10), we achieved optimal results on two of the three problems. MW4 is a three-objective problem where LLM4CMO performed poorly, consistent with previous analyses, which may be attributed to excessive diversity introduced by HOps. Nevertheless, it still significantly outperforms the URCMO algorithm, which also employs hybrid operators. For problems with completely separated UPF-CPF in MW (MW9, 11, 12), our algorithm showed suboptimal performance, likely due to our excessive focus on problems with difficult-to-explore feasible regions and the overly aggressive HOps designed by the LLM, which lacks local optimization capability for such irregular CMOPs. LLM4CMO performs well on other problems with partial overlaps. Although partial analysis on MW reveals some issues with the modules designed by the large model, the algorithm remains effective on most problems, further confirming the effectiveness and significant potential of LLMs in recommending operator combinations and assisting in algorithm design. By plotting the IGD variation curves of different algorithms on the same problems in MW, we observed that LLM4CMO demonstrates stable convergence capability.

Compared to MW, DASCMOP exhibits slower convergence to both UPF and CPF, resulting in a longer first stage. In terms of HV values, LLM4CMO achieved optimal performance on 4 problems, while CMOEAMS obtained 2 optimal results, and CCMO, CMEGL, and URCMO each achieved 1. Additionally, LLM4CMO demonstrated relatively strong performance on several other problems. We observed that the URCMO algorithm, which also employs multi-operators, uses switch point transformation to learn UPF-CPF relationships, but its applicability significantly decreases when transitioning from MW to DASCMOP. Our method utilizes dynamic distance discriminant convergence combined with one-time transformation and dynamic update mechanism, effectively addressing the ineffective utilization of operator combinations across different classifications due to insufficient learning.
    
\subsubsection{LIRCMOP, DOC and FCP}
LIRCMOP, DOC, and FCP represent complex CMOPs characterized by narrow feasible regions, coupled constraints involving both decision variables and objectives, or deceptive constraints. For the LIRCMOP test suite, LLM4CMO achieved optimal results in 11 out of 14 problems for the HV metric, while URCMO and CMOEAMS achieved 1 and 2, respectively. To analyze the reasons for our algorithm's strong performance, LIRCMOP problems contain numerous narrow and irregular regions with many decision variables, which can cause some algorithms to become trapped in local optima. Through HOps, we effectively addressed and optimized the issue of algorithms becoming trapped in local optima for narrow problems. In the Stage 2 algorithm design, we utilized popAux to generate reference vectors and partition angular regions to explore points farther from the boundaries. The results demonstrate improved performance in quickly finding boundary solutions within narrow optimization regions such as LIRCMOP1-3. On the other hand, the synergy of these two mechanisms introduces challenges in handling increased uncertainty when the number of objectives grows. That is, our algorithm may overemphasize diversity, making it more difficult to balance additional objectives, particularly as dimensionality increases. Overall, the two-stage algorithm design of HOps and LLM4CMO effectively addresses most LIRCMOP problems and achieves state-of-the-art performance among the compared algorithms. Additionally, the IGD convergence curves on LIRCMOP problems demonstrate the stable convergence of our algorithm, gradually achieving higher optimization levels as it transitions to Stage 2.
    
In addressing the DOC problem, the UPF-CPF mechanism proved ineffective. Therefore, we assessed the performance of the HOps module on several challenging DOC tasks. The LLM4CMO algorithm demonstrated performance comparable to C3M in terms of the HV metric, achieving optimal results in three cases, similar to C3M. Upon analyzing the PF produced by our method, we observed that while our algorithm occasionally performed below the average baseline on certain DOC problems, it consistently delivered superior PF quality compared to other methods. These outcomes confirm the effectiveness of both the HOps module and epsilon control strategies in tackling DOC problems. Nonetheless, the observed instability in the algorithm aligns with previous findings: the diversity-rich HOps recommended by the LLM can inadvertently introduce instability due to their aggressive exploration capabilities. This phenomenon underscores significant untapped potential in refining HOps generated by LLM. Indeed, even minimal interactions with the LLM yielded operator combinations whose performance markedly surpassed manually designed URCMO configurations.

For the FCP test suite, due to our introduction of the Opposite method specifically designed to handle CMOPs with deceptive constraints such as those in FCP, LLM4CMO outperformed existing algorithms on problems FCP1–FCP5. The Opposite method effectively mitigates the influence of deceptive constraints, thus enabling our algorithm to solve a broader spectrum of CMOPs.

\subsubsection{Real-World CMOPs}
We further conducted evaluations on real-world CMOPs from established benchmarks~\cite{kumar2021benchmark}, including Pressure Vessel Design (PVD)~\cite{narayanan1999improving}, Vibrating Platform Design (VPD)~\cite{narayanan1999improving}, Two Bar Truss Design (TBTD)~\cite{chiandussi2012comparison}, Gear Box Design (GBD)~\cite{kumar2021benchmark}, and synchronous optimal pulse-width modulation problems for inverters at different levels(SOPM)~\cite{rathore2010synchronous,rathore2012generalized,edpuganti2015optimal,edpuganti2015fundamental,edpuganti2017optimal}. Detailed descriptions of these problems are available in the supplementary materials. We compared the proposed LLM4CMO algorithm against 11 baseline algorithms using the HV metric. The complete results are presented in Table~\ref{tab:rwhv}. Across the 10 evaluated problems, LLM4CMO achieved superior performance on five instances, whereas CMOEAMS achieved optimal results on two problems. URCMO, CMOEMT, and CMEGL each attained optimal results on one problem. Additionally, we observed consistently strong performance of LLM4CMO on all problems except for PVD and VPD. For detailed sources and additional information, please refer to the supplementary materials (Sec.~\ref{sec:A1}).

\subsubsection{Process in LIRCMOP1}

We compared the operational processes of BICO, URCMO, and LLM4CMO on the LIRCMOP problem set. Observations indicate that, compared to BICO, both URCMO and LLM4CMO emphasize learning in the early evolutionary stages, subsequently shifting to exploring the PF boundaries in the mid-stage. BICO achieves rapid convergence by employing an archived auxiliary population mechanism that balances exploration and exploitation; however, it tends to suffer from limited diversity. Conversely, URCMO and LLM4CMO both utilize hybrid operator to maintain solution diversity, especially evident at the 25\% and 50\% stages of the evolutionary process. Notably, LLM4CMO initiates vector-based environmental selection at the 25\% stage. By the 50\% stage, LLM4CMO’s popAux (P2) effectively converges toward popMain (P1). In the later stages, LLM4CMO covers both the boundary points of the PF and a broader set of feasible solutions, demonstrating superior diversity compared to URCMO. These observations confirm the feasibility and effectiveness of the HOps approach recommended by the LLM. For further analysis of evolutionary processes across different problem types, please refer to the supplementary materials (Sec.~\ref{sec:A7}).

    \begin{figure*}
        \centering
        \subfloat[]{\includegraphics[width=0.2\textwidth]{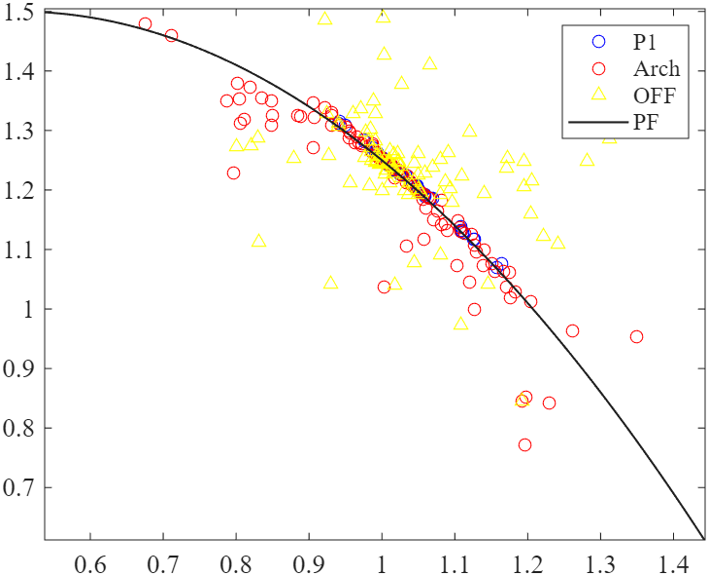}\label{fig:sub1BI1}}
        \hfill
        \subfloat[]{\includegraphics[width=0.2\textwidth]{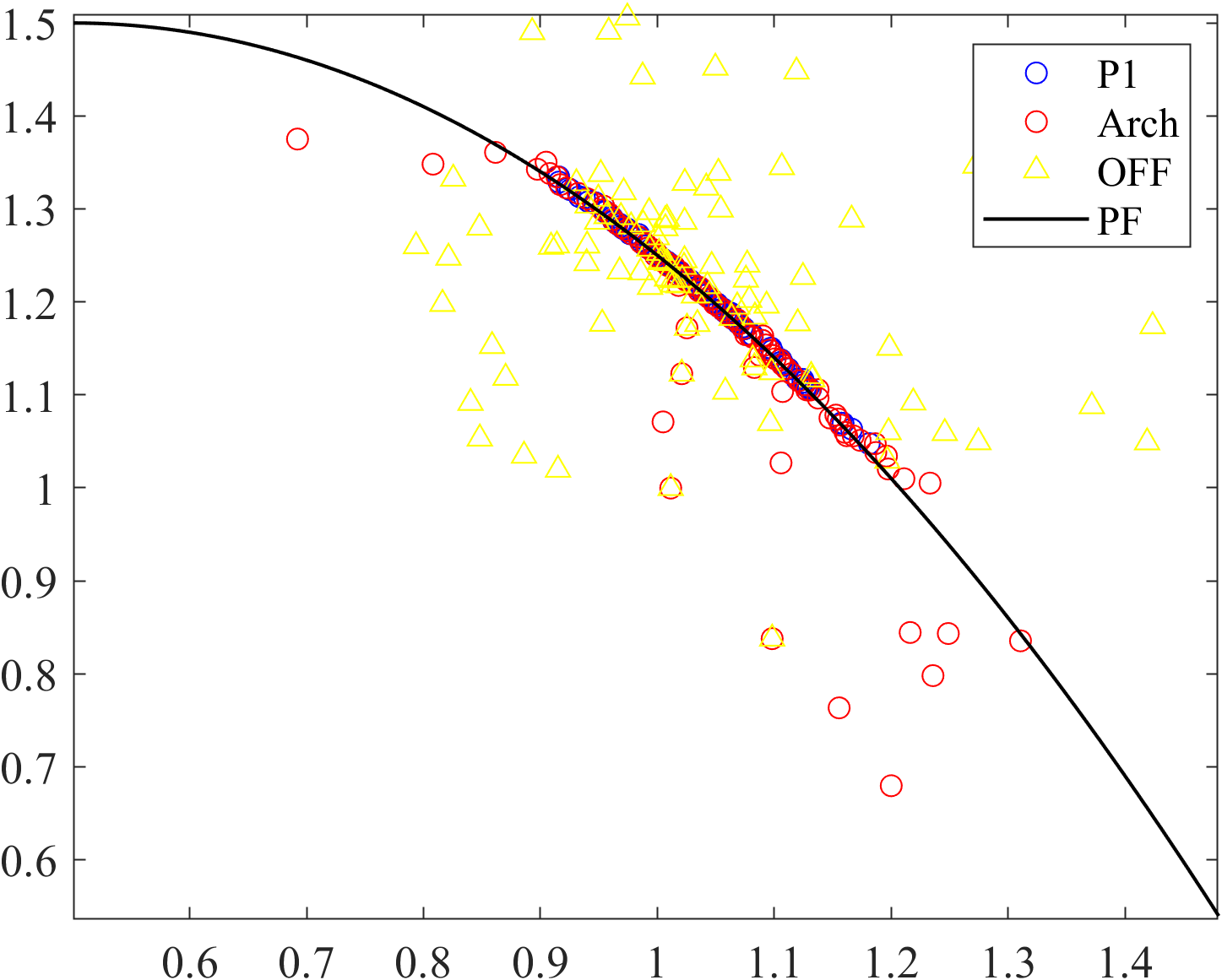}\label{fig:sub2BI2}}
        \hfill
        \subfloat[]{\includegraphics[width=0.2\textwidth]{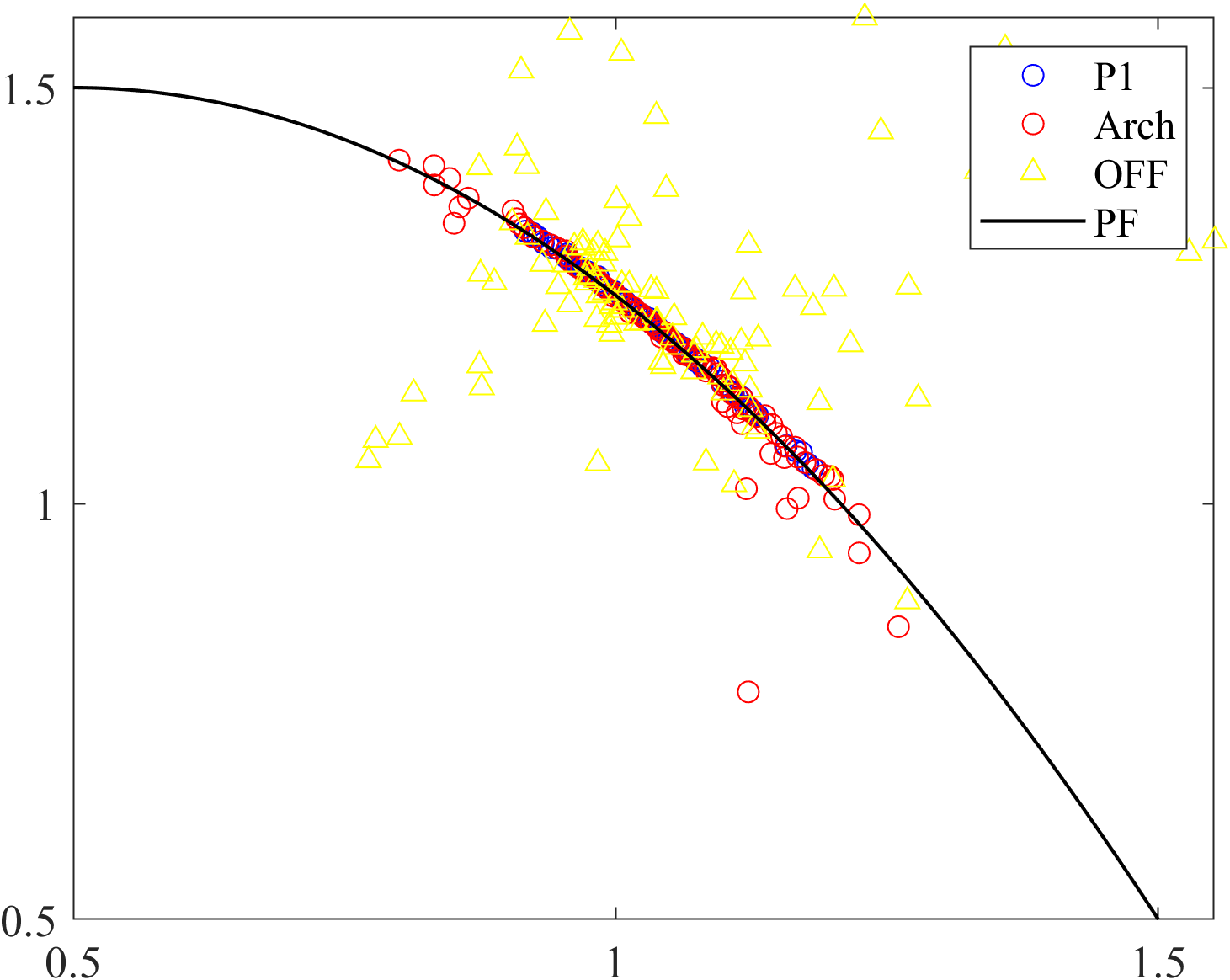}\label{fig:sub3BI3}}
        \hfill
        \subfloat[]{\includegraphics[width=0.2\textwidth]{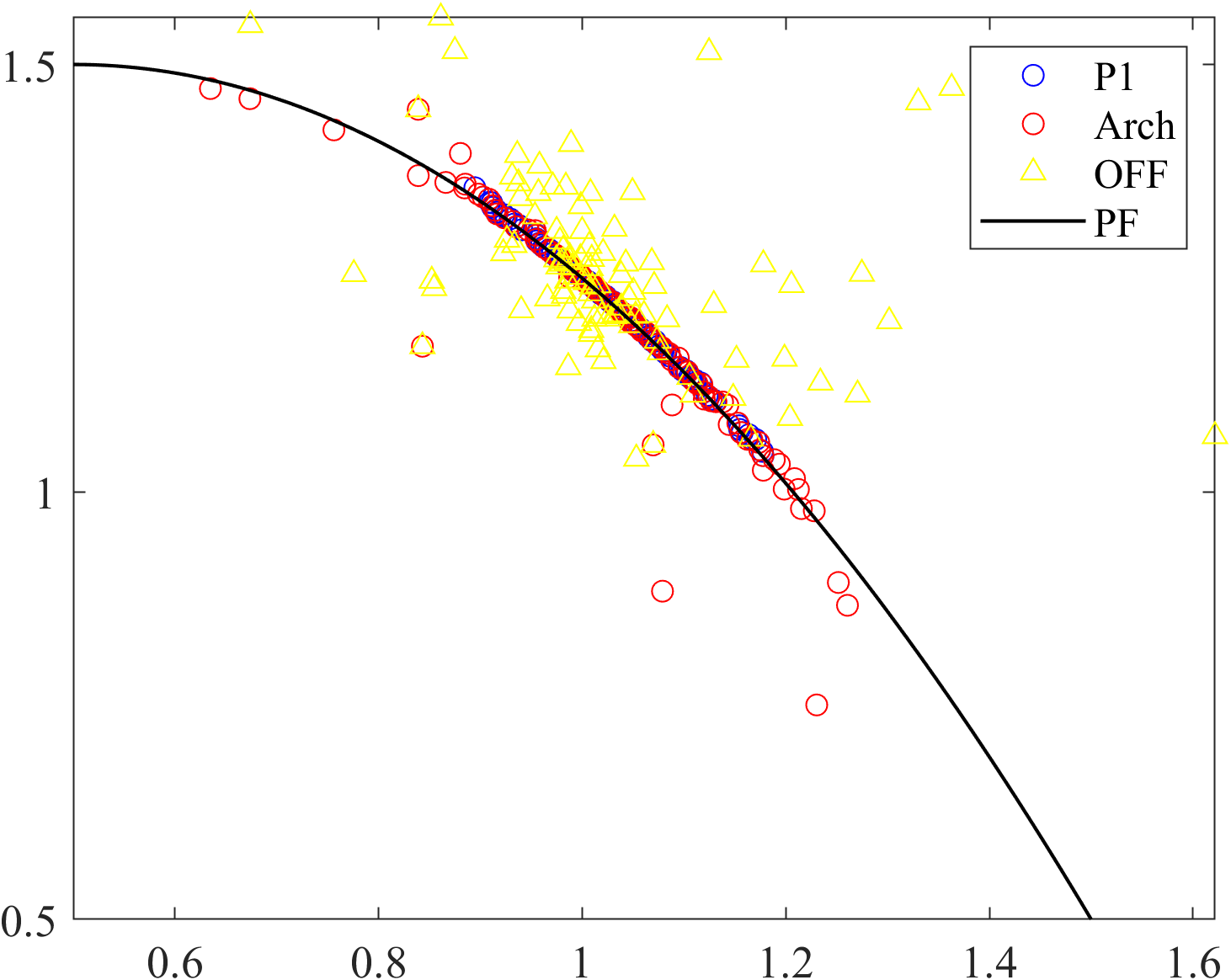}\label{fig:sub1BI5}}
        \\[1ex]
        \subfloat[]{\includegraphics[width=0.2\textwidth]{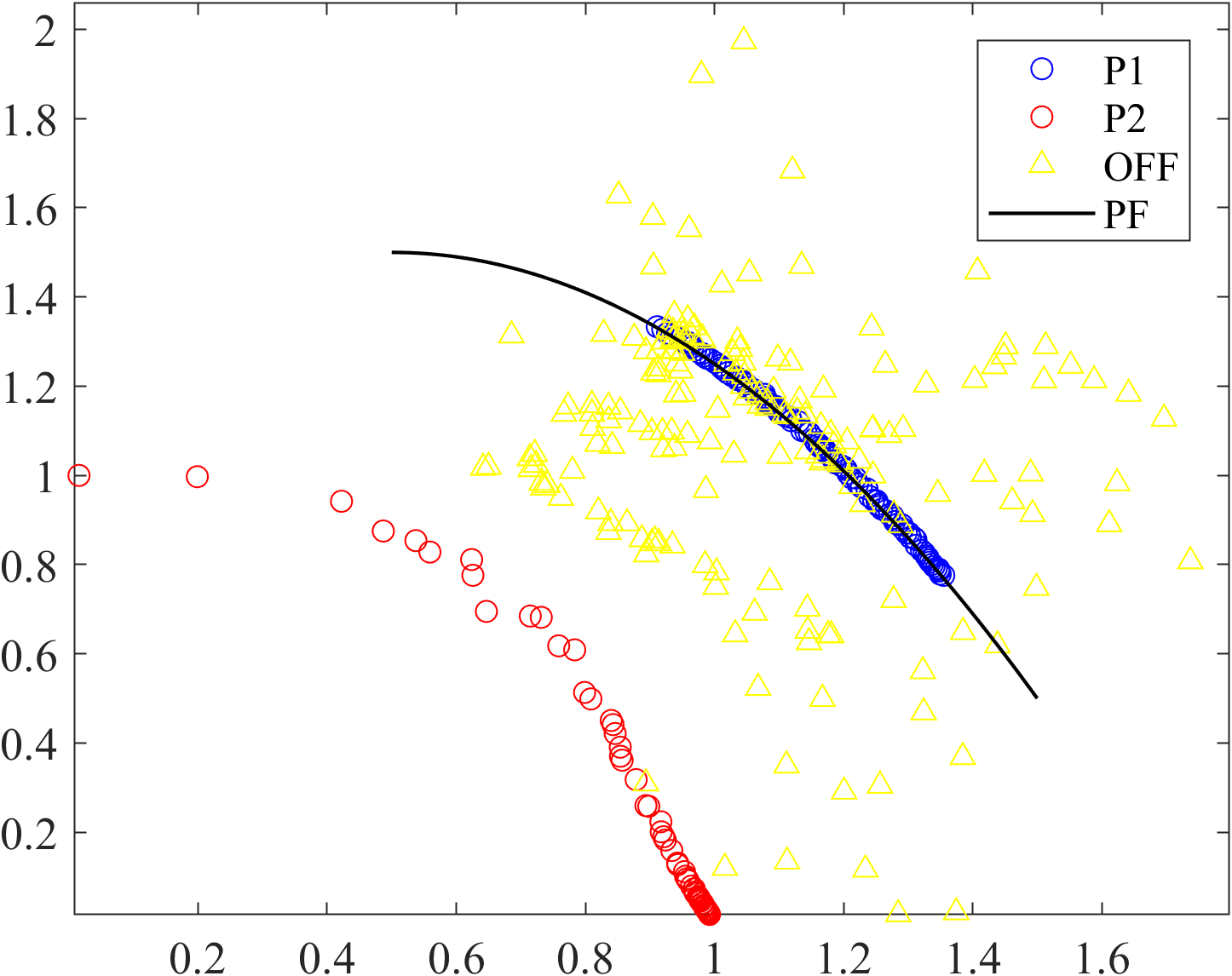}\label{fig:subU1}}
        \hfill
        \subfloat[]{\includegraphics[width=0.2\textwidth]{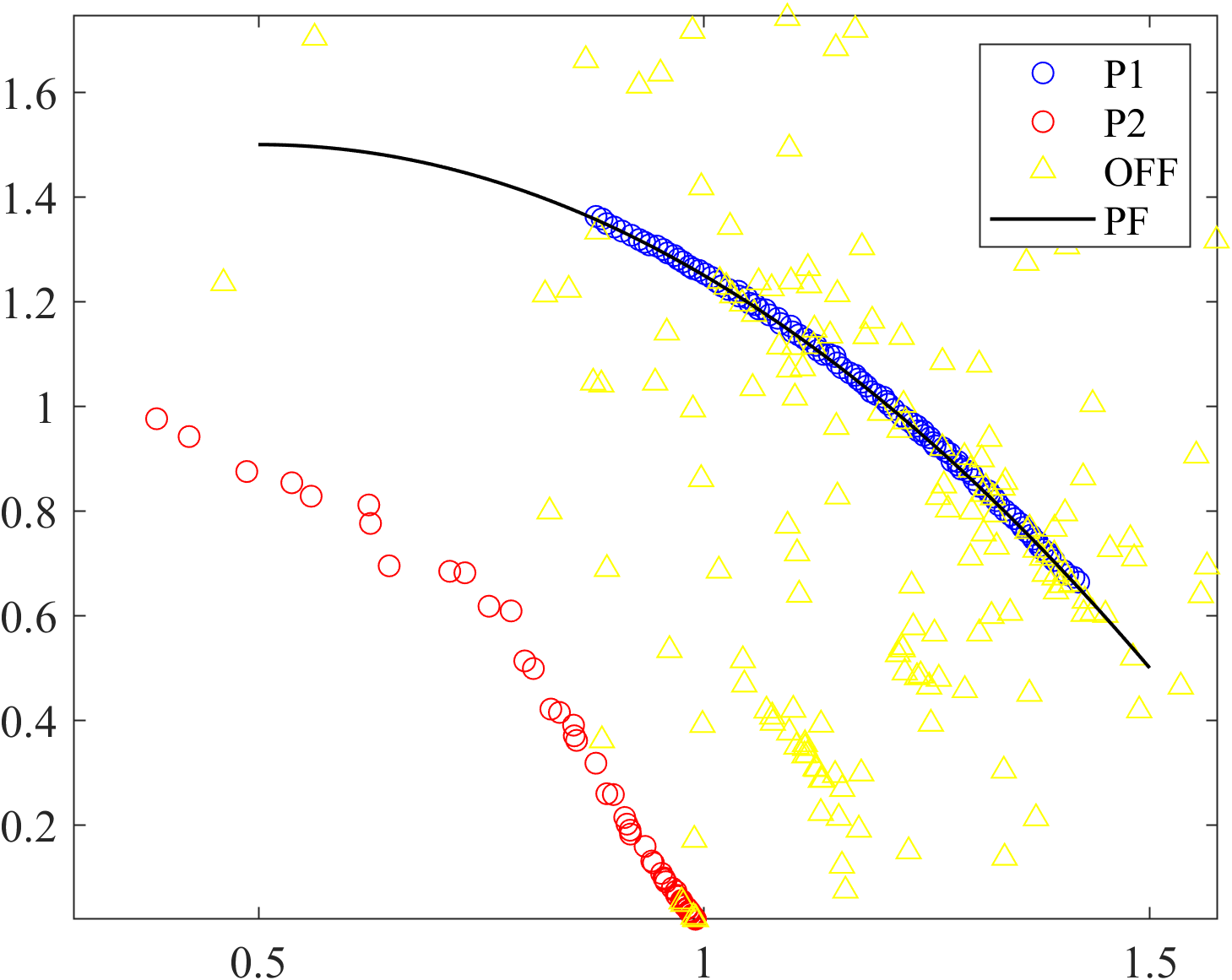}\label{fig:subU2}}
        \hfill
        \subfloat[]{\includegraphics[width=0.2\textwidth]{plt/U1.png}\label{fig:subU3}}
        \hfill
        \subfloat[]{\includegraphics[width=0.2\textwidth]{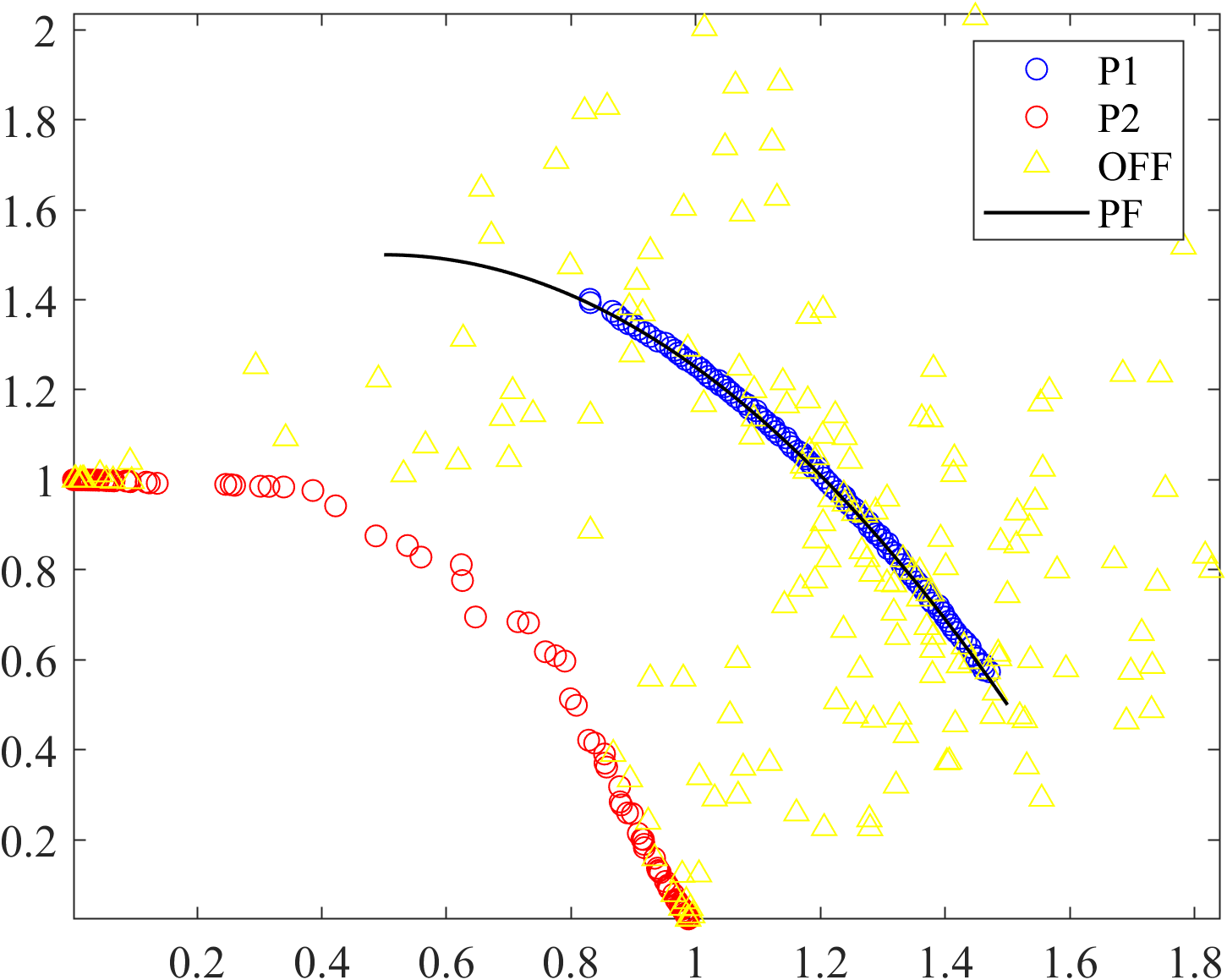}\label{fig:subU4}}
        \\[1ex]
        \subfloat[0\%]{\includegraphics[width=0.2\textwidth]{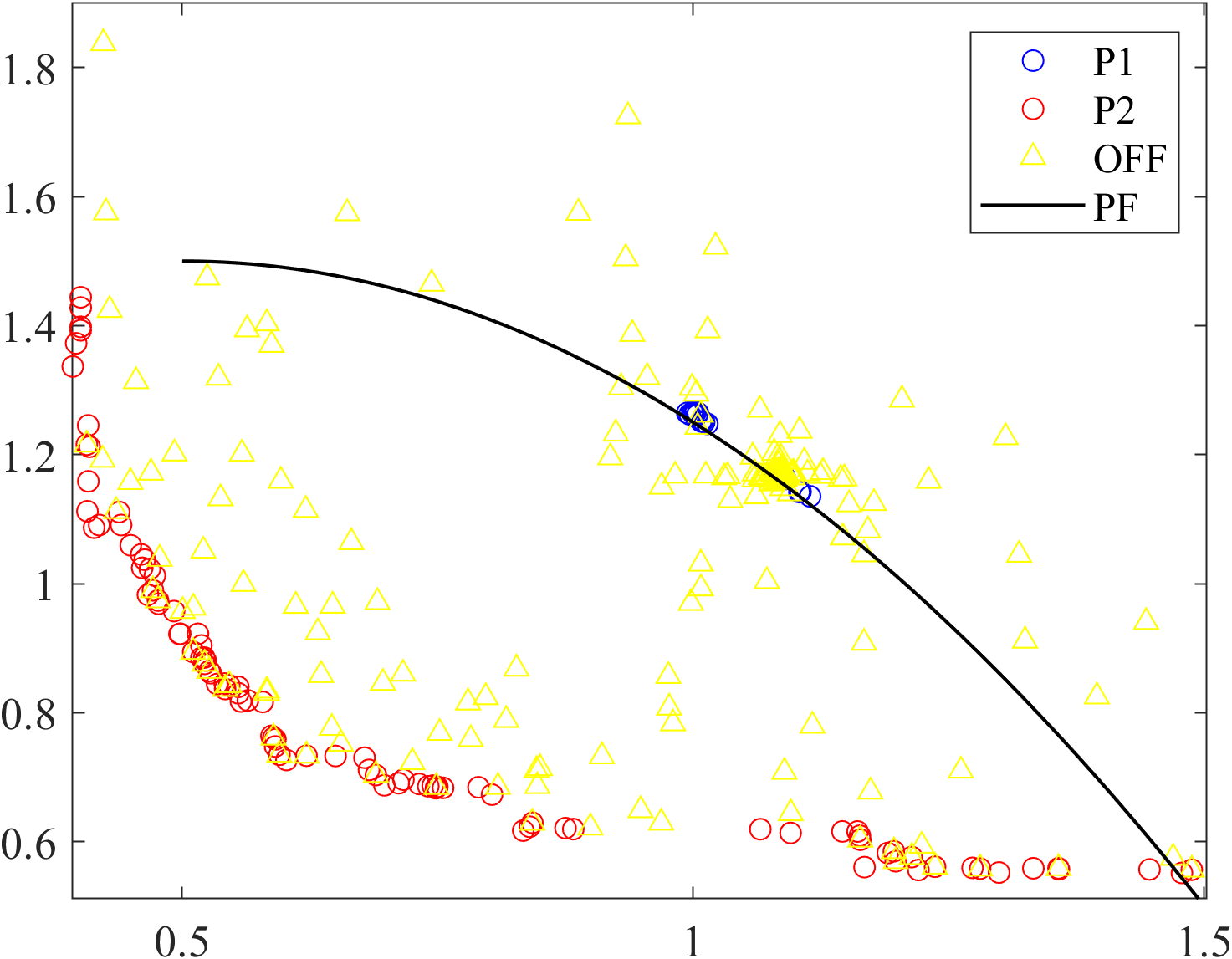}\label{fig:subL1}}
        \hfill
        \subfloat[25\%]{\includegraphics[width=0.2\textwidth]{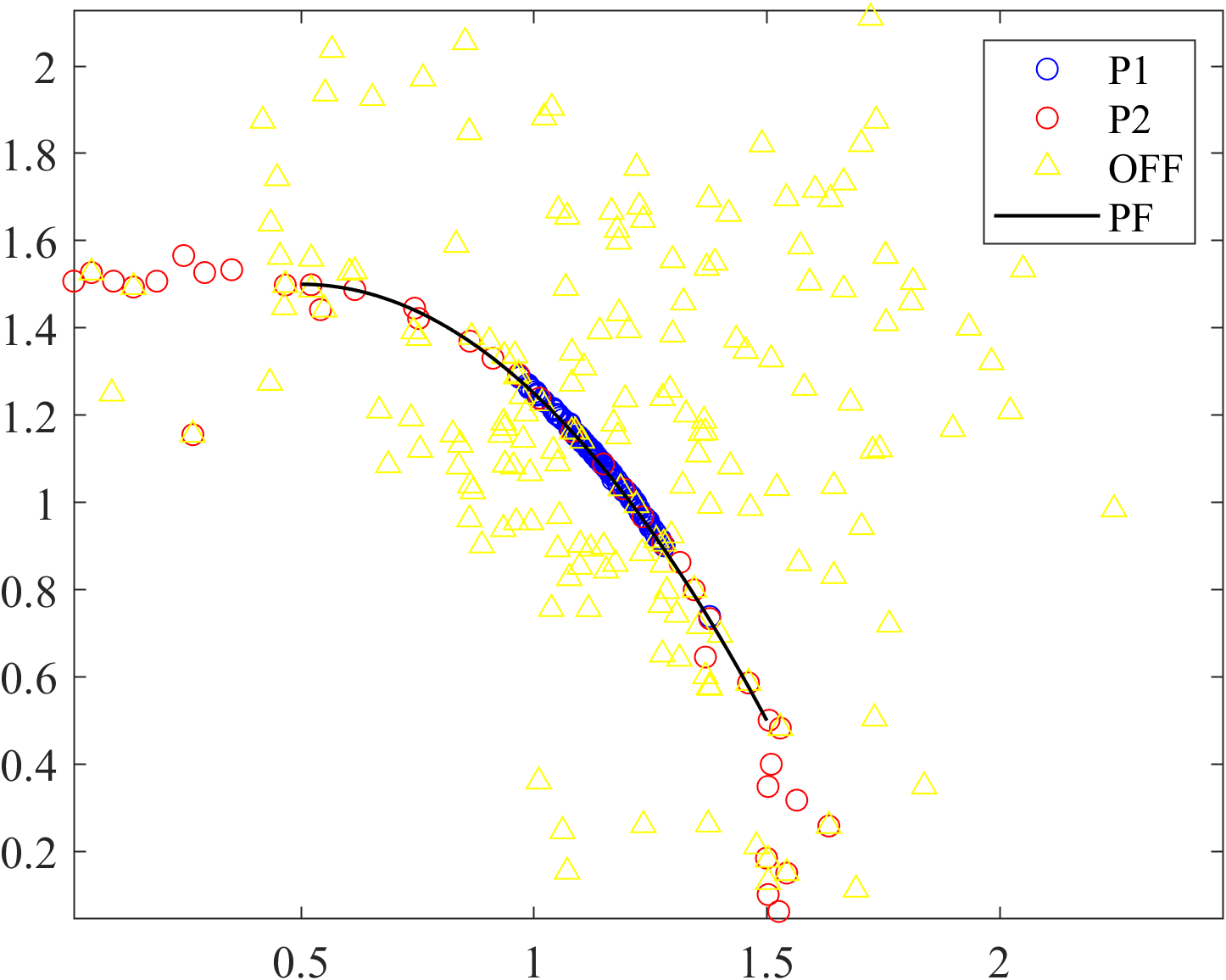}\label{fig:subL2}}
        \hfill
        \subfloat[50\%]{\includegraphics[width=0.2\textwidth]{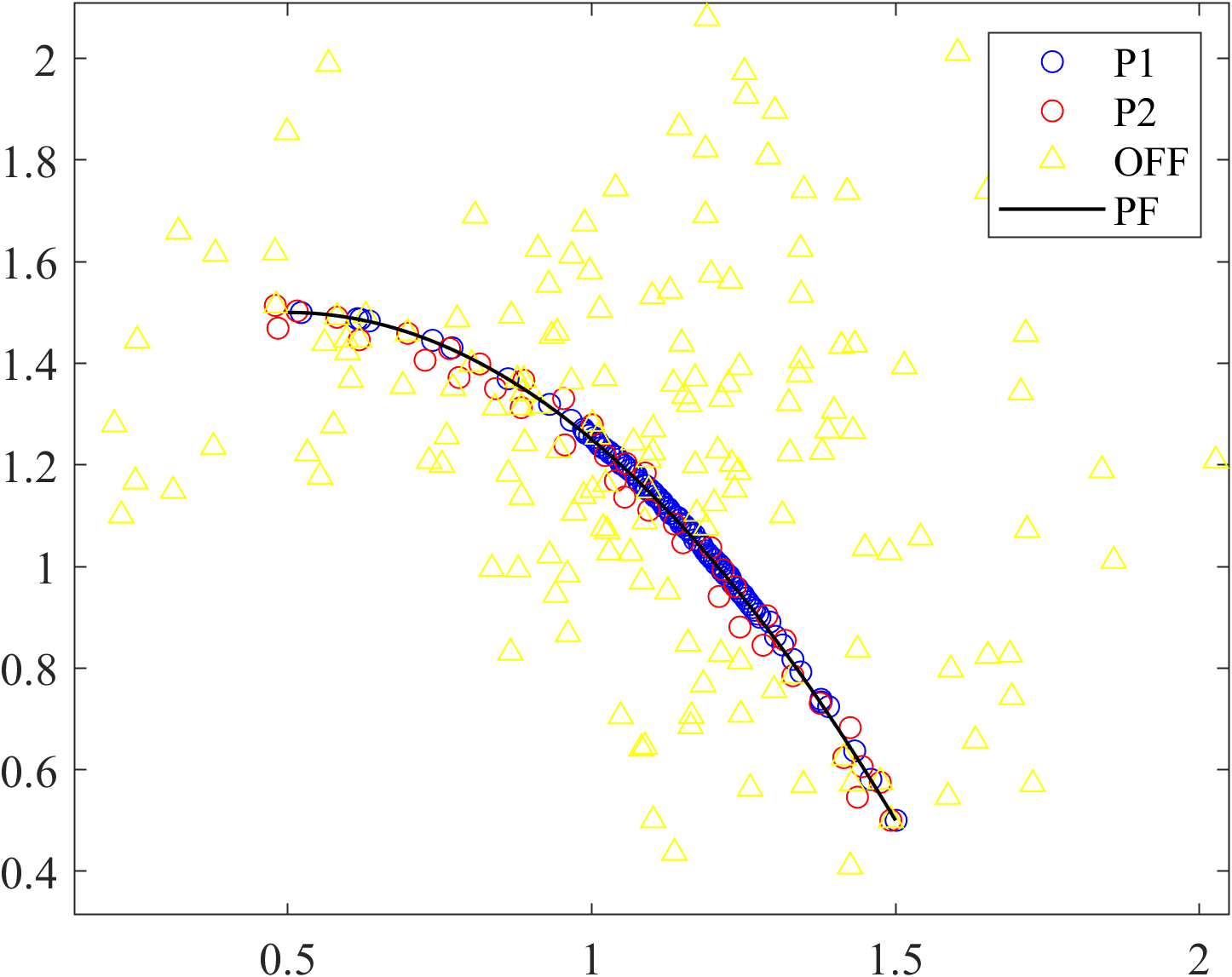}\label{fig:subL3}}
        \hfill
        \subfloat[75\%]{\includegraphics[width=0.2\textwidth]{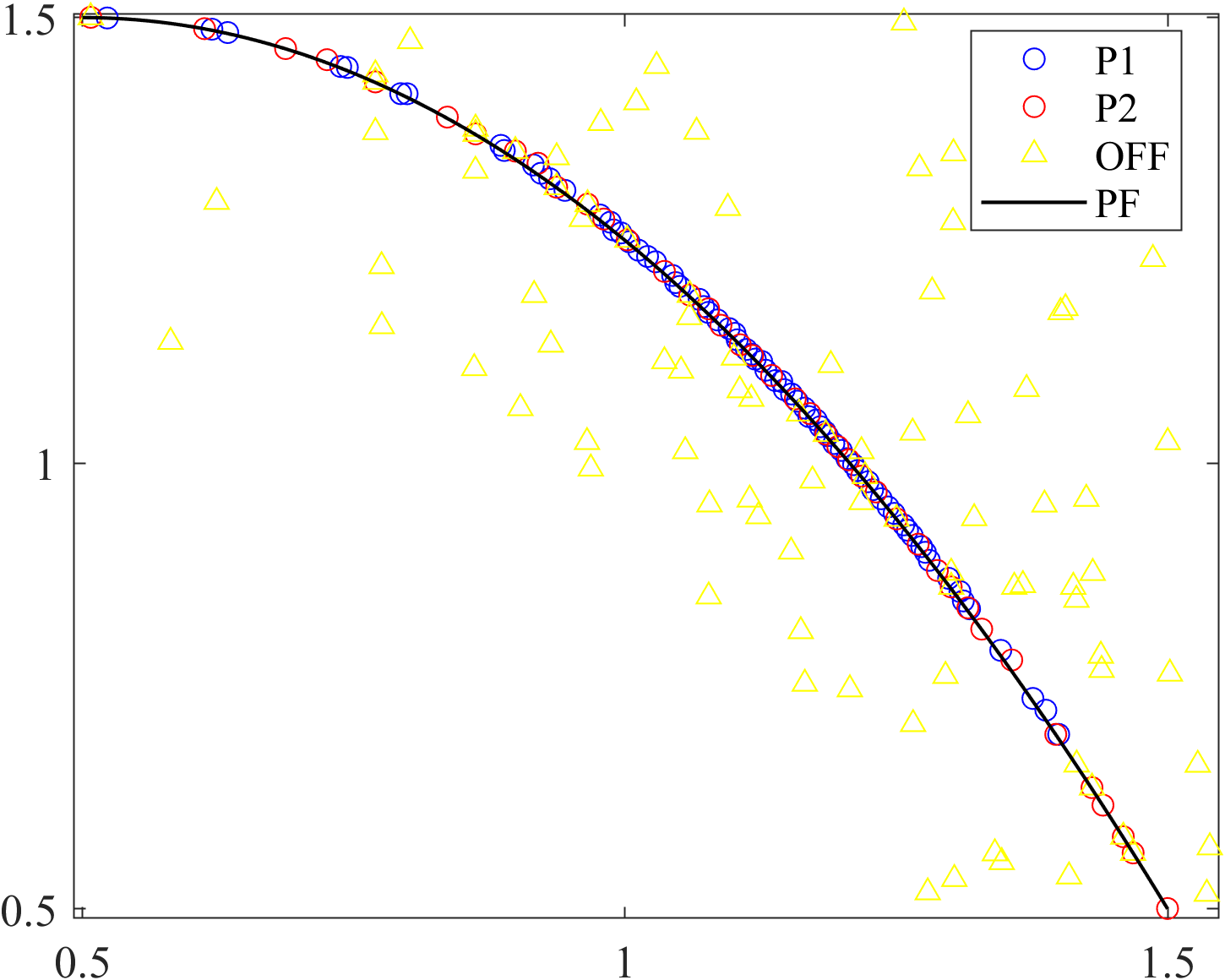}\label{fig:subL4}}
        
        \caption{The change process of pareto front, P1 (popMain), P2 (popAux), ARCH (achieve population) and OFF (offsping) of BiCO, URCMO and LLM4CMO on LIRCMOP1.}
        \label{fig:process}
    \end{figure*}
\subsection{Ablation Study}
For all other ablation experiment results, please refer to the supplementary materials (Sec.~\ref{sec:A4-6} and Sec.~\ref{sec:A4-7}).

\subsubsection{Results of the Base Framework}

We conducted ablation studies focusing on the opposite mechanism and the staged optimization phases. The results show that, for most problems, the inclusion of the opposite mechanism has no significant impact; however, a slight performance advantage was observed when it is enabled. While our adapted version of the opposite mechanism incorporates transformation techniques, its effectiveness remains generally consistent with the original implementation. Designed to address deceptive constraints—such as those found in the FCP test suite—the opposite mechanism plays a complementary role by mitigating substantial performance degradation caused by such constraints.

Importantly, when Stage 2 was altered by removing the epsilon-controlled three-phase structure, the algorithm's performance declined significantly on certain problems, particularly those in the LIRCMOP series. Although the vector-based and angular-partitioning environmental selection strategy was retained—remaining active for fully separable UPF-CPF problems like LIRCMOP1–3—the absence of the phased structure impaired performance. This outcome suggests that the epsilon decay function, discovered through LLM-based search, was inherently aligned with the original multi-phase design of Stage 2. Replacing it with a single-phase approach leads to inferior performance on LIRCMOP1–3 and similar problem types, highlighting the importance of the staged structure.

\begin{table}[htp!]
\centering
\caption{Wilcoxon rank sum test result and multi-problem wilcoxon signed rank test results with respect to HV and IGD on all 61 functions of the opposite module, epsilon control in the stage 2, hops included different UPF-CPF types from 1 to 4, and the DRA module.}
\label{tab:AB}
\resizebox{0.99\linewidth}{!}{%
\begin{tabular}{llllll}
\hline
LLM4CMO vs. & \multicolumn{1}{c}{HV(+/-/=)} & \multicolumn{1}{c}{$R^+$} & \multicolumn{1}{c}{$R^-$} & \multicolumn{1}{c}{$p$-value} & \multicolumn{1}{c}{level=0.05} \\ \hline
LLM4CMOWoRR & 30/5/26 & 1426 & 465 & 0.000168709 & YES \\
LLM4CMOWoS1C & 10/4/47 & 1082 & 809 & 0.218926479 & NO \\
LLM4CMOwoOP & 8/0/53 & 1195 & 696 & 0.039279306 & YES \\
LLM4CMOWo3P & 19/6/36 & 1327 & 564 & 0.002421513 & YES \\
LLM4CMOEps1 & 14/2/45 & 1208 & 683 & 0.031009586 & YES \\
LLM4CMOHOps-T1 & 26/15/20 & 1359 & 532 & 0.002977244 & YES \\
LLM4CMOHOps-T2 & 21/10/30 & 1294 & 597 & 0.01230769 & YES \\
LLM4CMOHOps-T3 & 30/3/28 & 1502 & 389 & 0.000015514 & YES \\
LLM4CMOHOps-T4 & 24/13/24 & 1503 & 388 & 6.21758E-05 & YES \\
LLM4CMOWoDRA & 6/0/55 & 1268 & 623 & 0.009359227 & YES \\ \hline
LLM4CMO vs. & \multicolumn{1}{c}{IGD(+/-/=)} & \multicolumn{1}{c}{$R^+$} & \multicolumn{1}{c}{$R^-$} & \multicolumn{1}{c}{$p$-value} & \multicolumn{1}{c}{level=0.05} \\ \hline
LLM4CMOWoRR & 27/6/28 & 1390 & 501 & 0.001409279 & YES \\
LLM4CMOWoS1C & 7/2/52 & 1163 & 728 & 0.118228233 & NO \\
LLM4CMOwoOP & 7/0/54 & 1224 & 667 & 0.045456762 & YES \\
LLM4CMOWo3P & 16/5/40 & 1255 & 636 & 0.026211143 & YES \\
LLM4CMOEps1 & 13/2/46 & 1343 & 548 & 0.004301606 & YES \\
LLM4CMOHOps-T1 & 26/16/19 & 1317 & 574 & 0.00762132 & YES \\
LLM4CMOHOps-T2 & 19/14/28 & 1188 & 703 & 0.081539403 & NO \\
LLM4CMOHOps-T3 & 27/4/30 & 1410 & 481 & 0.000848712 & YES \\
LLM4CMOHOps-T4 & 21/13/27 & 1342 & 549 & 0.0043999 & YES \\
LLM4CMOWoDRA & 5/1/55 & 1239 & 652 & 0.035018479 & YES \\ \hline
\end{tabular}%
}
\end{table}

In the design of the base framework, we found that employing a relaxed judgment condition for $R_s$ facilitates the algorithm’s transition to Stage 2 more efficiently. Compared to the previous hard threshold condition, the relaxed approach resulted in superior performance in nearly half of the test cases.

\subsubsection{Results of Core Modules}

We also performed before-and-after comparisons on the three core modules specifically designed through the LLM. The results indicate that while modifications to the DRA module had a relatively minor effect on overall performance, they still contributed positively. In contrast, replacing the original epsilon decay function led to performance improvements on 10 test functions, confirming its effectiveness. Among all modules, the HOps module had the most significant impact on performance.
To further evaluate HOps, we conducted tests on the operator combinations identified through LLM interaction. The results reveal that the overall performance of the four individual operator types was significantly lower than that of the complete HOps configuration. However, each template exhibited specific strengths when applied to certain problem types. This suggests that the HOps configurations derived via LLM are not entirely accurate for all type-specific applications. Nevertheless, the LLM4CMO we obtained and optimized through interaction with LLM has improved the core modules on the original framework, significantly enhancing the algorithm's performance, which preliminarily verifies the effectiveness of LLM interaction design.

\section{Conclusion} \label{sec5}

In this study, we conducted a preliminary exploration of the potential role of LLMs in the design of CMOEAs. By constructing an algorithm based on a dual-population, two-stage architecture and designing its key modules, we aimed to address the challenge that LLMs—when operating with limited domain knowledge—may struggle to adapt holistic algorithm frameworks effectively. To overcome this, we developed simple prompt templates, incorporated interactive feedback, and iteratively refined module designs through LLM-human interaction. This process enabled the rapid development of several core algorithm components.

The resulting algorithm, LLM4CMO, was benchmarked against eleven state-of-the-art algorithms across six representative test suites. Experimental results demonstrated that LLM4CMO consistently outperformed baseline methods on multiple performance metrics, highlighting the effectiveness of LLM-aided module design. These findings underscore the potential of LLMs to support and accelerate the development of optimization algorithms, particularly when combined with human expertise.

Our findings suggest that the integration of LLMs into algorithm design may represent a promising new paradigm—one that redefines traditional development workflows by leveraging the generative and analytical capabilities of LLMs.

\section{Limiations} \label{sec6}
Our work is based on the design of auxiliary algorithms for LLM interaction. Our algorithm is improved from some existing algorithms. Through the LLM-human interaction, we present a case study, in which the interaction method is used to design and optimize the 3 core modules of the CMOEA algorithm. Our experiments have demonstrated that the algorithm indeed has good performance. However, it must be acknowledged that the basic algorithm architecture is indeed complex, and the LLM-Human interaction paradigm has certain subjectivity and relies on the observations of designers, thus there is still a gap compared with the paradigm of LLM automatic complex algorithms. 

\bibliographystyle{IEEEtran}
\bibliography{./ref.bib}

\begin{thebibliography}{10}
\providecommand{\url}[1]{#1}
\csname url@samestyle\endcsname
\providecommand{\newblock}{\relax}
\providecommand{\bibinfo}[2]{#2}
\providecommand{\BIBentrySTDinterwordspacing}{\spaceskip=0pt\relax}
\providecommand{\BIBentryALTinterwordstretchfactor}{4}
\providecommand{\BIBentryALTinterwordspacing}{\spaceskip=\fontdimen2\font plus
\BIBentryALTinterwordstretchfactor\fontdimen3\font minus \fontdimen4\font\relax}
\providecommand{\BIBforeignlanguage}[2]{{%
\expandafter\ifx\csname l@#1\endcsname\relax
\typeout{** WARNING: IEEEtran.bst: No hyphenation pattern has been}%
\typeout{** loaded for the language `#1'. Using the pattern for}%
\typeout{** the default language instead.}%
\else
\language=\csname l@#1\endcsname
\fi
#2}}
\providecommand{\BIBdecl}{\relax}
\BIBdecl

\bibitem{zuo2014vehicle}
X.~Zuo, C.~Chen, W.~Tan, and M.~Zhou, ``Vehicle scheduling of an urban bus line via an improved multiobjective genetic algorithm,'' \emph{IEEE Transactions on Intelligent Transportation Systems}, vol.~16, no.~2, pp. 1030--1041, 2014.

\bibitem{mishra2019butterfly}
S.~Mishra, A.~Kumar, D.~Singh, and R.~Kumar~Misra, ``Butterfly optimizer for placement and sizing of distributed generation for feeder phase balancing,'' in \emph{Computational Intelligence: Theories, Applications and Future Directions-Volume II: ICCI-2017}.\hskip 1em plus 0.5em minus 0.4em\relax Springer, 2019, pp. 519--530.

\bibitem{saravanan2009evolutionary}
R.~Saravanan, S.~Ramabalan, N.~G.~R. Ebenezer, and C.~Dharmaraja, ``Evolutionary multi criteria design optimization of robot grippers,'' \emph{Applied Soft Computing}, vol.~9, no.~1, pp. 159--172, 2009.

\bibitem{kannan1994augmented}
B.~Kannan and S.~N. Kramer, ``An augmented lagrange multiplier based method for mixed integer discrete continuous optimization and its applications to mechanical design,'' \emph{Journal of Mechanical Design}, vol. 116, pp. 405--411, 1994.

\bibitem{tian2020coevolutionary}
Y.~Tian, T.~Zhang, J.~Xiao, X.~Zhang, and Y.~Jin, ``A coevolutionary framework for constrained multiobjective optimization problems,'' \emph{IEEE Transactions on Evolutionary Computation}, vol.~25, no.~1, pp. 102--116, 2020.

\bibitem{deb2002fast}
K.~Deb, A.~Pratap, S.~Agarwal, and T.~Meyarivan, ``A fast and elitist multiobjective genetic algorithm: {NSGA-II},'' \emph{IEEE Transactions on Evolutionary Computation}, vol.~6, no.~2, pp. 182--197, 2002.

\bibitem{zhang2007moea}
Q.~Zhang and H.~Li, ``{MOEA/D}: A multiobjective evolutionary algorithm based on decomposition,'' \emph{IEEE Transactions on Evolutionary Computation}, vol.~11, no.~6, pp. 712--731, 2007.

\bibitem{takahama2005constrained}
T.~Takahama and S.~Sakai, ``Constrained optimization by $\varepsilon$ constrained particle swarm optimizer with $\varepsilon$-level control,'' in \emph{Soft Computing as Transdisciplinary Science and Technology: Proceedings of the fourth IEEE International Workshop WSTST’05}.\hskip 1em plus 0.5em minus 0.4em\relax Springer, 2005, pp. 1019--1029.

\bibitem{xu2025handling}
B.~Xu, Y.~Zheng, W.~Li, X.~Gao, D.~Gong, J.~He, and Z.~Fan, ``Handling multiobjective optimization problems with complex constraints: A constraints grouping-based approach,'' \emph{IEEE Transactions on Systems, Man, and Cybernetics: Systems}, 2025.

\bibitem{ming2023constraint}
F.~Ming, W.~Gong, L.~Wang, and L.~Gao, ``A constraint-handling technique for decomposition-based constrained many-objective evolutionary algorithms,'' \emph{IEEE Transactions on Systems, Man, and Cybernetics: Systems}, vol.~53, no.~12, pp. 7783--7793, 2023.

\bibitem{li2024competitive}
Y.~Li, W.~Gong, Z.~Hu, and S.~Li, ``A competitive and cooperative evolutionary framework for ensemble of constraint handling techniques,'' \emph{IEEE Transactions on Systems, Man, and Cybernetics: Systems}, vol.~54, no.~4, pp. 2440--2451, 2024.

\bibitem{zou2021dual}
J.~Zou, R.~Sun, S.~Yang, and J.~Zheng, ``A dual-population algorithm based on alternative evolution and degeneration for solving constrained multi-objective optimization problems,'' \emph{Information Sciences}, vol. 579, pp. 89--102, 2021.

\bibitem{ming2021dual}
M.~Ming, A.~Trivedi, R.~Wang, D.~Srinivasan, and T.~Zhang, ``A dual-population-based evolutionary algorithm for constrained multiobjective optimization,'' \emph{IEEE Transactions on Evolutionary Computation}, vol.~25, no.~4, pp. 739--753, 2021.

\bibitem{liang2022utilizing}
J.~Liang, K.~Qiao, K.~Yu, B.~Qu, C.~Yue, W.~Guo, and L.~Wang, ``Utilizing the relationship between unconstrained and constrained pareto fronts for constrained multiobjective optimization,'' \emph{IEEE Transactions on Cybernetics}, vol.~53, no.~6, pp. 3873--3886, 2022.

\bibitem{liu2024coevolutionary}
S.~Liu, J.~Feng, S.~Yang, J.~Zheng, and Q.~Xiao, ``A coevolutionary algorithm with detection and supervision strategies for constrained multiobjective optimization,'' \emph{IEEE Transactions on Evolutionary Computation}, 2024.

\bibitem{liang2022survey}
J.~Liang, X.~Ban, K.~Yu, B.~Qu, K.~Qiao, C.~Yue, K.~Chen, and K.~C. Tan, ``A survey on evolutionary constrained multiobjective optimization,'' \emph{IEEE Transactions on Evolutionary Computation}, vol.~27, no.~2, pp. 201--221, 2022.

\bibitem{zhao2023survey}
W.~X. Zhao, K.~Zhou, J.~Li, T.~Tang, X.~Wang, Y.~Hou, Y.~Min, B.~Zhang, J.~Zhang, Z.~Dong \emph{et~al.}, ``A survey of large language models,'' \emph{arXiv preprint arXiv:2303.18223}, vol.~1, no.~2, 2023.

\bibitem{liu2024systematic}
F.~Liu, Y.~Yao, P.~Guo, Z.~Yang, Z.~Zhao, X.~Lin, X.~Tong, M.~Yuan, Z.~Lu, Z.~Wang \emph{et~al.}, ``A systematic survey on large language models for algorithm design,'' \emph{arXiv preprint arXiv:2410.14716}, 2024.

\bibitem{luo2025llm4sr}
Z.~Luo, Z.~Yang, Z.~Xu, W.~Yang, and X.~Du, ``{LLM4SR}: A survey on large language models for scientific research,'' \emph{arXiv preprint arXiv:2501.04306}, 2025.

\bibitem{wei2022chain}
J.~Wei, X.~Wang, D.~Schuurmans, M.~Bosma, F.~Xia, E.~Chi, Q.~V. Le, D.~Zhou \emph{et~al.}, ``Chain-of-thought prompting elicits reasoning in large language models,'' \emph{Advances in Neural Information Processing Systems}, vol.~35, pp. 24\,824--24\,837, 2022.

\bibitem{romera2024mathematical}
B.~Romera-Paredes, M.~Barekatain, A.~Novikov, M.~Balog, M.~P. Kumar, E.~Dupont, F.~J. Ruiz, J.~S. Ellenberg, P.~Wang, O.~Fawzi \emph{et~al.}, ``Mathematical discoveries from program search with large language models,'' \emph{Nature}, vol. 625, no. 7995, pp. 468--475, 2024.

\bibitem{liu2024evolution}
F.~Liu, X.~Tong, M.~Yuan, X.~Lin, F.~Luo, Z.~Wang, Z.~Lu, and Q.~Zhang, ``Evolution of heuristics: towards efficient automatic algorithm design using large language model,'' in \emph{Proceedings of the 41st International Conference on Machine Learning}, ser. ICML'24.\hskip 1em plus 0.5em minus 0.4em\relax JMLR.org, 2024.

\bibitem{liu2025large}
F.~Liu, X.~Lin, S.~Yao, Z.~Wang, X.~Tong, M.~Yuan, and Q.~Zhang, ``Large language model for multiobjective evolutionary optimization,'' in \emph{International Conference on Evolutionary Multi-Criterion Optimization}.\hskip 1em plus 0.5em minus 0.4em\relax Springer, 2025, pp. 178--191.

\bibitem{wang2024large}
Z.~Wang, S.~Liu, J.~Chen, and K.~C. Tan, ``Large language model-aided evolutionary search for constrained multiobjective optimization,'' in \emph{International Conference on Intelligent Computing}.\hskip 1em plus 0.5em minus 0.4em\relax Springer, 2024, pp. 218--230.

\bibitem{jain2013evolutionary}
H.~Jain and K.~Deb, ``An evolutionary many-objective optimization algorithm using reference-point based nondominated sorting approach, part {II}: Handling constraints and extending to an adaptive approach,'' \emph{IEEE Transactions on Evolutionary Computation}, vol.~18, no.~4, pp. 602--622, 2013.

\bibitem{fan2019push}
Z.~Fan, W.~Li, X.~Cai, H.~Li, C.~Wei, Q.~Zhang, K.~Deb, and E.~Goodman, ``Push and pull search for solving constrained multi-objective optimization problems,'' \emph{Swarm and Evolutionary Computation}, vol.~44, pp. 665--679, 2019.

\bibitem{fan2019improved}
Z.~Fan, W.~Li, X.~Cai, H.~Huang, Y.~Fang, Y.~You, J.~Mo, C.~Wei, and E.~Goodman, ``An improved epsilon constraint-handling method in {MOEA/D} for {CMOP}s with large infeasible regions,'' \emph{Soft Computing}, vol.~23, pp. 12\,491--12\,510, 2019.

\bibitem{liu2025constraint}
Z.~Liu, F.~Han, Q.~Ling, H.~Han, and J.~Jiang, ``Constraint-pareto dominance and diversity enhancement strategy based evolutionary algorithm for solving constrained multiobjective optimization problems,'' \emph{IEEE Transactions on Evolutionary Computation}, 2025.

\bibitem{fan2016angle}
Z.~Fan, W.~Li, X.~Cai, K.~Hu, H.~Lin, and H.~Li, ``Angle-based constrained dominance principle in {MOEA/D} for constrained multi-objective optimization problems,'' in \emph{2016 IEEE Congress on Evolutionary Computation (CEC)}.\hskip 1em plus 0.5em minus 0.4em\relax IEEE, 2016, pp. 460--467.

\bibitem{fan2019moea}
Z.~Fan, Y.~Fang, W.~Li, X.~Cai, C.~Wei, and E.~Goodman, ``{MOEA/D} with angle-based constrained dominance principle for constrained multi-objective optimization problems,'' \emph{Applied Soft Computing}, vol.~74, pp. 621--633, 2019.

\bibitem{qiao2024cooperative}
K.~Qiao, K.~Yu, C.~Yue, B.~Qu, M.~Liu, and J.~Liang, ``A cooperative multistep mutation strategy for multiobjective optimization problems with deceptive constraints,'' \emph{IEEE Transactions on Systems, Man, and Cybernetics: Systems}, vol.~54, no.~11, pp. 6670--6682, 2024.

\bibitem{ming2021novel}
M.~Ming, R.~Wang, H.~Ishibuchi, and T.~Zhang, ``A novel dual-stage dual-population evolutionary algorithm for constrained multiobjective optimization,'' \emph{IEEE Transactions on Evolutionary Computation}, vol.~26, no.~5, pp. 1129--1143, 2021.

\bibitem{liu2021handling}
Z.-Z. Liu, B.-C. Wang, and K.~Tang, ``Handling constrained multiobjective optimization problems via bidirectional coevolution,'' \emph{IEEE Transactions on Cybernetics}, vol.~52, no.~10, pp. 10\,163--10\,176, 2021.

\bibitem{ma2021multi}
H.~Ma, H.~Wei, Y.~Tian, R.~Cheng, and X.~Zhang, ``A multi-stage evolutionary algorithm for multi-objective optimization with complex constraints,'' \emph{Information Sciences}, vol. 560, pp. 68--91, 2021.

\bibitem{tian2021balancing}
Y.~Tian, Y.~Zhang, Y.~Su, X.~Zhang, K.~C. Tan, and Y.~Jin, ``Balancing objective optimization and constraint satisfaction in constrained evolutionary multiobjective optimization,'' \emph{IEEE Transactions on Cybernetics}, vol.~52, no.~9, pp. 9559--9572, 2021.

\bibitem{qiao2023evolutionary}
K.~Qiao, J.~Liang, Z.~Liu, K.~Yu, C.~Yue, and B.~Qu, ``Evolutionary multitasking with global and local auxiliary tasks for constrained multi-objective optimization,'' \emph{IEEE/CAA Journal of Automatica Sinica}, vol.~10, no.~10, pp. 1951--1964, 2023.

\bibitem{ming2022constrained}
F.~Ming, W.~Gong, L.~Wang, and L.~Gao, ``Constrained multiobjective optimization via multitasking and knowledge transfer,'' \emph{IEEE Transactions on Evolutionary Computation}, vol.~28, no.~1, pp. 77--89, 2022.

\bibitem{sun2022multistage}
R.~Sun, J.~Zou, Y.~Liu, S.~Yang, and J.~Zheng, ``A multistage algorithm for solving multiobjective optimization problems with multiconstraints,'' \emph{IEEE Transactions on Evolutionary Computation}, vol.~27, no.~5, pp. 1207--1219, 2022.

\bibitem{ming2024even}
F.~Ming, W.~Gong, and Y.~Jin, ``Even search in a promising region for constrained multi-objective optimization,'' \emph{IEEE/CAA Journal of Automatica Sinica}, vol.~11, no.~2, pp. 474--486, 2024.

\bibitem{ma2019evolutionary}
Z.~Ma and Y.~Wang, ``Evolutionary constrained multiobjective optimization: Test suite construction and performance comparisons,'' \emph{IEEE Transactions on Evolutionary Computation}, vol.~23, no.~6, pp. 972--986, 2019.

\bibitem{holland1992genetic}
J.~H. Holland, ``Genetic algorithms,'' \emph{Scientific american}, vol. 267, no.~1, pp. 66--73, 1992.

\bibitem{storn1997differential}
R.~Storn and K.~Price, ``Differential evolution--a simple and efficient heuristic for global optimization over continuous spaces,'' \emph{Journal of Global Optimization}, vol.~11, pp. 341--359, 1997.

\bibitem{zhang2009jade}
J.~Zhang and A.~C. Sanderson, ``{JADE}: adaptive differential evolution with optional external archive,'' \emph{IEEE Transactions on Evolutionary Computation}, vol.~13, no.~5, pp. 945--958, 2009.

\bibitem{bertsimas2024robust}
D.~Bertsimas and G.~Margaritis, ``Robust and adaptive optimization under a large language model lens,'' \emph{arXiv preprint arXiv:2501.00568}, 2024.

\bibitem{yang2023large}
C.~Yang, X.~Wang, Y.~Lu, H.~Liu, Q.~V. Le, D.~Zhou, and X.~Chen, ``Large language models as optimizers,'' \emph{arXiv preprint arXiv:2309.03409}, 2023.

\bibitem{garey1981approximation}
M.~R. Garey and D.~S. Johnson, ``Approximation algorithms for bin packing problems: A survey,'' in \emph{Analysis and design of algorithms in combinatorial optimization}.\hskip 1em plus 0.5em minus 0.4em\relax Springer, 1981, pp. 147--172.

\bibitem{reinelt1992fast}
G.~Reinelt, ``Fast heuristics for large geometric traveling salesman problems,'' \emph{ORSA Journal on Computing}, vol.~4, no.~2, pp. 206--217, 1992.

\bibitem{van2024llamea}
N.~van Stein and T.~B{\"a}ck, ``Llamea: A large language model evolutionary algorithm for automatically generating metaheuristics,'' \emph{IEEE Transactions on Evolutionary Computation}, 2024.

\bibitem{wolpert1997no}
D.~H. Wolpert and W.~G. Macready, ``No free lunch theorems for optimization,'' \emph{IEEE Transactions on Evolutionary Computation}, vol.~1, no.~1, pp. 67--82, 1997.

\bibitem{guo2025deepseek}
D.~Guo, D.~Yang, H.~Zhang, J.~Song, R.~Zhang, R.~Xu, Q.~Zhu, S.~Ma, P.~Wang, X.~Bi \emph{et~al.}, ``Deepseek-r1: Incentivizing reasoning capability in {LLM}s via reinforcement learning,'' \emph{arXiv preprint arXiv:2501.12948}, 2025.

\bibitem{zitzler2002multiobjective}
E.~Zitzler and L.~Thiele, ``Multiobjective evolutionary algorithms: a comparative case study and the strength pareto approach,'' \emph{IEEE Transactions on Evolutionary Computation}, vol.~3, no.~4, pp. 257--271, 2002.

\bibitem{bosman2003balance}
P.~A. Bosman and D.~Thierens, ``The balance between proximity and diversity in multiobjective evolutionary algorithms,'' \emph{IEEE Transactions on Evolutionary Computation}, vol.~7, no.~2, pp. 174--188, 2003.

\bibitem{mashwani2016multiobjective}
W.~K. Mashwani and A.~Salhi, ``Multiobjective evolutionary algorithm based on multimethod with dynamic resources allocation,'' \emph{Applied Soft Computing}, vol.~39, pp. 292--309, 2016.

\bibitem{fan2020difficulty}
Z.~Fan, W.~Li, X.~Cai, H.~Li, C.~Wei, Q.~Zhang, K.~Deb, and E.~Goodman, ``Difficulty adjustable and scalable constrained multiobjective test problem toolkit,'' \emph{Evolutionary Computation}, vol.~28, no.~3, pp. 339--378, 2020.

\bibitem{liu2019handling}
Z.-Z. Liu and Y.~Wang, ``Handling constrained multiobjective optimization problems with constraints in both the decision and objective spaces,'' \emph{IEEE Transactions on Evolutionary Computation}, vol.~23, no.~5, pp. 870--884, 2019.

\bibitem{yuan2021indicator}
J.~Yuan, H.-L. Liu, Y.-S. Ong, and Z.~He, ``Indicator-based evolutionary algorithm for solving constrained multiobjective optimization problems,'' \emph{IEEE Transactions on Evolutionary Computation}, vol.~26, no.~2, pp. 379--391, 2021.

\bibitem{kumar2021benchmark}
A.~Kumar, G.~Wu, M.~Z. Ali, Q.~Luo, R.~Mallipeddi, P.~N. Suganthan, and S.~Das, ``A benchmark-suite of real-world constrained multi-objective optimization problems and some baseline results,'' \emph{Swarm and Evolutionary Computation}, vol.~67, p. 100961, 2021.

\bibitem{tian2017platemo}
Y.~Tian, R.~Cheng, X.~Zhang, and Y.~Jin, ``Plat{EMO}: A {MATLAB} platform for evolutionary multi-objective optimization [educational forum],'' \emph{IEEE Computational Intelligence Magazine}, vol.~12, no.~4, pp. 73--87, 2017.

\bibitem{narayanan1999improving}
S.~Narayanan and S.~Azarm, ``On improving multiobjective genetic algorithms for design optimization,'' \emph{Structural Optimization}, vol.~18, pp. 146--155, 1999.

\bibitem{chiandussi2012comparison}
G.~Chiandussi, M.~Codegone, S.~Ferrero, and F.~E. Varesio, ``Comparison of multi-objective optimization methodologies for engineering applications,'' \emph{Computers \& Mathematics with Applications}, vol.~63, no.~5, pp. 912--942, 2012.

\bibitem{rathore2010synchronous}
A.~K. Rathore, J.~Holtz, and T.~Boller, ``Synchronous optimal pulsewidth modulation for low-switching-frequency control of medium-voltage multilevel inverters,'' \emph{IEEE Transactions on Industrial Electronics}, vol.~57, no.~7, pp. 2374--2381, 2010.

\bibitem{rathore2012generalized}
{Rathore, Akshay K and Holtz, Joachim and Boller, Till}, ``Generalized optimal pulsewidth modulation of multilevel inverters for low-switching-frequency control of medium-voltage high-power industrial {AC} drives,'' \emph{IEEE Transactions on Industrial Electronics}, vol.~60, no.~10, pp. 4215--4224, 2012.

\bibitem{edpuganti2015optimal}
A.~Edpuganti, A.~Dwivedi, A.~K. Rathore, and R.~K. Srivastava, ``Optimal pulsewidth modulation of cascade nine-level (9{L}) inverter for medium voltage high power industrial {AC} drives,'' in \emph{IECON 2015-41st Annual Conference of the IEEE Industrial Electronics Society}.\hskip 1em plus 0.5em minus 0.4em\relax IEEE, 2015, pp. 004\,259--004\,264.

\bibitem{edpuganti2015fundamental}
A.~Edpuganti and A.~K. Rathore, ``Fundamental switching frequency optimal pulsewidth modulation of medium-voltage cascaded seven-level inverter,'' \emph{IEEE Transactions on Industry Applications}, vol.~51, no.~4, pp. 3485--3492, 2015.

\bibitem{edpuganti2017optimal}
{Edpuganti, Amarendra and Rathore, Akshay Kumar}, ``Optimal pulsewidth modulation for common-mode voltage elimination scheme of medium-voltage modular multilevel converter-fed open-end stator winding induction motor drives,'' \emph{IEEE Transactions on Industrial Electronics}, vol.~64, no.~1, pp. 848--856, 2017.

\end{thebibliography}

\newpage


\title{Supplementary materials}
\maketitle
\tableofcontents  

\twocolumn
\section{Limiations} \label{sec6}
Our work is based on the design of auxiliary algorithms for LLM interaction. Our algorithm is improved from some existing algorithms. Through the LLM-human interaction, we present a case study, in which the interaction method is used to design and optimize the 3 core modules of the CMOEA algorithm. Our experiments have demonstrated that the algorithm indeed has good performance. However, it must be acknowledged that the basic algorithm architecture is indeed complex, and the LLM-Human interaction paradigm has certain subjectivity and relies on the observations of designers, thus there is still a gap compared with the paradigm of LLM automatic complex algorithms. 

\section{The description of CMOPs feature in 6 test suite and real-world CMOPs.} \label{sec:A1}
\begin{table}[hptb!]
    \centering
    \caption{The number of objectives (M) and decision variables (D) for each problem}
    \label{tab:Problems1}
    \resizebox{0.7\linewidth}{!}{%
    \begin{tabular}{ccc}
    \hline
    \textbf{Problem} & \textbf{M} & \textbf{D} \\ \hline
    DASCMOP1-6 & 2 & 30 \\
    DASCMOP7-9 & 3 & 30 \\ \hline
    LIRCMOP1-12 & 2 & 30 \\
    LIRCMOP13-14 & 3 & 30 \\ \hline
    MW1-3,MW5,MW6-7,MW11-13 & 2 & 15 \\
    MW4,MW8,MW14 & 3 & 15 \\ \hline
    DOC1 & 2 & 6 \\
    DOC2 & 2 & 16 \\
    DOC3 & 2 & 10 \\
    DOC4-5 & 2 & 8 \\
    DOC6-7 & 2 & 11 \\
    DOC8 & 3 & 10 \\
    DOC9 & 3 & 11 \\ \hline
    CF1-7 & 2 & 10 \\
    CF8-10 & 3 & 10 \\ \hline
    FCP1-5 & 2 & 30 \\ \hline
    \end{tabular}%
    }
\end{table}

\begin{table}[hptb!]
\centering
\caption{Description of eleven real-world CMOPs. M is the number of objectives, D is the number of decision variables, \textbf{ng} is the number of inequality constraints, and \textbf{nh} is the number of equality constraints}
\label{tab:RWCMOPs}
\resizebox{0.9\linewidth}{!}{%
\begin{tabular}{ccccc}
\hline
\textbf{Problem} & \textbf{M} & \textbf{D} & \textbf{ng} & \textbf{nh} \\ \hline
\textbf{Pressure Vessel Design} & 2 & 4 & 2 & 2 \\
\textbf{Vibrating Platform Design} & 2 & 5 & 5 & 0 \\
\textbf{Two Bar Truss Design} & 2 & 3 & 3 & 0 \\
\textbf{Gear Box Design} & 3 & 7 & 11 & 0 \\
\textbf{Synchronous Optimal Pulse-width Modulation of 3-level Inverters} & 2 & 25 & 24 & 0 \\
\textbf{Synchronous Optimal Pulse-width Modulation of 5-level Inverters} & 2 & 25 & 24 & 0 \\
\textbf{Synchronous Optimal Pulse-width Modulation of 7-level Inverters} & 2 & 25 & 24 & 0 \\
\textbf{Synchronous Optimal Pulse-width Modulation of 9-level Inverters} & 2 & 30 & 24 & 0 \\
\textbf{Synchronous Optimal Pulse-width Modulation of 11-level Inverters} & 2 & 30 & 29 & 0 \\
\textbf{Synchronous Optimal Pulse-width Modulation of 13-level Inverters} & 2 & 30 & 29 & 0 \\ \hline
\end{tabular}%
}
\end{table}

\section{The description of Operators} \label{sec:A2}

We summarize below the operator combinations and parameter settings used in our algorithm. For the SBX and PM operators, the crossover probability and distribution index are set to 1 and 20, respectively. The mutation probability is set to \(1/D\), where \(D\) is the dimensionality of the problem, and the mutation distribution index is set to 20. In Stage~2, the mutation probability remains at \(1/D\), but the distribution index is reduced to 1 to enhance local search capability.

For differential evolution operators, the scaling factor \(F\) and crossover rate \(CR\) for each individual are randomly selected from the sets \(F_{\text{set}} = \{0.6, 0.8, 1.0\}\) and \(CR_{\text{set}} = \{0.1, 0.2, 1.0\}\), respectively.

During the transition from Stage~1 to Stage~2, we first determine the UPF–CPF relationship type using a method inspired by URCMO~\cite{liang2022utilizing}. To simplify the classification, we merge Type~2 and Type~3 into a single category, as both represent overlapping UPF-CPF relationships. In our dynamic type discrimination process, types that change frequently across generations are considered unstable and thus treated as \textit{unclassifiable}; these cases are managed using a separate fallback strategy.

For DE/current-to-pbest/1:
\begin{equation}
    \label{eq:DE/pbest}
    \begin{cases}
        \mathbf{v}_{P_2,r_1} + F \cdot (\mathbf{x}_{pbest} -\mathbf{x}_{P_2,r_1}) + F \cdot (\mathbf{x}_{P_2,r_2} - \mathbf{x}_{P_2,r_3})\\
        \mathbf{u}_{P_2} = \mathbf{v}_{P_2}\\
        F = [0.6,0.8,1.0]
    \end{cases}
\end{equation}
where, $F$ is the scale factor, $\mathbf{x}_{pbest}$ is a random individual from top 100p\% individuals of $P_1$, $p$ is factor to control the numbers of reference solutions, $r_1,r_2,r_3$ is different integers ranging from 1 to size of $P_2$.

For DE/transfer/1:
DE/transfer combination information of $Pop_{aux}$ and $Pop_{main}$.
\begin{equation}
    \label{eq:DE/pbest}
    \begin{cases}
        \mathbf{v}_{P_1,r},\quad if\hspace{1mm}rand < CR\hspace{1mm}or\hspace{1mm}d = d_{rand} \\
        \mathbf{x}_{P_2,r},\quad otherwise \\
        CR =[0.1,0.2,0.3]
    \end{cases}
\end{equation}
where $\mathbf{x}_{P_1,r} and \mathbf{x}_{P_2,r}$ are solutions from $Pop_{main}$ and $Pop_{aux}$, $rand$ is a random number between 0 and 1, $CR$ is the crossover rate, $d = (1,2,...,D)$ is a random integer, and $d_{rand}$ is a random integer uniformly generated between 1 and $D$.

For DE/current-to-rand/1:
\begin{equation}
    \label{eq:DE/rand}
    \begin{cases}
        \mathbf{v}_{P,r_1} + \mathbf{R} \cdot \mathbf{x}_{P,r_1}  + F\cdot(\mathbf{x}_{P,r2} -\mathbf{x}_{P,r_3}) \\
        \mathbf{u}_{P} = \mathbf{v}_{P}\\
        F = [0.6,0.8,1.0]\\
    \end{cases}
\end{equation}
where $F$ is the scale factor, $\mathbf{R}$ is a $D-dim$ random vector and $\mathbf{x}{P,r}$ are the solutions from the input population. $r_1,r_2,r_3$ are different integers ranging from 1 to the size of $P_2$.

\section{Ablation Study} \label{sec:AS}
For all other ablation experiment results, please refer to the supplementary materials (Sec.~\ref{sec:A4-6} and Sec.~\ref{sec:A4-7}).

\subsubsection{Results of the Base Framework}

We conducted ablation studies focusing on the opposite mechanism and the staged optimization phases. The results show that, for most problems, the inclusion of the opposite mechanism has no significant impact; however, a slight performance advantage was observed when it is enabled. While our adapted version of the opposite mechanism incorporates transformation techniques, its effectiveness remains generally consistent with the original implementation. Designed to address deceptive constraints—such as those found in the FCP test suite—the opposite mechanism plays a complementary role by mitigating substantial performance degradation caused by such constraints.

Importantly, when Stage 2 was altered by removing the epsilon-controlled three-phase structure, the algorithm's performance declined significantly on certain problems, particularly those in the LIRCMOP series. Although the vector-based and angular-partitioning environmental selection strategy was retained—remaining active for fully separable UPF-CPF problems like LIRCMOP1–3—the absence of the phased structure impaired performance. This outcome suggests that the epsilon decay function, discovered through LLM-based search, was inherently aligned with the original multi-phase design of Stage 2. Replacing it with a single-phase approach leads to inferior performance on LIRCMOP1–3 and similar problem types, highlighting the importance of the staged structure.

\begin{table}[htp!]
\centering
\caption{Wilcoxon rank sum test result and multi-problem wilcoxon signed rank test results with respect to HV and IGD on all 61 functions of the opposite module, epsilon control in the stage 2, hops included different UPF-CPF types from 1 to 4, and the DRA module.}
\label{tab:AB}
\resizebox{0.99\linewidth}{!}{%
\begin{tabular}{llllll}
\hline
LLM4CMO vs. & \multicolumn{1}{c}{HV(+/-/=)} & \multicolumn{1}{c}{$R^+$} & \multicolumn{1}{c}{$R^-$} & \multicolumn{1}{c}{$p$-value} & \multicolumn{1}{c}{level=0.05} \\ \hline
LLM4CMOWoRR & 30/5/26 & 1426 & 465 & 0.000168709 & YES \\
LLM4CMOWoS1C & 10/4/47 & 1082 & 809 & 0.218926479 & NO \\
LLM4CMOwoOP & 8/0/53 & 1195 & 696 & 0.039279306 & YES \\
LLM4CMOWo3P & 19/6/36 & 1327 & 564 & 0.002421513 & YES \\
LLM4CMOEps1 & 14/2/45 & 1208 & 683 & 0.031009586 & YES \\
LLM4CMOHOps-T1 & 26/15/20 & 1359 & 532 & 0.002977244 & YES \\
LLM4CMOHOps-T2 & 21/10/30 & 1294 & 597 & 0.01230769 & YES \\
LLM4CMOHOps-T3 & 30/3/28 & 1502 & 389 & 0.000015514 & YES \\
LLM4CMOHOps-T4 & 24/13/24 & 1503 & 388 & 6.21758E-05 & YES \\
LLM4CMOWoDRA & 6/0/55 & 1268 & 623 & 0.009359227 & YES \\ \hline
LLM4CMO vs. & \multicolumn{1}{c}{IGD(+/-/=)} & \multicolumn{1}{c}{$R^+$} & \multicolumn{1}{c}{$R^-$} & \multicolumn{1}{c}{$p$-value} & \multicolumn{1}{c}{level=0.05} \\ \hline
LLM4CMOWoRR & 27/6/28 & 1390 & 501 & 0.001409279 & YES \\
LLM4CMOWoS1C & 7/2/52 & 1163 & 728 & 0.118228233 & NO \\
LLM4CMOwoOP & 7/0/54 & 1224 & 667 & 0.045456762 & YES \\
LLM4CMOWo3P & 16/5/40 & 1255 & 636 & 0.026211143 & YES \\
LLM4CMOEps1 & 13/2/46 & 1343 & 548 & 0.004301606 & YES \\
LLM4CMOHOps-T1 & 26/16/19 & 1317 & 574 & 0.00762132 & YES \\
LLM4CMOHOps-T2 & 19/14/28 & 1188 & 703 & 0.081539403 & NO \\
LLM4CMOHOps-T3 & 27/4/30 & 1410 & 481 & 0.000848712 & YES \\
LLM4CMOHOps-T4 & 21/13/27 & 1342 & 549 & 0.0043999 & YES \\
LLM4CMOWoDRA & 5/1/55 & 1239 & 652 & 0.035018479 & YES \\ \hline
\end{tabular}%
}
\end{table}

In the design of the base framework, we found that employing a relaxed judgment condition for $R_s$ facilitates the algorithm’s transition to Stage 2 more efficiently. Compared to the previous hard threshold condition, the relaxed approach resulted in superior performance in nearly half of the test cases.

\subsubsection{Results of Core Modules}

We also performed before-and-after comparisons on the three core modules specifically designed through the LLM. The results indicate that while modifications to the DRA module had a relatively minor effect on overall performance, they still contributed positively. In contrast, replacing the original epsilon decay function led to performance improvements on 10 test functions, confirming its effectiveness. Among all modules, the HOps module had the most significant impact on performance.
To further evaluate HOps, we conducted tests on the operator combinations identified through LLM interaction. The results reveal that the overall performance of the four individual operator types was significantly lower than that of the complete HOps configuration. However, each template exhibited specific strengths when applied to certain problem types. This suggests that the HOps configurations derived via LLM are not entirely accurate for all type-specific applications. Nevertheless, the LLM4CMO we obtained and optimized through interaction with LLM has improved the core modules on the original framework, significantly enhancing the algorithm's performance, which preliminarily verifies the effectiveness of LLM interaction design.


\section{Sensitivity analysis of the $\epsilon$ and $\epsilon_0$} \label{sec:SA}
We tested the first and last results obtained by the LLM-aided design on the HOps task, and examined the parameter combinations of $\epsilon_0$ and the switch points of phase1, phase2 and phase3 in stage 2. It can be seen that the results of the parameters selected by our model in the set of test functions verified during the design process are not the best, and even tend to be worse parameter selections. Moreover, from the perspective of the HV metrics results of the two groups of experiments on HOps1 and HOps2, the improvement of performance does not depend on the selection of $\epsilon$ and switch point parameters.
\begin{table}[htp!]
\centering
\caption{Results of the HV metrics based on the verification of HOps4's sensitivity to $\epsilon_0$.}
\label{tab:HOps1e0}
\resizebox{0.99\linewidth}{!}{%
\begin{tabular}{ccccc}
\hline
\textbf{Problem} & \textbf{$\epsilon_0$=0.4} & \textbf{$\epsilon_0$=0.6} & \textbf{$\epsilon_0$=0.8} & \textbf{HOps1(0.2)} \\ \hline
\textbf{CF1} & 5.6280e-1 (8.61e-4) = & 5.6288e-1 (7.73e-4) - & 5.6270e-1 (6.23e-4) - & {\color[HTML]{3333E9} 5.6323e-1 (6.41e-4)} \\
\textbf{CF2} & 6.7699e-1 (1.33e-3) = & 6.7737e-1 (7.20e-4) = & 6.7642e-1 (2.03e-3) - & {\color[HTML]{3333E9} 6.7739e-1 (7.43e-4)} \\
\textbf{CF3} & {\color[HTML]{3333E9} 2.5336e-1 (3.29e-2) =} & 2.4289e-1 (3.10e-2) = & 2.4375e-1 (3.96e-2) = & 2.4577e-1 (4.33e-2) \\
\textbf{DASCMOP1} & 1.9378e-1 (4.78e-2) = & 1.9628e-1 (5.00e-2) = & 1.9164e-1 (5.33e-2) = & {\color[HTML]{3333E9} 1.9651e-1 (4.85e-2)} \\
\textbf{DASCMOP2} & 3.5547e-1 (1.49e-4) = & 3.5289e-1 (1.43e-2) = & 3.5546e-1 (1.99e-4) = & {\color[HTML]{3333E9} 3.5547e-1 (1.71e-4)} \\
\textbf{DASCMOP3} & 2.1678e-1 (2.53e-2) = & {\color[HTML]{3333E9} 2.3013e-1 (3.76e-2) =} & 2.1241e-1 (1.65e-2) = & 2.2657e-1 (3.81e-2) \\
\textbf{MW3} & 5.4460e-1 (4.35e-4) = & 5.4463e-1 (3.85e-4) + & {\color[HTML]{3333E9} 5.4464e-1 (3.67e-4) =} & 5.4442e-1 (4.73e-4) \\
\textbf{MW4} & 8.4140e-1 (5.30e-4) = & 8.4133e-1 (5.04e-4) = & 8.4118e-1 (6.21e-4) = & {\color[HTML]{3333E9} 8.4146e-1 (4.04e-4)} \\
\textbf{MW5} & 3.2443e-1 (1.82e-4) = & 3.2442e-1 (1.92e-4) = & {\color[HTML]{3333E9} 3.2450e-1 (1.22e-4) +} & 3.2438e-1 (2.21e-4) \\
\textbf{MW9} & 3.9709e-1 (2.22e-3) = & 3.9700e-1 (1.80e-3) = & {\color[HTML]{3333E9} 3.9848e-1 (1.60e-3) +} & 3.9672e-1 (2.55e-3) \\
\textbf{LIRCMOP1} & 2.1774e-1 (1.23e-2) = & {\color[HTML]{3333E9} 2.1806e-1 (1.42e-2) +} & 2.1659e-1 (1.27e-2) = & 2.1144e-1 (1.39e-2) \\
\textbf{LIRCMOP5} & 2.8187e-1 (5.32e-2) = & 2.9150e-1 (2.66e-4) = & {\color[HTML]{3333E9} 2.9152e-1 (2.92e-4) =} & 2.8183e-1 (5.32e-2) \\
\textbf{LIRCMOP9} & 4.9389e-1 (7.27e-2) = & {\color[HTML]{3333E9} 5.0136e-1 (9.14e-2) +} & 4.8237e-1 (8.78e-2) = & 4.5704e-1 (7.30e-2) \\
\textbf{LIRCMOP11} & 6.8473e-1 (1.85e-2) = & 6.8556e-1 (2.09e-2) = & {\color[HTML]{3333E9} 6.8947e-1 (1.16e-2) +} & 6.6989e-1 (2.36e-2) \\ \hline
\textbf{+/-/=} & 0/0/14 & 3/1/10 & 3/2/9 &  \\ \hline
\end{tabular}%
}
\end{table}

\begin{table}[htp!]
\centering
\caption{Results of the HV metrics based on the verification of HOps1's sensitivity to $\epsilon_0$.}
\label{tab:HOps4e0}
\resizebox{0.99\linewidth}{!}{%
\begin{tabular}{ccccc}
\hline
\textbf{Problem} & \textbf{$\epsilon_0$=0.4} & \textbf{$\epsilon_0$=0.6} & \textbf{$\epsilon_0$=0.8} & \textbf{HOp4 (0.2)} \\ \hline
\textbf{CF1} & 5.6294e-1 (7.39e-4) - & 5.6279e-1 (8.95e-4) - & 5.6293e-1 (7.71e-4) - & {\color[HTML]{3333E9} 5.6349e-1 (4.06e-4)} \\
\textbf{CF2} & 6.7661e-1 (1.54e-3) - & 6.7725e-1 (9.91e-4) = & 6.7715e-1 (8.63e-4) = & {\color[HTML]{3333E9} 6.7747e-1 (7.76e-4)} \\
\textbf{CF3} & 2.5637e-1 (3.42e-2) = & 2.4761e-1 (4.05e-2) = & {\color[HTML]{3333E9} 2.6076e-1 (3.74e-2) =} & 2.5257e-1 (4.45e-2) \\
\textbf{DASCMOP1} & 2.1237e-1 (4.87e-4) = & {\color[HTML]{3333E9} 2.1243e-1 (4.23e-4) =} & 2.1238e-1 (3.07e-4) = & 2.1241e-1 (4.52e-4) \\
\textbf{DASCMOP2} & {\color[HTML]{3333E9} 3.5514e-1 (8.67e-5) =} & 3.5512e-1 (1.01e-4) = & 3.5509e-1 (7.35e-5) = & 3.5512e-1 (8.63e-5) \\
\textbf{DASCMOP3} & 3.1172e-1 (2.41e-3) = & {\color[HTML]{3333E9} 3.1179e-1 (2.48e-3) =} & 3.1134e-1 (3.35e-3) = & 3.0055e-1 (2.97e-2) \\
\textbf{MW3} & 5.4456e-1 (4.38e-4) = & 5.4455e-1 (5.39e-4) = & 5.4449e-1 (5.16e-4) = & {\color[HTML]{3333E9} 5.4457e-1 (4.00e-4)} \\
\textbf{MW4} & 8.4149e-1 (4.53e-4) = & 8.4149e-1 (5.88e-4) = & {\color[HTML]{3333E9} 8.4156e-1 (4.73e-4) +} & 8.4133e-1 (4.70e-4) \\
\textbf{MW5} & 3.2443e-1 (2.43e-4) = & 3.2444e-1 (2.12e-4) = & {\color[HTML]{3333E9} 3.2446e-1 (2.25e-4) =} & 3.2444e-1 (1.60e-4) \\
\textbf{MW9} & 3.9570e-1 (1.82e-3) = & 3.9596e-1 (1.54e-3) + & {\color[HTML]{3333E9} 3.9598e-1 (2.13e-3) +} & 3.9387e-1 (4.03e-3) \\
\textbf{LIRCMOP1} & {\color[HTML]{3333E9} 2.3670e-1 (6.94e-4) =} & 2.3636e-1 (1.02e-3) = & 2.3634e-1 (1.26e-3) = & 2.3642e-1 (1.02e-3) \\
\textbf{LIRCMOP5} & 2.9157e-1 (2.90e-4) = & 2.9156e-1 (3.22e-4) = & {\color[HTML]{3333E9} 2.9160e-1 (2.99e-4) +} & 2.9146e-1 (2.62e-4) \\
\textbf{LIRCMOP9} & 4.7706e-1 (7.62e-2) = & 4.9903e-1 (6.63e-2) = & {\color[HTML]{3333E9} 5.1873e-1 (6.06e-2) +} & 4.7458e-1 (7.36e-2) \\
\textbf{LIRCMOP11} & 6.8328e-1 (2.51e-2) = & {\color[HTML]{3333E9} 6.8714e-1 (2.02e-2) =} & 6.8137e-1 (3.53e-2) = & 6.7887e-1 (2.66e-2) \\ \hline
\textbf{+/-/=} & 0/2/12 & 1/1/12 & 4/1/9 &  \\ \hline
\end{tabular}%
}
\end{table}

\begin{table}[htp!]
\centering
\caption{Result of HV metrics for sensitivity test of three-stage switch parameters in HOps1. $S_1$ and $S_2$ represent switch points of phase 1 to 2 and phase 2 to phase 3. $tt$ is the threshold of activating opposition offspring. The original parameters are [0.195,0.005,0.0005].}
\label{tab:HOps1ecp}
\resizebox{0.99\linewidth}{!}{%
\begin{tabular}{ccccc}
\hline
\textbf{Problem} & \textbf{[$S_1$,$S_2$,$tt$] = {[}0.19,0.01,0.001{]}} & \textbf{[$S_1$,$S_2$,$tt$] = {[}0.18,0.02,0.002{]}} & \textbf{[$S_1$,$S_2$,$tt$] = {[}0.15,0.05,0.005{]}} & \textbf{HOps1} \\ \hline
\textbf{CF1} & 5.6316e-1 (6.15e-4) = & 5.6321e-1 (5.65e-4) = & 5.6297e-1 (7.03e-4) = & {\color[HTML]{3333E9} 5.6323e-1 (6.41e-4)} \\
\textbf{CF2} & 6.7733e-1 (6.86e-4) = & 6.7708e-1 (8.48e-4) = & 6.7717e-1 (8.67e-4) = & {\color[HTML]{3333E9} 6.7739e-1 (7.43e-4)} \\
\textbf{CF3} & {\color[HTML]{3333E9} 2.5950e-1 (3.12e-2) =} & 2.5742e-1 (3.17e-2) = & 2.4916e-1 (3.42e-2) = & 2.4577e-1 (4.33e-2) \\
\textbf{DASCMOP1} & 1.8457e-1 (5.68e-2) = & 1.9406e-1 (4.78e-2) - & 1.8232e-1 (6.15e-2) = & {\color[HTML]{3333E9} 1.9651e-1 (4.85e-2)} \\
\textbf{DASCMOP2} & 3.5294e-1 (1.46e-2) - & 3.5275e-1 (1.57e-2) - & {\color[HTML]{3333E9} 3.5560e-1 (1.43e-4) +} & 3.5547e-1 (1.71e-4) \\
\textbf{DASCMOP3} & 2.2075e-1 (3.11e-2) = & 2.2643e-1 (3.33e-2) = & 2.1528e-1 (1.63e-2) = & {\color[HTML]{3333E9} 2.2657e-1 (3.81e-2)} \\
\textbf{MW3} & 5.4442e-1 (4.87e-4) = & {\color[HTML]{3333E9} 5.4451e-1 (3.32e-4) =} & 5.4446e-1 (4.37e-4) = & 5.4442e-1 (4.73e-4) \\
\textbf{MW4} & 8.4129e-1 (1.08e-3) = & 8.4120e-1 (4.59e-4) - & 8.4124e-1 (4.94e-4) = & {\color[HTML]{3333E9} 8.4146e-1 (4.04e-4)} \\
\textbf{MW5} & 3.2435e-1 (2.87e-4) = & 3.2432e-1 (2.69e-4) = & {\color[HTML]{3333E9} 3.2442e-1 (1.82e-4) =} & 3.2438e-1 (2.21e-4) \\
\textbf{MW9} & 3.9716e-1 (3.34e-3) = & 3.9704e-1 (2.66e-3) = & {\color[HTML]{3333E9} 3.9756e-1 (1.94e-3) =} & 3.9672e-1 (2.55e-3) \\
\textbf{LIRCMOP1} & {\color[HTML]{3333E9} 2.1740e-1 (8.11e-3) =} & 2.1326e-1 (1.30e-2) = & 2.1444e-1 (1.29e-2) = & 2.1144e-1 (1.39e-2) \\
\textbf{LIRCMOP5} & 2.9156e-1 (2.69e-4) = & 2.9143e-1 (3.71e-4) = & {\color[HTML]{3333E9} 2.9159e-1 (2.29e-4) =} & 2.8183e-1 (5.32e-2) \\
\textbf{LIRCMOP9} & 4.7215e-1 (7.57e-2) = & {\color[HTML]{3333E9} 4.7441e-1 (7.42e-2) =} & 4.5379e-1 (1.07e-1) = & 4.5704e-1 (7.30e-2) \\
\textbf{LIRCMOP11} & 6.6747e-1 (3.44e-2) = & 6.7309e-1 (4.21e-2) = & {\color[HTML]{3333E9} 6.8191e-1 (2.14e-2) =} & 6.6989e-1 (2.36e-2) \\ \hline
\textbf{+/-/=} & 0/1/13 & 0/3/11 & 1/0/13 &  \\ \hline
\end{tabular}%
}
\end{table}

\begin{table}[htp!]
\centering
\caption{Result of HV metrics for sensitivity test of three-stage switch parameters in HOps4. $S_1$ and $S_2$ represent switch points of phase 1 to 2 and phase 2 to phase 3. $tt$ is the threshold of activating opposition offspring. The original parameters are [0.195,0.005,0.0005].}
\label{tab:HOps4ecp}
\resizebox{0.99\linewidth}{!}{%
\begin{tabular}{ccccc}
\hline
\textbf{Problem} & \textbf{[$S_1$,$S_2$,$tt$] = {[}0.19,0.01,0.001{]}} & \textbf{[$S_1$,$S_2$,$tt$] = {[}0.18,0.02,0.002{]}} & \textbf{[$S_1$,$S_2$,$tt$] = {[}0.15,0.05,0.005{]}} & \textbf{HOps4} \\ \hline
\textbf{CF1} & 5.6344e-1 (6.61e-4) = & 5.6337e-1 (6.07e-4) = & 5.6312e-1 (7.49e-4) = & {\color[HTML]{3333E9} 5.6349e-1 (4.06e-4)} \\
\textbf{CF2} & 6.7721e-1 (8.37e-4) = & 6.7710e-1 (7.83e-4) - & 6.7715e-1 (1.19e-3) = & {\color[HTML]{3333E9} 6.7747e-1 (7.76e-4)} \\
\textbf{CF3} & 2.5093e-1 (2.94e-2) = & {\color[HTML]{3333E9} 2.5537e-1 (5.08e-2) =} & 2.3483e-1 (3.85e-2) = & 2.5257e-1 (4.45e-2) \\
\textbf{DASCMOP1} & 2.1268e-1 (3.24e-4) + & {\color[HTML]{3333E9} 2.1269e-1 (3.19e-4) +} & 2.1268e-1 (2.73e-4) + & 2.1241e-1 (4.52e-4) \\
\textbf{DASCMOP2} & 3.5521e-1 (8.08e-5) + & {\color[HTML]{3333E9} 3.5522e-1 (7.85e-5) +} & 3.5521e-1 (1.04e-4) + & 3.5512e-1 (8.63e-5) \\
\textbf{DASCMOP3} & {\color[HTML]{3333E9} 3.1032e-1 (7.10e-3) +} & 2.9332e-1 (3.53e-2) = & 2.9443e-1 (3.53e-2) = & 3.0055e-1 (2.97e-2) \\
\textbf{MW3} & {\color[HTML]{3333E9} 5.4461e-1 (3.83e-4) =} & 5.4444e-1 (4.22e-4) = & 5.4458e-1 (4.18e-4) = & 5.4457e-1 (4.00e-4) \\
\textbf{MW4} & {\color[HTML]{3333E9} 8.4139e-1 (3.76e-4) =} & 8.4134e-1 (4.76e-4) = & 8.4128e-1 (4.56e-4) = & 8.4133e-1 (4.70e-4) \\
\textbf{MW5} & {\color[HTML]{3333E9} 3.2447e-1 (1.74e-4) =} & 3.2444e-1 (2.07e-4) = & 3.2440e-1 (2.21e-4) = & 3.2444e-1 (1.60e-4) \\
\textbf{MW9} & 3.9523e-1 (2.20e-3) = & 3.9517e-1 (2.44e-3) = & {\color[HTML]{3333E9} 3.9590e-1 (2.13e-3) +} & 3.9387e-1 (4.03e-3) \\
\textbf{LIRCMOP1} & 2.3592e-1 (2.05e-3) = & 2.3585e-1 (2.06e-3) = & 2.3499e-1 (2.55e-3) - & {\color[HTML]{3333E9} 2.3642e-1 (1.02e-3)} \\
\textbf{LIRCMOP5} & 2.9149e-1 (2.69e-4) = & {\color[HTML]{3333E9} 2.9154e-1 (2.16e-4) =} & 2.9151e-1 (3.25e-4) = & 2.9146e-1 (2.62e-4) \\
\textbf{LIRCMOP9} & 4.9411e-1 (7.30e-2) = & {\color[HTML]{3333E9} 4.9502e-1 (7.47e-2) =} & 4.8287e-1 (6.88e-2) = & 4.7458e-1 (7.36e-2) \\
\textbf{LIRCMOP11} & {\color[HTML]{3333E9} 6.8249e-1 (1.93e-2) =} & 6.7191e-1 (3.95e-2) = & 6.7912e-1 (2.63e-2) = & 6.7887e-1 (2.66e-2) \\ \hline
\textbf{+/-/=} & 3/0/11 & 2/1/11 & 3/1/10 &  \\ \hline
\end{tabular}%
}
\end{table}
\newpage
\section{Epsilon decay function}
 The epsilon decay function finally obtained through LLM-human interaction is presented as follows:
\begin{footnotesize} 
    \label{eq:epsilon final}
    \begin{align}
        \small
        \epsilon(FE) = 
        \begin{cases}
        \epsilon_0 \cdot \left(0.9 + 0.1 \cdot \frac{\ln\left(1 + a \cdot \left(1 - \frac{FE}{t_1}\right)\right)}{\ln(1 + a)}\right) & \text{if } FE \leq t_1, \\
        Peri &\text{if } t_1 < FE \leq t_2, \\[1em]
        0.005 \cdot \epsilon_0 \cdot \exp\left(-k \cdot \frac{FE - t_2}{maxFE - t_2}\right) & \text{if } FE > t_2,
        \end{cases}
    \end{align}
    \vspace{-1em}
\end{footnotesize} 

\noindent where
\begin{itemize}
    \item $\epsilon_0$ is initial value of epsilon decay process,
    \item $a = 15$ is the logarithmic curvature control parameter,
    \item $t_1 = FE_{\text{switch}} + 0.2 \cdot (maxFE - FE_{\text{switch}})$ is the end of Phase 1,
    \item $t_2 = \min\left(t_1 + 0.3 \cdot (maxFE - FE_{\text{switch}}), maxFE\right)$ is the end of Phase 2,
    \item $Peri$ is 
    \[\begin{cases} 
        b_3(t) + a_3(t) \cdot \sin\left(\frac{2\pi \cdot FE}{200}\right) & \text{if type} = 3, \\
        b_{1|2}(t) + a_{1|2}(t) \cdot \sin\left(\frac{2\pi \cdot FE}{150}\right) & \text{otherwise},
        \end{cases}\]
    \item $k = \ln\left(\frac{0.005 \cdot \epsilon_0}{10^{-8}}\right)$ is the decay rate for Phase 3,
    \item The baseline functions for Phase 2 are:
        \begin{align*}
            \begin{cases}
            b_3(t) &= 0.05\epsilon_0 - (0.05 - 0.005)\epsilon_0 \cdot t, \\
            b_{1|2}(t) &= 0.9\epsilon_0 - (0.9 - 0.005)\epsilon_0 \cdot t,
            \end{cases}
        \end{align*}
    \item The amplitude functions for Phase 2 are:
        \begin{align*}
            \begin{cases}
                 a_3(t) &= 0.005\epsilon_0 \cdot \exp(-8t), \\
                 a_{1|2}(t) &= 0.04\epsilon_0 \cdot \exp(-5t),\\
                 t = \frac{FE - t_1}{t_2 - t_1}
            \end{cases}
        \end{align*}
\end{itemize}

\clearpage
\onecolumn
\section{The Friedman test results.} \label{sec:A3}

In this section, we present the hypervolume (HV) results obtained by running LLM4CMO and eleven baseline algorithms 30 times on six benchmark test suites. We perform the Friedman test with a significance level of \( p < 0.05 \). To illustrate the comparative performance, we include average rank plots, HV median plots, failure rate plots, and the final algorithm ranking.

Based on the Friedman test results, LLM4CMO demonstrates the best overall performance in terms of HV median values. Regarding average rank sum, LLM4CMO achieves the top position, significantly outperforming the second-best algorithm, URCMO. The associated \( p \)-values are close to zero, indicating statistically significant differences between LLM4CMO and the other algorithms.

Although LLM4CMO performs slightly worse than URCMO in terms of failure rate, this does not diminish its overall advantage. Further analysis reveals that most failures occur on several problems in the DOC test suite, suggesting that HOps may not be optimally configured and are potentially over-biased toward promoting diversity.

\vspace{1.2cm}
\begin{figure}[hptb!]
    \centering
    \includegraphics[width=1.0\textwidth]{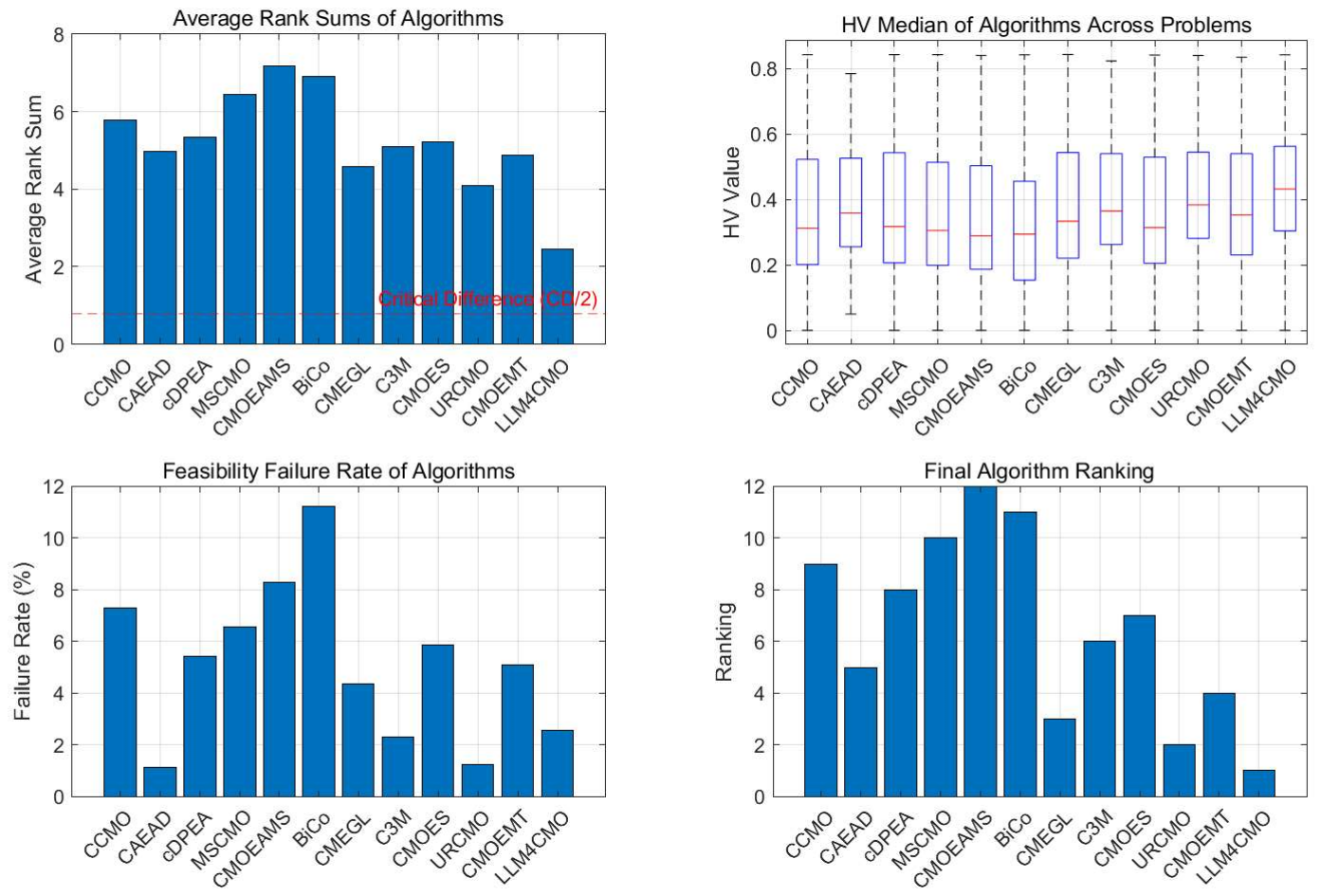}
    \caption{The Friedman test results of LLM4CMO and other 11 algorithms run 30 times on 6 test suites.}
    \label{fig:ft}
\end{figure}

\clearpage
\section{Others HV and IGD metrics tables on six test suite.} \label{sec:A4}
\subsection{The HV and IGD metrics tables on CF test suite.} \label{sec:A4-1}
\begin{table}[hptb!]
\centering
\caption{The results of mean and std HV metric obtained by twelve algorithms on the CF test suite.}
\label{tab:CFHV}
\resizebox{\textwidth}{!}{%
\begin{tabular}{ccccccccccccc}
\hline
\textbf{Problem} & \textbf{CCMO} & \textbf{CAEAD} & \textbf{cDPEA} & \textbf{MSCMO} & \textbf{CMOEAMS} & \textbf{BiCo} & \textbf{CMEGL} & \textbf{C3M} & \textbf{URCMO} & \textbf{CMOES} & \textbf{CMOEMT} & \textbf{LLM4CMO} \\ \hline
\textbf{CF1} & 5.6193e-1 (4.68e-4) - & {\color[HTML]{3333E9} 5.6567e-1 (9.49e-5) +} & 5.6163e-1 (6.56e-4) - & 5.5710e-1 (2.04e-3) - & 5.4751e-1 (1.53e-2) - & 5.5464e-1 (1.68e-3) - & 5.6279e-1 (3.20e-4) - & 5.6551e-1 (1.80e-4) + & 5.6335e-1 (4.13e-4) - & 5.6380e-1 (2.49e-4) = & 5.6288e-1 (5.17e-4) - & 5.6357e-1 (6.44e-4) \\
\textbf{CF2} & 6.4523e-1 (1.55e-2) - & 6.7453e-1 (1.13e-3) - & 6.4660e-1 (1.76e-2) - & 6.3557e-1 (2.08e-2) - & 6.1362e-1 (3.59e-2) - & 6.1804e-1 (2.58e-2) - & 6.4487e-1 (2.04e-2) - & 6.7568e-1 (5.93e-4) - & 6.7504e-1 (1.68e-3) - & 6.4935e-1 (1.46e-2) - & {\color[HTML]{3333E9} 6.7875e-1 (2.38e-4) +} & 6.7699e-1 (1.48e-3) \\
\textbf{CF3} & 1.9376e-1 (4.65e-2) - & 1.1529e-1 (4.31e-2) - & 1.7996e-1 (3.29e-2) - & 1.8551e-1 (4.26e-2) - & 1.8514e-1 (4.09e-2) - & 2.0479e-1 (3.63e-2) - & 1.8778e-1 (4.79e-2) - & 1.3231e-1 (4.60e-2) - & 1.9094e-1 (4.72e-2) - & 2.0222e-1 (3.36e-2) - & 2.2314e-1 (4.34e-2) - & {\color[HTML]{3333E9} 2.5405e-1 (3.44e-2)} \\
\textbf{CF4} & 4.3612e-1 (3.71e-2) - & 4.7832e-1 (1.11e-2) - & 4.3196e-1 (3.26e-2) - & 4.1839e-1 (3.28e-2) - & 4.1088e-1 (3.39e-2) - & 4.3410e-1 (1.94e-2) - & 4.2861e-1 (2.87e-2) - & 4.8900e-1 (1.14e-2) - & 5.0548e-1 (1.12e-2) - & 4.3927e-1 (2.56e-2) - & 4.9649e-1 (2.22e-2) - & {\color[HTML]{3333E9} 5.1364e-1 (7.07e-3)} \\
\textbf{CF5} & 2.5919e-1 (7.31e-2) - & 2.7900e-1 (8.14e-2) - & 2.7825e-1 (7.15e-2) - & 3.0107e-1 (6.46e-2) - & 2.5960e-1 (7.94e-2) - & 2.4651e-1 (8.89e-2) - & 2.9311e-1 (6.10e-2) - & 3.3784e-1 (7.42e-2) - & 3.1571e-1 (8.14e-2) - & 2.9922e-1 (6.73e-2) - & 3.6527e-1 (3.93e-2) - & {\color[HTML]{3333E9} 3.9535e-1 (6.10e-2)} \\
\hline
\end{tabular}%
}
\end{table}

\begin{table}[hptb!]
\centering
\caption{The results of mean and std IGD metric obtained by twelve algorithms on the CF test suite.}
\label{tab:CFIGD}
\resizebox{\textwidth}{!}{%
\begin{tabular}{ccccccccccccc}
\hline
\textbf{Problem} & \textbf{CCMO} & \textbf{CAEAD} & \textbf{cDPEA} & \textbf{MSCMO} & \textbf{CMOEAMS} & \textbf{BiCo} & \textbf{CMEGL} & \textbf{C3M} & \textbf{URCMO} & \textbf{CMOES} & \textbf{CMOEMT} & \textbf{LLM4CMO} \\ \hline
\textbf{CF1} & 3.3719e-3 (3.62e-4) - & {\color[HTML]{3333E9} 3.6231e-4 (7.80e-5) +} & 3.6273e-3 (5.55e-4) - & 7.1394e-3 (1.58e-3) - & 1.4994e-2 (1.28e-2) - & 8.9712e-3 (1.27e-3) - & 2.6623e-3 (2.51e-4) - & 4.9397e-4 (1.47e-4) + & 2.2589e-3 (3.37e-4) - & 1.8542e-3 (1.89e-4) = & 2.6528e-3 (4.28e-4) - & 2.0762e-3 (5.21e-4) \\
\textbf{CF2} & 2.7745e-2 (1.09e-2) - & 4.4803e-3 (7.53e-4) - & 2.5494e-2 (1.07e-2) - & 3.3130e-2 (1.60e-2) - & 4.7130e-2 (3.34e-2) - & 4.4662e-2 (1.58e-2) - & 2.6521e-2 (1.26e-2) - & 3.7914e-3 (2.38e-4) - & 4.7458e-3 (4.31e-4) - & 2.7012e-2 (1.20e-2) - & {\color[HTML]{3333E9} 2.9876e-3 (1.53e-4) +} & 3.4599e-3 (6.36e-4) \\
\textbf{CF3} & 2.5206e-1 (1.11e-1) - & 2.4979e-1 (6.38e-2) - & 2.4339e-1 (4.51e-2) - & 2.5303e-1 (1.09e-1) - & 2.5717e-1 (8.95e-2) - & 2.1991e-1 (7.73e-2) - & 2.3193e-1 (6.76e-2) - & 2.3387e-1 (6.34e-2) - & 2.1041e-1 (7.23e-2) - & 2.3325e-1 (6.32e-2) - & 1.9935e-1 (1.01e-1) - & {\color[HTML]{3333E9} 1.4396e-1 (6.07e-2)} \\
\textbf{CF4} & 8.2240e-2 (3.84e-2) - & 4.2418e-2 (7.96e-3) - & 8.2773e-2 (2.73e-2) - & 9.4791e-2 (3.07e-2) - & 1.1970e-1 (5.43e-2) - & 7.8961e-2 (1.79e-2) - & 8.8455e-2 (3.06e-2) - & 3.4684e-2 (7.96e-3) - & 2.4224e-2 (8.48e-3) - & 7.5696e-2 (2.33e-2) - & 3.2415e-2 (2.19e-2) - & {\color[HTML]{3333E9} 1.7778e-2 (6.34e-3)} \\
\textbf{CF5} & 3.5408e-1 (1.34e-1) - & 2.2607e-1 (9.11e-2) - & 3.0191e-1 (1.30e-1) - & 2.7538e-1 (1.17e-1) - & 3.5001e-1 (1.40e-1) - & 3.5708e-1 (1.56e-1) - & 2.8071e-1 (1.04e-1) - & 1.6156e-1 (8.56e-2) = & 2.7313e-1 (1.25e-1) - & 2.5178e-1 (1.15e-1) - & 1.7934e-1 (5.52e-2) - & {\color[HTML]{3333E9} 1.5043e-1 (8.61e-2)} \\
\hline
\end{tabular}%
}
\end{table}
\subsection{The HV and IGD metrics tables on DASCMOP and MW test suite.} \label{sec:A4-2}
\begin{table}[hptb!]
\centering
\caption{The results of mean and std HV metric obtained by twelve algorithms on the DASCMOP and MW test suite.}
\label{tab:DASMWHV}
\resizebox{\textwidth}{!}{%
\begin{tabular}{ccccccccccccc}
\hline
\textbf{Problem} & \textbf{CCMO} & \textbf{CAEAD} & \textbf{cDPEA} & \textbf{MSCMO} & \textbf{CMOEAMS} & \textbf{BiCo} & \textbf{CMEGL} & \textbf{C3M} & \textbf{URCMO} & \textbf{CMOES} & \textbf{CMOEMT} & \textbf{LLM4CMO} \\ \hline
\textbf{DASCMOP1} & 1.6324e-2 (1.75e-2) - & 2.1208e-1 (4.21e-4) - & 2.7306e-2 (3.01e-2) - & 2.1323e-2 (1.64e-2) - & 1.5311e-2 (1.80e-2) - & 9.6000e-3 (6.22e-3) - & 3.1825e-2 (2.35e-2) - & 2.1203e-1 (7.38e-4) - & 2.1232e-1 (3.62e-4) - & 2.4749e-2 (2.03e-2) - & 1.5753e-1 (6.69e-2) - & {\color[HTML]{3333E9} 2.1254e-1 (4.07e-4)} \\
\textbf{DASCMOP2} & 2.5957e-1 (2.91e-3) - & 3.5502e-1 (8.46e-5) - & 2.6299e-1 (3.93e-3) - & 2.6882e-1 (5.65e-3) - & 2.5711e-1 (4.85e-3) - & 2.5245e-1 (3.52e-3) - & 2.7138e-1 (3.88e-3) - & 3.5498e-1 (1.01e-4) - & 3.5500e-1 (1.21e-4) - & 2.6542e-1 (4.88e-3) - & 3.5017e-1 (1.53e-2) - & {\color[HTML]{3333E9} 3.5513e-1 (9.94e-5)} \\
\textbf{DASCMOP3} & 2.1564e-1 (1.56e-2) - & 3.1202e-1 (2.84e-4) - & 2.1937e-1 (1.48e-2) - & 2.1953e-1 (1.18e-2) - & 2.0938e-1 (2.39e-3) - & 2.1560e-1 (8.99e-3) - & 2.1980e-1 (1.14e-2) - & 2.9466e-1 (2.55e-2) - & 2.3407e-1 (4.35e-2) - & 2.2562e-1 (1.53e-2) - & 2.8111e-1 (4.48e-2) - & {\color[HTML]{3333E9} 3.1230e-1 (1.97e-4)} \\
\textbf{DASCMOP4} & 2.0367e-1 (2.14e-3) - & 1.9988e-1 (4.93e-3) - & 2.0303e-1 (2.92e-3) = & 2.0362e-1 (2.11e-3) = & 1.9737e-1 (1.28e-2) = & 1.8578e-1 (3.72e-2) - & {\color[HTML]{3333E9} 2.0412e-1 (2.36e-4) =} & 1.9645e-1 (2.64e-2) - & 1.9149e-1 (3.22e-2) - & 2.0412e-1 (2.08e-4) = & 2.0142e-1 (3.79e-3) - & 2.0396e-1 (4.43e-4) \\
\textbf{DASCMOP5} & 3.5157e-1 (1.45e-4) + & 3.4502e-1 (1.95e-2) - & 3.5131e-1 (1.94e-4) = & 3.5138e-1 (2.08e-4) = & {\color[HTML]{3333E9} 3.5175e-1 (6.27e-5) +} & 3.3531e-1 (6.09e-2) = & 3.5156e-1 (1.10e-4) + & 3.3855e-1 (5.59e-2) - & 3.1824e-1 (8.91e-2) - & 3.5150e-1 (1.03e-4) = & 3.5120e-1 (1.92e-4) - & 3.5132e-1 (3.81e-4) \\
\hline
\end{tabular}%
}

\end{table}
\begin{table}[hptb!]
\centering
\caption{The results of mean and std IGD metric obtained by twelve algorithms on the DASCMOP and MW test suite.}
\label{tab:DASMWIGD}
\resizebox{\textwidth}{!}{%
\begin{tabular}{ccccccccccccc}
\hline
\textbf{Problem} & \textbf{CCMO} & \textbf{CAEAD} & \textbf{cDPEA} & \textbf{MSCMO} & \textbf{CMOEAMS} & \textbf{BiCo} & \textbf{CMEGL} & \textbf{C3M} & \textbf{URCMO} & \textbf{CMOES} & \textbf{CMOEMT} & \textbf{LLM4CMO} \\ \hline
\textbf{DASCMOP1} & 7.0566e-1 (3.83e-2) - & 3.2799e-3 (2.96e-4) - & 6.5962e-1 (9.69e-2) - & 6.7553e-1 (3.69e-2) - & 7.1261e-1 (3.19e-2) - & 7.1985e-1 (3.40e-2) - & 6.7297e-1 (3.92e-2) - & 4.4103e-3 (1.10e-3) - & 2.9662e-3 (2.08e-4) = & 6.7404e-1 (4.62e-2) - & 2.0933e-1 (2.50e-1) - & {\color[HTML]{3333E9} 2.9463e-3 (2.21e-4)} \\
\textbf{DASCMOP2} & 2.4770e-1 (1.89e-2) - & 4.4759e-3 (1.03e-4) = & 2.3366e-1 (3.19e-2) - & 2.1302e-1 (2.50e-2) - & 2.5398e-1 (3.59e-2) - & 2.4120e-1 (2.10e-2) - & 2.0865e-1 (2.65e-2) - & 4.4578e-3 (1.04e-4) = & 4.4695e-3 (1.22e-4) = & 2.2396e-1 (2.45e-2) - & 1.5515e-2 (3.82e-2) - & {\color[HTML]{3333E9} 4.4413e-3 (1.24e-4)} \\
\textbf{DASCMOP3} & 3.3335e-1 (4.92e-2) - & 1.9240e-2 (1.20e-3) = & 2.9818e-1 (5.49e-2) - & 2.6694e-1 (4.01e-2) - & 3.4205e-1 (3.86e-2) - & 2.8708e-1 (4.51e-2) - & 2.6365e-1 (3.44e-2) - & 9.0441e-2 (1.06e-1) - & 2.5955e-1 (1.39e-1) - & 2.4752e-1 (3.35e-2) - & 1.1680e-1 (1.41e-1) = & {\color[HTML]{3333E9} 1.7329e-2 (3.44e-3)} \\
\textbf{DASCMOP4} & 1.3512e-3 (7.30e-4) - & 4.3903e-3 (3.61e-3) - & 1.6075e-3 (1.06e-3) = & 1.3850e-3 (7.25e-4) = & 2.8145e-2 (7.30e-2) = & 4.2202e-2 (9.01e-2) - & {\color[HTML]{3333E9} 1.1941e-3 (5.76e-5) =} & 1.5842e-2 (6.56e-2) - & 2.9426e-2 (8.10e-2) - & 1.1959e-3 (4.93e-5) = & 2.2122e-3 (1.33e-3) - & 1.3109e-3 (3.27e-4) \\
\textbf{DASCMOP5} & 2.7436e-3 (9.43e-5) + & 1.0795e-2 (2.93e-2) - & 2.8895e-3 (2.23e-4) = & 2.8284e-3 (1.15e-4) + & {\color[HTML]{3333E9} 2.6649e-3 (7.25e-5) +} & 2.9858e-2 (1.02e-1) - & 2.7321e-3 (5.29e-5) + & 2.6780e-2 (1.16e-1) - & 6.6056e-2 (1.77e-1) = & 2.7883e-3 (6.33e-5) + & 2.9854e-3 (1.67e-4) = & 2.9791e-3 (2.28e-4) \\
\hline
\end{tabular}%
}
\end{table}

\subsection{The HV and IGD metrics tables on LIRCMOP, DOC and FCP test suite.} \label{sec:A4-3}
\begin{table}[H]
\centering
\caption{The results of mean and std HV metric obtained by twelve algorithms on the LIRCMOP, DOC and FCP test suite.}
\label{tab:LIRDOCFCPHV}
\resizebox{\textwidth}{!}{%
\begin{tabular}{ccccccccccccc}
\hline
\textbf{Problem} & \textbf{CCMO} & \textbf{CAEAD} & \textbf{cDPEA} & \textbf{MSCMO} & \textbf{CMOEAMS} & \textbf{BiCo} & \textbf{CMEGL} & \textbf{C3M} & \textbf{URCMO} & \textbf{CMOES} & \textbf{CMOEMT} & \textbf{LLM4CMO} \\ \hline
\textbf{LIRCMOP1} & 1.4374e-1 (1.73e-2) - & 2.3348e-1 (3.51e-3) - & 1.6690e-1 (2.11e-2) - & 1.6251e-1 (2.35e-2) - & 1.0819e-1 (1.50e-2) - & 1.5358e-1 (7.62e-3) - & 1.9265e-1 (1.37e-2) - & 1.9296e-1 (2.58e-2) - & 2.0778e-1 (1.74e-2) - & 1.7221e-1 (1.56e-2) - & 2.3518e-1 (2.44e-3) - & {\color[HTML]{3333E9} 2.3694e-1 (7.05e-4)} \\
\textbf{LIRCMOP2} & 2.7719e-1 (2.04e-2) - & 3.5917e-1 (1.10e-3) - & 2.9253e-1 (1.44e-2) - & 2.9247e-1 (1.74e-2) - & 2.2727e-1 (1.86e-2) - & 2.8227e-1 (1.07e-2) - & 3.1924e-1 (9.96e-3) - & 3.4323e-1 (2.76e-2) - & 3.5758e-1 (1.29e-2) - & 3.0579e-1 (1.00e-2) - & 3.5680e-1 (8.04e-3) - & {\color[HTML]{3333E9} 3.6045e-1 (1.62e-4)} \\
\textbf{LIRCMOP3} & 1.3733e-1 (1.73e-2) - & 2.0612e-1 (2.19e-3) - & 1.4986e-1 (1.77e-2) - & 1.4633e-1 (2.07e-2) - & 1.0161e-1 (1.12e-2) - & 1.3529e-1 (9.38e-3) - & 1.4618e-1 (1.39e-2) - & 1.6988e-1 (2.06e-2) - & 1.9265e-1 (2.26e-2) = & 1.6425e-1 (1.43e-2) - & 2.0118e-1 (1.44e-2) - & {\color[HTML]{3333E9} 2.0853e-1 (4.39e-4)} \\
\textbf{LIRCMOP4} & 2.2028e-1 (2.60e-2) - & 3.1401e-1 (3.52e-3) - & 2.4634e-1 (2.23e-2) - & 2.5933e-1 (1.62e-2) - & 1.9487e-1 (1.83e-2) - & 2.3214e-1 (1.51e-2) - & 2.4073e-1 (1.23e-2) - & 2.6126e-1 (4.26e-2) - & 2.8482e-1 (3.87e-2) - & 2.6525e-1 (1.20e-2) - & 3.0999e-1 (1.56e-2) - & {\color[HTML]{3333E9} 3.1742e-1 (4.71e-4)} \\
\textbf{LIRCMOP5} & 1.6433e-1 (1.79e-2) - & 2.8688e-1 (2.72e-3) - & 1.5884e-1 (2.42e-2) - & 1.3046e-1 (7.53e-2) - & 1.5809e-1 (2.21e-2) - & 0.0000e+0 (0.00e+0) - & 1.6320e-1 (2.32e-2) - & 2.7946e-1 (5.28e-2) - & 2.9042e-1 (7.38e-4) - & 1.6545e-1 (1.63e-2) - & 2.8071e-1 (1.35e-2) = & {\color[HTML]{3333E9} 2.9164e-1 (2.42e-4)} \\
\hline
\end{tabular}%
}
\end{table}

\begin{table}[H]
\centering
\caption{The results of mean and std IGD metric obtained by twelve algorithms on the LIRCMOP, DOC and FCP test suite.}
\label{tab:LIRDOCFCPIGD}
\resizebox{\textwidth}{!}{%
\begin{tabular}{ccccccccccccc}
\hline
\textbf{Problem} & \textbf{CCMO} & \textbf{CAEAD} & \textbf{cDPEA} & \textbf{MSCMO} & \textbf{CMOEAMS} & \textbf{BiCo} & \textbf{CMEGL} & \textbf{C3M} & \textbf{URCMO} & \textbf{CMOES} & \textbf{CMOEMT} & \textbf{LLM4CMO} \\ \hline
\textbf{LIRCMOP1} & 1.9818e-1 (4.99e-2) - & 1.5111e-2 (7.16e-3) - & 1.4339e-1 (5.35e-2) - & 1.4603e-1 (6.28e-2) - & 3.1262e-1 (4.20e-2) - & 1.7179e-1 (1.69e-2) - & 7.7264e-2 (2.77e-2) - & 8.5607e-2 (5.18e-2) - & 6.4410e-2 (3.19e-2) - & 1.2496e-1 (3.56e-2) - & 1.2294e-2 (7.15e-3) - & {\color[HTML]{3333E9} 8.3722e-3 (1.07e-3)} \\
\textbf{LIRCMOP2} & 1.6274e-1 (4.00e-2) - & 6.4367e-3 (1.57e-3) - & 1.3112e-1 (3.61e-2) - & 1.5685e-1 (4.06e-2) - & 2.6841e-1 (3.93e-2) - & 1.3734e-1 (1.56e-2) - & 7.4232e-2 (1.76e-2) - & 3.0379e-2 (4.16e-2) - & 1.1367e-2 (2.37e-2) - & 1.0894e-1 (1.88e-2) - & 1.3597e-2 (2.11e-2) - & {\color[HTML]{3333E9} 4.7593e-3 (1.76e-4)} \\
\textbf{LIRCMOP3} & 1.8250e-1 (5.15e-2) - & 4.7566e-3 (3.13e-3) - & 1.4919e-1 (6.21e-2) - & 1.4920e-1 (5.55e-2) - & 3.2096e-1 (5.39e-2) - & 1.8616e-1 (2.63e-2) - & 1.5437e-1 (3.74e-2) - & 1.0117e-1 (7.81e-2) - & 3.5965e-2 (4.83e-2) = & 1.0396e-1 (3.32e-2) - & 1.4976e-2 (3.32e-2) - & {\color[HTML]{3333E9} 2.6231e-3 (7.87e-4)} \\
\textbf{LIRCMOP4} & 2.2974e-1 (5.21e-2) - & 8.0458e-3 (7.20e-3) - & 1.6754e-1 (4.64e-2) - & 1.6962e-1 (5.20e-2) - & 2.9286e-1 (3.26e-2) - & 1.9082e-1 (2.99e-2) - & 1.7580e-1 (2.90e-2) - & 1.1800e-1 (8.64e-2) - & 7.1155e-2 (8.05e-2) - & 1.2870e-1 (2.64e-2) - & 1.6894e-2 (4.09e-2) - & {\color[HTML]{3333E9} 2.2285e-3 (3.60e-4)} \\
\textbf{LIRCMOP5} & 2.7186e-1 (4.61e-2) - & 1.3783e-2 (5.38e-3) - & 2.9626e-1 (5.55e-2) - & 4.8278e-1 (4.12e-1) - & 2.9156e-1 (5.71e-2) - & 1.2194e+0 (3.99e-3) - & 2.7543e-1 (6.07e-2) - & 4.7838e-2 (2.08e-1) - & 8.0848e-3 (9.69e-4) - & 2.6931e-1 (4.30e-2) - & 2.9453e-2 (2.81e-2) - & {\color[HTML]{3333E9} 6.0089e-3 (4.86e-4)} \\
\hline
\end{tabular}%
}
\end{table}

\subsection{The result of HV metrics that test 12 algorithm on 11 real-world CMOPs.} \label{sec:A4-4}
\begin{table}[H]
    \centering
    \caption{The results of mean and std HV metric obtained by twelve algorithm on the ten real-world test instances.}
    \label{tab:rwhv}
    \resizebox{\textwidth}{!}{%
    \begin{tabular}{ccccccccccccc}
    \hline
    \textbf{Problem} & \textbf{CCMO} & \textbf{CAEAD} & \textbf{cDPEA} & \textbf{MSCMO} & \textbf{CMOEAMS} & \textbf{BiCo} & \textbf{CMEGL} & \textbf{C3M} & \textbf{URCMO} & \textbf{CMOES} & \textbf{CMOEMT} & \textbf{LLM4CMO} \\ \hline
    \textbf{PVD} & 6.0734e-1 (4.52e-4) - & 6.0817e-1 (3.95e-4) + & 6.0738e-1 (3.89e-4) - & 6.0714e-1 (5.21e-4) - & {\color[HTML]{3333E9} 9.9823e-1 (3.01e-3) +} & 6.0920e-1 (1.72e-4) + & 6.0734e-1 (4.93e-4) - & 6.0824e-1 (2.96e-4) + & 6.0800e-1 (3.69e-4) + & 6.0705e-1 (4.19e-4) - & 6.0719e-1 (5.00e-4) - & 6.0768e-1 (3.95e-4) \\
    \textbf{VPD} & 3.7192e-1 (6.71e-2) = & 3.9288e-1 (2.26e-5) + & 3.6997e-1 (6.12e-2) - & 3.9290e-1 (4.91e-5) + & 3.7598e-1 (5.86e-2) = & 3.9074e-1 (9.48e-3) - & 3.9291e-1 (6.91e-5) + & 3.9289e-1 (1.57e-5) + & 3.9288e-1 (2.23e-5) + & 3.8803e-1 (2.47e-2) = & {\color[HTML]{3333E9} 3.9292e-1 (2.17e-5) +} & 3.9279e-1 (2.28e-5) \\
    \textbf{TBTD} & 8.9931e-1 (3.78e-4) - & 8.9948e-1 (3.98e-4) - & 8.9924e-1 (5.34e-4) - & 8.9918e-1 (5.52e-4) - & 8.6404e-1 (6.97e-3) - & 8.9993e-1 (4.97e-4) - & 8.9950e-1 (3.77e-4) - & 8.9945e-1 (4.67e-4) - & 8.9920e-1 (7.58e-4) - & 8.9899e-1 (5.96e-4) - & 8.9803e-1 (1.05e-3) - & {\color[HTML]{3333E9} 9.6642e-1 (4.55e-2)} \\
    \textbf{GBD} & 8.7634e-2 (9.84e-5) - & 8.7185e-2 (2.10e-4) - & 8.7513e-2 (1.48e-4) - & 8.7561e-2 (1.64e-4) - & 7.6872e-2 (1.05e-2) - & 8.8007e-2 (1.09e-4) - & 8.7640e-2 (1.36e-4) - & 8.7215e-2 (2.39e-4) - & 8.7392e-2 (1.61e-4) - & 8.7465e-2 (1.65e-4) - & 8.7020e-2 (1.88e-4) - & {\color[HTML]{3333E9} 9.1692e-2 (8.39e-4)} \\
    \textbf{SOPM3-level} & 5.8866e-1 (1.13e-1) - & 7.5975e-1 (6.46e-2) = & 5.9154e-1 (1.56e-1) - & NaN (NaN) & NaN (NaN) & 5.9545e-1 (9.42e-2) - & 6.2336e-1 (8.47e-2) - & 6.5888e-1 (1.11e-1) - & {\color[HTML]{3333E9} 7.8399e-1 (9.24e-2) =} & 6.5938e-1 (7.51e-2) - & 6.9420e-1 (1.06e-1) = & 7.4883e-1 (9.26e-2) \\
\hline
    \end{tabular}%
    }
\end{table}

\subsection{The result of HV and IGD metrics about LLM4CMO on four different $N_s$.} \label{sec:A4-5}
$N_s$ is the final determined parameter, which is used to minimize the resource consumption of popAux and reduce runtimes without affecting the performance of popAux. Therefore, we did not test parameters with $N_s < 30$ and $N_s > 60$.
\begin{table}[H]
\centering
\caption{The result of mean and std HV metric obtained by LLM4CMO with different $N_s$ on six test suite.}
\label{tab: NSHV}
\resizebox{0.85\textwidth}{!}{%
\begin{tabular}{ccccc}
\hline
\textbf{Problem} & \textbf{NS=30} & \textbf{NS=40} & \textbf{NS=50} & \textbf{NS=60} \\ \hline
\textbf{CF1} & 5.6344e-1 (6.95e-4) = & 5.6357e-1 (6.44e-4) = & 5.6359e-1 (6.29e-4) = & {\color[HTML]{3333E9} 5.6361e-1 (5.27e-4)} \\
\textbf{CF2} & {\color[HTML]{3333E9} 6.7717e-1 (9.23e-4) =} & 6.7699e-1 (1.48e-3) = & 6.7668e-1 (1.98e-3) = & 6.7679e-1 (1.50e-3) \\
\textbf{CF3} & 2.5326e-1 (4.28e-2) = & 2.5405e-1 (3.44e-2) = & 2.3727e-1 (4.44e-2) = & {\color[HTML]{3333E9} 2.5414e-1 (2.87e-2)} \\
\textbf{CF4} & 5.1188e-1 (1.00e-2) = & {\color[HTML]{3333E9} 5.1364e-1 (7.07e-3) =} & 5.0906e-1 (1.29e-2) = & 5.0771e-1 (1.49e-2) \\
\textbf{CF5} & 4.0094e-1 (5.27e-2) + & 3.9535e-1 (6.10e-2) = & {\color[HTML]{3333E9} 4.0347e-1 (4.71e-2) +} & 3.6171e-1 (6.35e-2) \\
\textbf{CF6} & 6.7604e-1 (5.40e-3) = & {\color[HTML]{3333E9} 6.7913e-1 (4.95e-3) =} & 6.7810e-1 (5.59e-3) = & 6.7887e-1 (5.44e-3) \\
\textbf{CF7} & 5.9232e-1 (2.94e-2) = & {\color[HTML]{3333E9} 6.0011e-1 (3.04e-2) =} & 5.9150e-1 (3.87e-2) = & 5.8722e-1 (4.93e-2) \\
\hline
\end{tabular}%
}
\end{table}

\begin{table}[htpb!]
\centering
\caption{The result of mean and std IGD metric obtained by LLM4CMO with different $N_s$ on six test suite.}
\label{tab:NSIGD}
\resizebox{0.85\textwidth}{!}{%
\begin{tabular}{ccccc}
\hline
\textbf{Problem} & \textbf{Ns=30} & \textbf{Ns=40} & \textbf{Ns=50} & \textbf{Ns=60} \\ \hline
\textbf{CF1} & 2.1817e-3 (5.69e-4) = & 2.0762e-3 (5.21e-4) = & 2.0626e-3 (5.11e-4) = & {\color[HTML]{3333E9} 2.0446e-3 (4.28e-4)} \\
\textbf{CF2} & {\color[HTML]{3333E9} 3.4413e-3 (4.48e-4) +} & 3.4599e-3 (6.36e-4) + & 3.7326e-3 (6.95e-4) = & 4.0362e-3 (1.68e-3) \\
\textbf{CF3} & {\color[HTML]{3333E9} 1.3244e-1 (6.93e-2) =} & 1.4396e-1 (6.07e-2) = & 1.3991e-1 (6.86e-2) = & 1.3976e-1 (4.79e-2) \\
\textbf{CF4} & 1.9947e-2 (8.62e-3) = & {\color[HTML]{3333E9} 1.7778e-2 (6.34e-3) =} & 2.1906e-2 (1.14e-2) = & 2.2964e-2 (1.22e-2) \\
\textbf{CF5} & 1.5333e-1 (8.31e-2) + & 1.5043e-1 (8.61e-2) + & {\color[HTML]{3333E9} 1.4107e-1 (7.23e-2) =} & 2.0040e-1 (1.03e-1) \\
\textbf{CF6} & 3.2800e-2 (1.16e-2) - & {\color[HTML]{3333E9} 2.6238e-2 (7.44e-3) =} & 2.7071e-2 (1.16e-2) = & 2.6276e-2 (7.37e-3) \\
\textbf{CF7} & 9.2064e-2 (3.41e-2) = & {\color[HTML]{3333E9} 8.4939e-2 (3.05e-2) =} & 8.9079e-2 (3.98e-2) = & 9.4415e-2 (4.53e-2) \\
\hline
\end{tabular}%
}
\end{table}

\clearpage
\subsection{The ablation study results of HV and IGD about LLM4CMO basic algorithm framework.} \label{sec:A4-6}
LLM4CMOWoRR removes the relaxation module; LLM4CMOWoS1C presents popAux without $Off_2$ in Stage 1. LLM4CMOWoOP present removes the opposite mechanism. LLM4CMOWo3P presents remove the third phase of epsilon-control.
\begin{table}[htpb]
\centering
\centering
\caption{Ablation study results of HV values obtained by the LLM4CMO basic algorithm framework on 6 test suites. RR presents relax $R_s$, S1C presents popAux without $Off_2$ in stage 1, OP presents opposite population mechanism, and 3P presents epsilon control 3 phase in stage 2.}
\label{tab:AB1HV}
\resizebox{0.9\textwidth}{!}{%
\begin{tabular}{cccccc}
\hline
\textbf{Problem} & \textbf{LLM4CMOWoRR} & \textbf{LLM4CMOWoS1C} & \textbf{LLM4CMOwoOP} & \textbf{LLM4CMOWo3P} & \textbf{LLM4CMO} \\ \hline
\textbf{CF1} & 5.6340e-1 (4.06e-4) = & 5.6363e-1 (4.10e-4) = & 5.6362e-1 (4.65e-4) = & {\color[HTML]{3333E9} 5.6370e-1 (4.98e-4) =} & 5.6357e-1 (6.44e-4) \\
\textbf{CF2} & 6.6073e-1 (2.21e-2) - & 6.7703e-1 (8.97e-4) = & 6.7716e-1 (9.51e-4) = & {\color[HTML]{3333E9} 6.7729e-1 (1.00e-3) =} & 6.7699e-1 (1.48e-3) \\
\textbf{CF3} & 2.2517e-1 (4.92e-2) - & 2.4850e-1 (4.31e-2) = & 2.4995e-1 (4.21e-2) = & 2.4947e-1 (3.98e-2) = & {\color[HTML]{3333E9} 2.5405e-1 (3.44e-2)} \\
\textbf{CF4} & 4.8487e-1 (2.96e-2) - & 5.1164e-1 (8.24e-3) = & 5.0814e-1 (1.22e-2) = & 5.1057e-1 (1.05e-2) = & {\color[HTML]{3333E9} 5.1364e-1 (7.07e-3)} \\
\textbf{CF5} & 3.7507e-1 (7.05e-2) = & 3.8830e-1 (5.85e-2) = & {\color[HTML]{3333E9} 3.9861e-1 (4.88e-2) =} & 3.8784e-1 (6.21e-2) = & 3.9535e-1 (6.10e-2) \\
\textbf{CF6} & 6.7603e-1 (5.88e-3) - & 6.7608e-1 (5.26e-3) - & 6.7775e-1 (4.79e-3) = & 6.7743e-1 (6.32e-3) = & {\color[HTML]{3333E9} 6.7913e-1 (4.95e-3)} \\
\textbf{CF7} & 5.8036e-1 (4.43e-2) = & 5.8533e-1 (3.50e-2) = & 5.9246e-1 (3.39e-2) = & {\color[HTML]{3333E9} 6.0278e-1 (2.26e-2) =} & 6.0011e-1 (3.04e-2) \\
\textbf{CF8} & {\color[HTML]{3333E9} 3.6363e-1 (7.08e-2) +} & 2.9276e-1 (8.63e-2) = & 2.6542e-1 (9.72e-2) = & 2.8807e-1 (8.32e-2) = & 2.8384e-1 (9.31e-2) \\
\textbf{CF9} & 3.2074e-1 (1.81e-2) - & 3.4299e-1 (2.14e-2) - & 4.4702e-1 (2.97e-2) = & 4.4860e-1 (2.26e-2) = & {\color[HTML]{3333E9} 4.5111e-1 (1.96e-2)} \\
\textbf{CF10} & 1.7702e-1 (4.71e-2) = & 1.8571e-1 (4.05e-2) = & {\color[HTML]{3333E9} 1.9028e-1 (5.29e-2) =} & 1.8509e-1 (4.49e-2) = & 1.7561e-1 (3.15e-2) \\
\textbf{DASCMOP1} & 2.1253e-1 (3.90e-4) = & 2.1251e-1 (2.67e-4) = & 2.1241e-1 (4.17e-4) = & 2.1208e-1 (3.64e-4) - & {\color[HTML]{3333E9} 2.1254e-1 (4.07e-4)} \\
\textbf{DASCMOP2} & 3.5511e-1 (1.10e-4) = & 3.5516e-1 (7.97e-5) = & {\color[HTML]{3333E9} 3.5517e-1 (7.06e-5) =} & 3.5497e-1 (9.00e-5) - & 3.5513e-1 (9.94e-5) \\
\textbf{DASCMOP3} & 3.0924e-1 (1.65e-2) = & {\color[HTML]{3333E9} 3.1233e-1 (1.11e-4) =} & 3.1227e-1 (1.37e-4) = & 3.1188e-1 (4.31e-4) - & 3.1230e-1 (1.97e-4) \\
\textbf{DASCMOP4} & 2.0196e-1 (3.61e-3) - & 2.0372e-1 (4.16e-4) - & {\color[HTML]{3333E9} 2.0408e-1 (2.88e-4) =} & 2.0394e-1 (4.68e-4) = & 2.0396e-1 (4.43e-4) \\
\textbf{DASCMOP5} & 3.5126e-1 (2.03e-4) - & 3.5097e-1 (3.97e-4) - & 3.5128e-1 (3.69e-4) = & 3.5131e-1 (3.95e-4) = & {\color[HTML]{3333E9} 3.5132e-1 (3.81e-4)} \\
\textbf{DASCMOP6} & 3.1015e-1 (5.03e-3) - & 3.1244e-1 (1.03e-4) = & 3.0252e-1 (3.65e-2) = & 3.1232e-1 (3.39e-4) = & {\color[HTML]{3333E9} 3.1245e-1 (1.91e-4)} \\
\textbf{DASCMOP7} & 2.8764e-1 (6.58e-4) - & 2.8575e-1 (1.84e-3) - & 2.8792e-1 (1.32e-3) = & 2.8754e-1 (1.82e-3) = & {\color[HTML]{3333E9} 2.8793e-1 (1.30e-3)} \\
\textbf{DASCMOP8} & {\color[HTML]{3333E9} 2.0662e-1 (5.56e-4) =} & 2.0527e-1 (1.51e-3) - & 2.0588e-1 (2.46e-3) = & 2.0611e-1 (2.42e-3) = & 2.0621e-1 (1.62e-3) \\
\textbf{DASCMOP9} & 1.7438e-1 (3.61e-2) - & 2.0463e-1 (3.93e-4) - & 2.0469e-1 (3.76e-4) - & 2.0360e-1 (5.46e-4) - & {\color[HTML]{3333E9} 2.0492e-1 (2.73e-4)} \\
\textbf{MW1} & 4.8992e-1 (6.69e-4) - & {\color[HTML]{3333E9} 4.9010e-1 (1.42e-5) =} & 4.8992e-1 (4.56e-4) = & 4.9002e-1 (2.96e-4) = & 4.9010e-1 (1.73e-5) \\
\textbf{MW2} & 5.8239e-1 (3.35e-5) = & {\color[HTML]{3333E9} 5.8240e-1 (3.42e-5) =} & 5.8240e-1 (3.21e-5) = & 5.8240e-1 (3.80e-5) = & 5.8239e-1 (3.88e-5) \\
\textbf{MW3} & 5.4445e-1 (5.04e-4) = & 5.4447e-1 (5.12e-4) = & 5.4438e-1 (4.35e-4) - & 5.4445e-1 (4.94e-4) = & {\color[HTML]{3333E9} 5.4464e-1 (2.78e-4)} \\
\textbf{MW4} & 8.4071e-1 (5.53e-4) - & 8.4146e-1 (5.01e-4) = & {\color[HTML]{3333E9} 8.4147e-1 (5.91e-4) =} & 8.4128e-1 (5.53e-4) = & 8.4143e-1 (3.62e-4) \\
\textbf{MW5} & 3.2446e-1 (2.67e-4) = & {\color[HTML]{3333E9} 3.2450e-1 (1.06e-4) =} & 3.2442e-1 (1.82e-4) = & 3.2426e-1 (6.02e-4) + & 3.2423e-1 (9.33e-4) \\
\textbf{MW6} & 3.2849e-1 (2.25e-5) = & 3.2850e-1 (2.10e-5) = & {\color[HTML]{3333E9} 3.2850e-1 (1.63e-5) =} & 3.2849e-1 (2.00e-5) = & 3.2850e-1 (1.56e-5) \\
\textbf{MW7} & {\color[HTML]{3333E9} 4.1217e-1 (3.86e-4) +} & 4.1199e-1 (4.09e-4) = & 4.1172e-1 (8.39e-4) = & 4.1167e-1 (6.85e-4) = & 4.1180e-1 (5.40e-4) \\
\textbf{MW8} & 5.4340e-1 (6.48e-3) - & {\color[HTML]{3333E9} 5.5324e-1 (9.58e-4) =} & 5.5297e-1 (1.87e-3) = & 5.5239e-1 (2.64e-3) = & 5.5314e-1 (1.08e-3) \\
\textbf{MW9} & 3.9522e-1 (2.94e-3) + & {\color[HTML]{3333E9} 3.9592e-1 (2.20e-3) +} & 3.9431e-1 (3.08e-3) = & 3.9346e-1 (2.84e-3) = & 3.9420e-1 (2.52e-3) \\
\textbf{MW10} & 4.5396e-1 (2.85e-3) = & 4.5297e-1 (6.30e-3) = & 4.5474e-1 (1.32e-4) = & {\color[HTML]{3333E9} 4.5476e-1 (1.19e-4) =} & 4.5441e-1 (2.03e-3) \\
\textbf{MW11} & 4.4740e-1 (1.77e-4) = & {\color[HTML]{3333E9} 4.4741e-1 (1.52e-4) =} & 4.4732e-1 (2.20e-4) = & 4.4710e-1 (2.19e-4) - & 4.4733e-1 (2.64e-4) \\
\textbf{MW12} & 6.0380e-1 (2.66e-4) = & 6.0384e-1 (2.21e-4) + & 6.0376e-1 (3.44e-4) = & {\color[HTML]{3333E9} 6.0386e-1 (4.10e-4) +} & 6.0368e-1 (2.55e-4) \\
\textbf{MW13} & 4.7640e-1 (2.16e-3) = & 4.7642e-1 (2.18e-3) = & 4.7624e-1 (2.15e-3) = & 4.7585e-1 (2.99e-3) = & {\color[HTML]{3333E9} 4.7680e-1 (2.09e-4)} \\
\textbf{MW14} & 4.7002e-1 (2.94e-3) - & {\color[HTML]{3333E9} 4.7544e-1 (1.72e-3) +} & 4.7395e-1 (2.53e-3) = & 4.7426e-1 (1.39e-3) = & 4.7404e-1 (1.84e-3) \\
\textbf{LIRCMOP1} & 2.3662e-1 (4.13e-4) - & 2.3687e-1 (4.39e-4) = & 2.3683e-1 (4.49e-4) = & 2.3532e-1 (5.00e-4) - & {\color[HTML]{3333E9} 2.3694e-1 (7.05e-4)} \\
\textbf{LIRCMOP2} & 3.5970e-1 (5.33e-4) - & {\color[HTML]{3333E9} 3.6049e-1 (1.35e-4) =} & 3.6040e-1 (2.33e-4) = & 3.5829e-1 (6.98e-4) - & 3.6045e-1 (1.62e-4) \\
\textbf{LIRCMOP3} & 2.0739e-1 (5.69e-4) - & 2.0844e-1 (3.77e-4) = & 2.0845e-1 (4.82e-4) = & 2.0681e-1 (4.82e-4) - & {\color[HTML]{3333E9} 2.0853e-1 (4.39e-4)} \\
\textbf{LIRCMOP4} & 3.1596e-1 (7.74e-4) - & 3.1733e-1 (6.90e-4) = & 3.1732e-1 (6.40e-4) = & 3.1501e-1 (5.94e-4) - & {\color[HTML]{3333E9} 3.1742e-1 (4.71e-4)} \\
\textbf{LIRCMOP5} & 2.9138e-1 (2.58e-4) - & 2.9158e-1 (2.17e-4) = & 2.9159e-1 (3.37e-4) = & 2.9150e-1 (2.64e-4) - & {\color[HTML]{3333E9} 2.9164e-1 (2.42e-4)} \\
\textbf{LIRCMOP6} & 1.9681e-1 (2.21e-4) - & {\color[HTML]{3333E9} 1.9702e-1 (1.59e-4) =} & 1.9693e-1 (1.77e-4) = & 1.9672e-1 (3.30e-4) - & 1.9695e-1 (1.61e-4) \\
\textbf{LIRCMOP7} & 2.9034e-1 (1.06e-2) - & 2.9240e-1 (7.85e-3) = & 2.9044e-1 (9.66e-3) = & 2.6608e-1 (2.50e-2) - & {\color[HTML]{3333E9} 2.9380e-1 (1.76e-3)} \\
\textbf{LIRCMOP8} & 2.8825e-1 (1.59e-2) - & 2.9434e-1 (3.70e-4) - & 2.9454e-1 (3.76e-4) - & 2.7963e-1 (2.41e-2) - & {\color[HTML]{3333E9} 2.9476e-1 (2.28e-4)} \\
\textbf{LIRCMOP9} & 4.7358e-1 (8.20e-2) - & {\color[HTML]{3333E9} 5.4449e-1 (3.61e-2) =} & 5.2541e-1 (6.34e-2) = & 5.2201e-1 (5.32e-2) = & 5.3447e-1 (4.58e-2) \\
\textbf{LIRCMOP10} & 6.9628e-1 (2.93e-2) - & 7.0253e-1 (1.85e-2) = & 7.0559e-1 (6.22e-3) = & 7.0674e-1 (4.60e-4) = & {\color[HTML]{3333E9} 7.0679e-1 (5.08e-4)} \\
\textbf{LIRCMOP11} & 6.8149e-1 (2.38e-2) - & 6.9170e-1 (8.54e-3) = & 6.9060e-1 (1.03e-2) = & 6.7184e-1 (4.39e-2) - & {\color[HTML]{3333E9} 6.9284e-1 (6.15e-3)} \\
\textbf{LIRCMOP12} & 6.0792e-1 (1.95e-2) - & {\color[HTML]{3333E9} 6.1783e-1 (8.24e-3) =} & 6.1536e-1 (1.27e-2) = & 5.9912e-1 (3.61e-2) - & 6.1281e-1 (1.45e-2) \\
\textbf{LIRCMOP13} & {\color[HTML]{3333E9} 5.5371e-1 (9.46e-4) +} & 5.4694e-1 (2.34e-3) + & 5.4437e-1 (4.38e-3) = & 5.4625e-1 (4.32e-3) + & 5.4405e-1 (3.81e-3) \\
\textbf{LIRCMOP14} & {\color[HTML]{3333E9} 5.5262e-1 (1.07e-3) +} & 5.5031e-1 (1.87e-3) - & 5.5089e-1 (1.31e-3) = & 5.5030e-1 (1.76e-3) - & 5.5142e-1 (1.30e-3) \\
\textbf{DOC1} & {\color[HTML]{3333E9} 3.4455e-1 (8.15e-4) =} & 3.1098e-1 (8.78e-2) = & 3.3618e-1 (4.63e-2) = & 3.3582e-1 (4.63e-2) = & 3.1094e-1 (8.78e-2) \\
\textbf{DOC2} & 6.2102e-1 (1.96e-3) = & 6.2085e-1 (2.78e-3) = & {\color[HTML]{3333E9} 6.2169e-1 (1.72e-3) =} & 6.2014e-1 (2.73e-3) = & 6.2066e-1 (2.84e-3) \\
\textbf{DOC3} & 2.9448e-1 (1.00e-1) = & 3.1780e-1 (6.01e-2) = & {\color[HTML]{3333E9} 3.2866e-1 (3.63e-3) =} & 2.9730e-1 (8.91e-2) - & 3.1736e-1 (6.02e-2) \\
\textbf{DOC4} & 5.4226e-1 (1.43e-2) = & 5.4659e-1 (1.19e-2) = & {\color[HTML]{3333E9} 5.4790e-1 (7.77e-3) =} & 5.3511e-1 (1.45e-2) - & 5.4603e-1 (1.09e-2) \\
\textbf{DOC5} & {\color[HTML]{3333E9} 3.4031e-1 (2.32e-1) =} & 2.6435e-1 (2.51e-1) = & 2.7437e-1 (2.46e-1) = & 2.7051e-1 (2.48e-1) = & 3.0943e-1 (2.42e-1) \\
\textbf{DOC6} & 0.0000e+0 (0.00e+0) - & 0.0000e+0 (0.00e+0) - & {\color[HTML]{3333E9} 5.5587e-1 (5.24e-3) =} & 5.2499e-1 (1.03e-1) = & 4.6415e-1 (2.05e-1) \\
\textbf{DOC7} & 7.5434e-2 (1.91e-1) = & 1.0845e-1 (2.21e-1) = & 1.0269e-1 (2.10e-1) = & {\color[HTML]{3333E9} 1.0934e-1 (2.22e-1) =} & 9.9085e-2 (2.11e-1) \\
\textbf{DOC8} & 2.1490e-2 (1.18e-1) - & 7.0669e-1 (3.36e-2) = & {\color[HTML]{3333E9} 7.1843e-1 (2.72e-2) =} & 5.4837e-1 (1.99e-1) - & 7.1633e-1 (2.76e-2) \\
\textbf{DOC9} & {\color[HTML]{3333E9} 0.0000e+0 (0.00e+0) =} & 0.0000e+0 (0.00e+0) = & 0.0000e+0 (0.00e+0) = & 0.0000e+0 (0.00e+0) = & 0.0000e+0 (0.00e+0) \\
\textbf{FCP1} & 5.8171e-1 (1.19e-4) = & 5.8176e-1 (1.39e-4) = & NaN (NaN) & {\color[HTML]{3333E9} 5.8224e-1 (4.53e-5) +} & 5.8176e-1 (1.15e-4) \\
\textbf{FCP2} & 4.3148e-1 (8.94e-5) - & 4.3153e-1 (7.15e-5) = & NaN (NaN) & {\color[HTML]{3333E9} 4.3157e-1 (7.01e-5) =} & 4.3154e-1 (7.91e-5) \\
\textbf{FCP3} & 3.4717e-1 (1.21e-4) = & 3.4727e-1 (6.79e-5) = & NaN (NaN) & {\color[HTML]{3333E9} 3.4731e-1 (7.05e-5) +} & 3.4723e-1 (8.60e-5) \\
\textbf{FCP4} & 6.3462e-1 (6.43e-5) - & 6.3467e-1 (4.00e-5) = & NaN (NaN) & {\color[HTML]{3333E9} 6.3469e-1 (3.65e-5) +} & 6.3467e-1 (3.52e-5) \\
\textbf{FCP5} & 4.7959e-1 (2.20e-4) = & 4.7950e-1 (3.40e-4) = & 2.6215e-1 (5.22e-5) - & 4.7951e-1 (2.48e-4) = & {\color[HTML]{3333E9} 4.7961e-1 (2.13e-4)} \\ \hline
\textbf{+/-/=} & 5/30/26 & 4/10/47 & 0/8/53 & 6/19/36 &  \\ \hline
\end{tabular}%
}
\end{table}

\begin{table}[htpb!]
\centering
\caption{Ablation study results of IGD values obtained by LLM4CMO basic algorithm framework on 6 test suite. RR presents relax $R_s$, S1C presents popAux without $Off_2$ in stage 1, OP presents opposite population mechanism and 3P presents epsilon constrol 3 phase in stage 2.}
\label{tab:AB1IGD}
\resizebox{0.9\textwidth}{!}{%
\begin{tabular}{cccccc}
\hline
\textbf{Problem} & \textbf{LLM4CMOWoRR} & \textbf{LLM4CMOWoS1C} & \textbf{LLM4CMOwoOP} & \textbf{LLM4CMOWo3P} & \textbf{LLM4CMO} \\ \hline
\textbf{CF1} & 2.2156e-3 (3.27e-4) = & 2.0250e-3 (3.33e-4) = & 2.0351e-3 (3.84e-4) = & {\color[HTML]{3333E9} 1.9690e-3 (4.05e-4) =} & 2.0762e-3 (5.21e-4) \\
\textbf{CF2} & 1.6843e-2 (1.65e-2) - & {\color[HTML]{3333E9} 3.2922e-3 (2.93e-4) =} & 3.4157e-3 (4.50e-4) = & 3.6472e-3 (9.59e-4) = & 3.4599e-3 (6.36e-4) \\
\textbf{CF3} & 1.9427e-1 (8.04e-2) - & 1.3663e-1 (6.66e-2) = & 1.3630e-1 (6.44e-2) = & {\color[HTML]{3333E9} 1.2345e-1 (6.40e-2) =} & 1.4396e-1 (6.07e-2) \\
\textbf{CF4} & 3.9764e-2 (2.33e-2) - & 2.0666e-2 (7.07e-3) = & 2.3004e-2 (1.03e-2) - & 2.0814e-2 (8.66e-3) = & {\color[HTML]{3333E9} 1.7778e-2 (6.34e-3)} \\
\textbf{CF5} & 1.6901e-1 (9.35e-2) = & 1.5442e-1 (7.99e-2) = & 1.5162e-1 (7.29e-2) = & 1.5947e-1 (8.53e-2) = & {\color[HTML]{3333E9} 1.5043e-1 (8.61e-2)} \\
\textbf{CF6} & 2.9642e-2 (8.37e-3) = & 3.1403e-2 (1.26e-2) = & 2.8557e-2 (9.12e-3) = & 2.8971e-2 (1.18e-2) = & {\color[HTML]{3333E9} 2.6238e-2 (7.44e-3)} \\
\textbf{CF7} & 1.0586e-1 (5.87e-2) = & 1.1310e-1 (5.13e-2) - & 9.9724e-2 (5.85e-2) = & 8.5539e-2 (2.90e-2) = & {\color[HTML]{3333E9} 8.4939e-2 (3.05e-2)} \\
\textbf{CF8} & {\color[HTML]{3333E9} 1.6933e-1 (8.93e-2) +} & 2.2571e-1 (9.55e-2) = & 2.5698e-1 (1.07e-1) = & 2.2382e-1 (8.71e-2) = & 2.3201e-1 (8.92e-2) \\
\textbf{CF9} & 1.4061e-1 (1.92e-2) - & 1.2753e-1 (1.63e-2) - & 9.1231e-2 (2.63e-2) = & 9.0359e-2 (1.69e-2) = & {\color[HTML]{3333E9} 8.8228e-2 (1.64e-2)} \\
\textbf{CF10} & 3.1762e-1 (8.69e-2) = & 3.2607e-1 (5.63e-2) = & {\color[HTML]{3333E9} 3.1208e-1 (5.82e-2) =} & 3.1638e-1 (6.60e-2) = & 3.2466e-1 (6.34e-2) \\
\textbf{DASCMOP1} & 3.0520e-3 (2.50e-4) - & 2.9671e-3 (2.55e-4) = & 3.0225e-3 (1.91e-4) = & 3.3867e-3 (3.40e-4) - & {\color[HTML]{3333E9} 2.9463e-3 (2.21e-4)} \\
\textbf{DASCMOP2} & 4.4877e-3 (1.15e-4) = & 4.4239e-3 (1.08e-4) = & {\color[HTML]{3333E9} 4.4174e-3 (9.29e-5) =} & 4.6167e-3 (1.13e-4) - & 4.4413e-3 (1.24e-4) \\
\textbf{DASCMOP3} & 2.6988e-2 (5.46e-2) = & 1.7495e-2 (3.16e-3) = & 1.8619e-2 (2.98e-3) = & {\color[HTML]{3333E9} 1.6622e-2 (3.31e-3) =} & 1.7329e-2 (3.44e-3) \\
\textbf{DASCMOP4} & 2.3534e-3 (1.82e-3) - & 1.3804e-3 (1.26e-4) - & {\color[HTML]{3333E9} 1.2388e-3 (1.33e-4) =} & 1.3115e-3 (2.33e-4) = & 1.3109e-3 (3.27e-4) \\
\textbf{DASCMOP5} & {\color[HTML]{3333E9} 2.8717e-3 (1.36e-4) =} & 3.2160e-3 (3.17e-4) - & 3.0255e-3 (2.51e-4) = & 2.9999e-3 (2.77e-4) = & 2.9791e-3 (2.28e-4) \\
\textbf{DASCMOP6} & 2.4181e-2 (1.52e-2) = & 1.7847e-2 (2.96e-3) = & 3.7074e-2 (7.08e-2) = & {\color[HTML]{3333E9} 1.7694e-2 (3.04e-3) =} & 1.8202e-2 (2.84e-3) \\
\textbf{DASCMOP7} & 3.1329e-2 (8.58e-4) = & 3.2991e-2 (1.96e-3) - & {\color[HTML]{3333E9} 3.1028e-2 (1.18e-3) =} & 3.1827e-2 (1.48e-3) = & 3.1474e-2 (1.36e-3) \\
\textbf{DASCMOP8} & 4.0074e-2 (1.13e-3) = & {\color[HTML]{3333E9} 3.9501e-2 (1.58e-3) =} & 4.0150e-2 (2.46e-3) = & 4.0894e-2 (3.74e-3) = & 4.0000e-2 (1.39e-3) \\
\textbf{DASCMOP9} & 1.6756e-1 (1.58e-1) - & 4.0470e-2 (8.82e-4) = & {\color[HTML]{3333E9} 4.0176e-2 (9.35e-4) =} & 4.0819e-2 (1.00e-3) = & 4.0441e-2 (1.10e-3) \\
\textbf{MW1} & 1.6945e-3 (2.92e-4) = & {\color[HTML]{3333E9} 1.6283e-3 (1.02e-5) =} & 1.6962e-3 (1.81e-4) = & 1.6565e-3 (9.92e-5) = & 1.6293e-3 (1.31e-5) \\
\textbf{MW2} & 3.7370e-3 (2.11e-5) = & {\color[HTML]{3333E9} 3.7308e-3 (2.37e-5) =} & 3.7328e-3 (2.19e-5) = & 3.7325e-3 (2.51e-5) = & 3.7350e-3 (2.39e-5) \\
\textbf{MW3} & 4.8114e-3 (2.33e-4) = & 4.8409e-3 (2.56e-4) = & 4.8860e-3 (1.99e-4) - & 4.8212e-3 (2.30e-4) = & {\color[HTML]{3333E9} 4.7352e-3 (1.53e-4)} \\
\textbf{MW4} & 4.1514e-2 (3.19e-4) - & {\color[HTML]{3333E9} 4.1167e-2 (4.08e-4) =} & 4.1202e-2 (4.52e-4) = & 4.1360e-2 (4.68e-4) = & 4.1203e-2 (3.50e-4) \\
\textbf{MW5} & 5.2946e-4 (5.27e-4) = & {\color[HTML]{3333E9} 4.5995e-4 (3.39e-4) =} & 5.2604e-4 (2.42e-4) = & 7.5580e-4 (6.86e-4) - & 6.9019e-4 (9.26e-4) \\
\textbf{MW6} & 2.7523e-3 (1.81e-5) = & 2.7552e-3 (2.10e-5) = & 2.7582e-3 (2.20e-5) = & {\color[HTML]{3333E9} 2.7517e-3 (1.63e-5) =} & 2.7532e-3 (1.95e-5) \\
\textbf{MW7} & 4.4538e-3 (2.26e-4) = & {\color[HTML]{3333E9} 4.4344e-3 (3.35e-4) =} & 4.5605e-3 (3.50e-4) = & 4.5518e-3 (3.43e-4) = & 4.5298e-3 (3.80e-4) \\
\textbf{MW8} & 4.3932e-2 (1.50e-3) - & {\color[HTML]{3333E9} 4.2938e-2 (5.87e-4) =} & 4.3025e-2 (6.79e-4) = & 4.3144e-2 (6.61e-4) = & 4.3129e-2 (6.01e-4) \\
\textbf{MW9} & 5.1843e-3 (8.32e-4) + & {\color[HTML]{3333E9} 4.9949e-3 (2.68e-4) +} & 5.4750e-3 (1.38e-3) = & 5.6761e-3 (1.65e-3) = & 5.4156e-3 (5.73e-4) \\
\textbf{MW10} & 3.8505e-3 (1.40e-3) = & 4.7117e-3 (4.85e-3) = & {\color[HTML]{3333E9} 3.4780e-3 (4.77e-5) =} & 3.4901e-3 (6.03e-5) = & 3.6641e-3 (1.00e-3) \\
\textbf{MW11} & {\color[HTML]{3333E9} 6.2451e-3 (1.76e-4) =} & 6.2962e-3 (1.84e-4) = & 6.3412e-3 (1.45e-4) = & 6.3907e-3 (1.97e-4) = & 6.3137e-3 (1.71e-4) \\
\textbf{MW12} & 5.0870e-3 (1.37e-4) + & 5.1373e-3 (1.29e-4) = & 5.1092e-3 (1.60e-4) = & {\color[HTML]{3333E9} 5.0568e-3 (1.61e-4) +} & 5.1894e-3 (1.50e-4) \\
\textbf{MW13} & 1.1248e-2 (2.33e-3) = & 1.1224e-2 (2.38e-3) = & 1.1542e-2 (2.58e-3) = & 1.1870e-2 (3.30e-3) = & {\color[HTML]{3333E9} 1.0763e-2 (2.41e-4)} \\
\textbf{MW14} & 1.0100e-1 (2.77e-3) - & 9.7211e-2 (2.05e-3) = & 9.8110e-2 (2.19e-3) = & {\color[HTML]{3333E9} 9.7144e-2 (1.29e-3) =} & 9.7820e-2 (1.96e-3) \\
\textbf{LIRCMOP1} & 9.0804e-3 (1.53e-3) - & 8.7145e-3 (1.32e-3) = & 8.7622e-3 (1.56e-3) = & 1.1352e-2 (1.15e-3) - & {\color[HTML]{3333E9} 8.3722e-3 (1.07e-3)} \\
\textbf{LIRCMOP2} & 6.1772e-3 (9.86e-4) - & {\color[HTML]{3333E9} 4.7283e-3 (1.76e-4) =} & 4.8793e-3 (4.20e-4) = & 8.7830e-3 (1.31e-3) - & 4.7593e-3 (1.76e-4) \\
\textbf{LIRCMOP3} & 4.1306e-3 (9.03e-4) - & 2.6686e-3 (6.92e-4) = & 2.7103e-3 (8.73e-4) = & 5.1230e-3 (1.04e-3) - & {\color[HTML]{3333E9} 2.6231e-3 (7.87e-4)} \\
\textbf{LIRCMOP4} & 3.4938e-3 (8.20e-4) - & 2.2866e-3 (5.34e-4) = & 2.3043e-3 (5.06e-4) = & 4.9232e-3 (8.73e-4) - & {\color[HTML]{3333E9} 2.2285e-3 (3.60e-4)} \\
\textbf{LIRCMOP5} & 6.4282e-3 (5.06e-4) - & 6.1246e-3 (4.54e-4) = & 6.0670e-3 (5.60e-4) = & 6.2958e-3 (5.25e-4) - & {\color[HTML]{3333E9} 6.0089e-3 (4.86e-4)} \\
\textbf{LIRCMOP6} & 5.8731e-3 (4.66e-4) - & {\color[HTML]{3333E9} 5.5501e-3 (2.81e-4) =} & 5.6523e-3 (3.12e-4) = & 6.0250e-3 (4.50e-4) - & 5.6623e-3 (2.80e-4) \\
\textbf{LIRCMOP7} & 1.5734e-2 (2.73e-2) - & 1.1916e-2 (2.03e-2) = & 1.5639e-2 (2.21e-2) = & 7.8801e-2 (6.70e-2) - & {\color[HTML]{3333E9} 8.5201e-3 (4.72e-3)} \\
\textbf{LIRCMOP8} & 2.2882e-2 (4.30e-2) - & 7.4642e-3 (5.04e-4) - & 7.1730e-3 (4.01e-4) = & 4.8327e-2 (7.68e-2) - & {\color[HTML]{3333E9} 7.0061e-3 (2.11e-4)} \\
\textbf{LIRCMOP9} & 2.4422e-1 (2.01e-1) - & {\color[HTML]{3333E9} 8.0694e-2 (1.15e-1) =} & 1.2490e-1 (1.69e-1) = & 1.4616e-1 (1.53e-1) = & 1.1833e-1 (1.39e-1) \\
\textbf{LIRCMOP10} & 2.7173e-2 (6.36e-2) - & 1.5274e-2 (4.13e-2) = & 8.6127e-3 (1.71e-2) = & 5.3120e-3 (5.34e-4) = & {\color[HTML]{3333E9} 5.2473e-3 (7.75e-4)} \\
\textbf{LIRCMOP11} & 2.6099e-2 (4.25e-2) - & 7.2199e-3 (1.77e-2) = & 9.5068e-3 (2.13e-2) = & 4.1047e-2 (6.99e-2) - & {\color[HTML]{3333E9} 4.8358e-3 (1.28e-2)} \\
\textbf{LIRCMOP12} & 2.9985e-2 (4.19e-2) - & {\color[HTML]{3333E9} 8.6019e-3 (1.80e-2) =} & 1.4315e-2 (2.86e-2) = & 5.0199e-2 (7.87e-2) - & 1.9978e-2 (3.29e-2) \\
\textbf{LIRCMOP13} & {\color[HTML]{3333E9} 9.3813e-2 (9.19e-4) +} & 9.6504e-2 (1.50e-3) + & 9.8305e-2 (2.64e-3) = & 9.7438e-2 (2.66e-3) + & 9.8376e-2 (2.07e-3) \\
\textbf{LIRCMOP14} & {\color[HTML]{3333E9} 9.6684e-2 (1.12e-3) +} & 9.7934e-2 (1.35e-3) = & 9.7800e-2 (8.96e-4) = & 9.7875e-2 (1.05e-3) = & 9.7740e-2 (1.25e-3) \\
\textbf{DOC1} & {\color[HTML]{3333E9} 4.8801e-3 (2.90e-4) =} & 1.0289e-1 (2.54e-1) = & 2.9321e-2 (1.34e-1) = & 2.9458e-2 (1.35e-1) = & 1.0292e-1 (2.54e-1) \\
\textbf{DOC2} & 4.1719e-3 (1.29e-3) = & 4.3531e-3 (2.06e-3) = & {\color[HTML]{3333E9} 3.7600e-3 (1.16e-3) =} & 4.7316e-3 (2.03e-3) = & 4.5134e-3 (2.12e-3) \\
\textbf{DOC3} & 1.4273e+1 (5.40e+1) - & 4.3748e-2 (1.88e-1) = & {\color[HTML]{3333E9} 9.3501e-3 (4.04e-3) =} & 1.3298e+0 (6.88e+0) + & 1.7471e+0 (9.51e+0) \\
\textbf{DOC4} & 1.8931e-2 (1.26e-2) = & 1.5102e-2 (1.05e-2) = & {\color[HTML]{3333E9} 1.4069e-2 (6.64e-3) =} & 2.5022e-2 (1.28e-2) - & 1.5676e-2 (9.49e-3) \\
\textbf{DOC5} & {\color[HTML]{3333E9} 4.0523e+1 (6.15e+1) =} & 5.6132e+1 (6.28e+1) = & 4.9564e+1 (6.20e+1) = & 5.4936e+1 (6.33e+1) = & 4.4857e+1 (6.16e+1) \\
\textbf{DOC6} & 2.1241e+1 (6.86e+0) - & 2.0602e+1 (1.01e+1) - & {\color[HTML]{3333E9} 2.6544e-3 (1.36e-4) =} & 4.5662e-2 (2.24e-1) + & 2.2888e-1 (5.97e-1) \\
\textbf{DOC7} & 2.1731e+0 (1.73e+0) = & 2.2315e+0 (1.55e+0) = & 1.9372e+0 (1.53e+0) = & {\color[HTML]{3333E9} 1.8656e+0 (1.40e+0) =} & 2.2145e+0 (1.66e+0) \\
\textbf{DOC8} & 3.1884e+1 (3.83e+1) - & 1.2707e-1 (2.32e-2) = & {\color[HTML]{3333E9} 1.1796e-1 (1.86e-2) =} & 2.7680e-1 (2.85e-1) - & 1.2051e-1 (2.00e-2) \\
\textbf{DOC9} & {\color[HTML]{3333E9} 1.1858e-1 (9.85e-2) +} & 2.0811e-1 (8.87e-2) = & 1.9041e-1 (9.24e-2) = & 1.8713e-1 (8.08e-2) - & 1.5729e-1 (8.17e-2) \\
\textbf{FCP1} & 3.2837e-2 (4.30e-4) = & 3.2709e-2 (3.83e-4) = & NaN (NaN) & {\color[HTML]{3333E9} 3.1629e-2 (1.72e-4) +} & 3.2598e-2 (3.27e-4) \\
\textbf{FCP2} & 2.8217e-2 (1.55e-3) - & 2.7228e-2 (5.87e-4) = & NaN (NaN) & {\color[HTML]{3333E9} 2.7033e-2 (5.89e-4) =} & 2.7052e-2 (5.19e-4) \\
\textbf{FCP3} & 3.5566e-2 (3.00e-4) - & {\color[HTML]{3333E9} 3.5338e-2 (2.33e-4) =} & NaN (NaN) & 3.5430e-2 (3.94e-4) = & 3.5384e-2 (2.51e-4) \\
\textbf{FCP4} & 2.5337e-2 (3.87e-4) = & 2.5235e-2 (2.53e-4) = & NaN (NaN) & {\color[HTML]{3333E9} 2.5225e-2 (3.55e-4) =} & 2.5262e-2 (3.28e-4) \\
\textbf{FCP5} & 1.4458e-2 (6.48e-4) = & 1.4876e-2 (1.38e-3) = & 4.6770e+0 (1.59e-3) - & {\color[HTML]{3333E9} 1.4313e-2 (7.35e-4) =} & 1.4698e-2 (1.17e-3) \\ \hline
\textbf{+/-/=} & 6/27/28 & 2/7/52 & 0/7/54 & 5/16/40 &  \\ \hline
\end{tabular}%
}
\end{table}

\clearpage

\subsection{The ablation study results of HV and IGD about HOps, epsilon decay function and DRA of LLM4CMO obtaned by LLM-aided design.} \label{sec:A4-7}
LLM4CMOEps1 is a complete comparison with the original Eps, and LLM4CMOHOps-T1 to -T2 indicates that the algorithm only uses HOps under the UPF-CPF relationship types of full coincidence, partial overlap, complete separation, and unclear. LLM4CMOWoDRA represent remove DRA in LLM4CMO.

\begin{table}[htpb!]
\centering
\caption{Core modules ablation study results of HV values obtained by LLM4CMO on 6 test suite. Eps1 presents epsilon decay function use initial setting, HOps-T1, HOps-T1 and HOps-T2 present the HOps for UPF-CPF relationship type 1, 2 and 3, HOps-T4 present the HOps for UPF-CPF relationship type is unclear. DRA presents DRA mechnism.}
\label{tab:AB2HV}
\resizebox{\textwidth}{!}{%
\begin{tabular}{cccccccc}
\hline
\textbf{Problem} & \textbf{LLM4CMOEps1} & \textbf{LLM4CMOHOps-T1} & \textbf{LLM4CMOHOps-T2} & \textbf{LLM4CMOHOps-T3} & \textbf{LLM4CMOHOps-T4} & \textbf{LLM4CMOWoDRA} & \textbf{LLM4CMO} \\ \hline
\textbf{CF1} & 5.6353e-1 (4.46e-4) = & 5.6348e-1 (4.13e-4) = & 5.6363e-1 (4.94e-4) = & {\color[HTML]{3333E9} 5.6393e-1 (7.08e-4) +} & 5.6371e-1 (4.15e-4) = & 5.6369e-1 (4.30e-4) = & 5.6357e-1 (6.44e-4) \\
\textbf{CF2} & {\color[HTML]{3333E9} 6.7715e-1 (1.19e-3) =} & 6.7666e-1 (7.41e-4) - & 6.7682e-1 (1.18e-3) = & 6.7668e-1 (1.34e-3) = & 6.7682e-1 (8.83e-4) = & 6.7541e-1 (4.90e-3) = & 6.7699e-1 (1.48e-3) \\
\textbf{CF3} & {\color[HTML]{3333E9} 2.6178e-1 (2.55e-2) =} & 2.5369e-1 (2.80e-2) = & 2.4766e-1 (4.74e-2) = & 2.2145e-1 (3.94e-2) - & 2.3557e-1 (3.69e-2) = & 2.4475e-1 (3.76e-2) = & 2.5405e-1 (3.44e-2) \\
\textbf{CF4} & 5.0970e-1 (1.23e-2) = & 5.0570e-1 (1.09e-2) - & 5.1142e-1 (1.12e-2) = & 5.0045e-1 (1.38e-2) - & 5.0652e-1 (1.09e-2) - & 5.1029e-1 (8.66e-3) = & {\color[HTML]{3333E9} 5.1364e-1 (7.07e-3)} \\
\textbf{CF5} & 3.9524e-1 (4.89e-2) = & 3.7610e-1 (5.88e-2) = & {\color[HTML]{3333E9} 3.9774e-1 (6.89e-2) =} & 3.9108e-1 (7.65e-2) = & 3.7986e-1 (7.33e-2) = & 3.8580e-1 (6.37e-2) = & 3.9535e-1 (6.10e-2) \\
\textbf{CF6} & 6.7643e-1 (5.36e-3) - & 6.7531e-1 (4.56e-3) - & 6.7866e-1 (4.79e-3) = & 6.7840e-1 (3.95e-3) = & 6.7879e-1 (4.67e-3) = & 6.7751e-1 (5.92e-3) = & {\color[HTML]{3333E9} 6.7913e-1 (4.95e-3)} \\
\textbf{CF7} & 5.9573e-1 (1.59e-2) = & 5.9118e-1 (2.94e-2) = & 5.9551e-1 (2.84e-2) = & 5.7232e-1 (6.23e-2) - & 5.9678e-1 (2.07e-2) = & 5.9738e-1 (2.99e-2) = & {\color[HTML]{3333E9} 6.0011e-1 (3.04e-2)} \\
\textbf{CF8} & 2.9120e-1 (7.96e-2) = & 3.0624e-1 (8.14e-2) = & {\color[HTML]{3333E9} 3.2287e-1 (7.23e-2) =} & 1.7808e-1 (6.23e-2) - & 2.8343e-1 (6.95e-2) = & 2.7996e-1 (8.16e-2) = & 2.8384e-1 (9.31e-2) \\
\textbf{CF9} & 3.1664e-1 (2.35e-2) - & 3.2944e-1 (1.48e-2) - & 3.2614e-1 (1.64e-2) - & 3.2997e-1 (1.58e-2) - & 3.1427e-1 (2.06e-2) - & 3.2331e-1 (1.70e-2) - & {\color[HTML]{3333E9} 4.5111e-1 (1.96e-2)} \\
\textbf{CF10} & 1.7638e-1 (3.82e-2) = & {\color[HTML]{3333E9} 2.1605e-1 (6.10e-2) +} & 2.0469e-1 (6.42e-2) = & 1.1538e-1 (3.54e-2) - & 1.7198e-1 (3.57e-2) = & 1.8373e-1 (5.52e-2) = & 1.7561e-1 (3.15e-2) \\
\textbf{DASCMOP1} & 2.1245e-1 (3.68e-4) = & 1.7134e-1 (6.89e-2) - & 1.9822e-1 (4.22e-2) - & 2.1241e-1 (3.96e-4) = & 2.1190e-1 (5.98e-4) - & 2.1245e-1 (3.50e-4) = & {\color[HTML]{3333E9} 2.1254e-1 (4.07e-4)} \\
\textbf{DASCMOP2} & 3.5516e-1 (7.22e-5) = & 3.4296e-1 (2.74e-2) - & 3.5001e-1 (2.06e-2) - & 3.5517e-1 (9.29e-5) = & {\color[HTML]{3333E9} 3.5544e-1 (1.12e-4) +} & 3.5514e-1 (8.63e-5) = & 3.5513e-1 (9.94e-5) \\
\textbf{DASCMOP3} & 2.9447e-1 (3.32e-2) - & 2.2263e-1 (2.83e-2) - & 2.2645e-1 (3.37e-2) - & {\color[HTML]{3333E9} 3.1232e-1 (1.41e-4) =} & 3.0113e-1 (3.14e-2) - & 3.1123e-1 (5.88e-3) = & 3.1230e-1 (1.97e-4) \\
\textbf{DASCMOP4} & 2.0169e-1 (1.33e-2) = & {\color[HTML]{3333E9} 2.0422e-1 (1.45e-4) +} & 2.0406e-1 (2.12e-4) = & 2.0384e-1 (2.69e-4) - & 2.0410e-1 (2.45e-4) = & 2.0393e-1 (2.79e-4) = & 2.0396e-1 (4.43e-4) \\
\textbf{DASCMOP5} & 3.5129e-1 (3.69e-4) = & 3.5150e-1 (1.42e-4) = & 3.5144e-1 (2.24e-4) = & 3.5101e-1 (3.53e-4) - & {\color[HTML]{3333E9} 3.5150e-1 (1.14e-4) =} & 3.5136e-1 (3.28e-4) = & 3.5132e-1 (3.81e-4) \\
\textbf{DASCMOP6} & 2.9307e-1 (5.27e-2) = & 3.1249e-1 (1.32e-4) = & {\color[HTML]{3333E9} 3.1252e-1 (9.43e-5) =} & 3.0763e-1 (2.56e-2) - & 3.1251e-1 (6.85e-5) = & 3.1250e-1 (1.39e-4) = & 3.1245e-1 (1.91e-4) \\
\textbf{DASCMOP7} & 2.8788e-1 (1.54e-3) = & {\color[HTML]{3333E9} 2.8851e-1 (1.61e-4) =} & 2.8843e-1 (3.41e-4) = & 2.8492e-1 (1.06e-3) - & 2.8834e-1 (3.08e-4) = & 2.8700e-1 (2.13e-3) = & 2.8793e-1 (1.30e-3) \\
\textbf{DASCMOP8} & 2.0678e-1 (1.08e-3) = & {\color[HTML]{3333E9} 2.0711e-1 (3.28e-4) =} & 2.0686e-1 (7.49e-4) = & 2.0418e-1 (1.19e-3) - & 2.0700e-1 (3.77e-4) = & 2.0623e-1 (1.36e-3) = & 2.0621e-1 (1.62e-3) \\
\textbf{DASCMOP9} & 2.0479e-1 (3.91e-4) = & 2.0647e-1 (6.17e-4) + & {\color[HTML]{3333E9} 2.0703e-1 (3.81e-4) +} & 2.0479e-1 (3.98e-4) = & 2.0700e-1 (3.99e-4) + & 2.0469e-1 (5.28e-4) - & 2.0492e-1 (2.73e-4) \\
\textbf{MW1} & 4.9001e-1 (3.44e-4) = & 4.9009e-1 (1.65e-5) = & 4.9000e-1 (3.60e-5) - & 4.8901e-1 (2.02e-4) - & 4.9007e-1 (2.76e-5) - & 4.8990e-1 (7.22e-4) = & {\color[HTML]{3333E9} 4.9010e-1 (1.73e-5)} \\
\textbf{MW2} & 5.8240e-1 (3.02e-5) = & {\color[HTML]{3333E9} 5.8240e-1 (3.22e-5) =} & 5.8221e-1 (8.01e-5) - & 5.8154e-1 (2.38e-4) - & 5.8232e-1 (3.44e-5) - & 5.8239e-1 (3.56e-5) = & 5.8239e-1 (3.88e-5) \\
\textbf{MW3} & 5.4445e-1 (4.61e-4) = & 5.4452e-1 (4.29e-4) = & 5.4456e-1 (3.14e-4) = & 5.4262e-1 (4.72e-4) - & 5.4428e-1 (6.28e-4) - & 5.4442e-1 (5.07e-4) - & {\color[HTML]{3333E9} 5.4464e-1 (2.78e-4)} \\
\textbf{MW4} & 8.4050e-1 (5.20e-3) = & 8.4145e-1 (4.17e-4) = & 8.3839e-1 (9.82e-4) - & 8.1613e-1 (2.94e-3) - & 8.4018e-1 (7.81e-4) - & {\color[HTML]{3333E9} 8.4151e-1 (4.03e-4) =} & 8.4143e-1 (3.62e-4) \\
\textbf{MW5} & 3.2446e-1 (1.97e-4) = & {\color[HTML]{3333E9} 3.2448e-1 (1.44e-4) =} & 3.2441e-1 (1.48e-4) = & 3.2332e-1 (2.74e-4) - & 3.2446e-1 (1.96e-4) = & 3.2445e-1 (2.22e-4) = & 3.2423e-1 (9.33e-4) \\
\textbf{MW6} & 3.2850e-1 (2.14e-5) = & 3.2850e-1 (1.97e-5) = & 3.2849e-1 (1.77e-5) = & 3.2814e-1 (1.01e-4) - & {\color[HTML]{3333E9} 3.2851e-1 (1.74e-5) +} & 3.2850e-1 (1.91e-5) = & 3.2850e-1 (1.56e-5) \\
\textbf{MW7} & {\color[HTML]{3333E9} 4.1184e-1 (8.05e-4) =} & 4.1160e-1 (6.44e-4) = & 4.1180e-1 (6.12e-4) = & 4.1082e-1 (5.21e-4) - & 4.1164e-1 (6.64e-4) = & 4.1155e-1 (7.24e-4) = & 4.1180e-1 (5.40e-4) \\
\textbf{MW8} & {\color[HTML]{3333E9} 5.5329e-1 (7.92e-4) =} & 5.5316e-1 (1.03e-3) = & 5.5102e-1 (2.11e-3) - & 5.2859e-1 (7.22e-3) - & 5.5216e-1 (8.41e-4) - & 5.5299e-1 (1.11e-3) = & 5.5314e-1 (1.08e-3) \\
\textbf{MW9} & 3.9437e-1 (2.16e-3) = & 3.9724e-1 (2.06e-3) + & 3.9649e-1 (2.99e-3) + & 3.9378e-1 (3.60e-3) = & {\color[HTML]{3333E9} 3.9728e-1 (2.13e-3) +} & 3.9428e-1 (2.38e-3) = & 3.9420e-1 (2.52e-3) \\
\textbf{MW10} & 4.5477e-1 (1.17e-4) = & {\color[HTML]{3333E9} 4.5498e-1 (1.10e-4) +} & 4.5440e-1 (2.02e-3) = & 4.5101e-1 (4.15e-3) - & 4.5485e-1 (1.18e-4) + & 4.5475e-1 (1.02e-4) = & 4.5441e-1 (2.03e-3) \\
\textbf{MW11} & 4.4728e-1 (2.20e-4) = & 4.4751e-1 (2.72e-4) + & 4.4758e-1 (2.30e-4) + & 4.4735e-1 (1.98e-4) = & {\color[HTML]{3333E9} 4.4763e-1 (2.25e-4) +} & 4.4734e-1 (1.84e-4) = & 4.4733e-1 (2.64e-4) \\
\textbf{MW12} & 6.0385e-1 (3.48e-4) = & {\color[HTML]{3333E9} 6.0500e-1 (1.91e-4) +} & 6.0463e-1 (2.81e-4) + & 6.0375e-1 (3.08e-4) = & 6.0492e-1 (1.84e-4) + & 6.0370e-1 (3.70e-4) = & 6.0368e-1 (2.55e-4) \\
\textbf{MW13} & 4.7613e-1 (2.98e-3) - & 4.7683e-1 (1.04e-4) = & 4.7678e-1 (3.87e-4) = & 4.7275e-1 (4.55e-3) - & {\color[HTML]{3333E9} 4.7685e-1 (7.54e-5) =} & 4.7618e-1 (2.61e-3) = & 4.7680e-1 (2.09e-4) \\
\textbf{MW14} & 4.7393e-1 (2.09e-3) = & {\color[HTML]{3333E9} 4.7453e-1 (2.23e-3) =} & 4.7377e-1 (1.54e-3) = & 4.6293e-1 (2.99e-3) - & 4.7283e-1 (1.99e-3) - & 4.7429e-1 (1.94e-3) = & 4.7404e-1 (1.84e-3) \\
\textbf{LIRCMOP1} & 2.3672e-1 (9.23e-4) = & 1.9140e-1 (2.08e-2) - & 1.9731e-1 (1.75e-2) - & 2.3691e-1 (5.66e-4) = & 2.0381e-1 (1.72e-2) - & 2.3690e-1 (3.78e-4) = & {\color[HTML]{3333E9} 2.3694e-1 (7.05e-4)} \\
\textbf{LIRCMOP2} & 3.5963e-1 (4.42e-4) - & 3.2786e-1 (1.96e-2) - & 3.4096e-1 (1.32e-2) - & 3.6045e-1 (1.24e-4) = & 3.4520e-1 (1.36e-2) - & {\color[HTML]{3333E9} 3.6049e-1 (1.24e-4) =} & 3.6045e-1 (1.62e-4) \\
\textbf{LIRCMOP3} & 2.0726e-1 (6.07e-4) - & 1.7124e-1 (1.44e-2) - & 1.7808e-1 (1.50e-2) - & 2.0859e-1 (3.41e-4) = & 1.8084e-1 (1.64e-2) - & {\color[HTML]{3333E9} 2.0870e-1 (2.66e-4) =} & 2.0853e-1 (4.39e-4) \\
\textbf{LIRCMOP4} & 3.1635e-1 (4.62e-4) - & 2.8548e-1 (1.79e-2) - & 2.9123e-1 (1.16e-2) - & 3.1739e-1 (4.29e-4) = & 2.9289e-1 (9.13e-3) - & {\color[HTML]{3333E9} 3.1746e-1 (4.05e-4) =} & 3.1742e-1 (4.71e-4) \\
\textbf{LIRCMOP5} & 2.9138e-1 (5.73e-4) - & 2.9101e-1 (5.09e-4) - & {\color[HTML]{3333E9} 2.9172e-1 (3.93e-4) =} & 2.9133e-1 (2.61e-4) - & 2.9163e-1 (3.15e-4) = & 2.9149e-1 (3.42e-4) = & 2.9164e-1 (2.42e-4) \\
\textbf{LIRCMOP6} & 1.9677e-1 (2.92e-4) - & 1.8328e-1 (4.98e-2) - & 1.9042e-1 (3.60e-2) = & 1.9642e-1 (2.93e-4) - & 1.9217e-1 (2.59e-2) = & 1.9687e-1 (2.78e-4) = & {\color[HTML]{3333E9} 1.9695e-1 (1.61e-4)} \\
\textbf{LIRCMOP7} & 2.7905e-1 (1.93e-2) - & 2.7079e-1 (2.02e-2) - & 2.8247e-1 (1.83e-2) - & 2.9029e-1 (1.35e-2) - & 2.8455e-1 (1.60e-2) - & 2.9090e-1 (1.10e-2) = & {\color[HTML]{3333E9} 2.9380e-1 (1.76e-3)} \\
\textbf{LIRCMOP8} & 2.7914e-1 (2.18e-2) - & 2.8177e-1 (1.92e-2) - & 2.8582e-1 (1.73e-2) = & 2.9280e-1 (8.46e-3) - & 2.8876e-1 (1.60e-2) = & 2.9313e-1 (7.91e-3) - & {\color[HTML]{3333E9} 2.9476e-1 (2.28e-4)} \\
\textbf{LIRCMOP9} & 4.7529e-1 (7.06e-2) - & 4.7909e-1 (5.79e-2) - & 4.8361e-1 (7.05e-2) - & {\color[HTML]{3333E9} 5.5425e-1 (3.99e-2) +} & 5.2680e-1 (5.25e-2) = & 5.3289e-1 (6.41e-2) = & 5.3447e-1 (4.58e-2) \\
\textbf{LIRCMOP10} & 7.0686e-1 (3.91e-4) = & 7.0357e-1 (9.29e-3) - & {\color[HTML]{3333E9} 7.0690e-1 (3.66e-4) =} & 7.0676e-1 (2.87e-4) = & 7.0660e-1 (4.67e-4) - & 7.0024e-1 (2.44e-2) = & 7.0679e-1 (5.08e-4) \\
\textbf{LIRCMOP11} & 6.8387e-1 (2.33e-2) = & 6.8157e-1 (3.36e-2) - & 6.8193e-1 (2.36e-2) = & 6.9278e-1 (6.13e-3) - & 6.9089e-1 (9.48e-3) = & {\color[HTML]{3333E9} 6.9286e-1 (6.14e-3) =} & 6.9284e-1 (6.15e-3) \\
\textbf{LIRCMOP12} & 5.9360e-1 (3.71e-2) - & 5.9335e-1 (4.10e-2) - & 5.9417e-1 (3.53e-2) = & {\color[HTML]{3333E9} 6.1862e-1 (7.51e-3) +} & 6.0747e-1 (2.09e-2) = & 6.1405e-1 (1.51e-2) = & 6.1281e-1 (1.45e-2) \\
\textbf{LIRCMOP13} & 5.4621e-1 (3.37e-3) + & {\color[HTML]{3333E9} 5.4696e-1 (2.27e-3) +} & 5.3786e-1 (3.41e-3) - & 5.0621e-1 (1.50e-2) - & 5.4028e-1 (2.12e-3) - & 5.4592e-1 (3.09e-3) = & 5.4405e-1 (3.81e-3) \\
\textbf{LIRCMOP14} & 5.5061e-1 (2.16e-3) = & {\color[HTML]{3333E9} 5.5347e-1 (1.14e-3) +} & 5.4938e-1 (1.73e-3) - & 5.3434e-1 (5.87e-3) - & 5.5062e-1 (1.45e-3) - & 5.5169e-1 (1.42e-3) = & 5.5142e-1 (1.30e-3) \\
\textbf{DOC1} & 3.3633e-1 (4.64e-2) = & 3.4449e-1 (2.54e-3) + & 3.4527e-1 (5.42e-4) + & 3.3626e-1 (4.63e-2) = & {\color[HTML]{3333E9} 3.4554e-1 (7.62e-4) +} & 3.4467e-1 (6.68e-4) = & 3.1094e-1 (8.78e-2) \\
\textbf{DOC2} & 6.2079e-1 (2.09e-3) = & NaN (NaN) & 6.0728e-1 (3.40e-2) - & 6.2100e-1 (2.64e-3) = & NaN (NaN) & {\color[HTML]{3333E9} 6.2112e-1 (1.76e-3) =} & 6.2066e-1 (2.84e-3) \\
\textbf{DOC3} & {\color[HTML]{3333E9} 3.2920e-1 (4.23e-3) =} & 0.0000e+0 (0.00e+0) - & 0.0000e+0 (0.00e+0) - & 3.2610e-1 (8.76e-3) = & 0.0000e+0 (0.00e+0) - & 3.1599e-1 (6.00e-2) = & 3.1736e-1 (6.02e-2) \\
\textbf{DOC4} & 5.4119e-1 (1.30e-2) = & 4.8135e-1 (8.48e-2) - & 5.3963e-1 (9.89e-3) - & 5.4545e-1 (1.32e-2) = & 5.1088e-1 (2.92e-2) - & 5.4521e-1 (9.33e-3) = & {\color[HTML]{3333E9} 5.4603e-1 (1.09e-2)} \\
\textbf{DOC5} & 2.3721e-1 (2.50e-1) = & 0.0000e+0 (0.00e+0) = & 1.6420e-1 (2.37e-1) = & 2.2123e-1 (2.50e-1) = & 0.0000e+0 (0.00e+0) = & 1.9320e-1 (2.45e-1) = & {\color[HTML]{3333E9} 3.0943e-1 (2.42e-1)} \\
\textbf{DOC6} & 0.0000e+0 (0.00e+0) - & 0.0000e+0 (0.00e+0) - & 0.0000e+0 (0.00e+0) - & 0.0000e+0 (0.00e+0) - & 0.0000e+0 (0.00e+0) - & 0.0000e+0 (0.00e+0) - & {\color[HTML]{3333E9} 4.6415e-1 (2.05e-1)} \\
\textbf{DOC7} & 1.4834e-1 (2.50e-1) = & 0.0000e+0 (0.00e+0) - & {\color[HTML]{3333E9} 1.8719e-1 (2.34e-1) =} & 7.3360e-2 (1.90e-1) = & 3.1902e-3 (1.57e-2) = & 1.8390e-2 (1.01e-1) - & 9.9085e-2 (2.11e-1) \\
\textbf{DOC8} & 7.3190e-1 (2.14e-2) + & 2.2405e-2 (1.13e-1) - & {\color[HTML]{3333E9} 7.8926e-1 (7.44e-3) +} & 7.1786e-1 (1.74e-2) = & 1.5809e-2 (7.21e-2) - & 7.0411e-1 (4.51e-2) = & 7.1633e-1 (2.76e-2) \\
\textbf{DOC9} & {\color[HTML]{3333E9} 0.0000e+0 (0.00e+0) =} & NaN (NaN) & NaN (NaN) & 0.0000e+0 (0.00e+0) = & NaN (NaN) & 0.0000e+0 (0.00e+0) = & 0.0000e+0 (0.00e+0) \\
\textbf{FCP1} & 5.8178e-1 (8.50e-5) = & {\color[HTML]{3333E9} 5.8214e-1 (8.84e-5) +} & 5.8205e-1 (7.92e-5) + & 5.8175e-1 (1.46e-4) = & 5.8199e-1 (1.12e-4) + & 5.8175e-1 (1.16e-4) = & 5.8176e-1 (1.15e-4) \\
\textbf{FCP2} & 4.3154e-1 (8.06e-5) = & {\color[HTML]{3333E9} 4.3183e-1 (4.67e-5) +} & 4.3182e-1 (5.29e-5) + & 4.3155e-1 (9.44e-5) = & 4.3182e-1 (6.73e-5) + & 4.3152e-1 (8.29e-5) = & 4.3154e-1 (7.91e-5) \\
\textbf{FCP3} & 3.4725e-1 (7.44e-5) = & {\color[HTML]{3333E9} 3.4738e-1 (6.80e-5) +} & 3.4725e-1 (9.79e-5) = & 3.4723e-1 (8.18e-5) = & 3.4733e-1 (7.32e-5) + & 3.4726e-1 (7.33e-5) = & 3.4723e-1 (8.60e-5) \\
\textbf{FCP4} & 6.3468e-1 (3.57e-5) = & 6.3471e-1 (3.94e-5) + & {\color[HTML]{3333E9} 6.3474e-1 (3.03e-5) +} & 6.3468e-1 (3.39e-5) = & 6.3471e-1 (4.23e-5) + & 6.3468e-1 (4.56e-5) = & 6.3467e-1 (3.52e-5) \\
\textbf{FCP5} & 4.7955e-1 (3.17e-4) = & 4.7979e-1 (4.19e-4) + & {\color[HTML]{3333E9} 4.7989e-1 (2.10e-4) +} & 4.7961e-1 (2.45e-4) = & 4.7977e-1 (8.57e-4) + & 4.7959e-1 (3.43e-4) = & 4.7961e-1 (2.13e-4) \\ \hline
\textbf{+/-/=} & 2/14/45 & 15/26/20 & 10/20/30 & 3/30/28 & 13/24/24 & 0/6/55 &  \\ \hline
\end{tabular}%
}
\end{table}

\begin{table}[htpb!]
\centering
\caption{Core modules ablation study results of IGD values obtained by LLM4CMO on 6 test suite. Eps1 presents epsilon decay function use initial setting, HOps-T1, HOps-T1 and HOps-T2 present the HOps for UPF-CPF relationship type 1, 2 and 3, HOps-T4 present the HOps for UPF-CPF relationship type is unclear. DRA presents DRA mechnism.}
\label{tab:AB2IGD}
\resizebox{\textwidth}{!}{%
\begin{tabular}{cccccccc}
\hline
\textbf{Problem} & \textbf{LLM4CMOWoEps} & \textbf{LLM4CMOHOps-T1} & \textbf{LLM4CMOHOps-T2} & \textbf{LLM4CMOHOps-T3} & \textbf{LLM4CMOHOps-T4} & \textbf{LLM4CMOWoDRA} & \textbf{LLM4CMO} \\ \hline
\textbf{CF1} & 2.1067e-3 (3.56e-4) = & 2.1511e-3 (3.39e-4) = & 2.0280e-3 (4.04e-4) = & {\color[HTML]{3333E9} 1.7895e-3 (5.78e-4) +} & 1.9675e-3 (3.41e-4) = & 1.9808e-3 (3.52e-4) = & 2.0762e-3 (5.21e-4) \\
\textbf{CF2} & 3.6653e-3 (1.07e-3) = & 3.8195e-3 (5.29e-4) - & {\color[HTML]{3333E9} 3.1695e-3 (4.65e-4) +} & 3.1971e-3 (2.36e-4) + & 3.3776e-3 (5.47e-4) = & 4.4778e-3 (3.21e-3) - & 3.4599e-3 (6.36e-4) \\
\textbf{CF3} & 1.2739e-1 (4.99e-2) = & {\color[HTML]{3333E9} 1.2646e-1 (5.41e-2) =} & 1.3572e-1 (8.06e-2) = & 1.6156e-1 (5.67e-2) = & 1.7309e-1 (6.59e-2) = & 1.4454e-1 (6.60e-2) = & 1.4396e-1 (6.07e-2) \\
\textbf{CF4} & 2.1974e-2 (1.02e-2) = & 2.4574e-2 (9.36e-3) - & 2.0743e-2 (9.80e-3) = & 2.7560e-2 (1.05e-2) - & 2.4359e-2 (8.94e-3) - & 2.0831e-2 (7.20e-3) - & {\color[HTML]{3333E9} 1.7778e-2 (6.34e-3)} \\
\textbf{CF5} & 1.5682e-1 (7.42e-2) = & 1.8361e-1 (8.71e-2) = & 1.5543e-1 (1.10e-1) = & 1.5239e-1 (1.28e-1) = & 1.7751e-1 (1.18e-1) = & 1.6473e-1 (9.39e-2) = & {\color[HTML]{3333E9} 1.5043e-1 (8.61e-2)} \\
\textbf{CF6} & 3.0688e-2 (9.78e-3) = & 3.0570e-2 (7.31e-3) - & 2.5194e-2 (6.77e-3) = & {\color[HTML]{3333E9} 2.4594e-2 (6.19e-3) =} & 2.5622e-2 (7.66e-3) = & 3.0555e-2 (1.21e-2) = & 2.6238e-2 (7.44e-3) \\
\textbf{CF7} & 9.3916e-2 (2.81e-2) = & 9.8091e-2 (3.92e-2) = & 9.9867e-2 (4.24e-2) = & 9.7851e-2 (4.50e-2) = & 9.0689e-2 (2.86e-2) = & 8.8645e-2 (3.34e-2) = & {\color[HTML]{3333E9} 8.4939e-2 (3.05e-2)} \\
\textbf{CF8} & 2.2063e-1 (8.73e-2) = & 2.0383e-1 (7.63e-2) = & {\color[HTML]{3333E9} 1.8897e-1 (5.91e-2) +} & 3.1132e-1 (8.82e-2) - & 2.2404e-1 (7.35e-2) = & 2.4305e-1 (9.12e-2) = & 2.3201e-1 (8.92e-2) \\
\textbf{CF9} & 1.4259e-1 (2.66e-2) - & 1.3250e-1 (1.79e-2) - & 1.3518e-1 (1.76e-2) - & 1.3008e-1 (1.56e-2) - & 1.4547e-1 (2.14e-2) - & 1.3646e-1 (1.79e-2) - & {\color[HTML]{3333E9} 8.8228e-2 (1.64e-2)} \\
\textbf{CF10} & 3.2817e-1 (6.35e-2) = & {\color[HTML]{3333E9} 2.9474e-1 (7.42e-2) =} & 3.0051e-1 (7.17e-2) = & 3.9468e-1 (6.63e-2) - & 3.3926e-1 (6.44e-2) = & 3.3957e-1 (7.39e-2) = & 3.2466e-1 (6.34e-2) \\
\textbf{DASCMOP1} & {\color[HTML]{3333E9} 2.9372e-3 (2.06e-4) =} & 1.5781e-1 (2.69e-1) - & 6.9818e-2 (2.03e-1) = & 2.9716e-3 (1.78e-4) = & 3.6289e-3 (5.48e-4) - & 2.9962e-3 (1.87e-4) = & 2.9463e-3 (2.21e-4) \\
\textbf{DASCMOP2} & 4.4090e-3 (9.22e-5) = & 2.9617e-2 (5.94e-2) - & 1.8047e-2 (5.36e-2) - & 4.4443e-3 (1.29e-4) = & {\color[HTML]{3333E9} 4.2785e-3 (1.62e-4) +} & 4.4288e-3 (1.12e-4) = & 4.4413e-3 (1.24e-4) \\
\textbf{DASCMOP3} & 9.3731e-2 (1.13e-1) - & 2.9755e-1 (8.42e-2) - & 2.8923e-1 (1.01e-1) - & 1.8531e-2 (2.92e-3) = & 5.3107e-2 (9.86e-2) - & 2.3449e-2 (2.68e-2) = & {\color[HTML]{3333E9} 1.7329e-2 (3.44e-3)} \\
\textbf{DASCMOP4} & 8.2188e-3 (3.84e-2) = & {\color[HTML]{3333E9} 1.1571e-3 (1.58e-5) +} & 1.2060e-3 (3.59e-5) = & 1.3413e-3 (8.88e-5) - & 1.1911e-3 (4.59e-5) = & 1.2544e-3 (1.03e-4) = & 1.3109e-3 (3.27e-4) \\
\textbf{DASCMOP5} & 3.0066e-3 (2.75e-4) = & 2.8561e-3 (1.31e-4) + & 2.9307e-3 (2.30e-4) = & 3.2113e-3 (2.02e-4) - & {\color[HTML]{3333E9} 2.8208e-3 (6.61e-5) +} & 2.9718e-3 (1.60e-4) = & 2.9791e-3 (2.28e-4) \\
\textbf{DASCMOP6} & 5.4384e-2 (9.82e-2) = & 1.7845e-2 (2.99e-3) = & {\color[HTML]{3333E9} 1.6409e-2 (3.52e-3) =} & 2.6805e-2 (4.55e-2) = & 1.7953e-2 (2.98e-3) = & 1.7718e-2 (3.17e-3) = & 1.8202e-2 (2.84e-3) \\
\textbf{DASCMOP7} & 3.1616e-2 (1.52e-3) = & 3.0837e-2 (7.40e-4) + & {\color[HTML]{3333E9} 3.0748e-2 (1.12e-3) =} & 3.2969e-2 (1.70e-3) - & 3.1203e-2 (8.78e-4) = & 3.1798e-2 (1.73e-3) = & 3.1474e-2 (1.36e-3) \\
\textbf{DASCMOP8} & 4.0027e-2 (1.11e-3) = & 4.0129e-2 (9.27e-4) = & 3.9867e-2 (9.61e-4) = & {\color[HTML]{3333E9} 3.9692e-2 (1.51e-3) =} & 4.0183e-2 (1.14e-3) = & 4.0148e-2 (1.45e-3) = & 4.0000e-2 (1.39e-3) \\
\textbf{DASCMOP9} & 4.0179e-2 (9.73e-4) = & 4.0211e-2 (1.90e-3) = & {\color[HTML]{3333E9} 3.8268e-2 (7.45e-4) +} & 4.0011e-2 (1.15e-3) = & 3.8664e-2 (1.01e-3) + & 4.0426e-2 (9.15e-4) = & 4.0441e-2 (1.10e-3) \\
\textbf{MW1} & 1.6728e-3 (1.64e-4) = & {\color[HTML]{3333E9} 1.6265e-3 (9.83e-6) =} & 1.6506e-3 (1.40e-5) - & 2.0442e-3 (8.90e-5) - & 1.6386e-3 (1.14e-5) - & 1.7083e-3 (2.59e-4) = & 1.6293e-3 (1.31e-5) \\
\textbf{MW2} & 3.7340e-3 (2.41e-5) = & {\color[HTML]{3333E9} 3.7278e-3 (2.36e-5) =} & 3.8030e-3 (5.53e-5) - & 4.2857e-3 (1.74e-4) - & 3.7530e-3 (1.99e-5) - & 3.7366e-3 (2.79e-5) = & 3.7350e-3 (2.39e-5) \\
\textbf{MW3} & 4.8753e-3 (2.61e-4) - & 4.8248e-3 (2.03e-4) = & 4.8074e-3 (1.82e-4) = & 5.5683e-3 (2.30e-4) - & 4.9223e-3 (2.96e-4) - & 4.8774e-3 (2.34e-4) - & {\color[HTML]{3333E9} 4.7352e-3 (1.53e-4)} \\
\textbf{MW4} & 4.1765e-2 (3.09e-3) = & 4.1297e-2 (4.20e-4) = & 4.2501e-2 (6.18e-4) - & 5.6133e-2 (1.86e-3) - & 4.1739e-2 (5.38e-4) - & {\color[HTML]{3333E9} 4.1157e-2 (3.08e-4) =} & 4.1203e-2 (3.50e-4) \\
\textbf{MW5} & 4.7311e-4 (2.75e-4) = & {\color[HTML]{3333E9} 4.5143e-4 (2.21e-4) =} & 5.8500e-4 (2.47e-4) = & 1.7931e-3 (3.12e-4) - & 5.3335e-4 (5.44e-4) = & 4.9602e-4 (2.94e-4) = & 6.9019e-4 (9.26e-4) \\
\textbf{MW6} & 2.7562e-3 (2.02e-5) = & {\color[HTML]{3333E9} 2.7468e-3 (1.77e-5) =} & 2.7717e-3 (2.39e-5) - & 2.8497e-3 (3.63e-5) - & 2.7700e-3 (2.67e-5) - & 2.7601e-3 (2.01e-5) = & 2.7532e-3 (1.95e-5) \\
\textbf{MW7} & {\color[HTML]{3333E9} 4.4175e-3 (2.94e-4) =} & 4.6844e-3 (3.81e-4) = & 4.5189e-3 (3.00e-4) = & 4.8086e-3 (2.84e-4) - & 4.5753e-3 (3.43e-4) = & 4.5679e-3 (3.66e-4) = & 4.5298e-3 (3.80e-4) \\
\textbf{MW8} & 4.3219e-2 (6.53e-4) = & 4.3292e-2 (6.36e-4) = & 4.2915e-2 (7.21e-4) = & 4.8226e-2 (2.71e-3) - & {\color[HTML]{3333E9} 4.2777e-2 (6.54e-4) +} & 4.3100e-2 (5.45e-4) = & 4.3129e-2 (6.01e-4) \\
\textbf{MW9} & 5.4144e-3 (5.75e-4) = & {\color[HTML]{3333E9} 4.6297e-3 (2.63e-4) +} & 4.9702e-3 (6.40e-4) + & 5.8352e-3 (1.70e-3) = & 4.7469e-3 (3.77e-4) + & 5.3756e-3 (5.73e-4) = & 5.4156e-3 (5.73e-4) \\
\textbf{MW10} & 3.4795e-3 (5.24e-5) = & {\color[HTML]{3333E9} 3.4063e-3 (6.45e-5) +} & 3.6625e-3 (9.97e-4) = & 5.1809e-3 (2.26e-3) - & 3.4618e-3 (5.44e-5) = & 3.5024e-3 (5.56e-5) = & 3.6641e-3 (1.00e-3) \\
\textbf{MW11} & 6.3593e-3 (1.39e-4) = & 6.1177e-3 (1.57e-4) + & 6.1139e-3 (1.31e-4) + & 6.3131e-3 (1.54e-4) = & {\color[HTML]{3333E9} 6.0535e-3 (1.20e-4) +} & 6.3643e-3 (1.52e-4) = & 6.3137e-3 (1.71e-4) \\
\textbf{MW12} & 5.1129e-3 (1.73e-4) = & 4.7144e-3 (1.19e-4) + & 4.7919e-3 (1.48e-4) + & 5.1520e-3 (1.37e-4) = & {\color[HTML]{3333E9} 4.7060e-3 (9.13e-5) +} & 5.1673e-3 (1.93e-4) = & 5.1894e-3 (1.50e-4) \\
\textbf{MW13} & 1.1561e-2 (3.12e-3) = & {\color[HTML]{3333E9} 1.0672e-2 (1.39e-4) =} & 1.0829e-2 (5.21e-4) = & 1.5574e-2 (5.80e-3) - & 1.0719e-2 (1.30e-4) = & 1.1448e-2 (2.86e-3) = & 1.0763e-2 (2.41e-4) \\
\textbf{MW14} & 9.7775e-2 (2.18e-3) = & {\color[HTML]{3333E9} 9.7063e-2 (1.88e-3) =} & 9.8538e-2 (2.37e-3) = & 1.0315e-1 (3.14e-3) - & 9.8385e-2 (1.99e-3) = & 9.7653e-2 (2.07e-3) = & 9.7820e-2 (1.96e-3) \\
\textbf{LIRCMOP1} & 8.6700e-3 (1.30e-3) = & 9.0970e-2 (4.10e-2) - & 7.5483e-2 (3.29e-2) - & 8.4619e-3 (1.09e-3) = & 6.8070e-2 (3.01e-2) - & {\color[HTML]{3333E9} 8.2387e-3 (9.45e-4) =} & 8.3722e-3 (1.07e-3) \\
\textbf{LIRCMOP2} & 6.3179e-3 (6.57e-4) - & 6.1583e-2 (2.88e-2) - & 3.5262e-2 (2.06e-2) - & 4.7567e-3 (1.92e-4) = & 2.8870e-2 (1.89e-2) - & {\color[HTML]{3333E9} 4.7546e-3 (1.12e-4) =} & 4.7593e-3 (1.76e-4) \\
\textbf{LIRCMOP3} & 4.2622e-3 (8.26e-4) - & 8.7192e-2 (4.22e-2) - & 6.7408e-2 (3.87e-2) - & 2.5406e-3 (6.25e-4) = & 6.6603e-2 (4.67e-2) - & {\color[HTML]{3333E9} 2.4054e-3 (5.14e-4) =} & 2.6231e-3 (7.87e-4) \\
\textbf{LIRCMOP4} & 3.2973e-3 (6.02e-4) - & 7.1349e-2 (3.59e-2) - & 6.4749e-2 (2.96e-2) - & 2.2496e-3 (3.95e-4) = & 5.3914e-2 (2.29e-2) - & {\color[HTML]{3333E9} 2.1985e-3 (4.38e-4) =} & 2.2285e-3 (3.60e-4) \\
\textbf{LIRCMOP5} & 6.3863e-3 (7.86e-4) - & 7.0690e-3 (7.26e-4) - & {\color[HTML]{3333E9} 5.7238e-3 (4.41e-4) +} & 6.4739e-3 (4.42e-4) - & 5.9433e-3 (5.11e-4) = & 6.2898e-3 (5.79e-4) = & 6.0089e-3 (4.86e-4) \\
\textbf{LIRCMOP6} & 5.9831e-3 (5.65e-4) - & 9.5906e-2 (3.40e-1) - & 5.0226e-2 (2.44e-1) = & 6.5177e-3 (5.24e-4) - & 2.5916e-2 (1.11e-1) = & 5.8035e-3 (5.27e-4) = & {\color[HTML]{3333E9} 5.6623e-3 (2.80e-4)} \\
\textbf{LIRCMOP7} & 4.3140e-2 (4.76e-2) - & 6.5058e-2 (5.11e-2) - & 3.5637e-2 (4.56e-2) - & 1.7428e-2 (3.59e-2) - & 3.0319e-2 (3.98e-2) - & 1.5631e-2 (2.75e-2) = & {\color[HTML]{3333E9} 8.5201e-3 (4.72e-3)} \\
\textbf{LIRCMOP8} & 4.8842e-2 (6.89e-2) - & 3.9151e-2 (4.87e-2) - & 2.9409e-2 (4.45e-2) = & 1.1603e-2 (2.24e-2) - & 2.2789e-2 (4.26e-2) = & 1.0960e-2 (2.06e-2) = & {\color[HTML]{3333E9} 7.0061e-3 (2.11e-4)} \\
\textbf{LIRCMOP9} & 2.6226e-1 (1.82e-1) - & 2.8653e-1 (1.38e-1) - & 2.5805e-1 (1.79e-1) - & {\color[HTML]{3333E9} 2.9809e-2 (9.16e-2) +} & 1.3158e-1 (1.45e-1) = & 9.1471e-2 (1.65e-1) = & 1.1833e-1 (1.39e-1) \\
\textbf{LIRCMOP10} & {\color[HTML]{3333E9} 5.1300e-3 (5.01e-4) =} & 1.2707e-2 (2.49e-2) - & 5.1524e-3 (5.04e-4) = & 5.4103e-3 (3.04e-4) - & 5.4267e-3 (7.73e-4) - & 1.9201e-2 (5.30e-2) = & 5.2473e-3 (7.75e-4) \\
\textbf{LIRCMOP11} & 2.1254e-2 (4.10e-2) = & 2.5944e-2 (5.61e-2) - & 2.5896e-2 (4.26e-2) = & 5.0239e-3 (1.27e-2) - & 9.5201e-3 (2.14e-2) = & {\color[HTML]{3333E9} 4.8083e-3 (1.28e-2) =} & 4.8358e-3 (1.28e-2) \\
\textbf{LIRCMOP12} & 6.3002e-2 (7.93e-2) - & 6.2094e-2 (8.63e-2) - & 6.0641e-2 (7.55e-2) - & {\color[HTML]{3333E9} 7.1214e-3 (1.64e-2) +} & 3.0648e-2 (4.38e-2) = & 1.7335e-2 (3.42e-2) = & 1.9978e-2 (3.29e-2) \\
\textbf{LIRCMOP13} & 9.7085e-2 (1.94e-3) + & {\color[HTML]{3333E9} 9.6835e-2 (1.35e-3) +} & 1.0198e-1 (1.75e-3) - & 1.2328e-1 (1.43e-2) - & 1.0003e-1 (1.43e-3) - & 9.7333e-2 (1.69e-3) + & 9.8376e-2 (2.07e-3) \\
\textbf{LIRCMOP14} & 9.8263e-2 (1.85e-3) = & {\color[HTML]{3333E9} 9.6551e-2 (9.46e-4) +} & 9.8313e-2 (1.24e-3) = & 1.0920e-1 (5.37e-3) - & 9.7863e-2 (1.11e-3) = & 9.7652e-2 (1.04e-3) = & 9.7740e-2 (1.25e-3) \\
\textbf{DOC1} & 2.9324e-2 (1.34e-1) = & 6.0140e-3 (1.87e-3) + & 5.1161e-3 (3.70e-4) + & 2.9327e-2 (1.34e-1) = & 5.1250e-3 (3.01e-4) + & {\color[HTML]{3333E9} 4.8034e-3 (1.75e-4) =} & 1.0292e-1 (2.54e-1) \\
\textbf{DOC2} & 4.3432e-3 (1.50e-3) = & NaN (NaN) & 1.3766e-2 (2.37e-2) - & 4.2475e-3 (1.94e-3) = & NaN (NaN) & {\color[HTML]{3333E9} 4.0887e-3 (1.20e-3) =} & 4.5134e-3 (2.12e-3) \\
\textbf{DOC3} & {\color[HTML]{3333E9} 9.0094e-3 (6.36e-3) =} & 6.6769e+2 (2.49e+2) - & 3.2353e+2 (1.97e+2) - & 1.3186e-2 (1.36e-2) = & 5.6579e+2 (2.21e+2) - & 2.4948e+0 (1.36e+1) = & 1.7471e+0 (9.51e+0) \\
\textbf{DOC4} & 1.9814e-2 (1.13e-2) = & 7.6780e-2 (9.46e-2) - & 2.0934e-2 (8.64e-3) - & 1.6202e-2 (1.17e-2) = & 4.6382e-2 (2.69e-2) - & 1.6257e-2 (8.12e-3) = & {\color[HTML]{3333E9} 1.5676e-2 (9.49e-3)} \\
\textbf{DOC5} & 5.2072e+1 (5.86e+1) = & 1.3556e+2 (4.35e-1) - & 7.3214e+1 (5.66e+1) = & 7.2028e+1 (6.61e+1) = & 1.3054e+2 (0.00e+0) = & 7.3657e+1 (6.34e+1) = & {\color[HTML]{3333E9} 4.4857e+1 (6.16e+1)} \\
\textbf{DOC6} & 2.3359e+1 (2.61e+1) - & 2.2811e+1 (1.62e+1) - & 2.1622e+1 (9.95e+0) - & 2.1219e+1 (1.15e+1) - & 2.4167e+1 (1.38e+1) - & 1.8239e+1 (6.81e+0) - & {\color[HTML]{3333E9} 2.2888e-1 (5.97e-1)} \\
\textbf{DOC7} & 1.7141e+0 (1.39e+0) = & 4.5840e+0 (1.95e+0) - & {\color[HTML]{3333E9} 1.5637e+0 (1.76e+0) =} & 2.2216e+0 (1.49e+0) = & 2.5538e+0 (1.40e+0) = & 2.5354e+0 (1.36e+0) = & 2.2145e+0 (1.66e+0) \\
\textbf{DOC8} & 1.0826e-1 (1.45e-2) + & 4.2373e+1 (5.86e+1) - & {\color[HTML]{3333E9} 7.0157e-2 (5.52e-3) +} & 1.1907e-1 (1.13e-2) = & 3.0161e+1 (4.35e+1) - & 1.2812e-1 (3.15e-2) = & 1.2051e-1 (2.00e-2) \\
\textbf{DOC9} & 1.7955e-1 (8.87e-2) = & 1.2936e-1 (8.82e-2) + & {\color[HTML]{3333E9} 1.1419e-1 (8.29e-2) +} & 1.5688e-1 (7.91e-2) = & 1.3399e-1 (8.40e-2) = & 1.5411e-1 (7.73e-2) = & 1.5729e-1 (8.17e-2) \\
\textbf{FCP1} & 3.2536e-2 (2.83e-4) = & {\color[HTML]{3333E9} 3.1839e-2 (3.52e-4) +} & 3.1931e-2 (2.36e-4) + & 3.2680e-2 (4.42e-4) = & 3.2267e-2 (5.15e-4) + & 3.2671e-2 (3.51e-4) = & 3.2598e-2 (3.27e-4) \\
\textbf{FCP2} & 2.7325e-2 (6.82e-4) = & {\color[HTML]{3333E9} 2.6147e-2 (5.42e-4) +} & 2.6302e-2 (5.62e-4) + & 2.7264e-2 (6.10e-4) = & 2.6489e-2 (8.16e-4) + & 2.7410e-2 (6.97e-4) = & 2.7052e-2 (5.19e-4) \\
\textbf{FCP3} & 3.5348e-2 (2.42e-4) = & {\color[HTML]{3333E9} 3.4921e-2 (2.51e-4) +} & 3.5139e-2 (2.63e-4) + & 3.5408e-2 (2.05e-4) = & 3.4948e-2 (3.14e-4) + & 3.5292e-2 (2.51e-4) = & 3.5384e-2 (2.51e-4) \\
\textbf{FCP4} & 2.5221e-2 (3.24e-4) = & 2.4991e-2 (2.50e-4) + & {\color[HTML]{3333E9} 2.4833e-2 (3.31e-4) +} & 2.5178e-2 (3.83e-4) = & 2.4961e-2 (4.02e-4) + & 2.5139e-2 (3.56e-4) = & 2.5262e-2 (3.28e-4) \\
\textbf{FCP5} & 1.5196e-2 (2.14e-3) = & 1.4130e-2 (1.86e-3) + & 1.4991e-2 (2.96e-3) = & 1.4401e-2 (6.21e-4) = & {\color[HTML]{3333E9} 1.3598e-2 (8.97e-4) +} & 1.4513e-2 (7.90e-4) = & 1.4698e-2 (1.17e-3) \\ \hline
\textbf{+/-/=} & 2/13/46 & 16/26/19 & 14/19/28 & 4/27/30 & 13/21/27 & 1/5/55 &  \\ \hline
\end{tabular}%
}
\vspace{5cm}
\end{table}

\clearpage
\section{The convergence curve of IGD values on MW and LIRCMOP.} \label{sec:A5}

\subsection{The convergence curve of IGD values on LIRCMOP.} \label{sec:A5-1}
\subsubsection{Convergence}

\begin{figure}[htpb!]
    \centering
    \subfloat[]{\resizebox{0.24\textwidth}{!}{\includegraphics{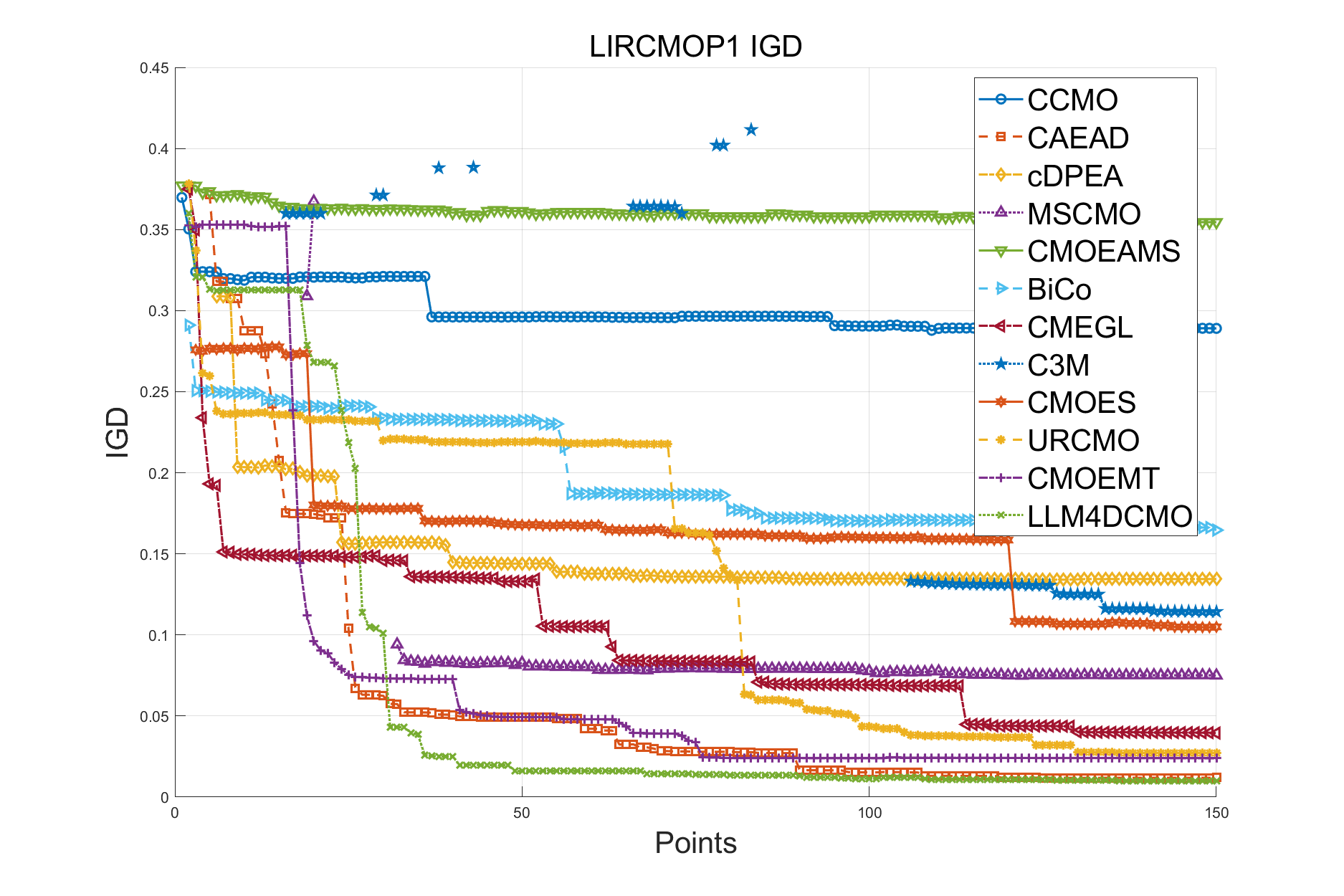}}\label{fig:sub1l}}
    \hfill
    \subfloat[]{\resizebox{0.24\textwidth}{!}{\includegraphics{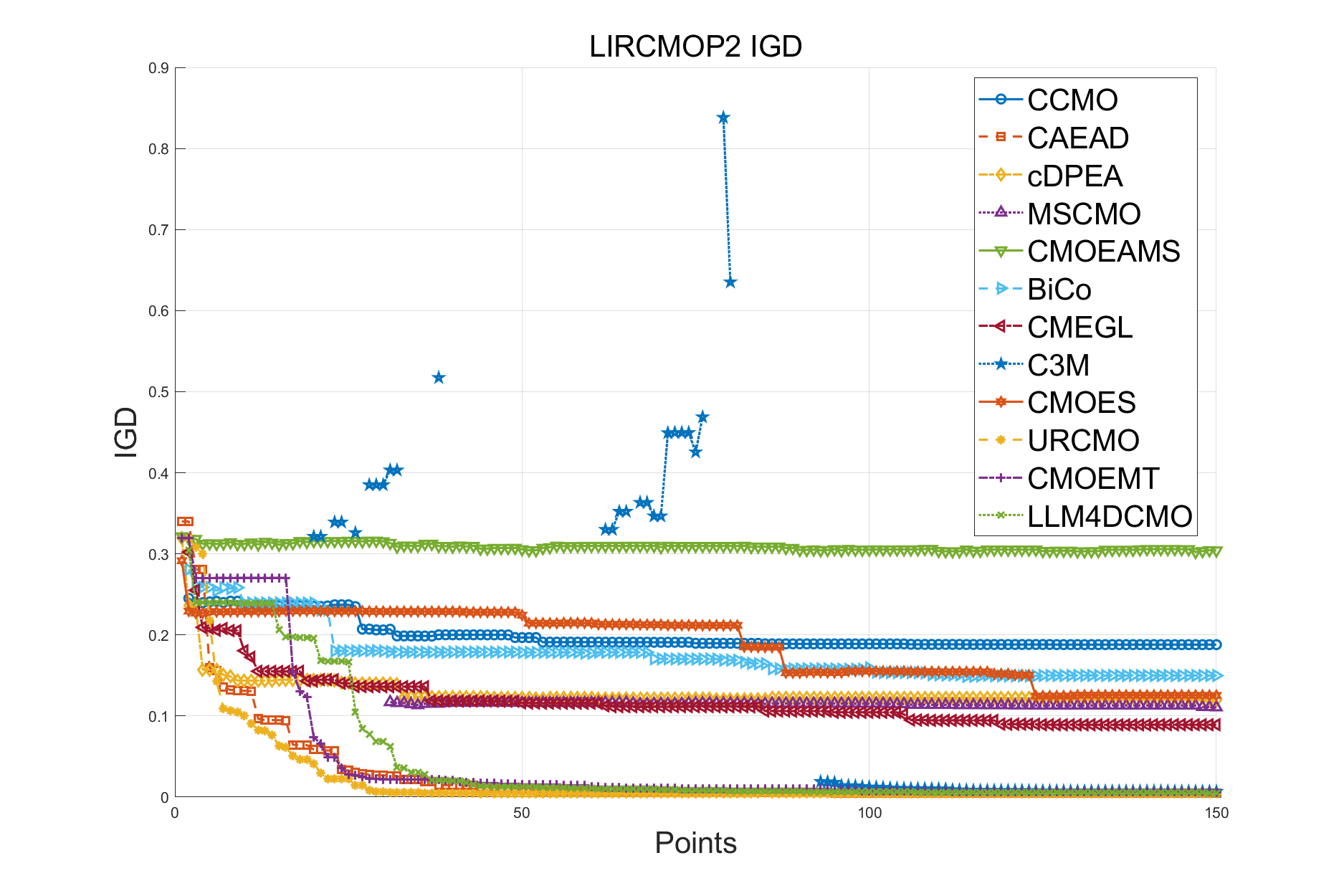}}\label{fig:sub2l}}
    \hfill
    \subfloat[]{\resizebox{0.24\textwidth}{!}{\includegraphics{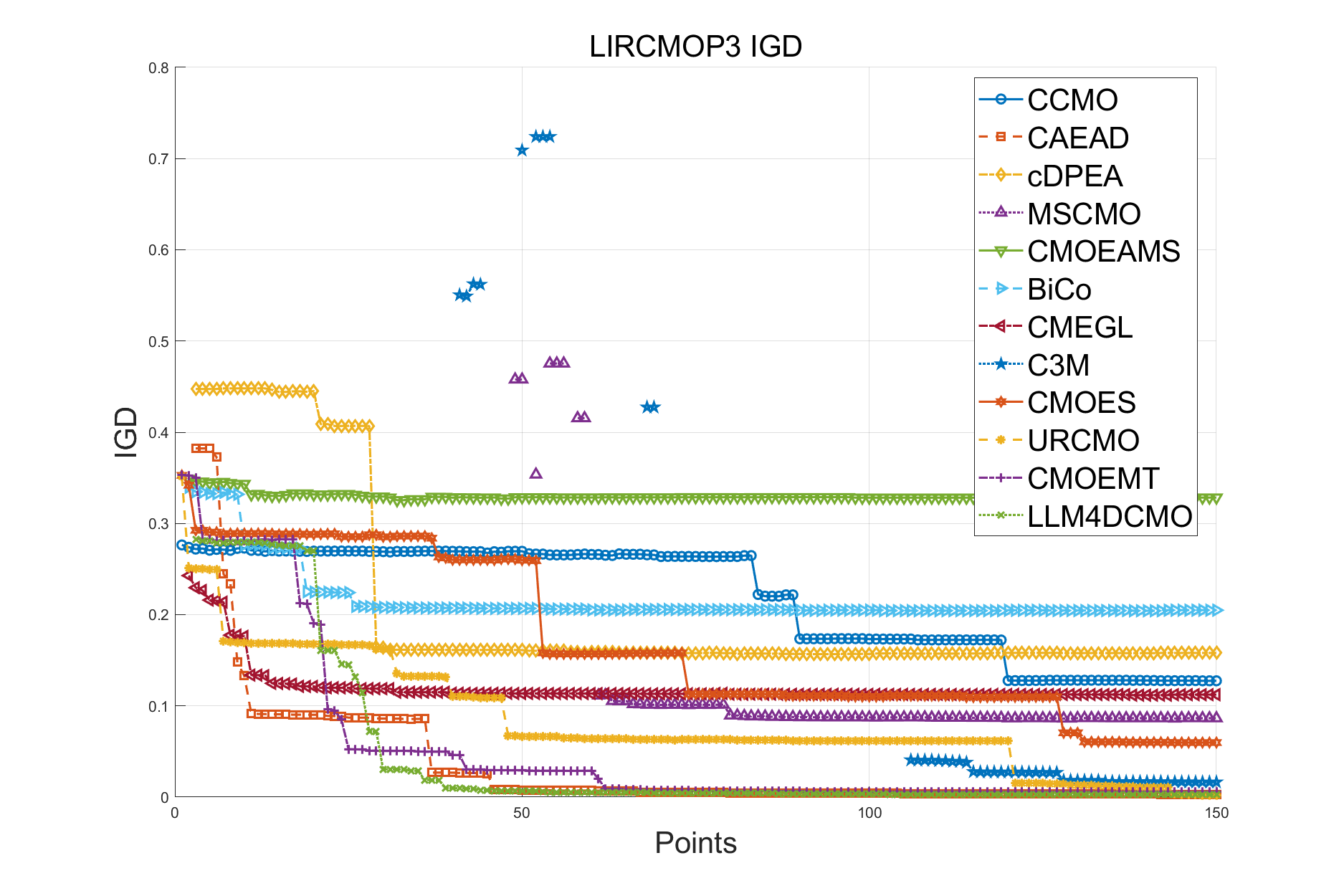}}\label{fig:sub3l}}
    \hfill
    \subfloat[]{\resizebox{0.24\textwidth}{!}{\includegraphics{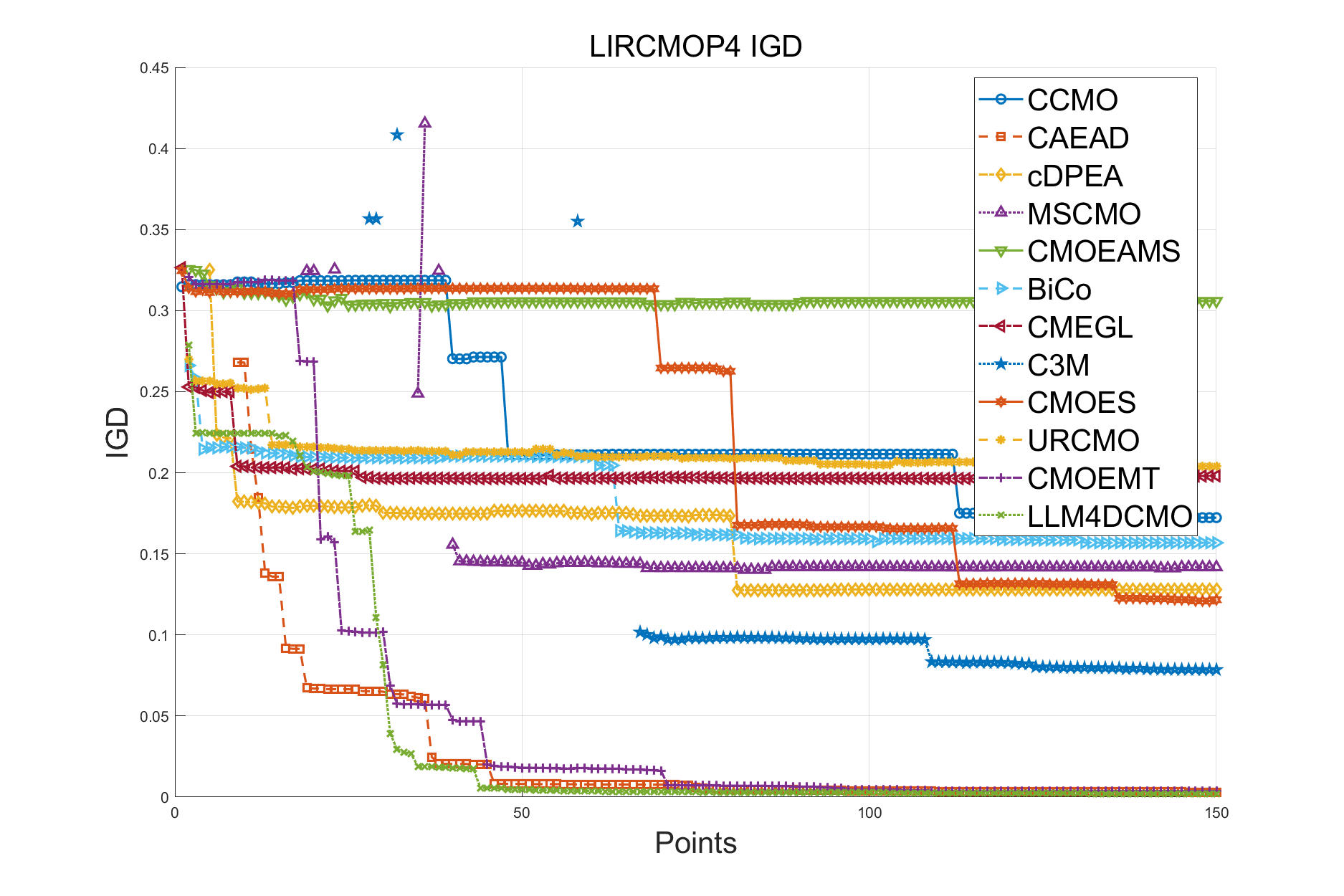}}\label{fig:sub4l}}\\
    \subfloat[]{\resizebox{0.24\textwidth}{!}{\includegraphics{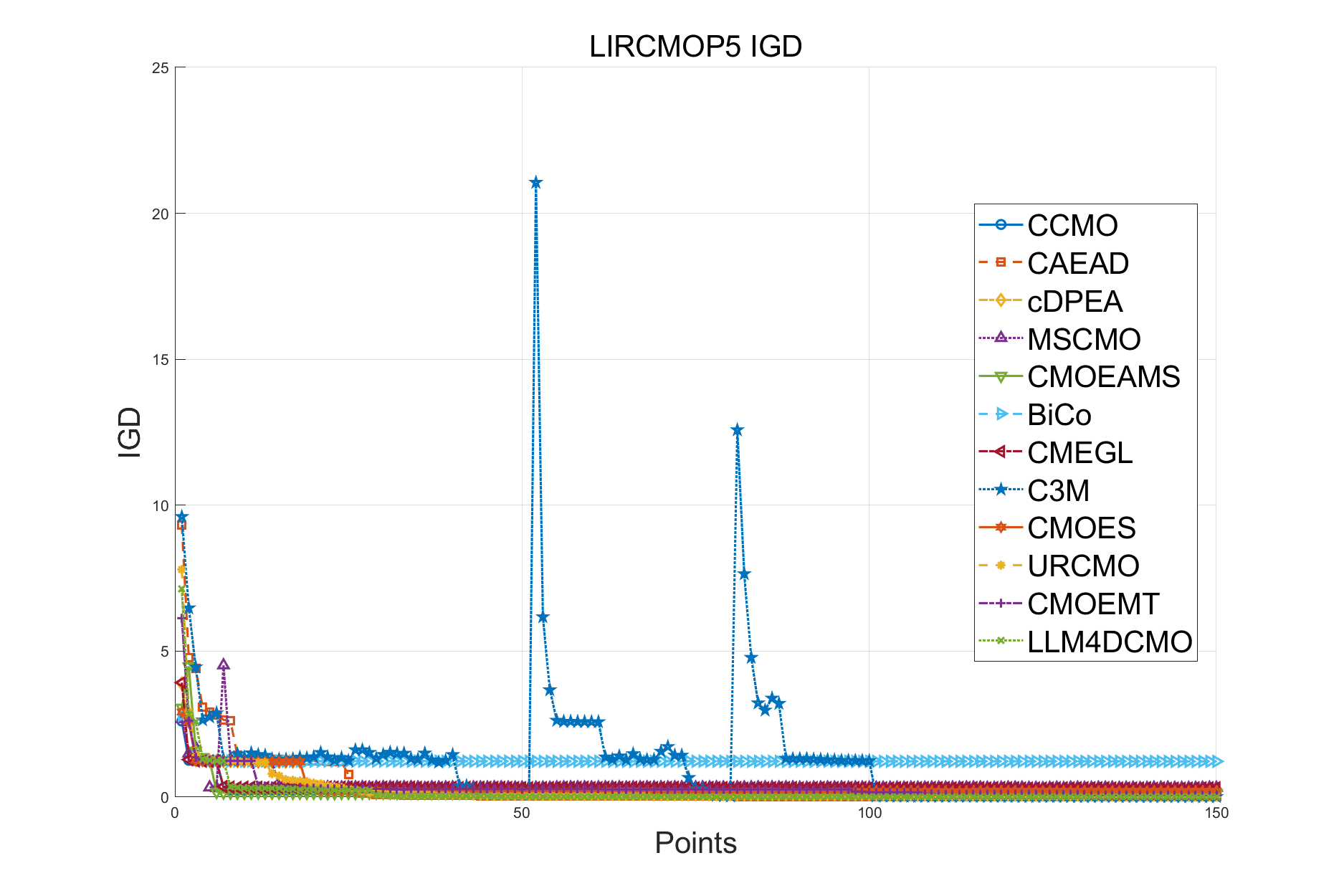}}\label{fig:sub5l}}
    \hfill
    \subfloat[]{\resizebox{0.24\textwidth}{!}{\includegraphics{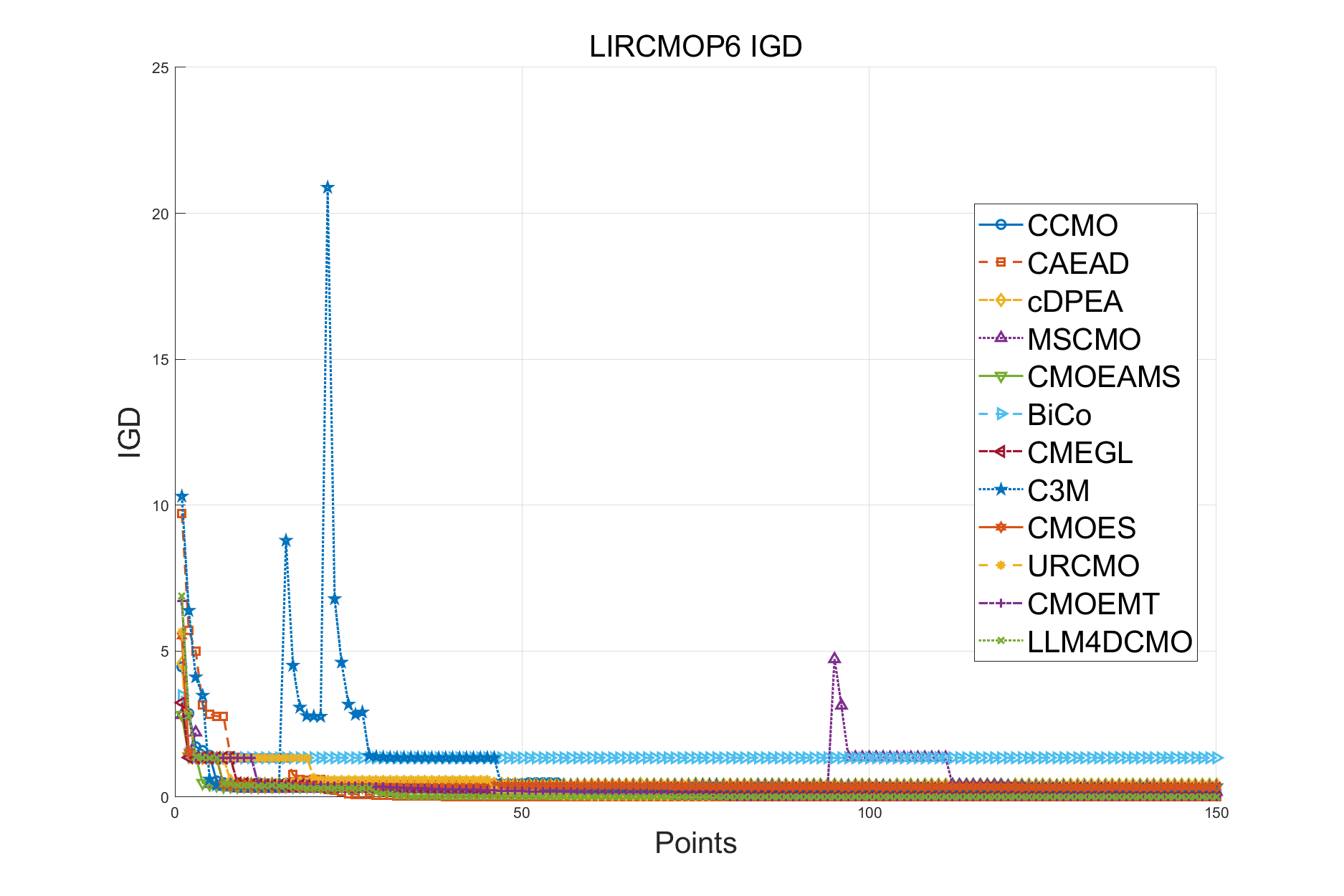}}\label{fig:sub6l}}
    \hfill
    \subfloat[]{\resizebox{0.24\textwidth}{!}{\includegraphics{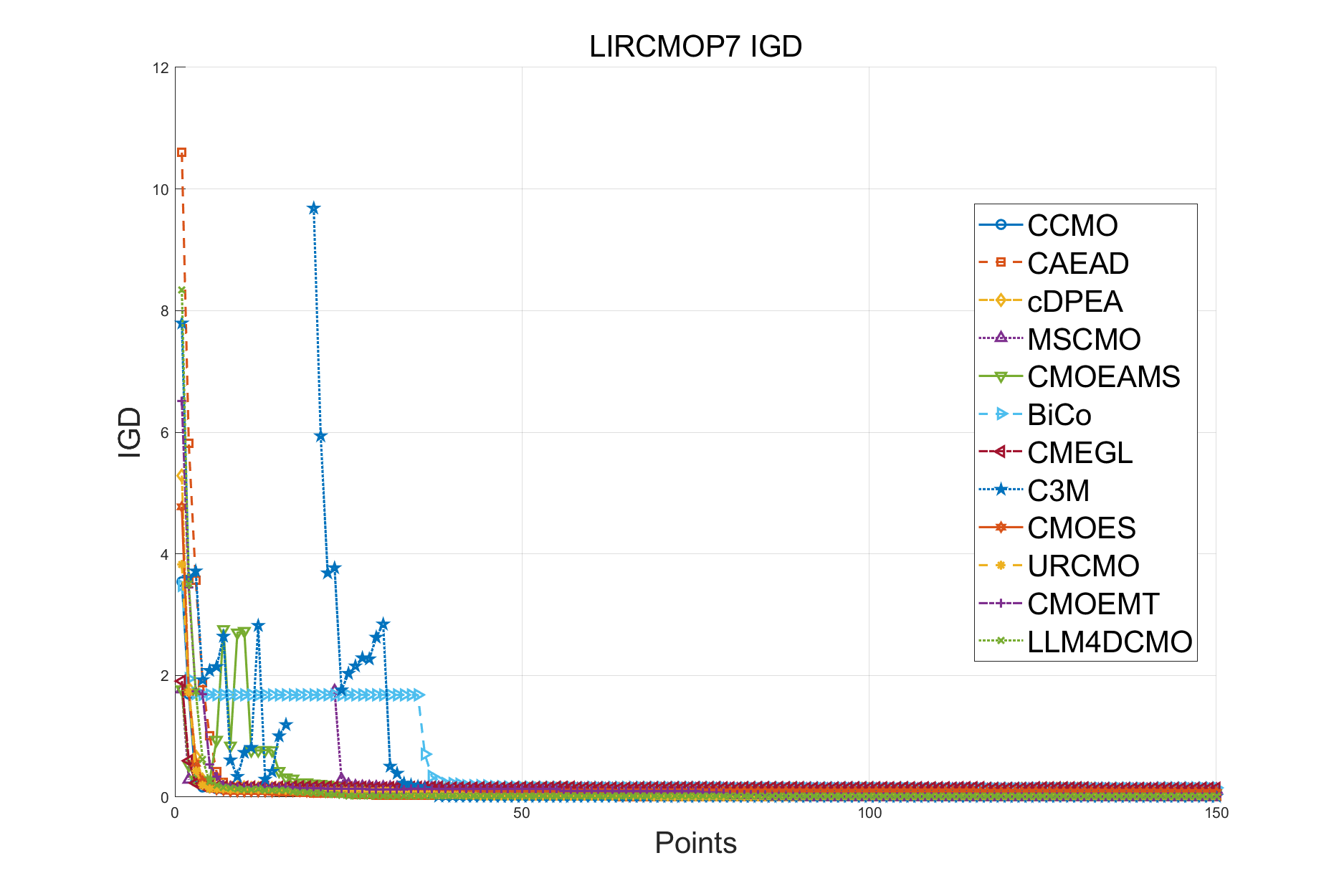}}\label{fig:sub7l}}
    \hfill
    \subfloat[]{\resizebox{0.24\textwidth}{!}{\includegraphics{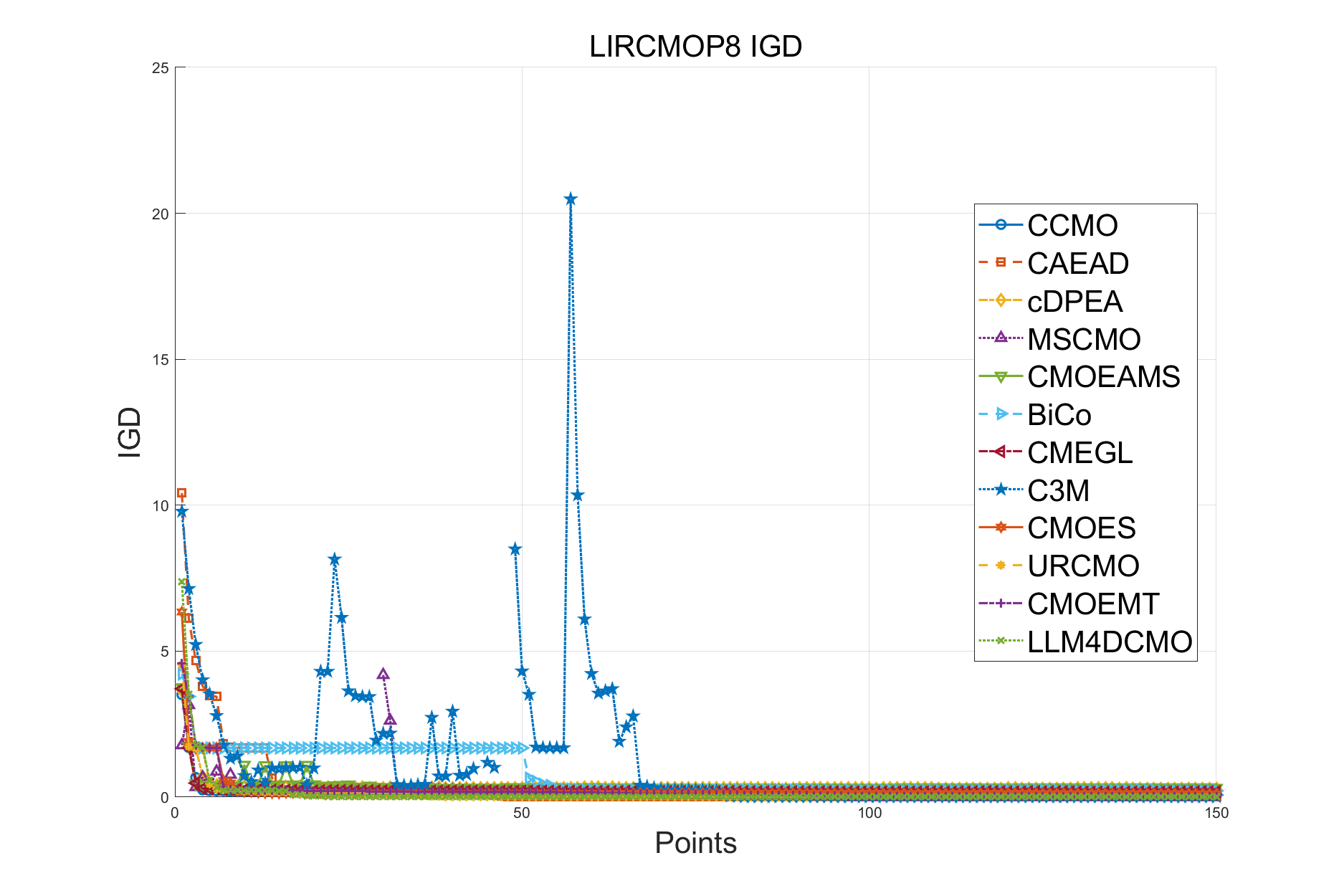}}\label{fig:sub8l}}\\
    \subfloat[]{\resizebox{0.24\textwidth}{!}{\includegraphics{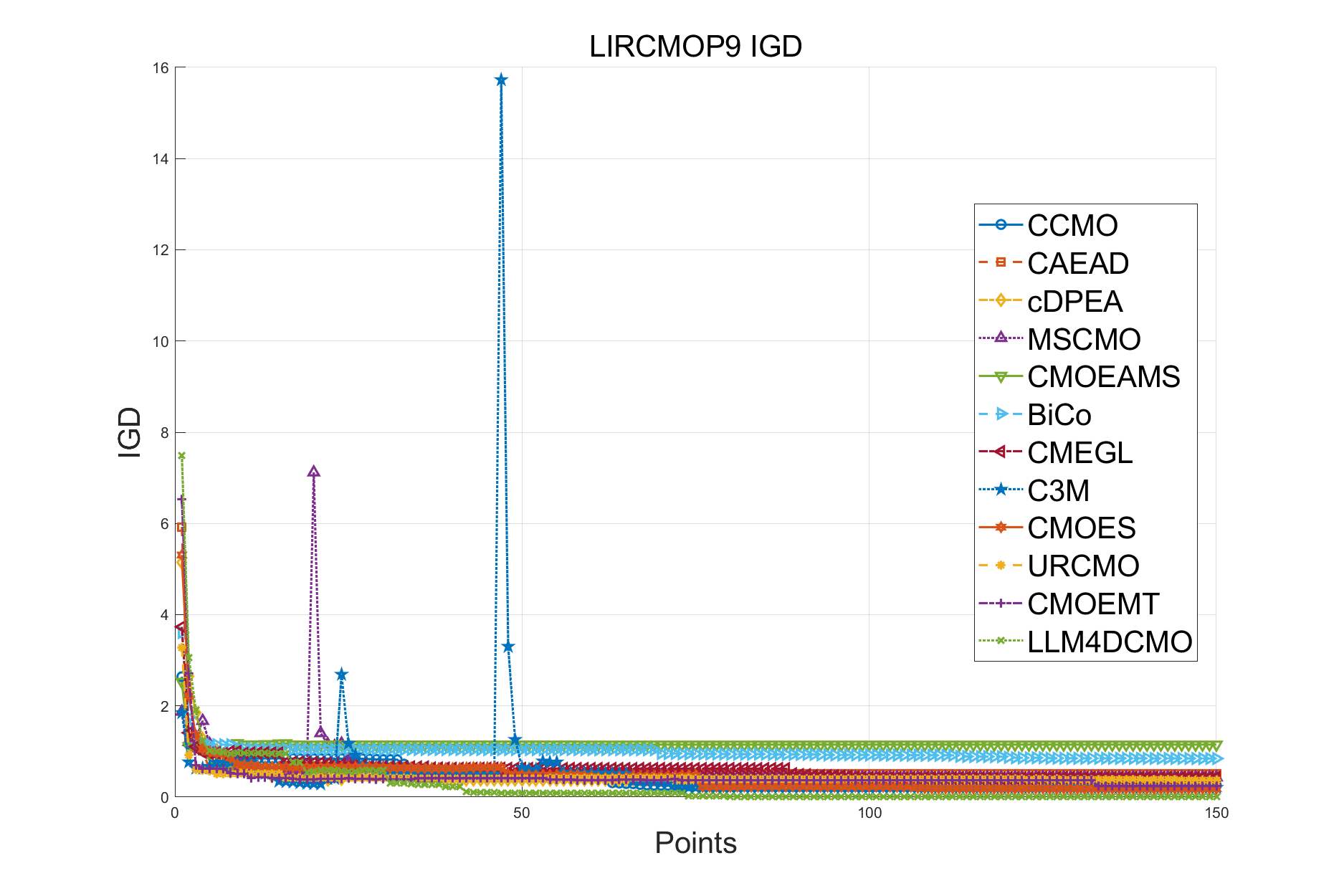}}\label{fig:sub9l}}
    \hfill
    \subfloat[]{\resizebox{0.24\textwidth}{!}{\includegraphics{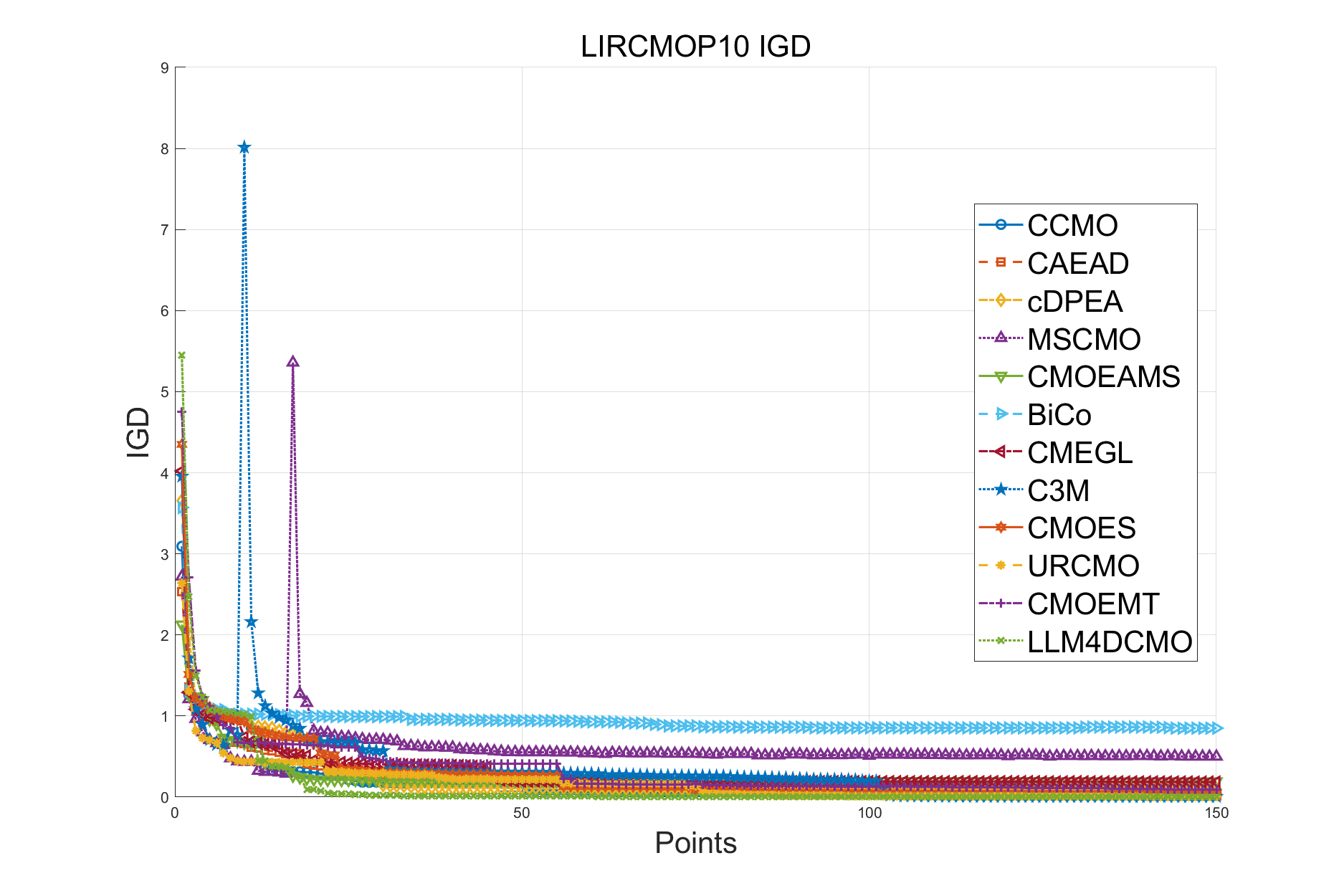}}\label{fig:sub10l}}
    \hfill
    \subfloat[]{\resizebox{0.24\textwidth}{!}{\includegraphics{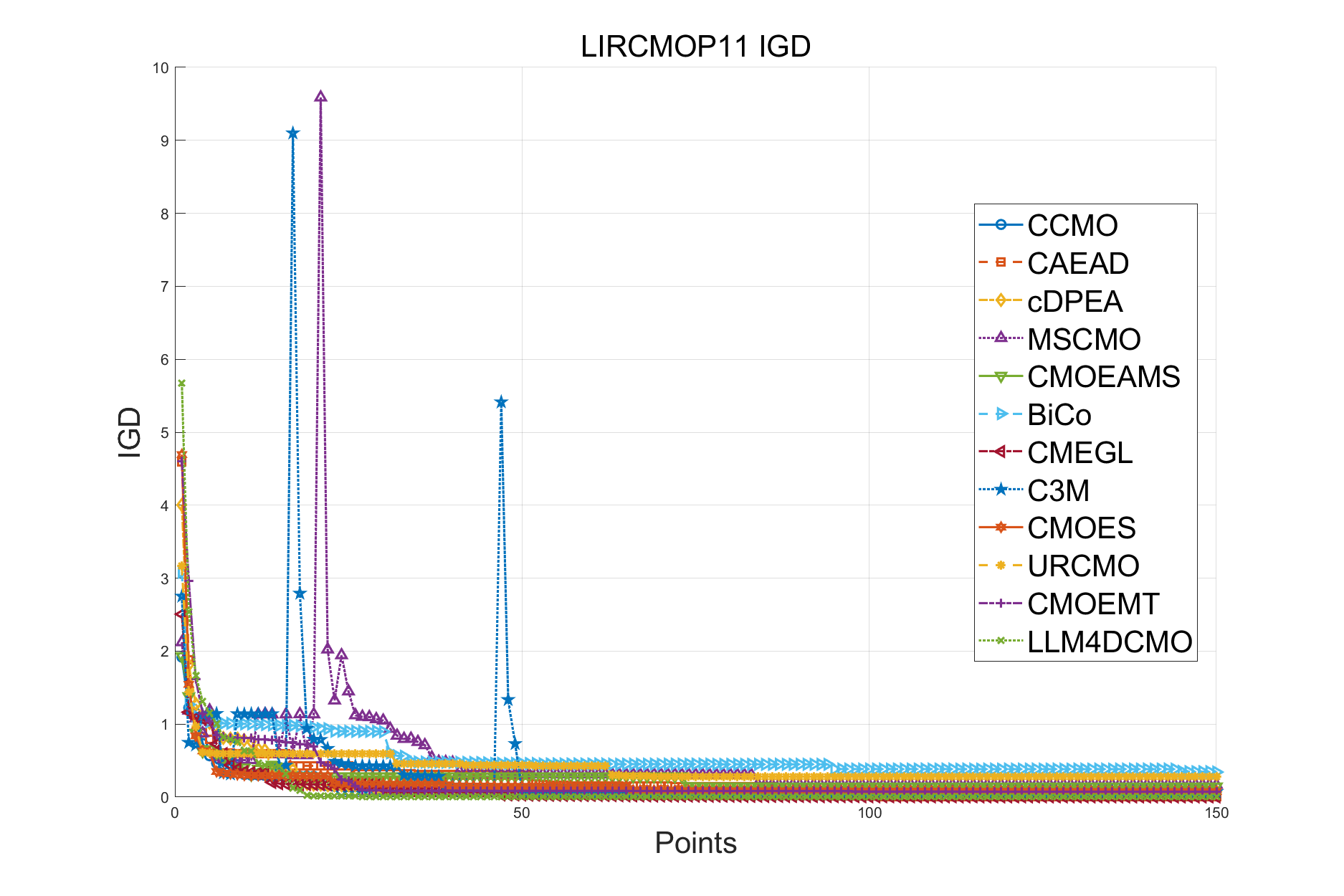}}\label{fig:sub11l}}
    \hfill
    \subfloat[]{\resizebox{0.24\textwidth}{!}{\includegraphics{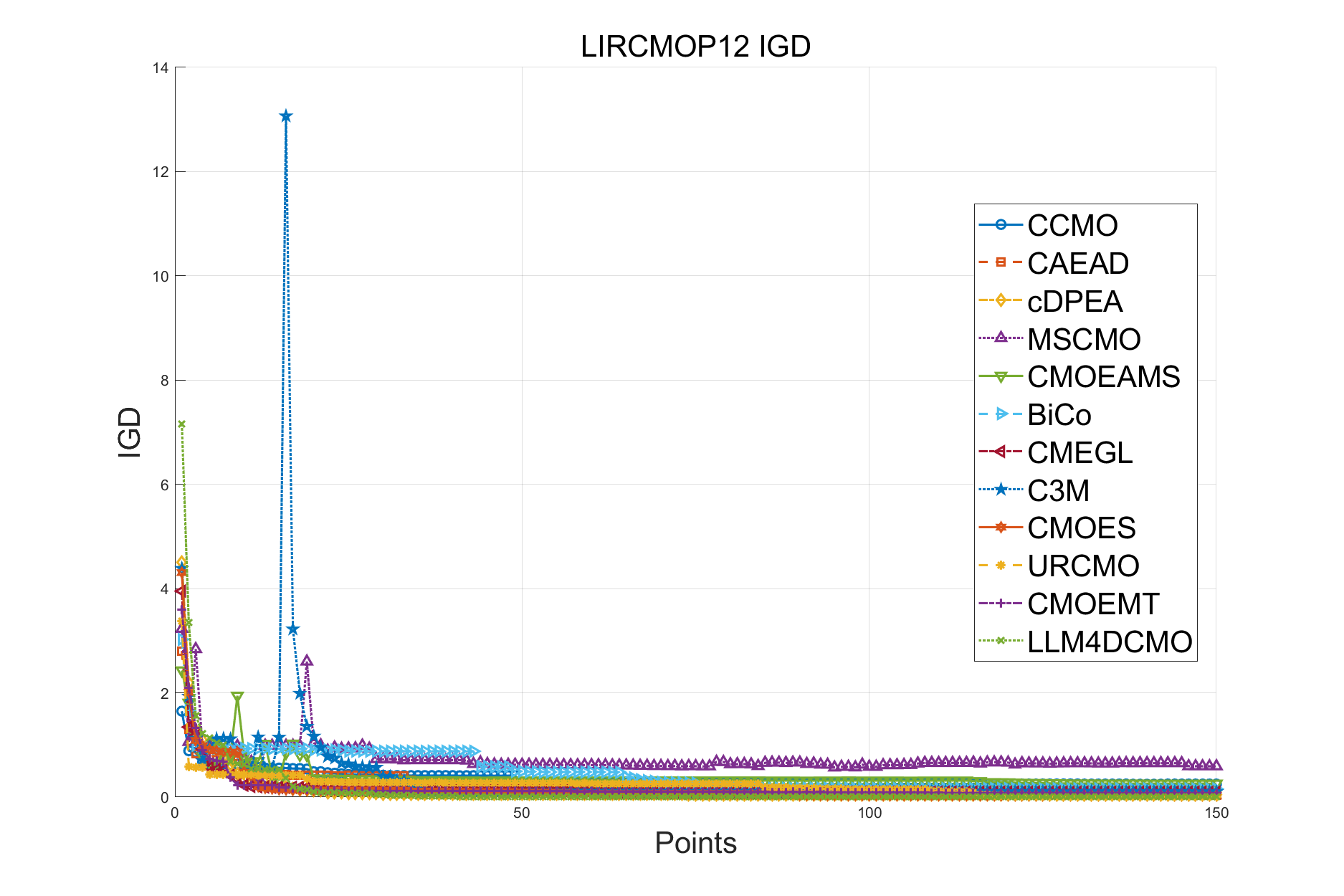}}\label{fig:sub12l}}\\
    \subfloat[]{\resizebox{0.24\textwidth}{!}{\includegraphics{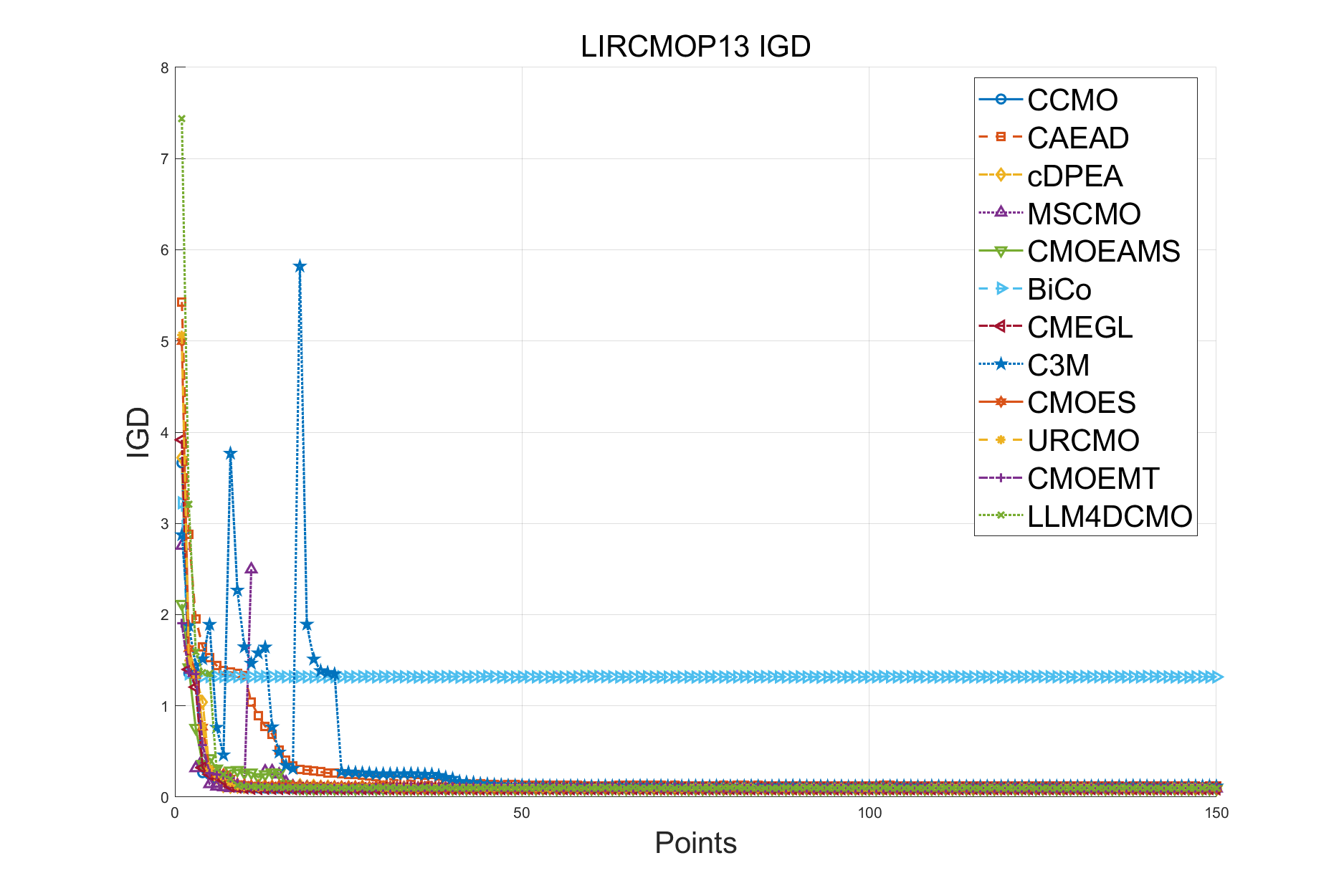}}\label{fig:sub13l}}
    \subfloat[]{\resizebox{0.24\textwidth}{!}{\includegraphics{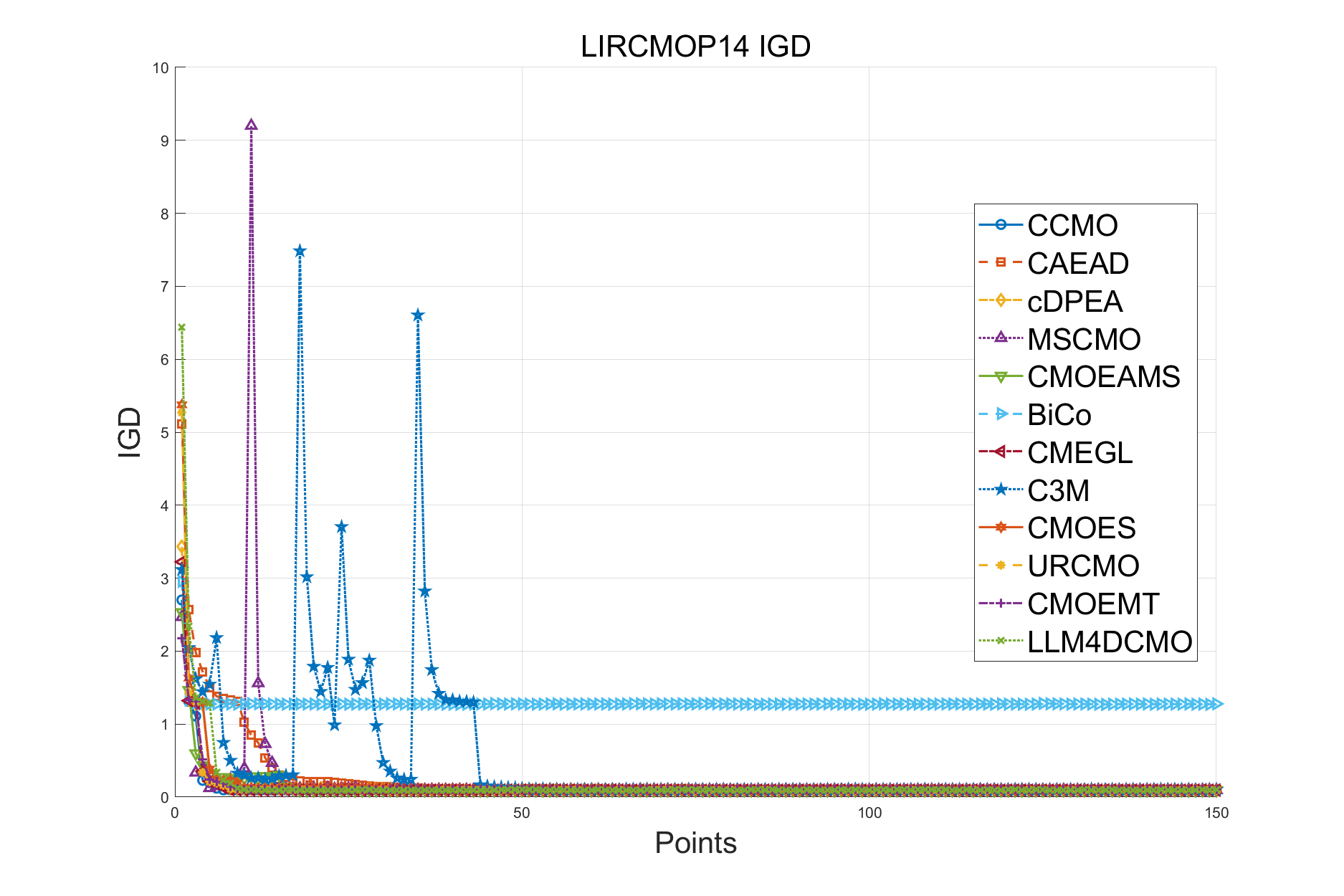}}\label{fig:sub14l}}
    \hfill
    \hfill
    \caption{The convergence curves of IGD metric on LIRCMOP test suite.}
    \label{fig:covLIRs}
\end{figure}

\begin{figure}[htbp!]
    \centering
    \subfloat[]{\includegraphics[width=0.24\textwidth]{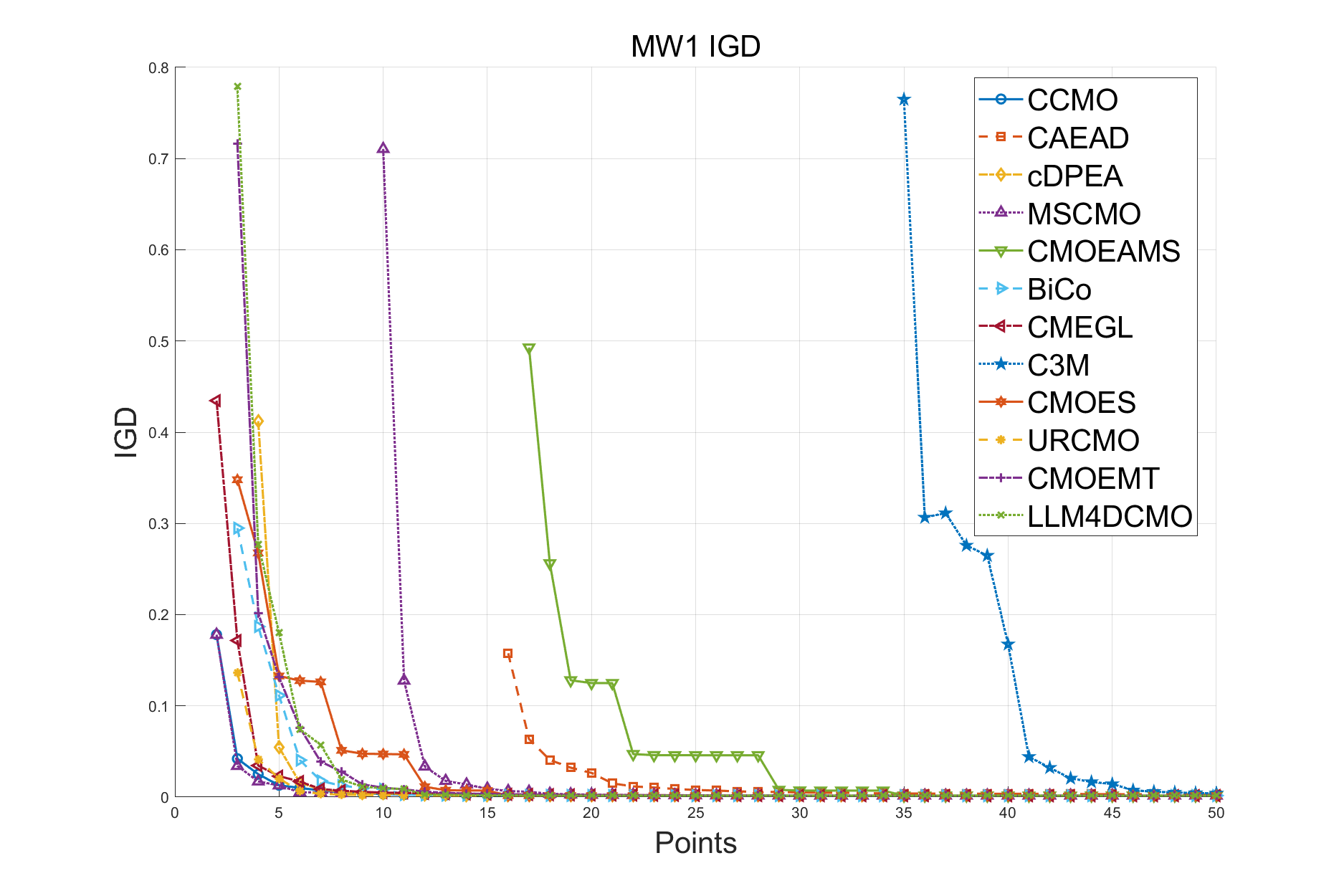}\label{fig:sub1m}}
    \hfill
    \subfloat[]{\includegraphics[width=0.24\textwidth]{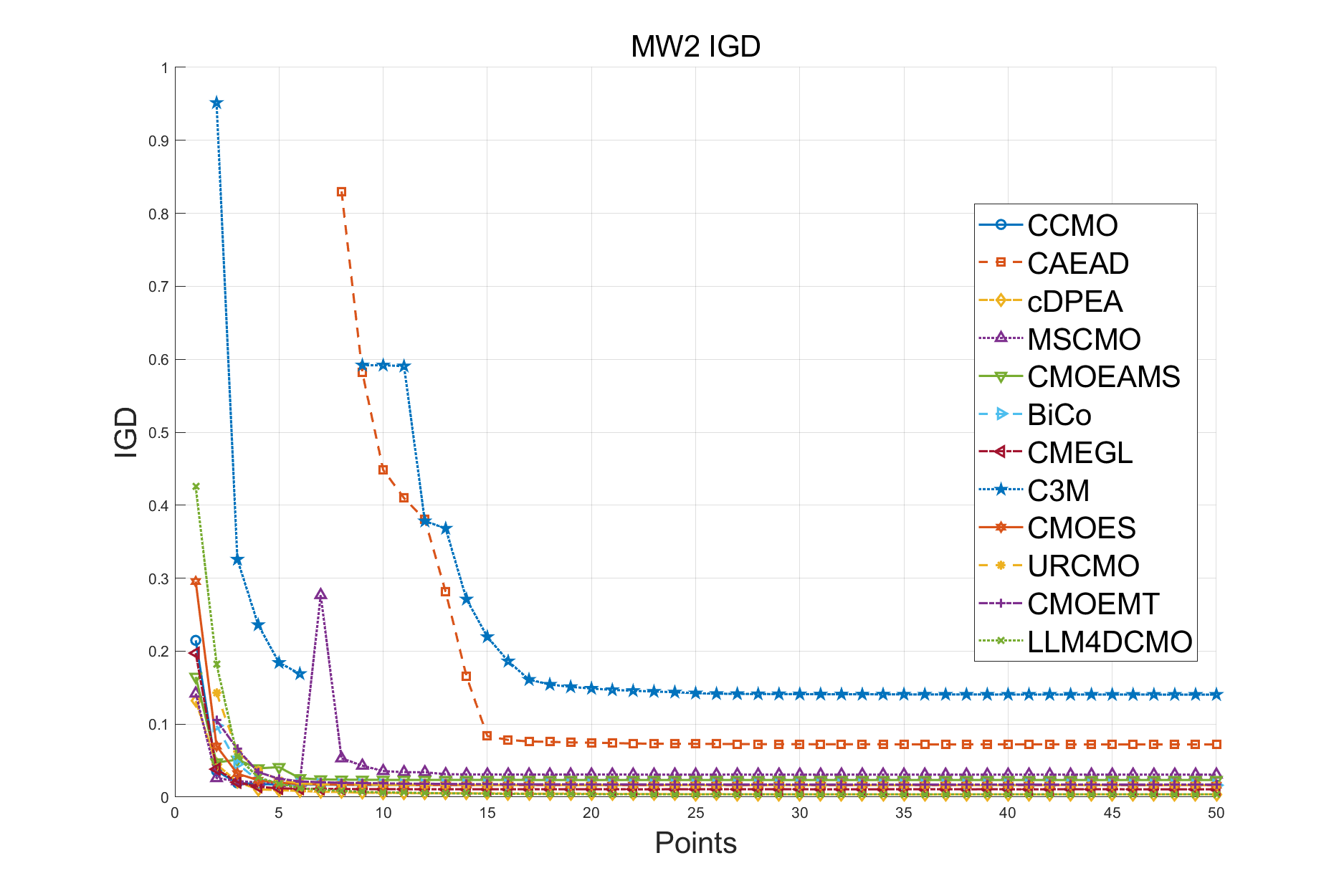}\label{fig:sub2m}}
    \hfill
    \subfloat[]{\includegraphics[width=0.24\textwidth]{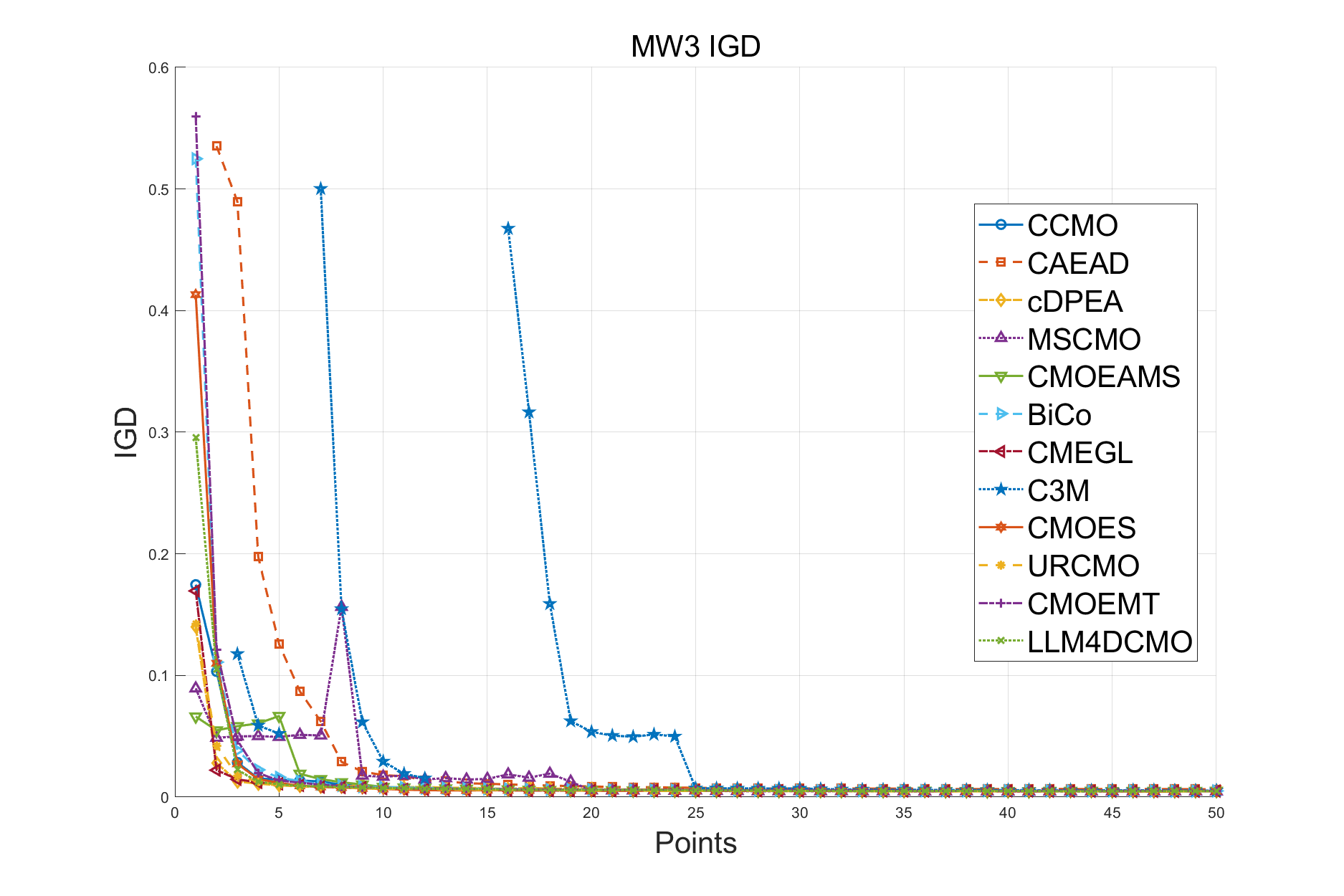}\label{fig:sub3m}}
    \hfill
    \subfloat[]{\includegraphics[width=0.24\textwidth]{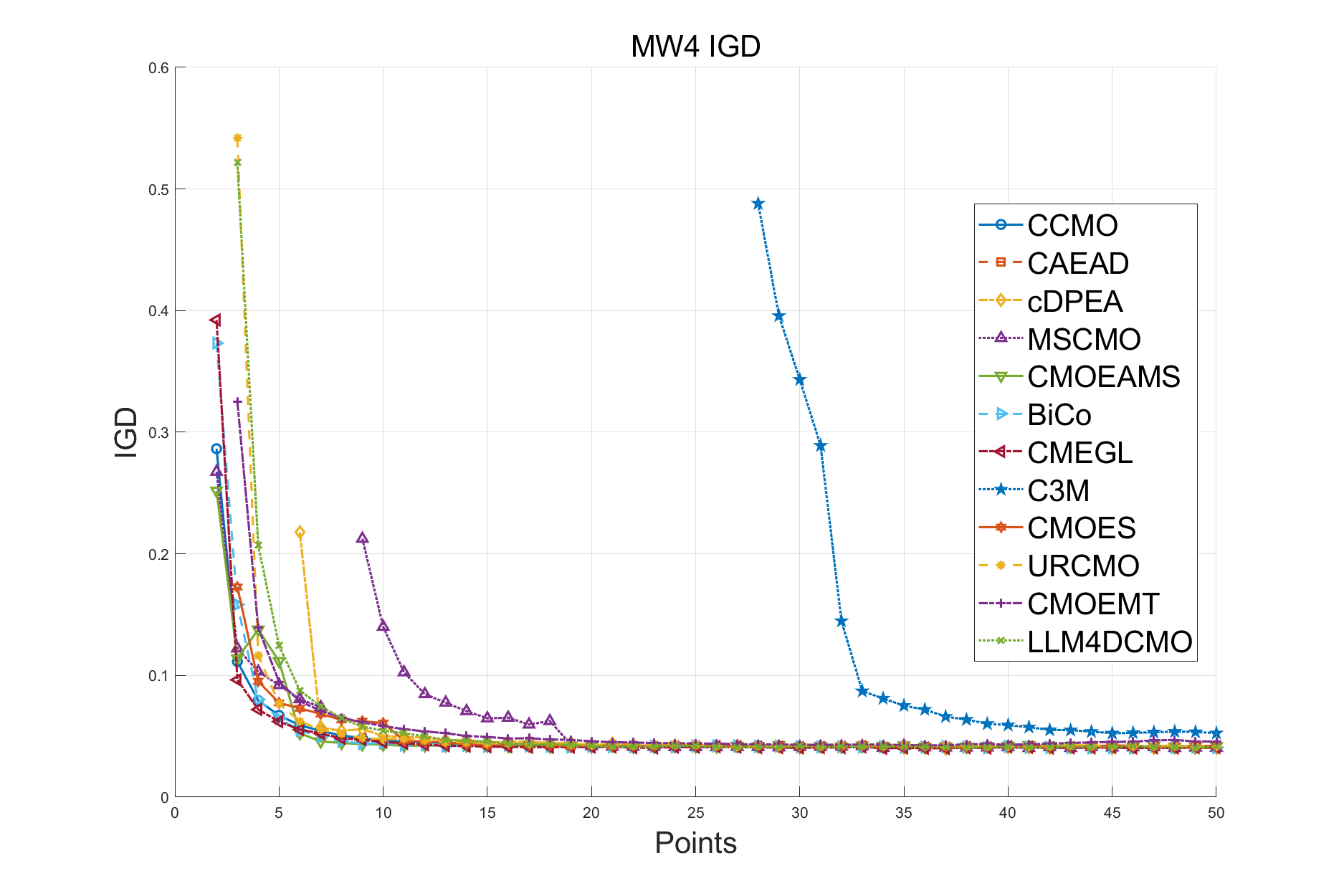}\label{fig:sub4m}}
    \\[1ex]
    
    \subfloat[]{\includegraphics[width=0.24\textwidth]{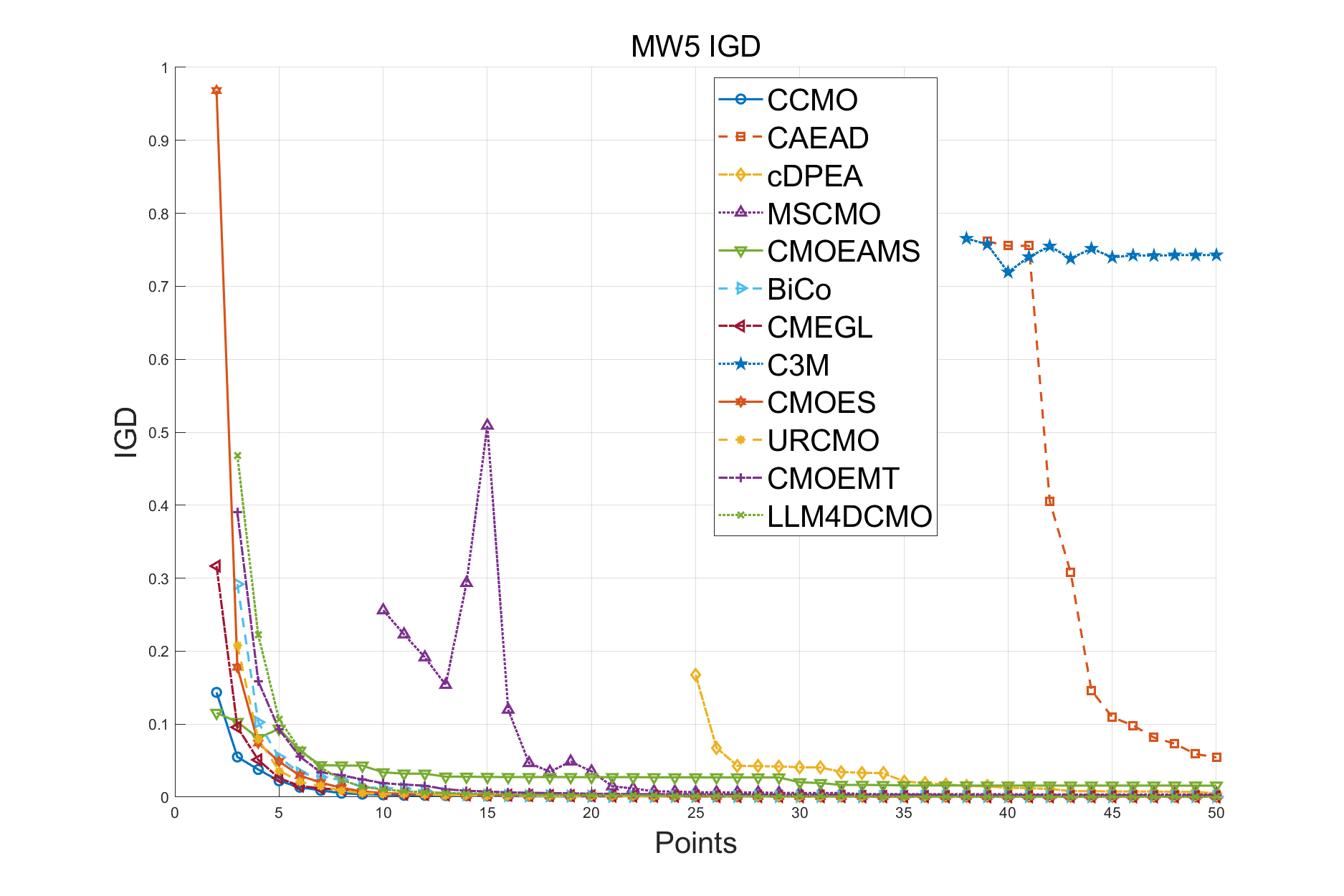}\label{fig:sub5m}}
    \hfill
    \subfloat[]{\includegraphics[width=0.24\textwidth]{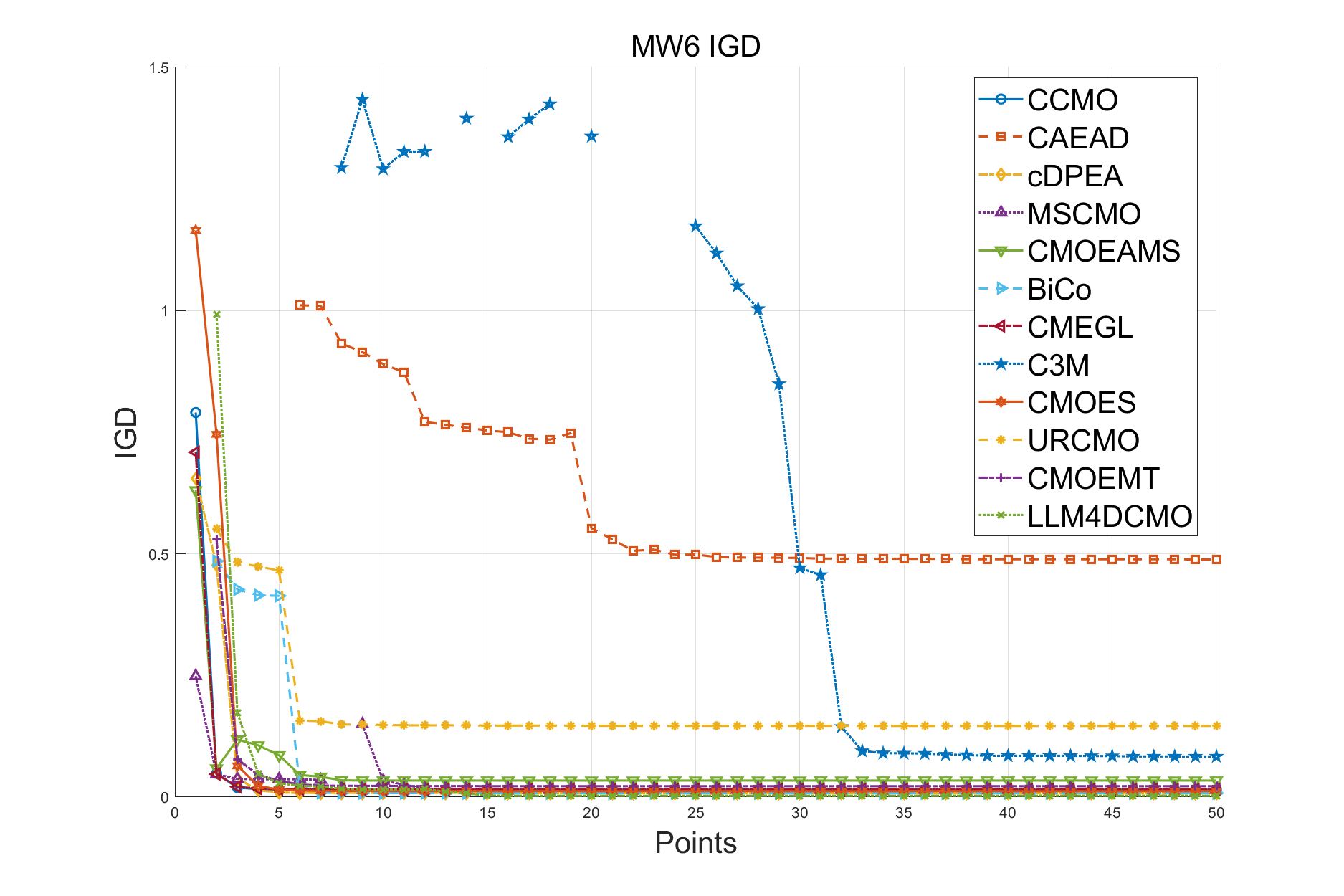}\label{fig:sub6m}}
    \hfill
    \subfloat[]{\includegraphics[width=0.24\textwidth]{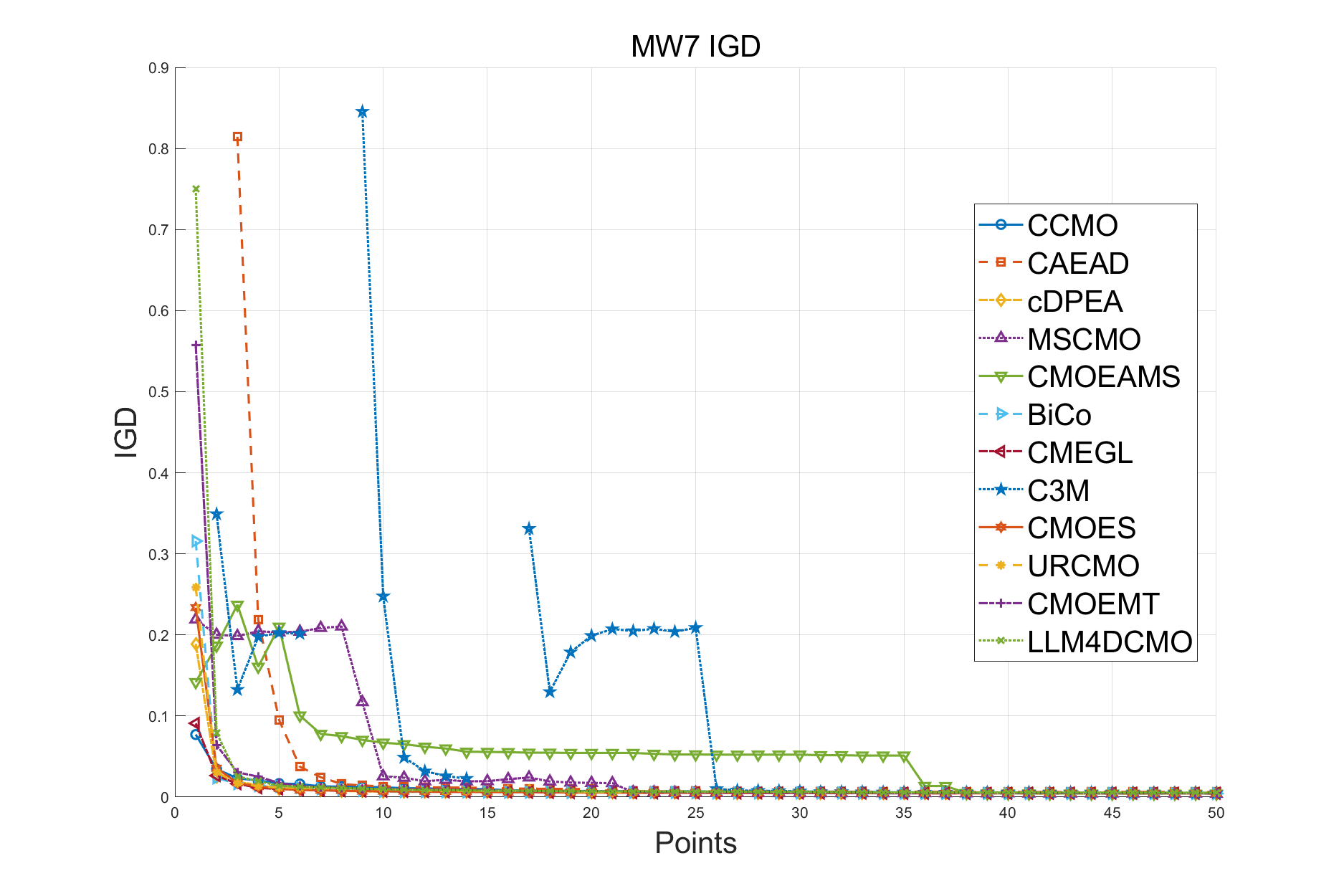}\label{fig:sub7m}}
    \hfill
    \subfloat[]{\includegraphics[width=0.24\textwidth]{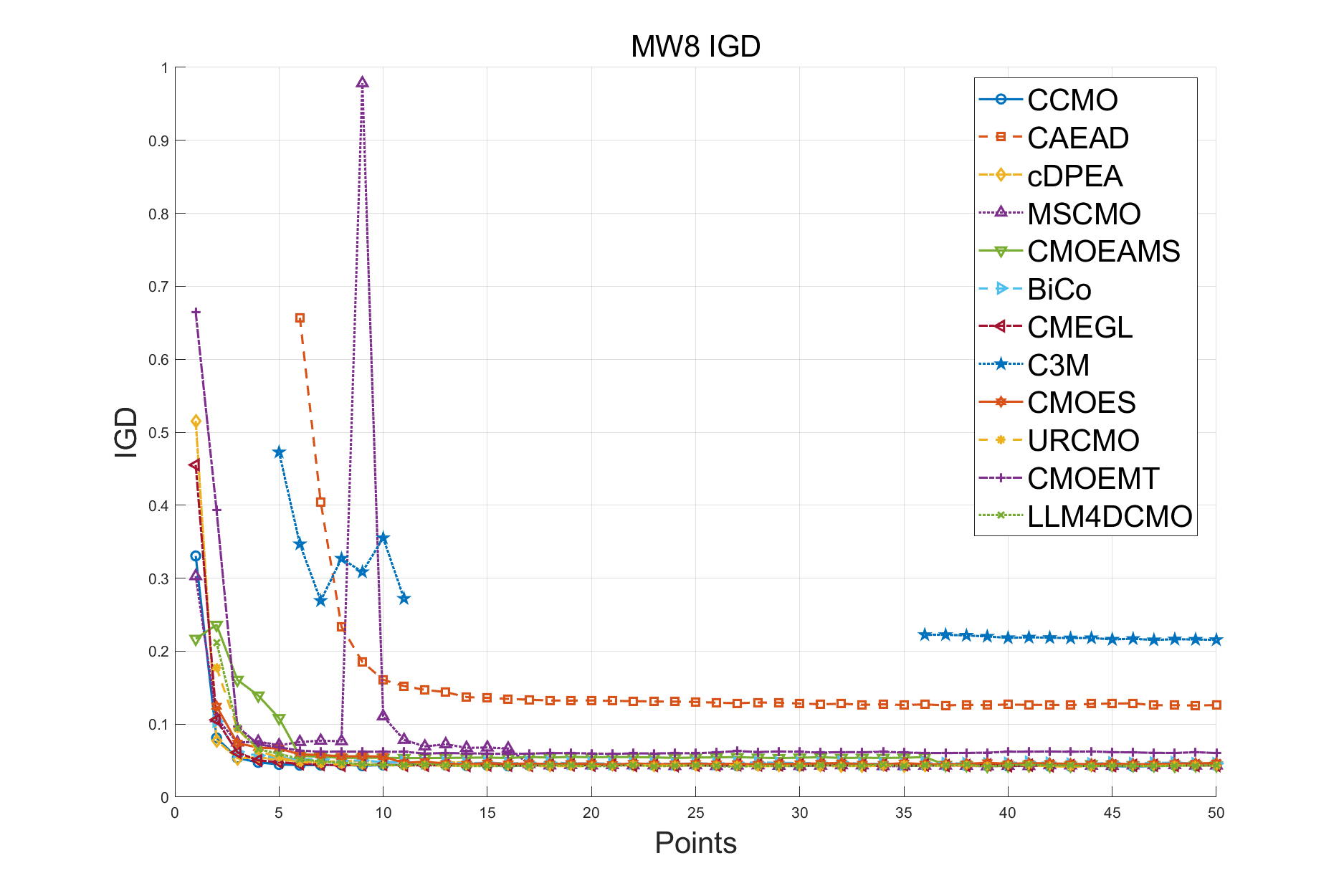}\label{fig:sub8m}}
    \\[1ex]
    
    \subfloat[]{\includegraphics[width=0.24\textwidth]{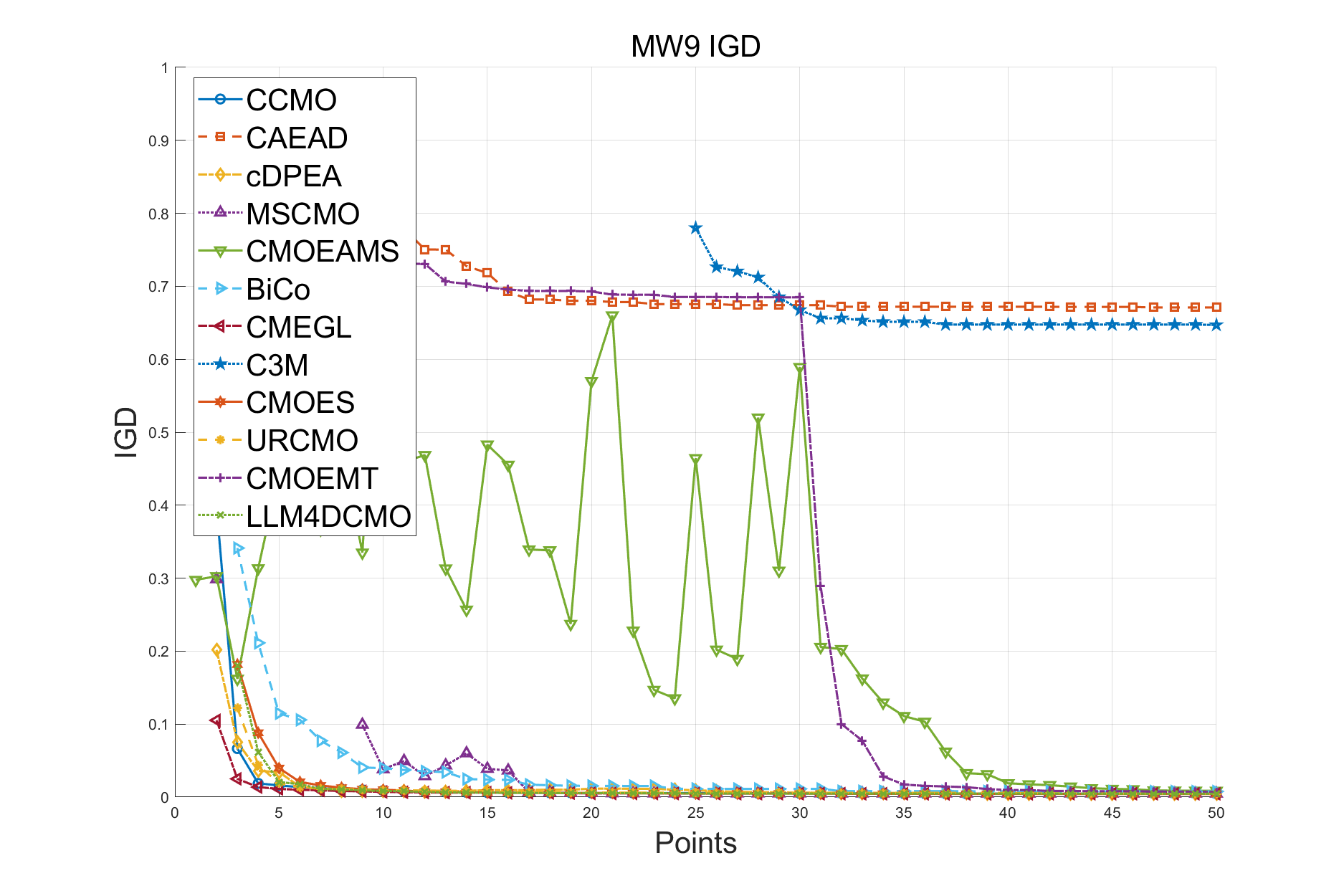}\label{fig:sub9m}}
    \hfill
    \subfloat[]{\includegraphics[width=0.24\textwidth]{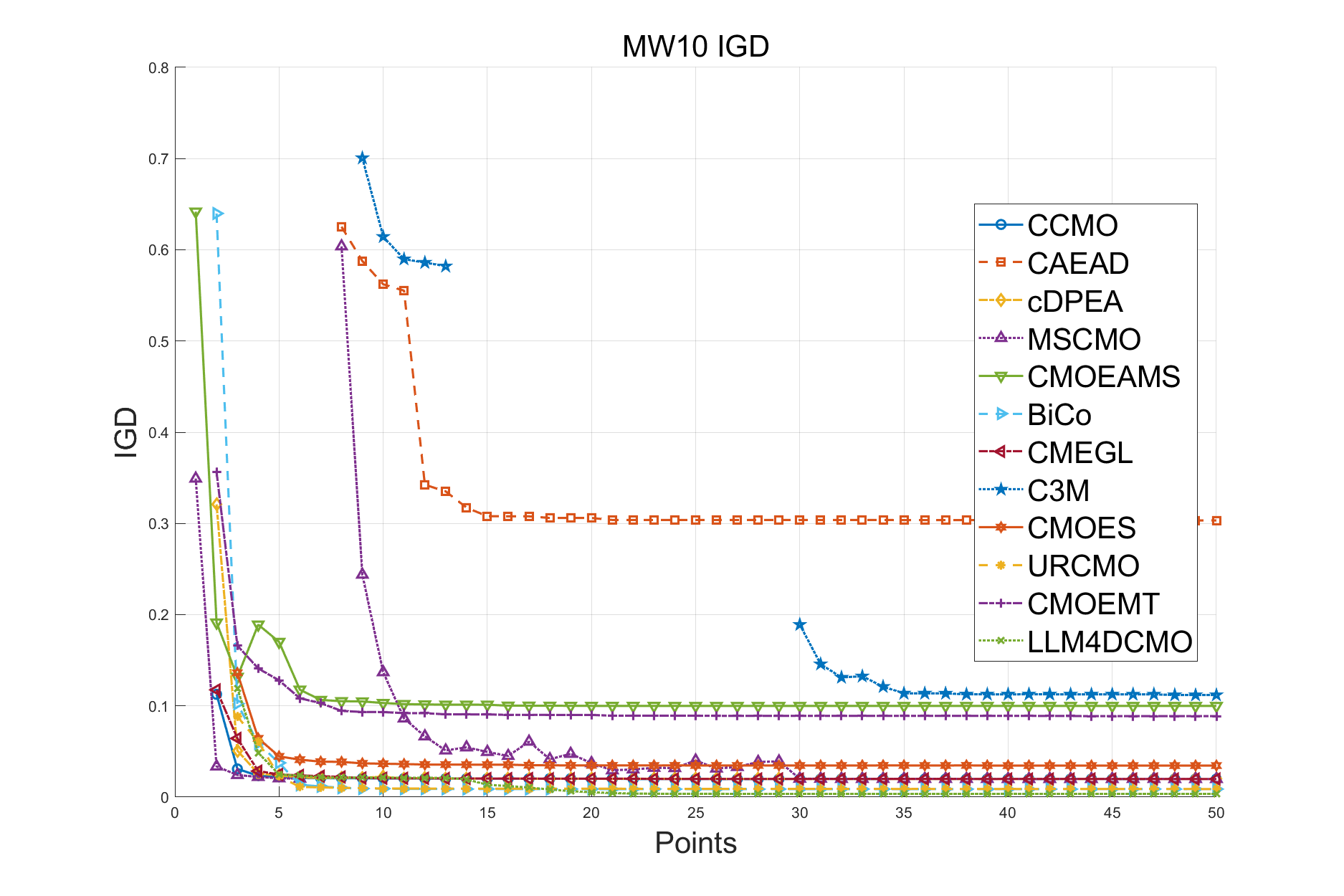}\label{fig:sub10m}}
    \hfill
    \subfloat[]{\includegraphics[width=0.24\textwidth]{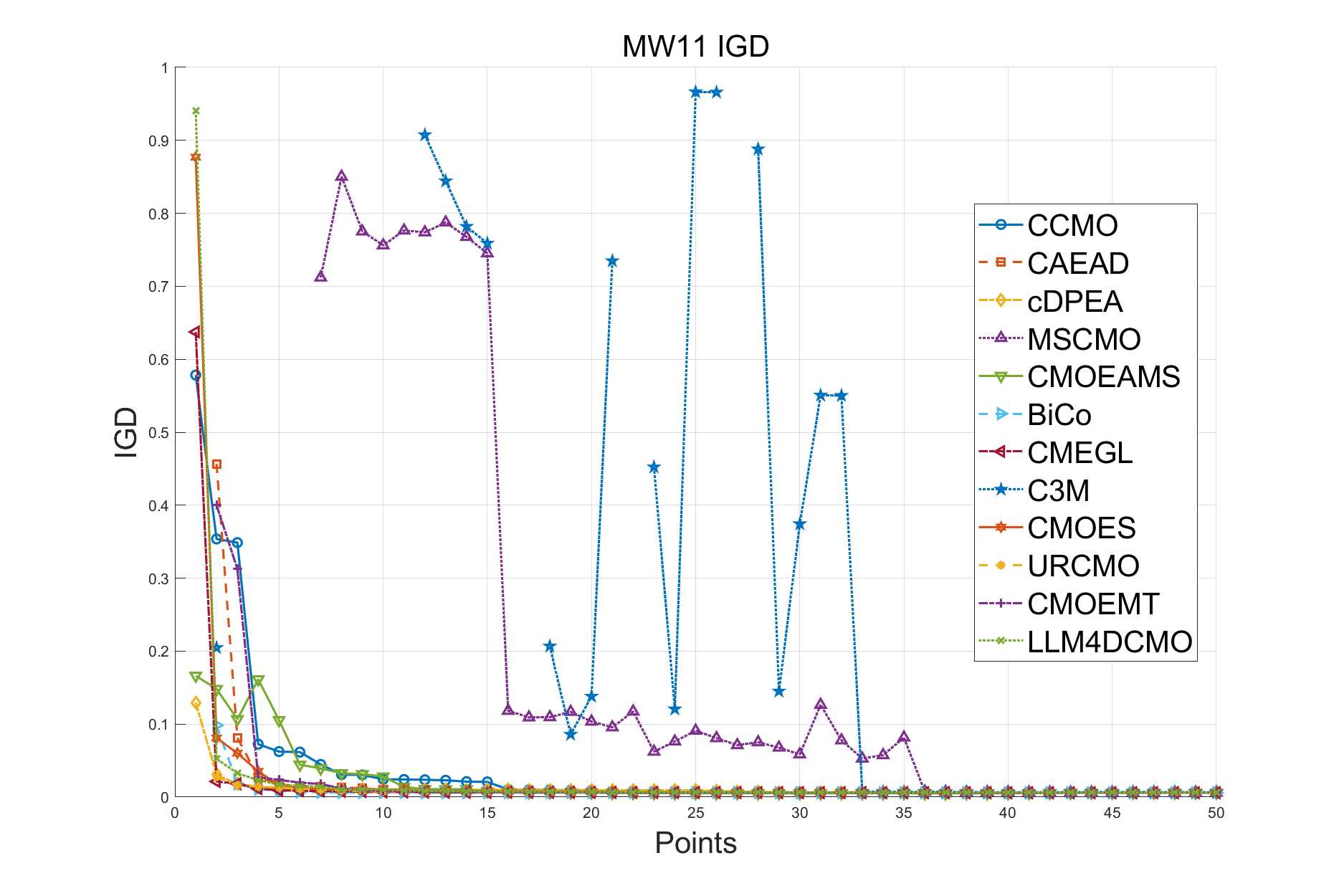}\label{fig:sub11m}}
    \hfill
    \subfloat[]{\includegraphics[width=0.24\textwidth]{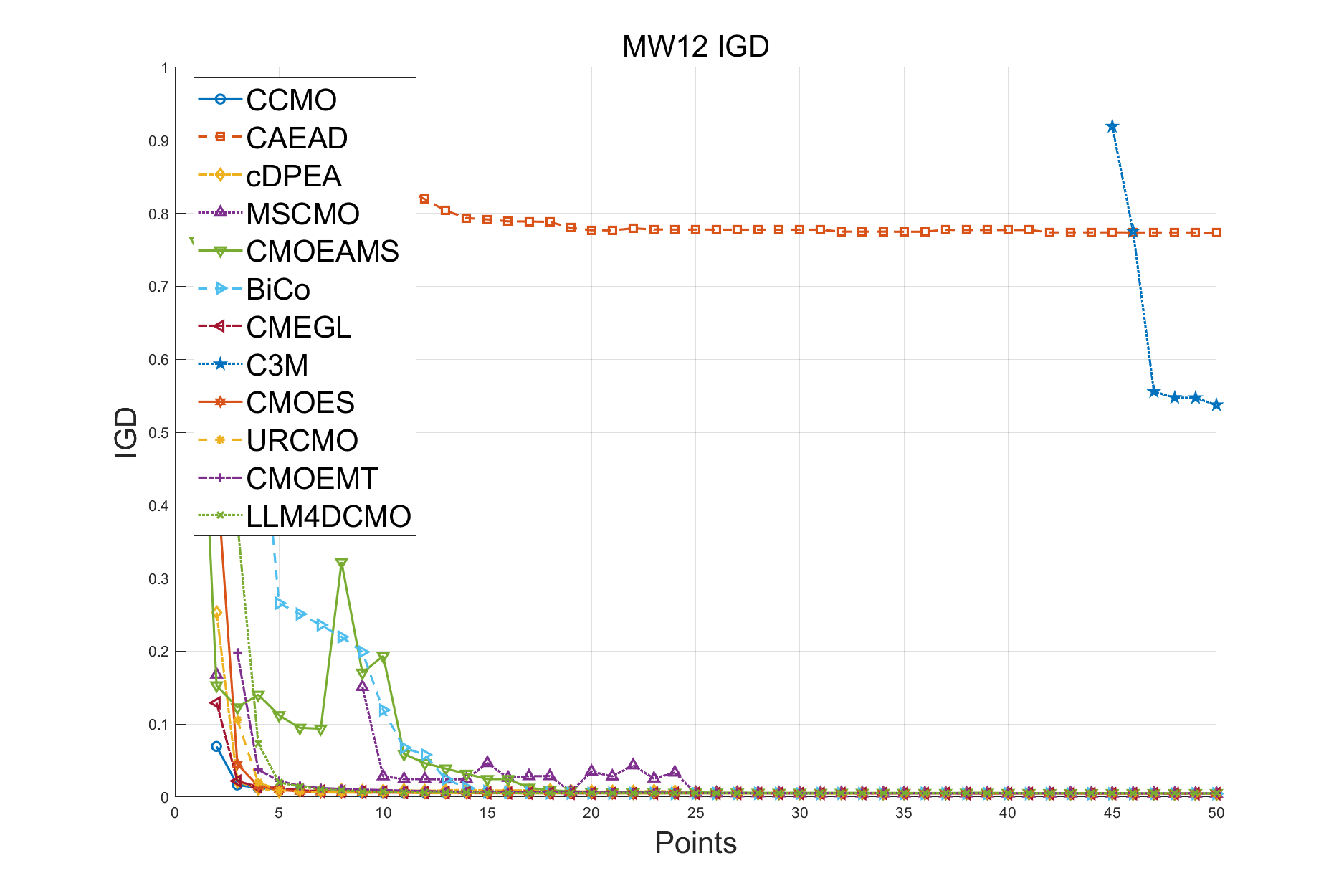}\label{fig:sub12m}}
    \\[1ex]
    
    \subfloat[]{\includegraphics[width=0.24\textwidth]{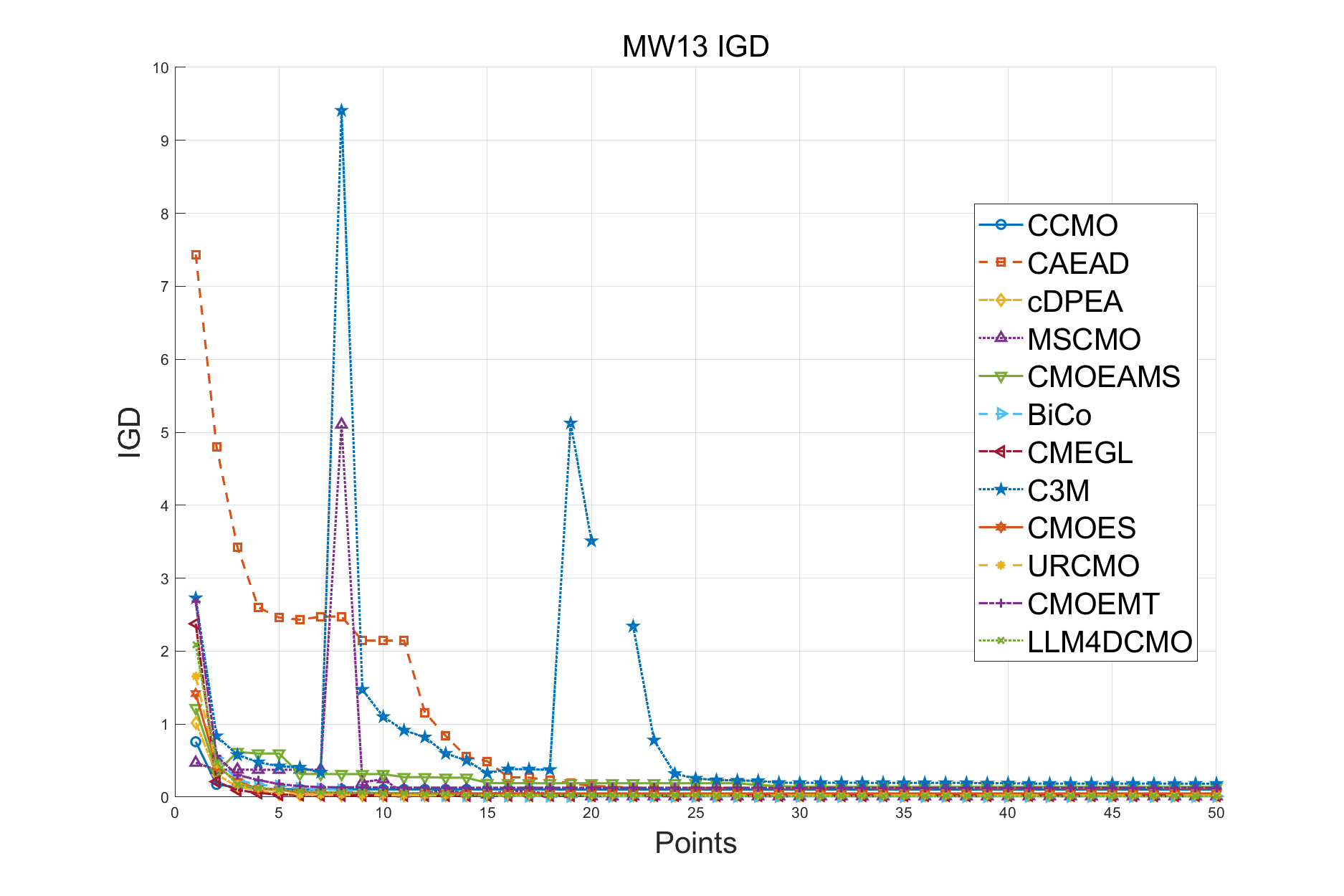}\label{fig:sub13m}}
    \subfloat[]{\includegraphics[width=0.24\textwidth]{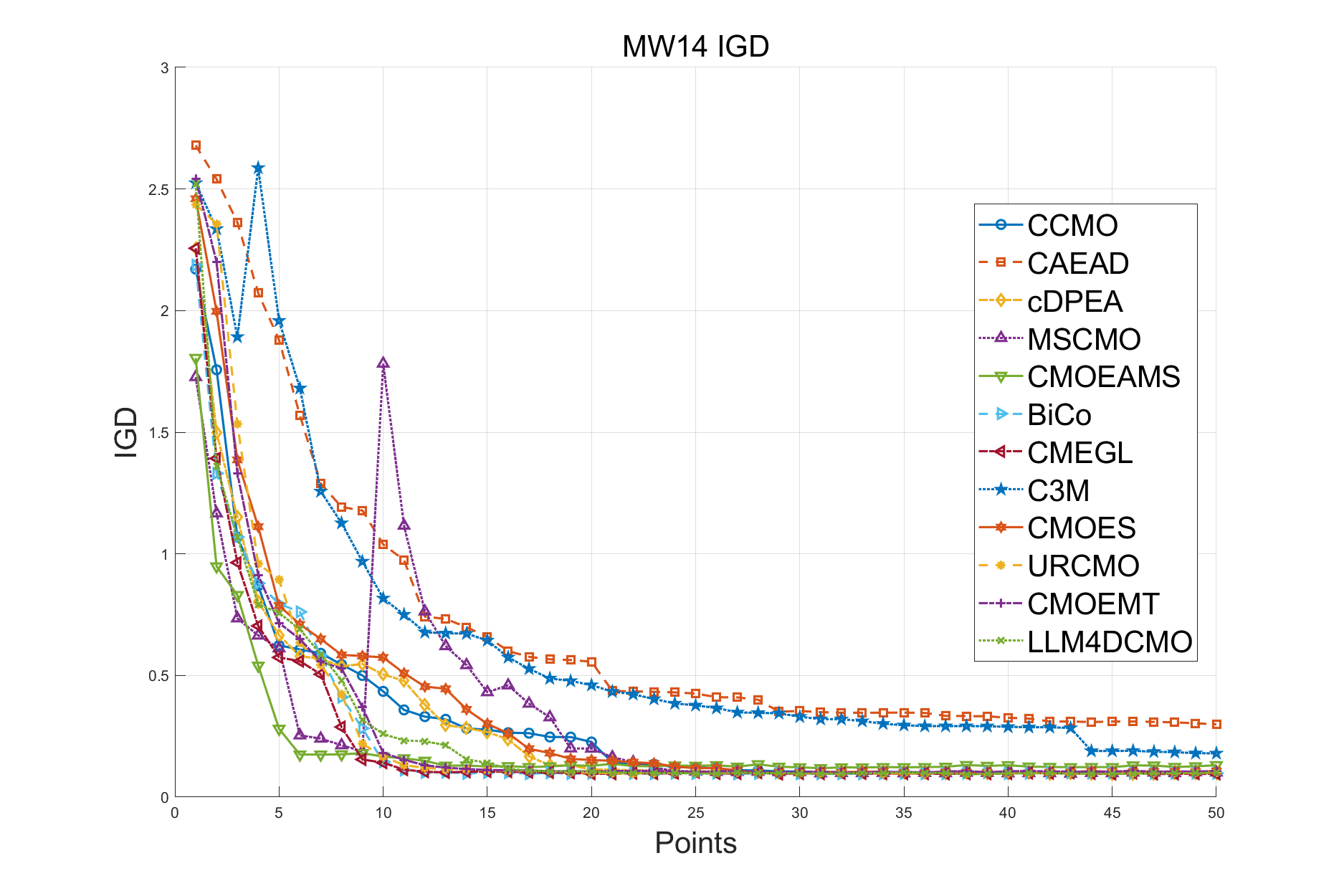}\label{fig:sub14m}}
    \hfill
    \hfill
    \caption{The convergence curve of IGD metric on MW test suite.}
    \label{fig:covMW}
\end{figure}

We plotted the convergence curves of IGD values for LLM4CMO and 11 baseline algorithms on the MW and LIRCMOP test suites, as presented in Fig. \ref{fig:covMW} and Fig. \ref{fig:covLIRs}. Although LLM4CMO may not consistently achieve rapid early-stage convergence compared to some baseline algorithms, it demonstrates stable and steady convergence throughout the process. The IGD values begin decreasing gradually at the onset of the second optimization stage, reflecting the combined influence of our dynamic type-learning mechanism and the relaxed convergence criteria of the UPF and CPF methods, which moderately prolong the learning phase. Subsequently, the HOps designed by the LLM and the epsilon function enable the algorithm to rapidly achieve high-quality convergence. 

\clearpage
\subsection{The convergence curve of IGD values on MW.} \label{sec:A5-2}

\section{Runtimes} \label{sec:Runtimes}
As shown in Table~\ref{tab:runtimes}, the running speed of our algorithm is at a moderate level. Although LLM4CMO, like URCMO, employs hybrid operators, its actual runtime is notably lower than that of URCMO and most other algorithms. Specifically, LLM4CMO's execution time is approximately one-third longer than that of CMOEAMS, which exhibits the shortest runtime, yet it remains within the same order of magnitude. Considering both computational efficiency and optimization performance, LLM4CMO achieves a favorable balance between runtime and solution quality.
\begin{table*}[htpb!]
    \centering
    \caption{Total runtime(s) for all algorithms run one time on the partly CF, DAS, and all LIRCMOP and MW.}
    \label{tab:runtimes}
    \resizebox{0.99\textwidth}{!}{
    \begin{tabular}{lcccccccccccc}
        \hline
        \textbf{Problem} & \textbf{URCMO} & \textbf{CCMO} & \textbf{CAEAD} & \textbf{cDPEA} & \textbf{BiCo} & \textbf{C3M} & \textbf{CMEGL} & \textbf{CMOES} & \textbf{CMOEMT} & \textbf{CMOEAMS} & \textbf{MSCMO} & \textbf{LLM4CMO} \\ \hline
        \textbf{CF123} & 1.7E+02 & 8.59E+01 & 6.10E+01 & 1.27E+02 & 3.36E+01 & 5.36E+01 & 9.69E+01 & 5.00E+02 & 3.90E+02 & 5.30E+01 & 6.83E+01 & 9.46E+01 \\
        \textbf{DAS123} & 5.76E+01 & 5.69E+01 & 3.40E+01 & 8.83E+01 & 4.22E+01 & 1.37E+02 & 9.65E+01 & 3.62E+02 & 1.65E+02 & 3.84E+01 & 8.13E+01 & 3.82E+01 \\
        \textbf{LIRCMOP} & 8.09E+02 & 6.53E+02 & 3.83E+02 & 9.24E+02 & 3.80E+02 & 4.74E+02 & 9.07E+02 & 6.12E+03 & 1.64E+03 & 2.84E+02 & 4.20E+02 & 4.50E+02 \\
        \textbf{MW} & 4.89E+02 & 3.70E+02 & 1.26E+02 & 4.47E+02 & 1.65E+02 & 1.90E+02 & 4.63E+02 & 3.16E+03 & 9.81E+02 & 1.98E+02 & 2.70E+02 & 2.74E+02 \\ \hline
        \textbf{Total} & 1.46E+03 & 1.17E+03 & 6.04E+02 & 1.59E+03 & 6.20E+02 & 8.55E+02 & 1.56E+03 & 1.01E+04 & 3.17E+03 & 5.73E+02 & 8.39E+02 & 8.57E+02 \\ \hline
    \end{tabular}%
    }
\end{table*}

\clearpage

\section{The Pareto Fronts, Feasible regions and Feasible solutions of six test suites.} \label{sec:A6}
In this section, we have plotted the result graphs of 12 algorithms on all 6 test suites, and compared different algorithms to demonstrate the differences among them.
\begin{figure}[hptb!]
    \centering
    \resizebox{0.85\textwidth}{!}{
        \includegraphics[angle=-90]{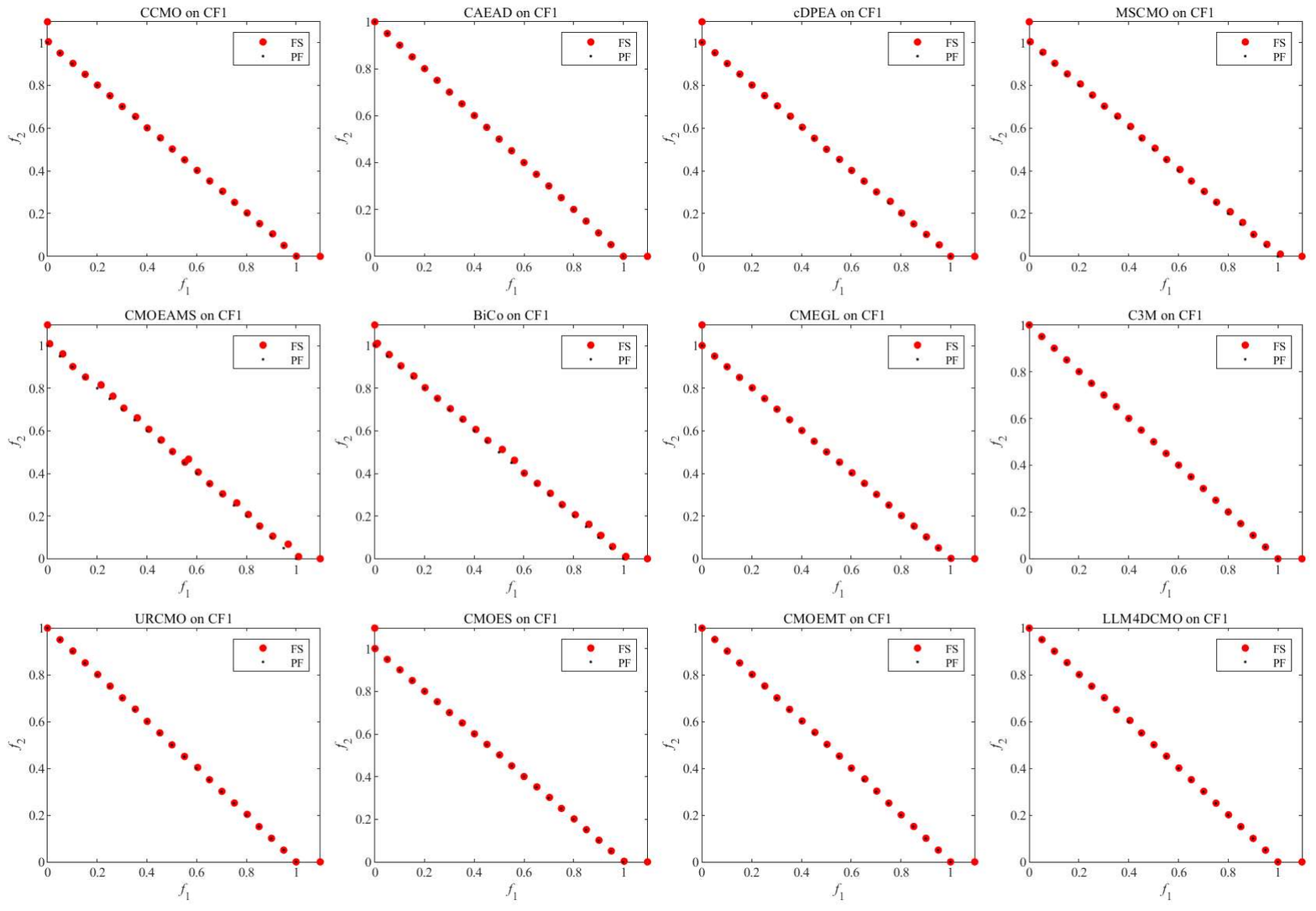}
    }
    \hfill 
    \resizebox{0.85\textwidth}{!}{
        \includegraphics[angle=-90]{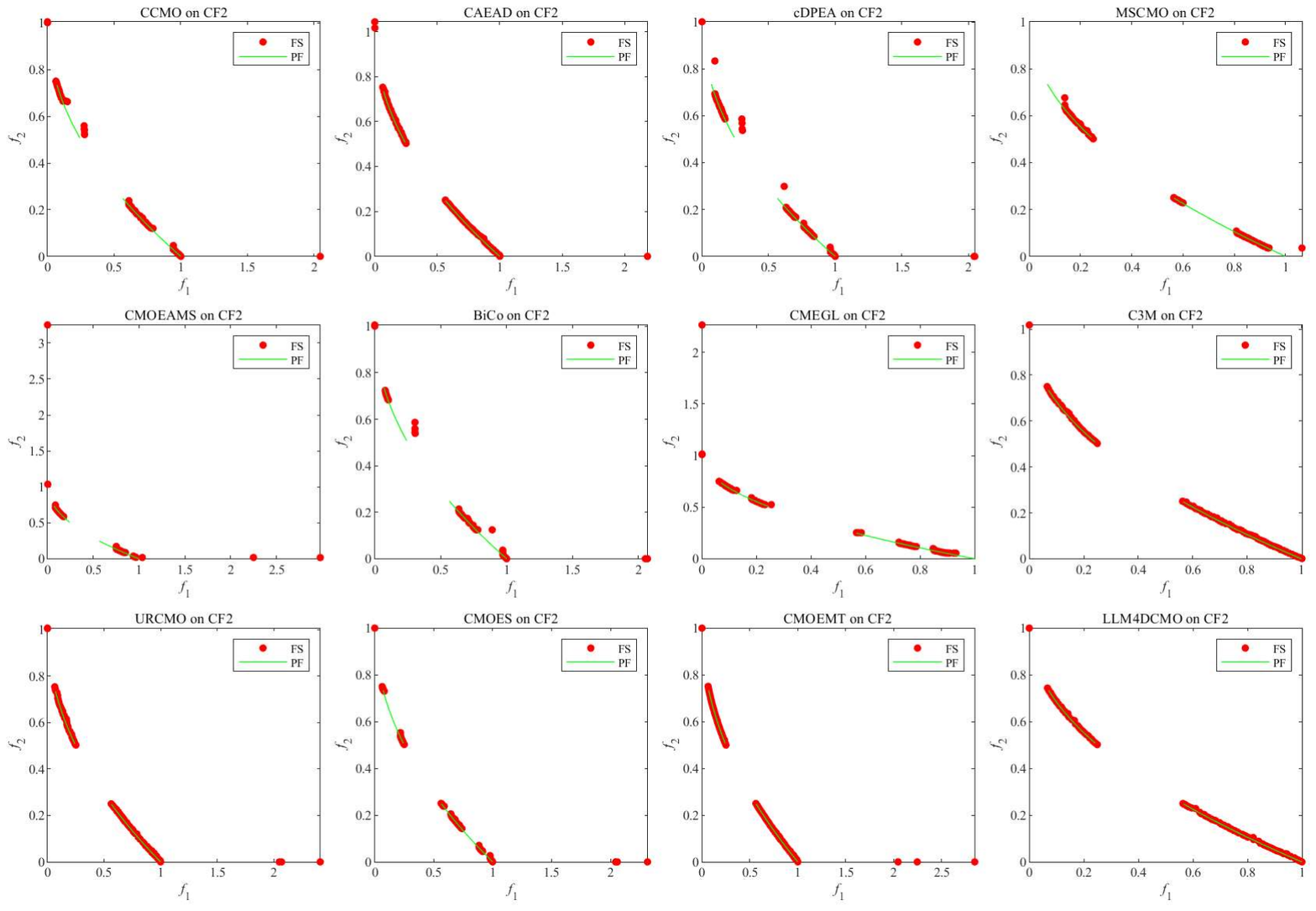}
    }
    \caption{The Pareto Fronts/Feasible regions and Feasible solutions of CF1 and CF2}
    \label{fig:doc9-fcp1-comparison}
\end{figure}

\foreach \j in {3,5,7,9}{
\begin{figure}[hptb!]
    \centering
    \resizebox{0.95\textwidth}{!}{
        \includegraphics[angle=-90]{plt/pareto/CF\j.pdf}
    }
    \vspace{0.3cm} 
    \resizebox{0.95\textwidth}{!}{
        \includegraphics[angle=-90]{plt/pareto/CF\the\numexpr\j+1\relax.pdf}
    }
    \caption{The Pareto Fronts/Feasible regions and Feasible solutions of CF\j{} and CF\the\numexpr\j+1\relax{}}
    \label{fig:cf-combined-\j}
\end{figure}
}
\foreach \j in {1,3,5,7}{
\begin{figure}[hptb!]
    \centering
    \resizebox{0.95\textwidth}{!}{
        \includegraphics[angle=-90]{plt/pareto/DAS\j.pdf}
    }
    \vspace{0.3cm}
    \resizebox{0.95\textwidth}{!}{
        \includegraphics[angle=-90]{plt/pareto/DAS\the\numexpr\j+1\relax.pdf}
    }
    \caption{The Pareto Fronts/Feasible regions and Feasible solutions of DAS\j{} and DAS\the\numexpr\j+1\relax{}}
    \label{fig:das-combined-\j}
\end{figure}
}

\begin{figure}[hptb!]
    \centering
    \resizebox{0.95\textwidth}{!}{
        \includegraphics[angle=-90]{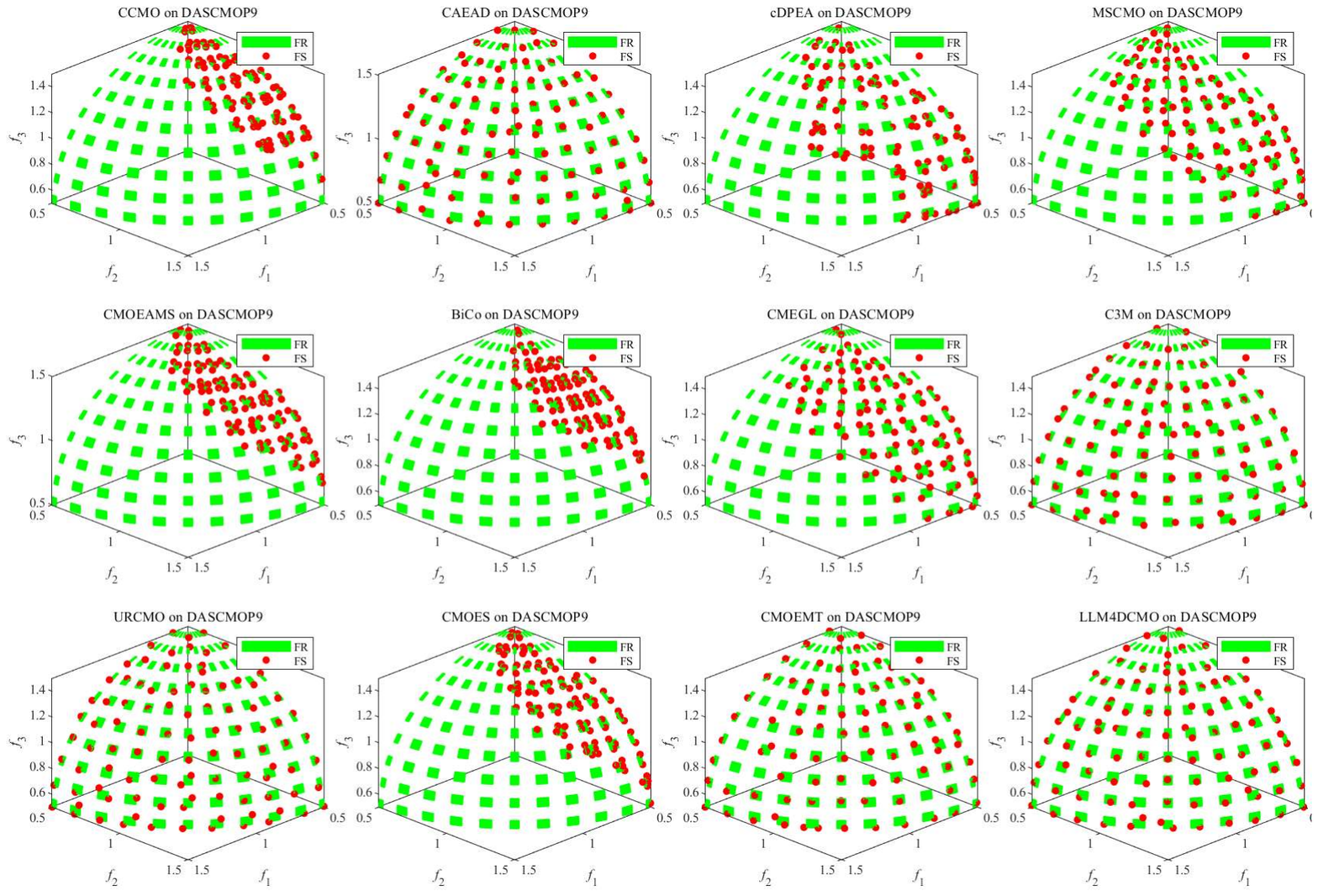}
    }
    \caption{The Feasible regions and Feasible solutions of DAS9}
    \label{fig:paretoDAS9}
\end{figure}
\foreach \j in {1,3,5,7,9,11,13}{
\begin{figure}[hptb!]
    \centering
    \resizebox{0.95\textwidth}{!}{
        \includegraphics[angle=-90]{plt/pareto/MW\j.pdf}
    }
    \vspace{0.3cm}
    \resizebox{0.95\textwidth}{!}{
        \includegraphics[angle=-90]{plt/pareto/MW\the\numexpr\j+1\relax.pdf}
    }
    \caption{The Pareto Fronts, Feasible regions and Feasible solutions of MW\j{} and MW\the\numexpr\j+1\relax{}}
    \label{fig:mw-combined-\j}
\end{figure}
}
\foreach \j in {1,3,5,7,9,11,13}{
\begin{figure}[hptb!]
    \centering
    \resizebox{0.95\textwidth}{!}{
        \includegraphics[angle=-90]{plt/pareto/LIR\j.pdf}
    }
    \vspace{0.3cm}
    \resizebox{0.95\textwidth}{!}{
        \includegraphics[angle=-90]{plt/pareto/LIR\the\numexpr\j+1\relax.pdf}
    }
    \caption{The Pareto Fronts/Feasible regions and Feasible solutions of LIR\j{} and LIR\the\numexpr\j+1\relax{}}
    \label{fig:lir-combined-\j}
\end{figure}
}
\foreach \j in {1,3,5,7}{
\begin{figure}[hptb!]
    \centering
    \resizebox{0.95\textwidth}{!}{
        \includegraphics[angle=-90]{plt/pareto/DOC\j.pdf}
    }
    \vspace{0.3cm}
    \resizebox{0.95\textwidth}{!}{
        \includegraphics[angle=-90]{plt/pareto/DOC\the\numexpr\j+1\relax.pdf}
    }
    \caption{The Pareto Fronts/Feasible regions and Feasible solutions of DOC\j{} and DOC\the\numexpr\j+1\relax{}}
    \label{fig:doc-combined-\j}
\end{figure}
}

\begin{figure}[hptb!]
    \centering
    \resizebox{0.98\textwidth}{!}{
        \includegraphics[angle=-90]{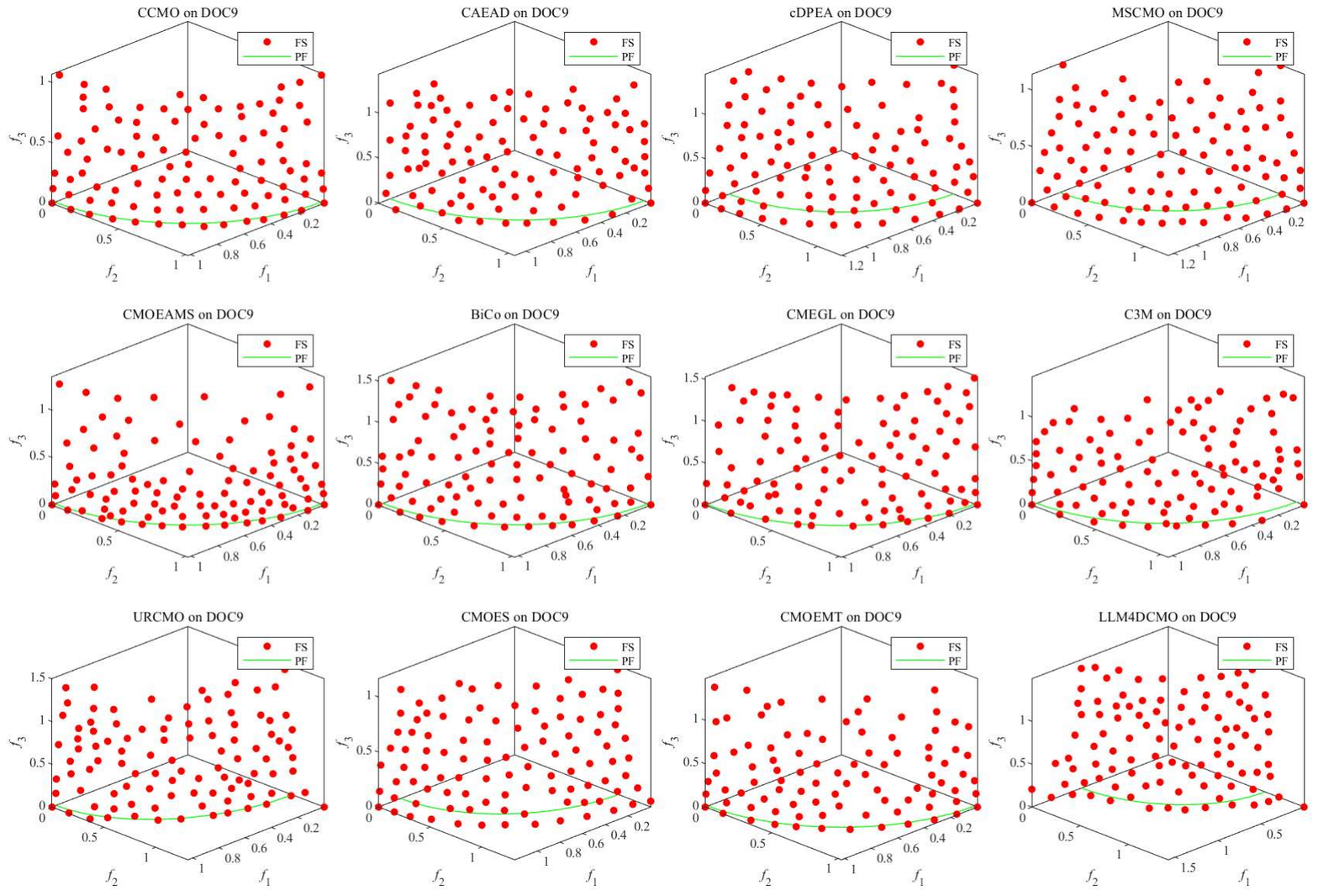}
    }
    \hfill 
    \resizebox{0.98\textwidth}{!}{
        \includegraphics[angle=-90]{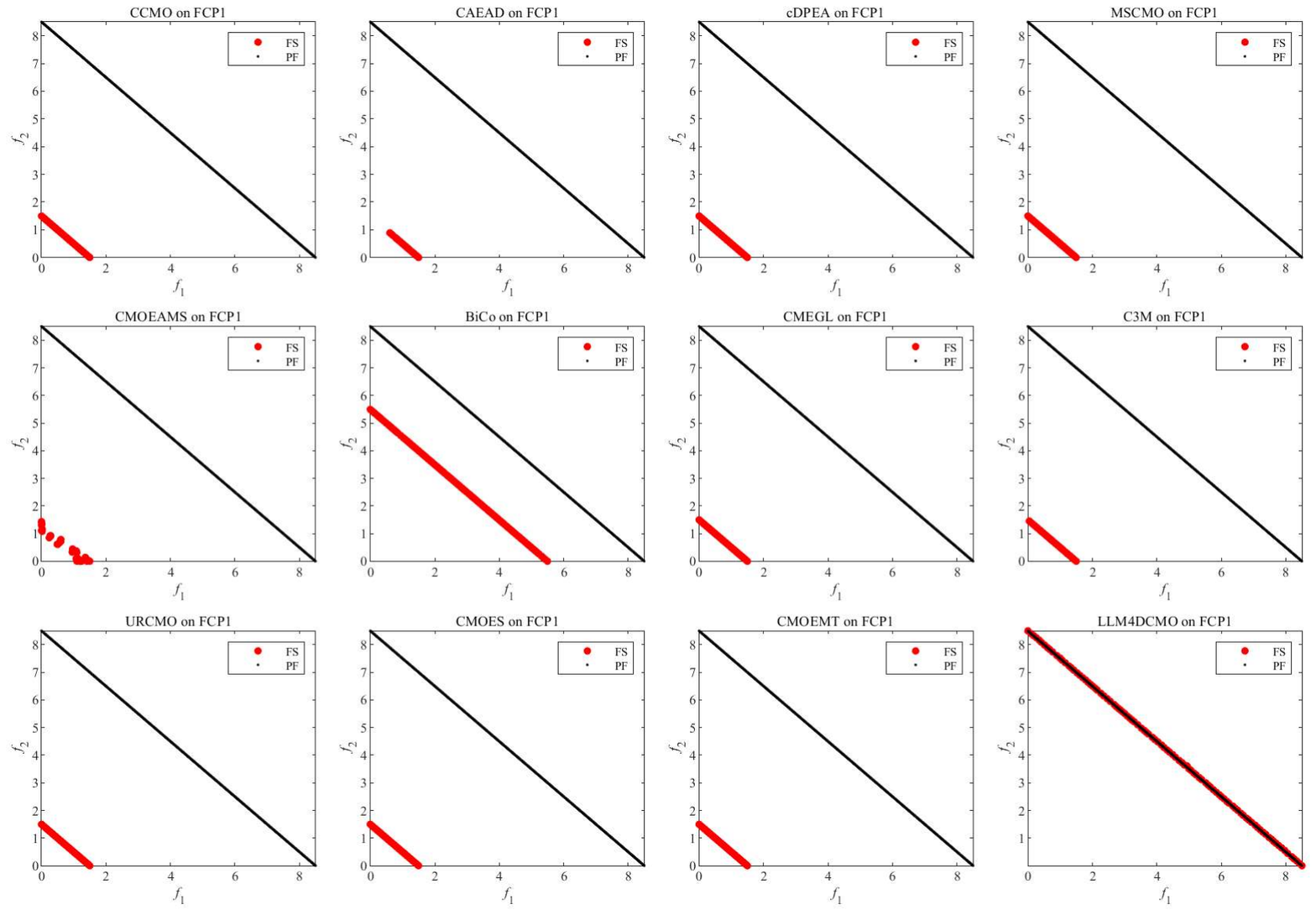}
    }
    \caption{The Pareto Fronts/Feasible regions and Feasible solutions of DOC9 and FCP1}
    \label{fig:doc9-fcp1-comparison}
\end{figure}

\foreach \j in {2,4}{
\begin{figure}[hptb!]
    \centering
    \resizebox{0.95\textwidth}{!}{
        \includegraphics[angle=-90]{plt/pareto/FCP\j.pdf}
    }
    \vspace{0.3cm}
    \resizebox{0.95\textwidth}{!}{
        \includegraphics[angle=-90]{plt/pareto/FCP\the\numexpr\j+1\relax.pdf}
    }
    \caption{The Pareto Fronts/Feasible regions and Feasible solutions of FCP\j{} and FCP\the\numexpr\j+1\relax{}}
    \label{fig:fcp-combined-\j}
\end{figure}
}
\clearpage
\section{The evolutionary process of Bico , URCMO and LLM4CMO on LIRCMOP1, LIRCMOP5, LIRCMOP9, LIRCMOP11.}
In this section, we selected four problems from LIRCMOP that are classified into traditional UPF-CPF relationship types 1 to 4, and plotted the evolutionary processes of Bico, URCMO, and LLM4CMO. LIRCMOP1 corresponds to the type-4 case of complete separation, LIRCMOP5 corresponds to the type-1 case of complete overlap, LIRCMOP9 corresponds to type-2 where CPF partially contains UPF, and LIRCMOP11 corresponds to type-3 where CPF partially contains with partial separation.

For the evolutionary results of the LIRCMOP1 problem, it can be seen that LLM4CMO has achieved strong diversity in the mid-term, which benefits from the combined effects of HOps, epsilon control, and environmental selection strategies.

For LIRCMOP5, similar to the results on URCMO, our popAux uses a smaller number but still has strong exploration ability, gradually exploring the entire PF region as the stage grows, which indicates the effectiveness of HOps in type-1 problems.

In the LIRCMOP9 problem where UPF-CPF is type-2, it can be observed that our algorithm maintains good diversity in the mid-term and almost converges to the optimal PF, while Bico falls into a local optimal solution and URCMO also has the problem of slow convergence. On the one hand, this is due to insufficient information obtained in the learning stage, which is reflected in the diversity of the feasible region explored by solutions in the early stage. On the other hand, it may be that the operator combination has insufficient ability to explore the feasible region. Our algorithm converges quickly and has high solution quality on this problem.

In the LIRCMOP11 problem, URCMO and LLM4CMO show similar performances, while BiCO still falls into a local optimum and struggles to explore the global optimal region.

\begin{figure}[H]
    \centering
    \begin{minipage}{0.3\textwidth}
        \centering
        \includegraphics[width=\textwidth]{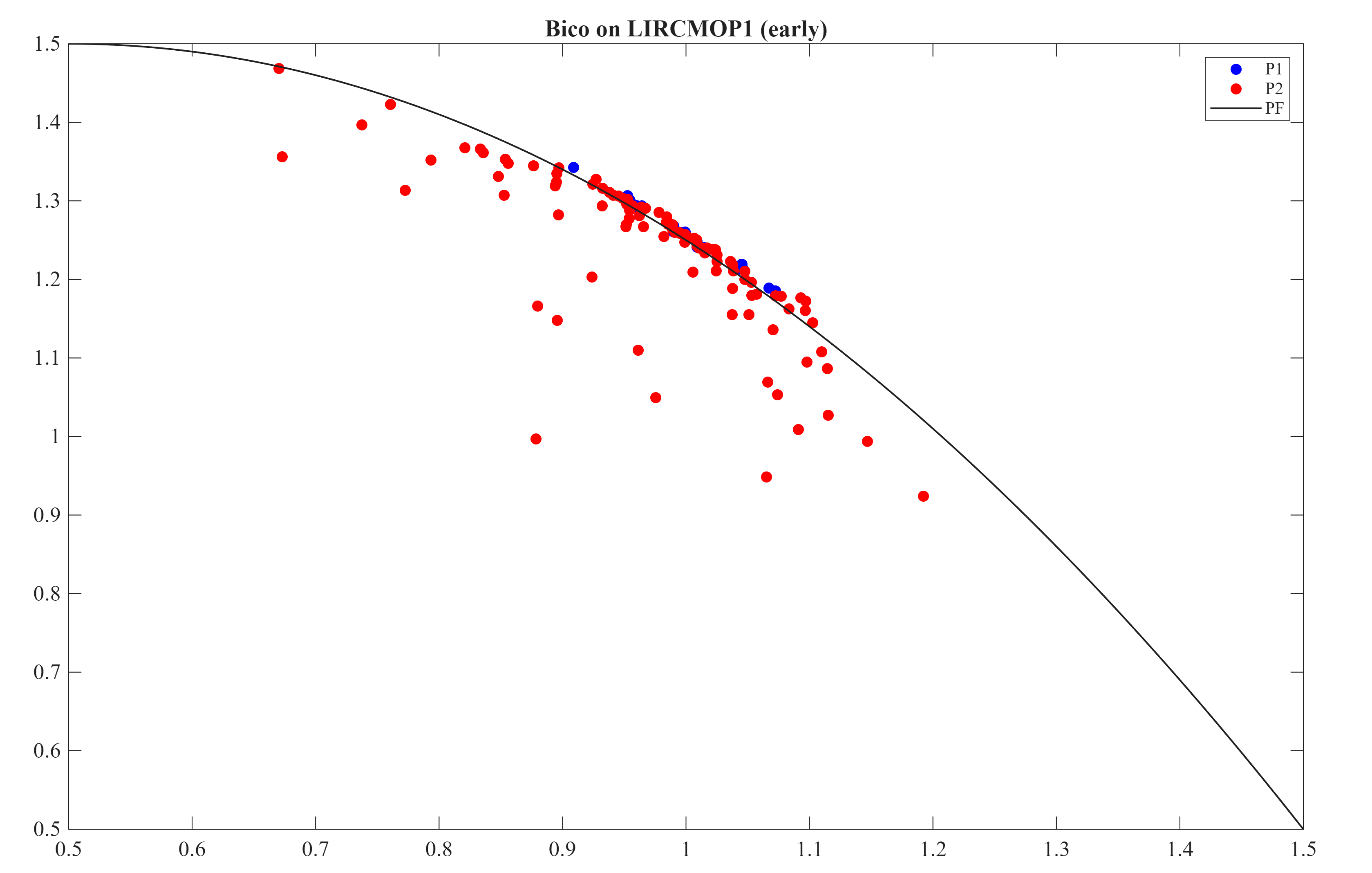}
    \end{minipage}
    \hfill
    \begin{minipage}{0.3\textwidth}
        \centering
        \includegraphics[width=\textwidth]{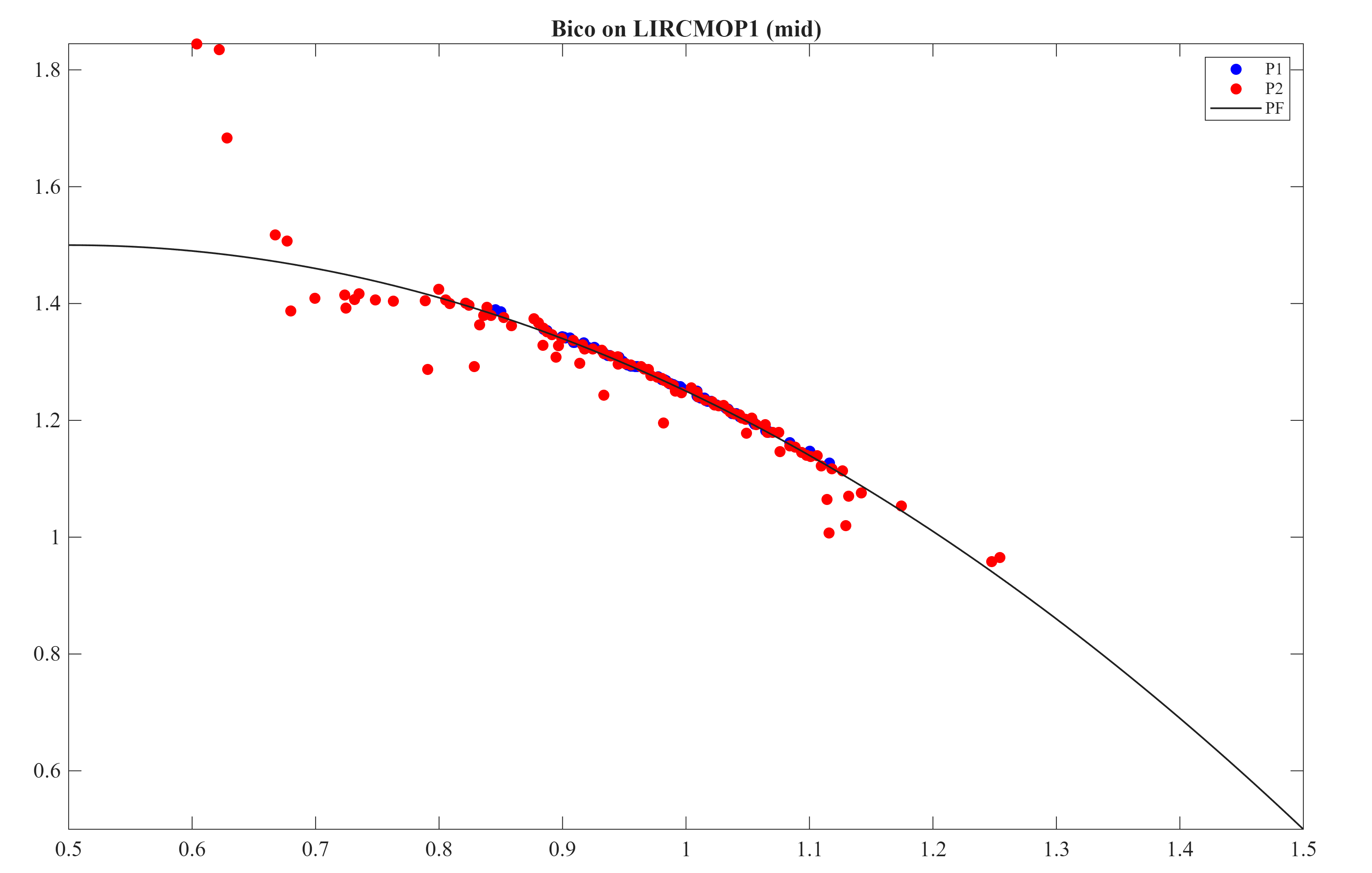}
    \end{minipage}
    \hfill
    \begin{minipage}{0.3\textwidth}
        \centering
        \includegraphics[width=\textwidth]{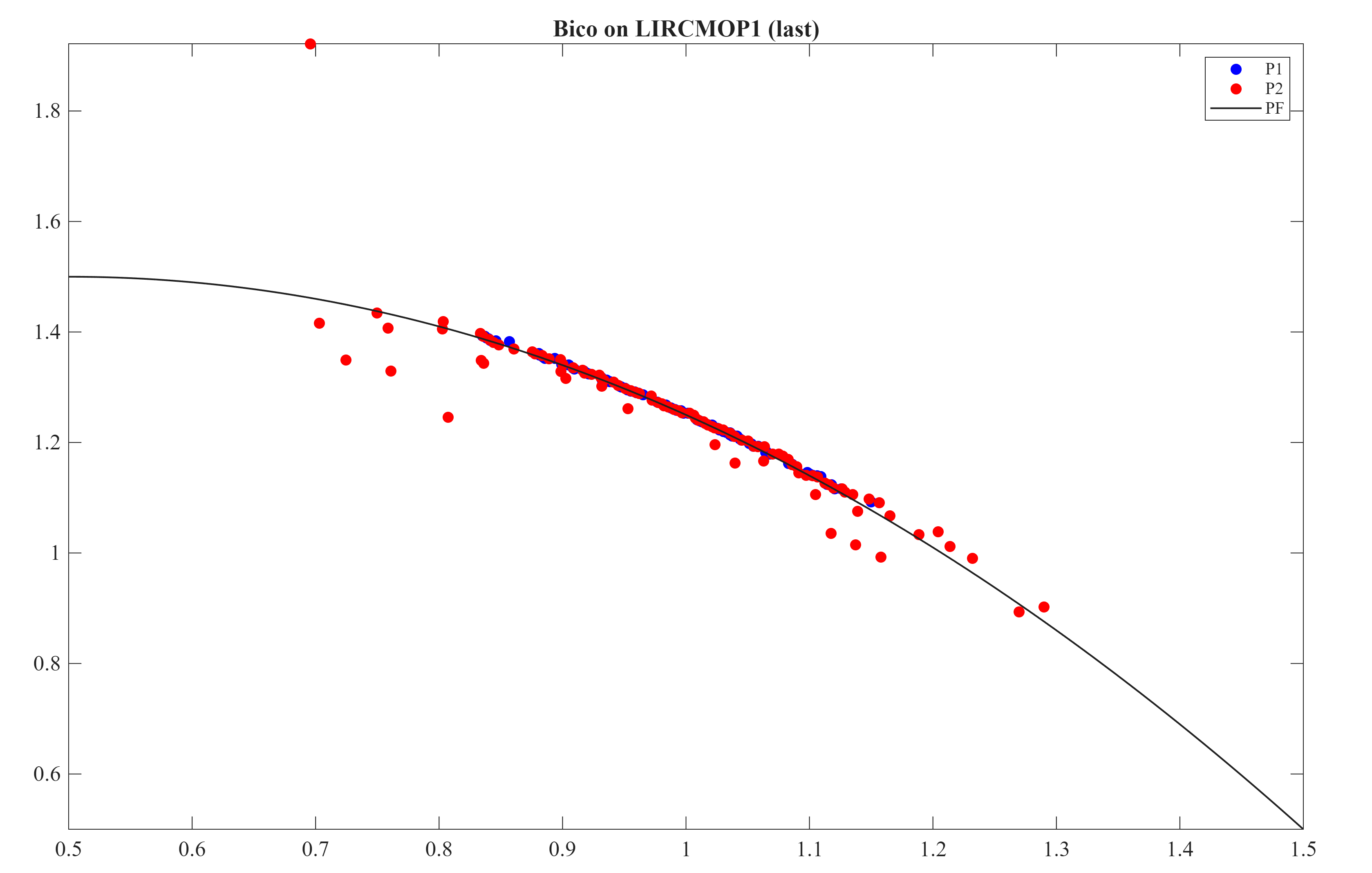}
    \end{minipage}
    
    
    \begin{minipage}{0.3\textwidth}
        \centering
        \includegraphics[width=\textwidth]{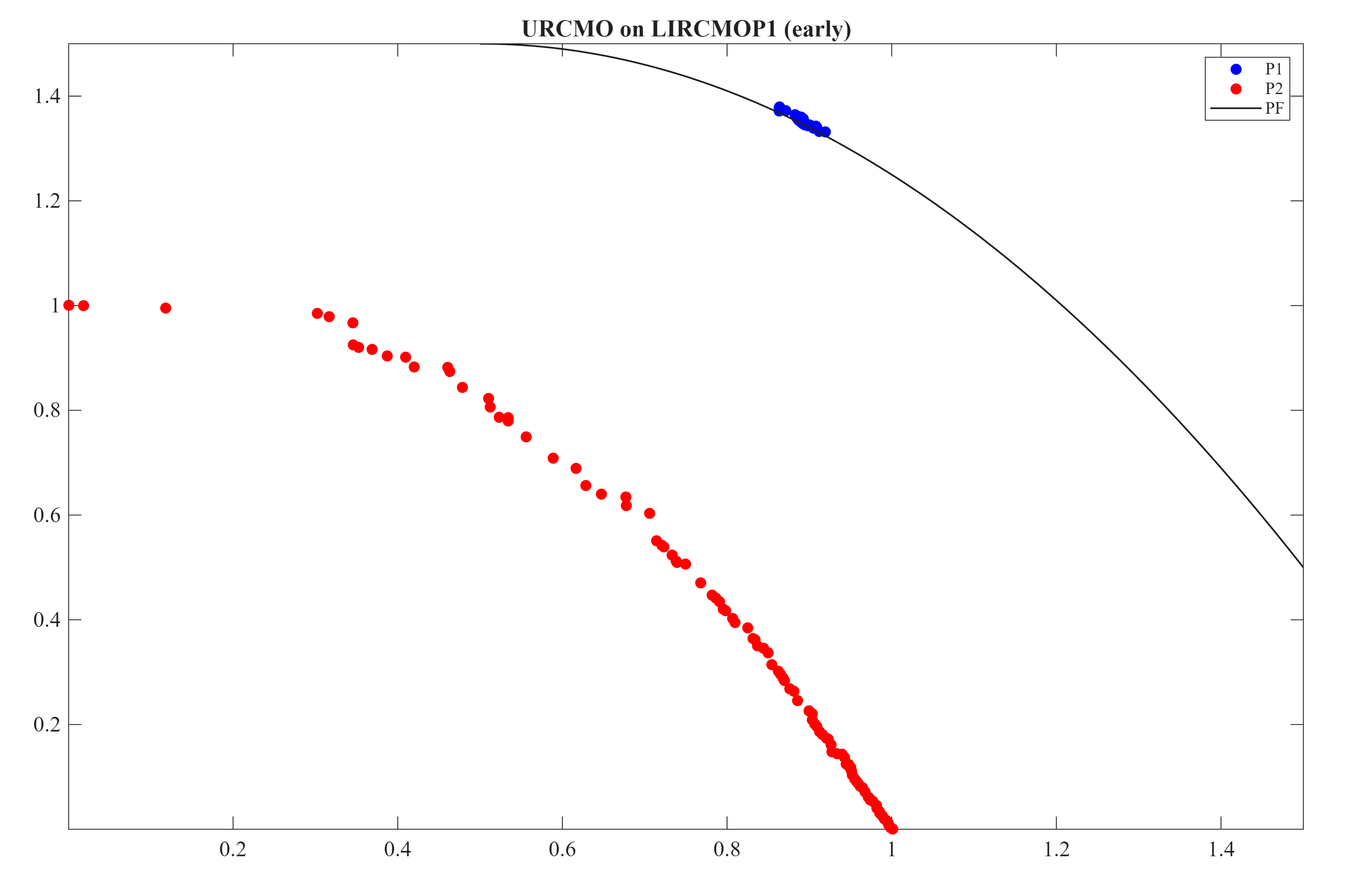}
    \end{minipage}
    \hfill
    \begin{minipage}{0.3\textwidth}
        \centering
        \includegraphics[width=\textwidth]{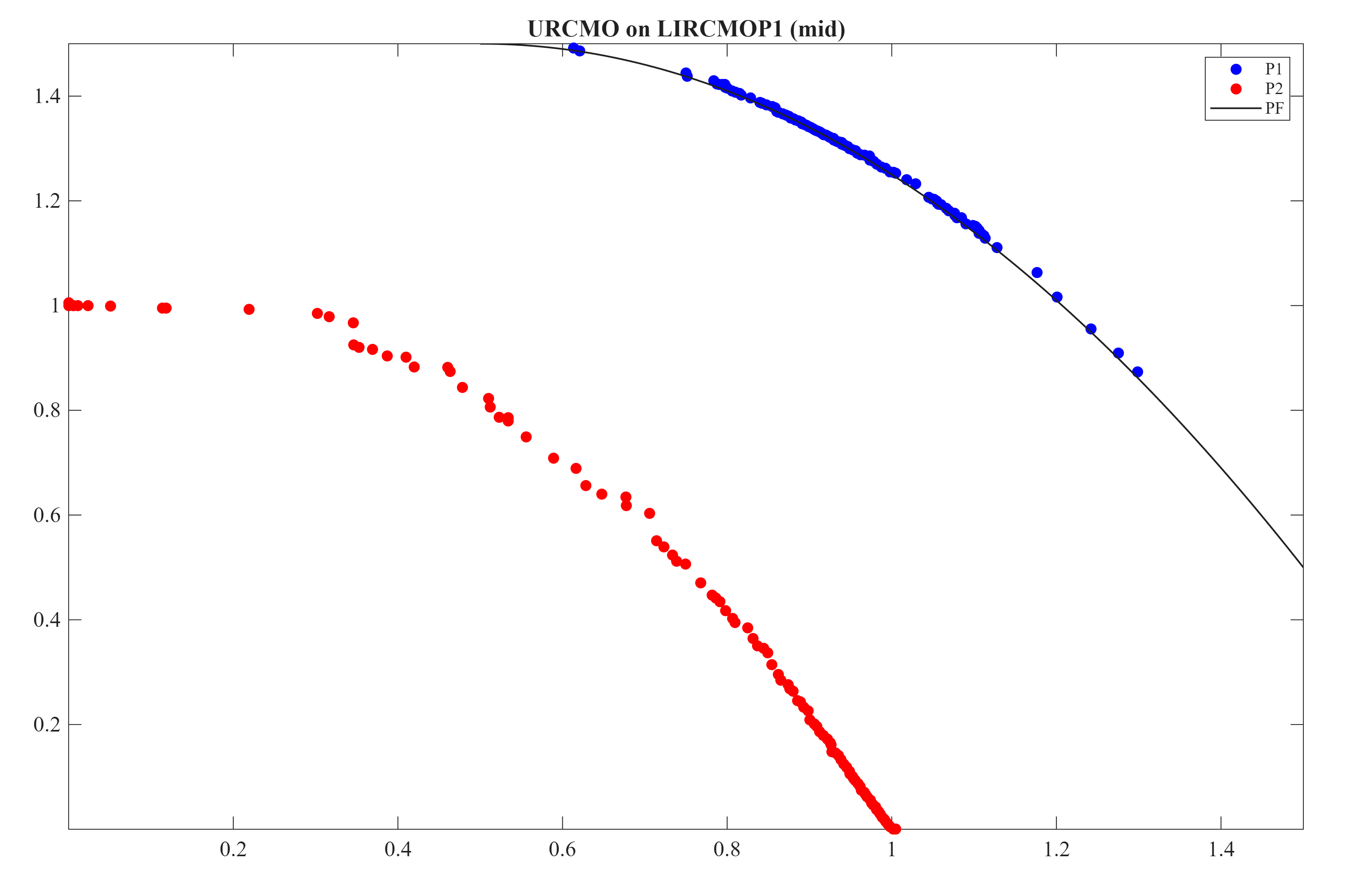}
    \end{minipage}
    \hfill
    \begin{minipage}{0.3\textwidth}
        \centering
        \includegraphics[width=\textwidth]{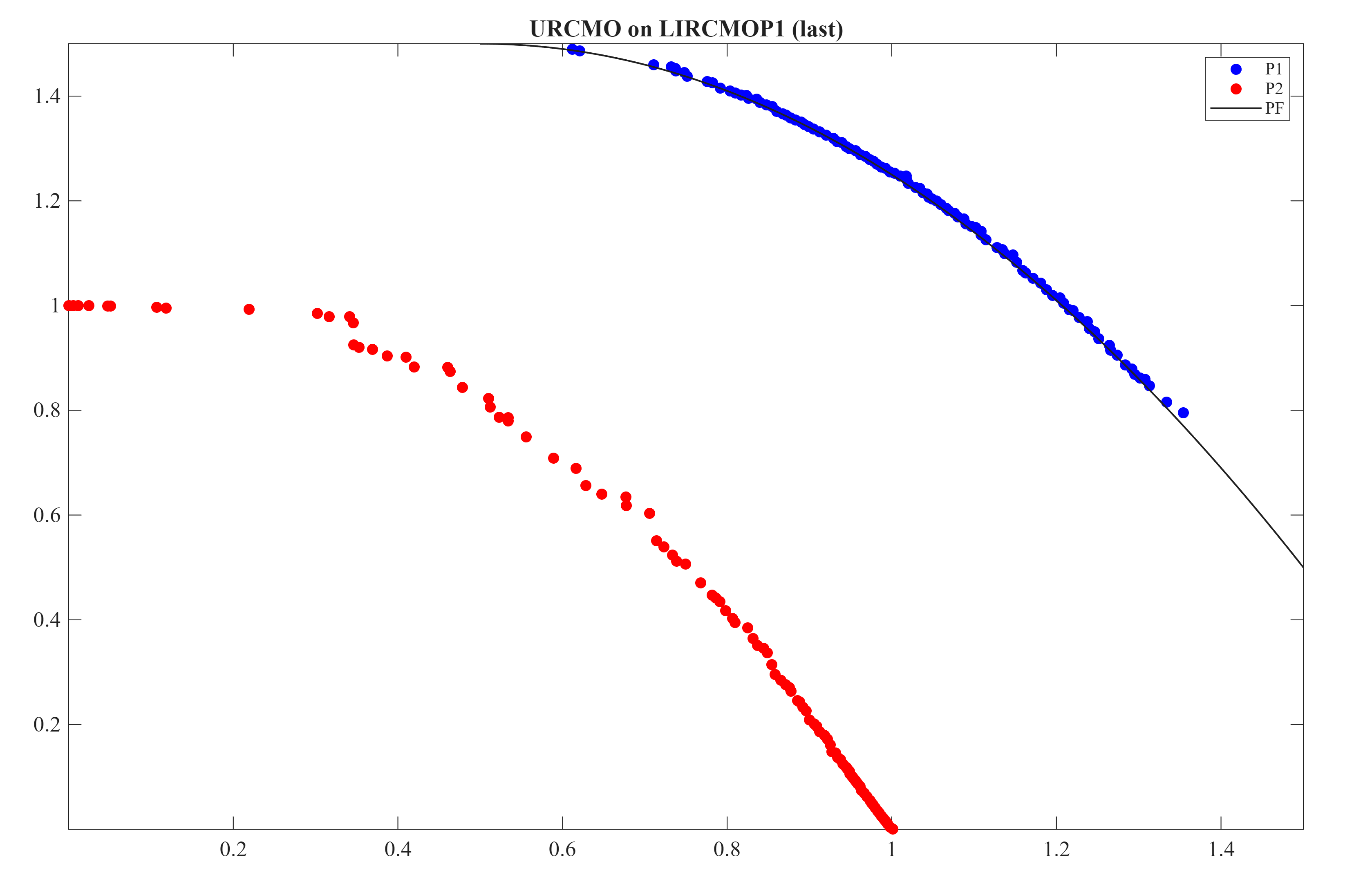}
    \end{minipage}
    
    
    \begin{minipage}{0.3\textwidth}
        \centering
        \includegraphics[width=\textwidth]{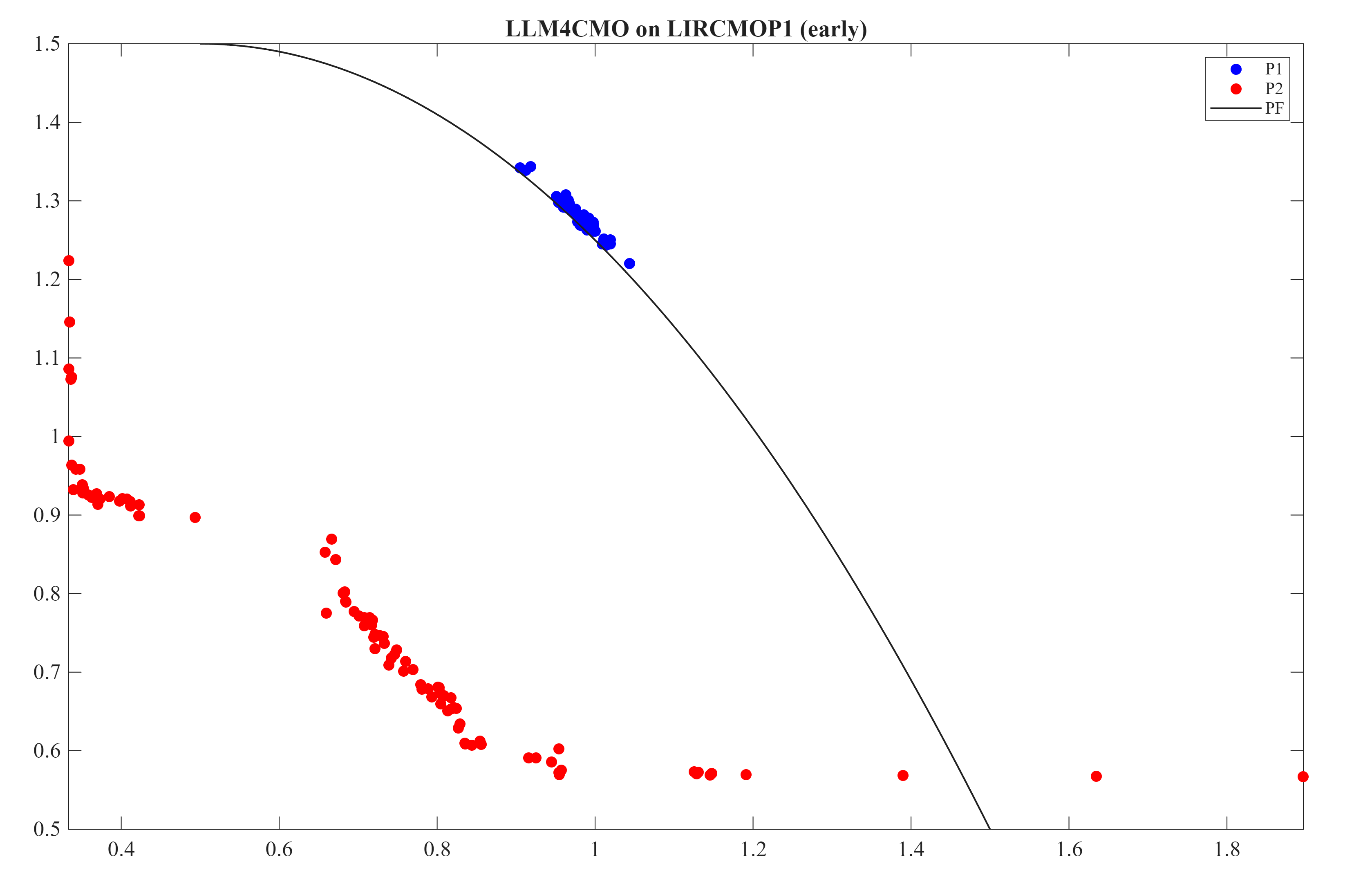}
    \end{minipage}
    \hfill
    \begin{minipage}{0.3\textwidth}
        \centering
        \includegraphics[width=\textwidth]{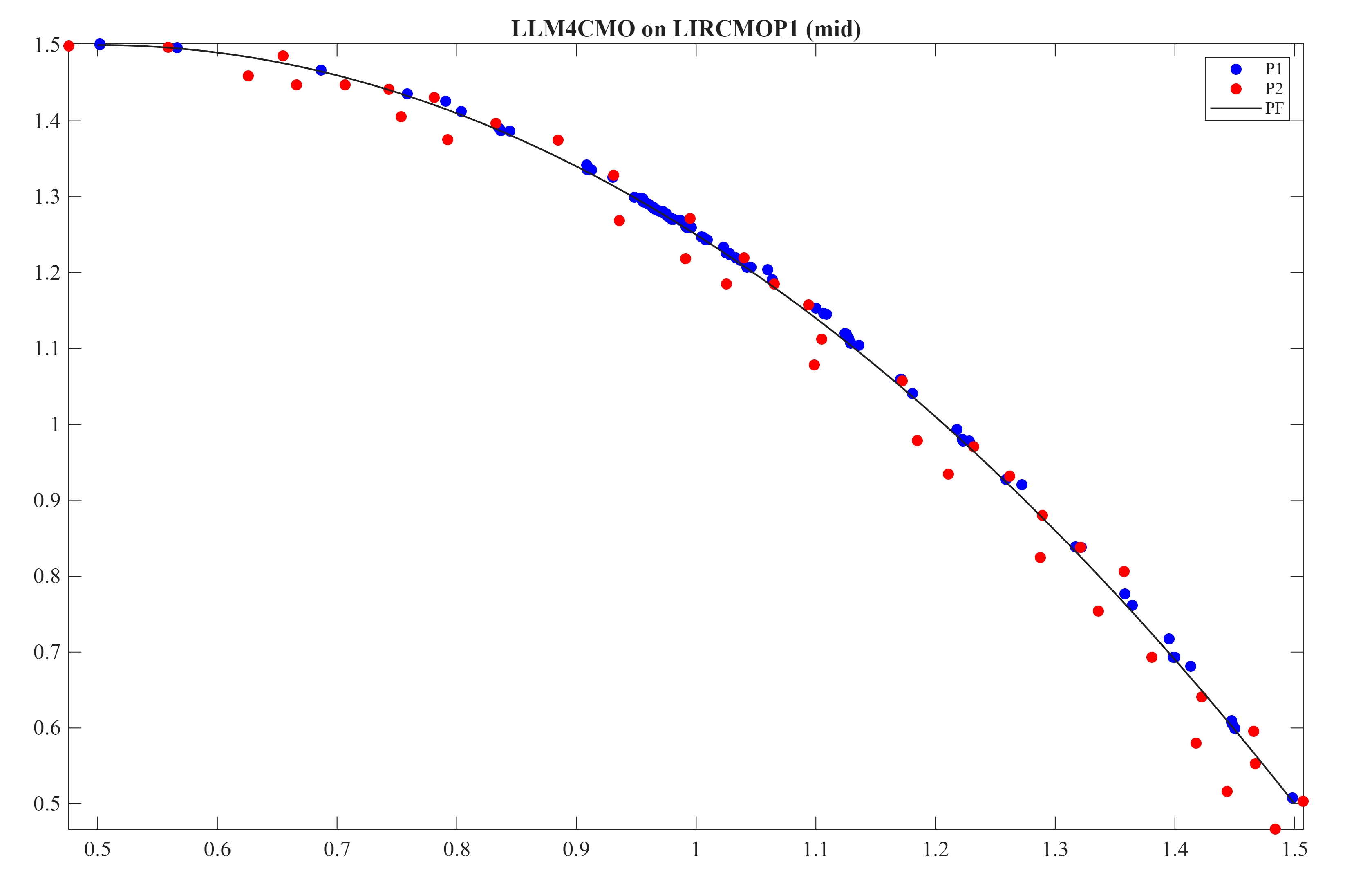}
    \end{minipage}
    \hfill
    \begin{minipage}{0.3\textwidth}
        \centering
        \includegraphics[width=\textwidth]{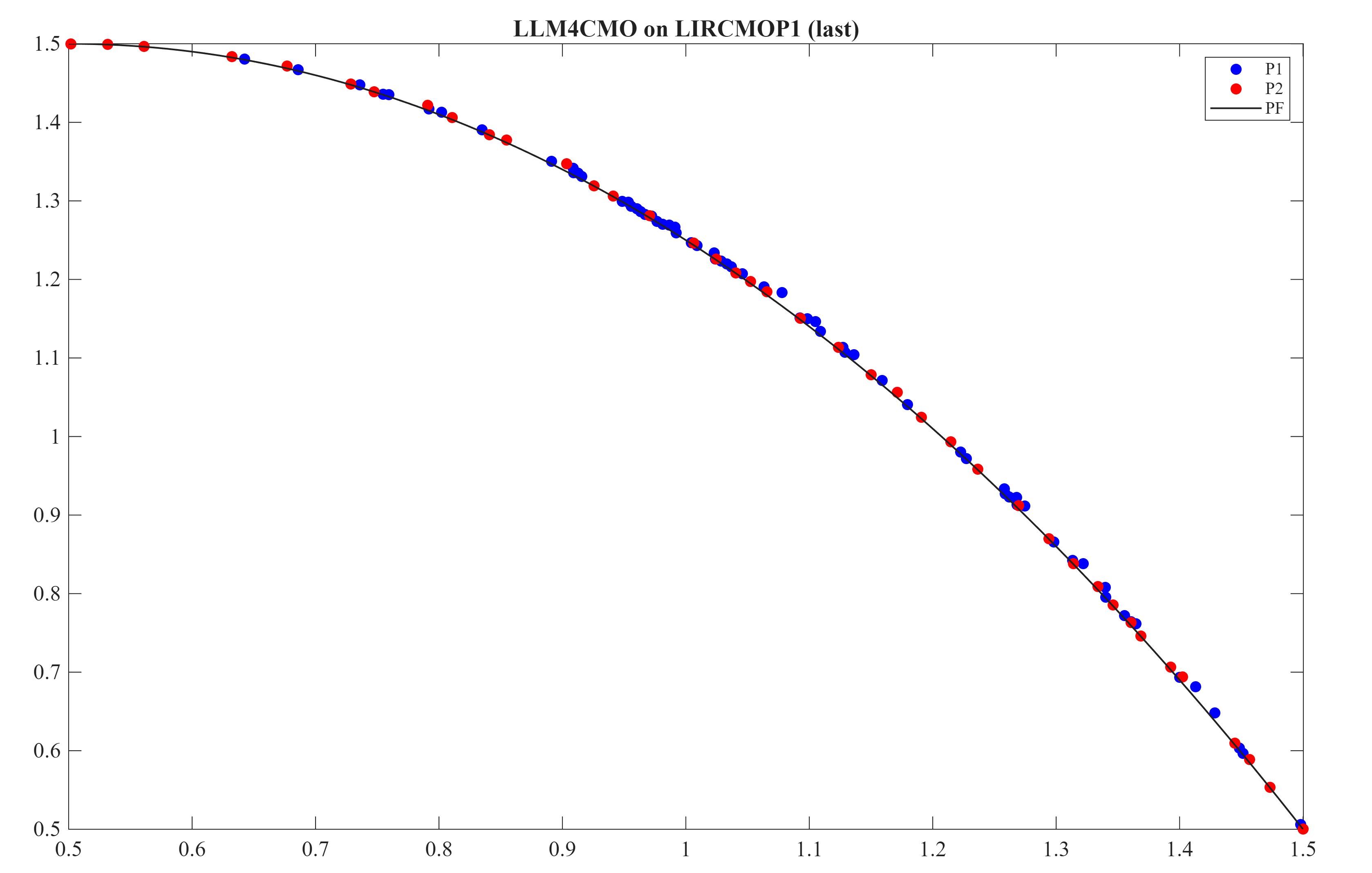}
    \end{minipage}
    
    \caption{The process charge of P1, P2 and of Bico, URCMO and LLM4CMOon LIRCMOP1(Type-4).}
    \label{fig:3ALIR1}
\end{figure}

\begin{figure}[htbp]
    \centering
    \begin{minipage}{0.3\textwidth}
        \centering
        \includegraphics[width=\textwidth]{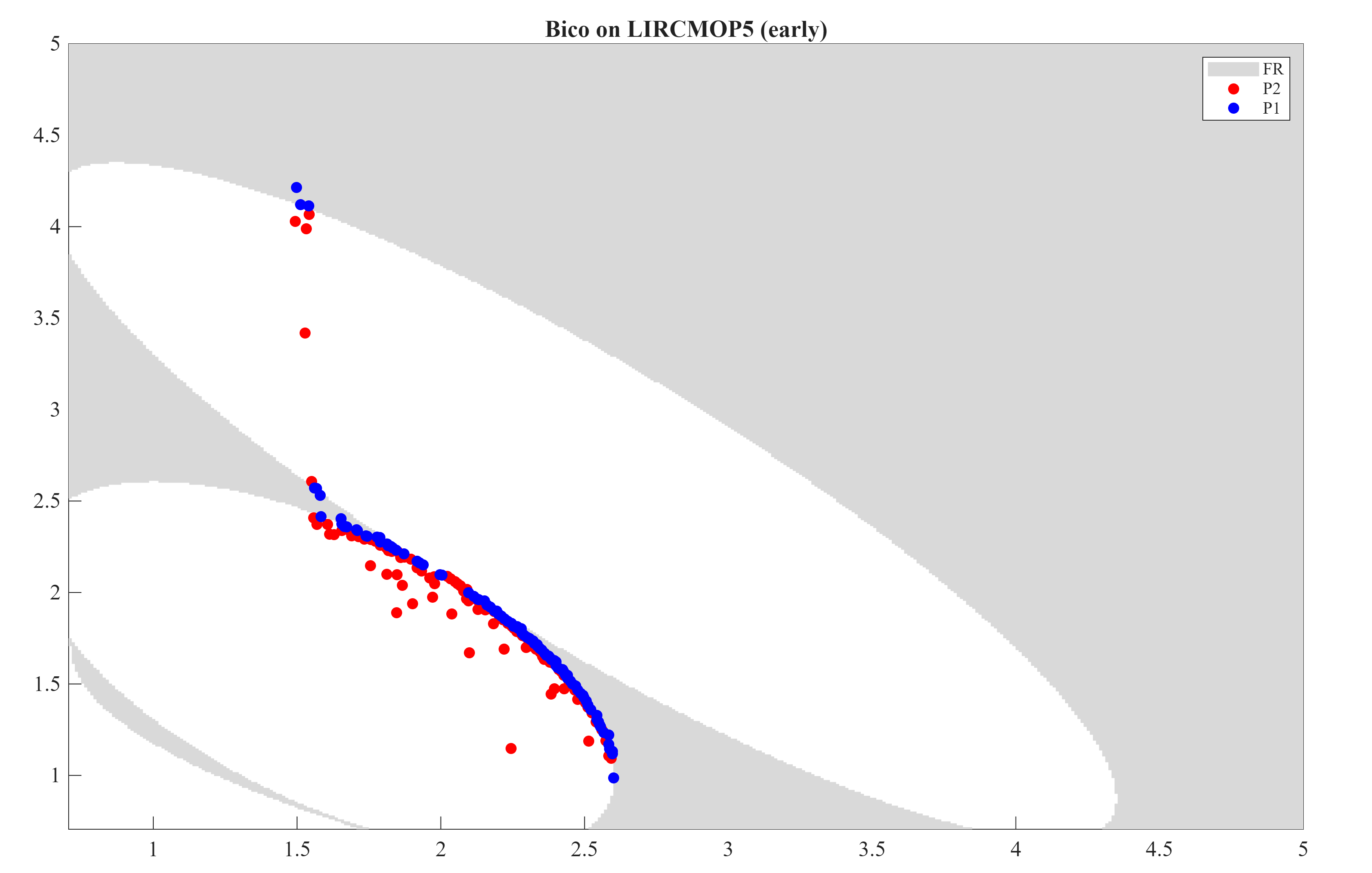}
    \end{minipage}
    \hfill
    \begin{minipage}{0.3\textwidth}
        \centering
        \includegraphics[width=\textwidth]{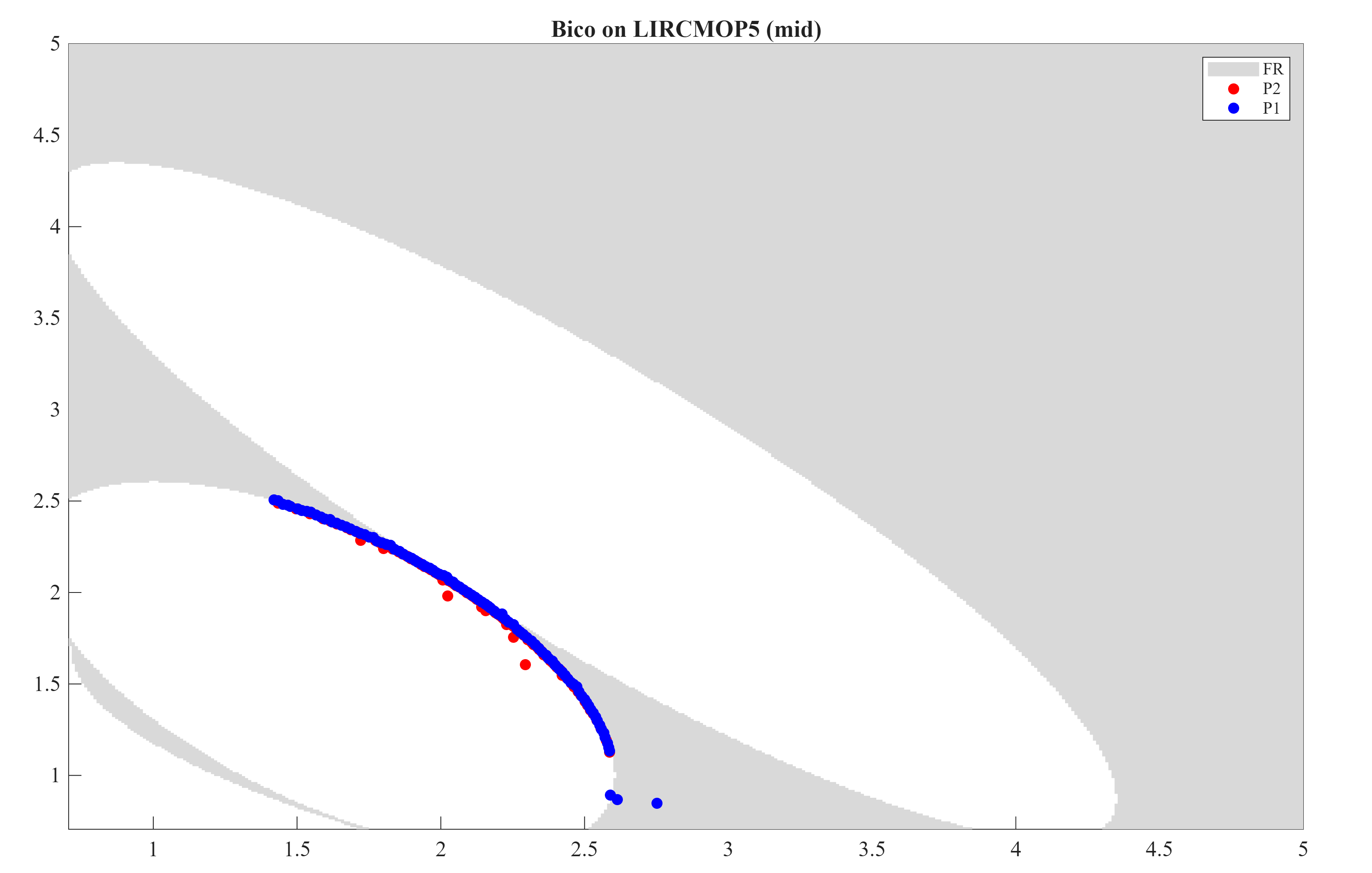}
    \end{minipage}
    \hfill
    \begin{minipage}{0.3\textwidth}
        \centering
        \includegraphics[width=\textwidth]{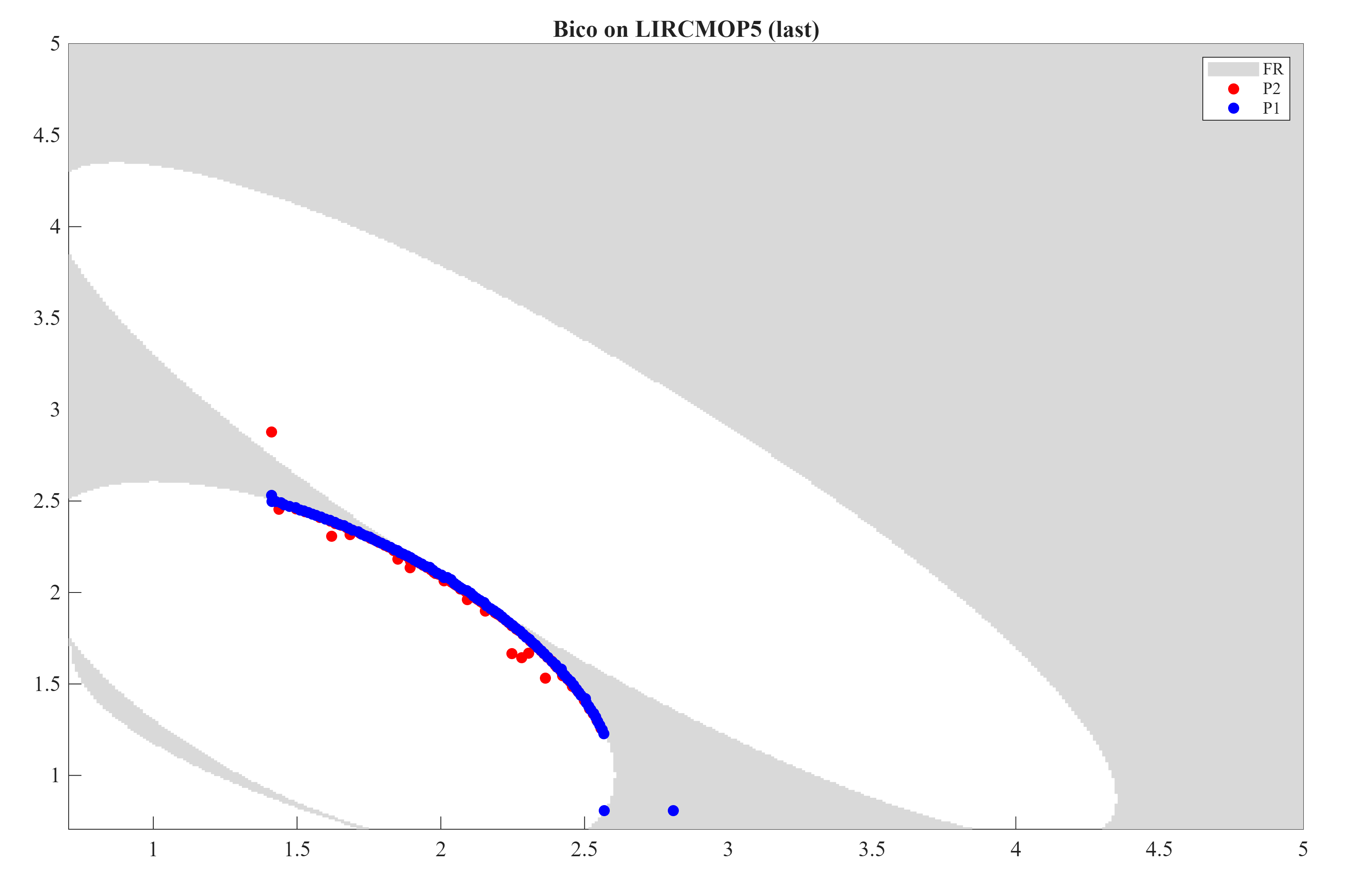}
    \end{minipage}
    
    
    \begin{minipage}{0.3\textwidth}
        \centering
        \includegraphics[width=\textwidth]{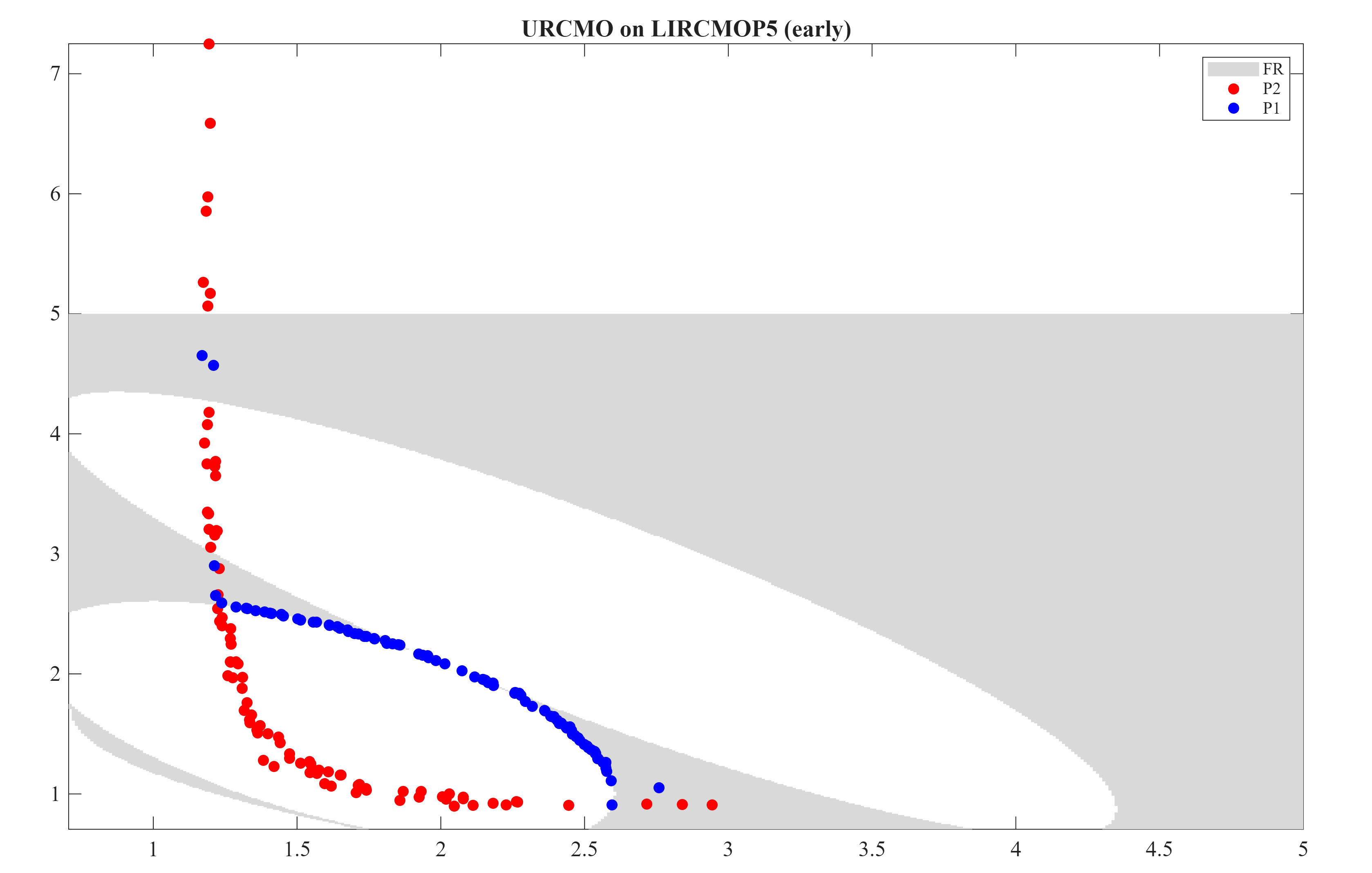}
    \end{minipage}
    \hfill
    \begin{minipage}{0.3\textwidth}
        \centering
        \includegraphics[width=\textwidth]{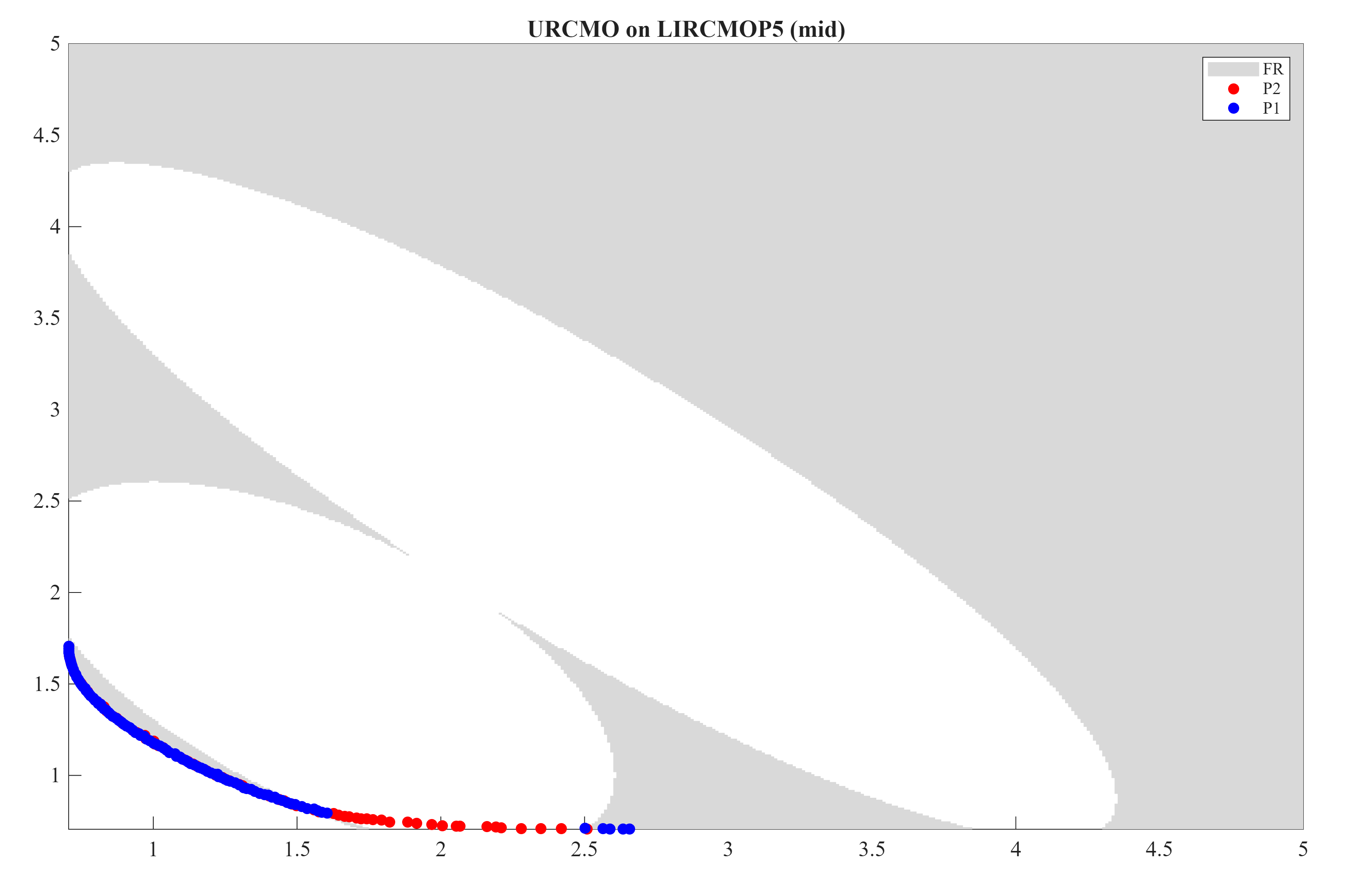}
    \end{minipage}
    \hfill
    \begin{minipage}{0.3\textwidth}
        \centering
        \includegraphics[width=\textwidth]{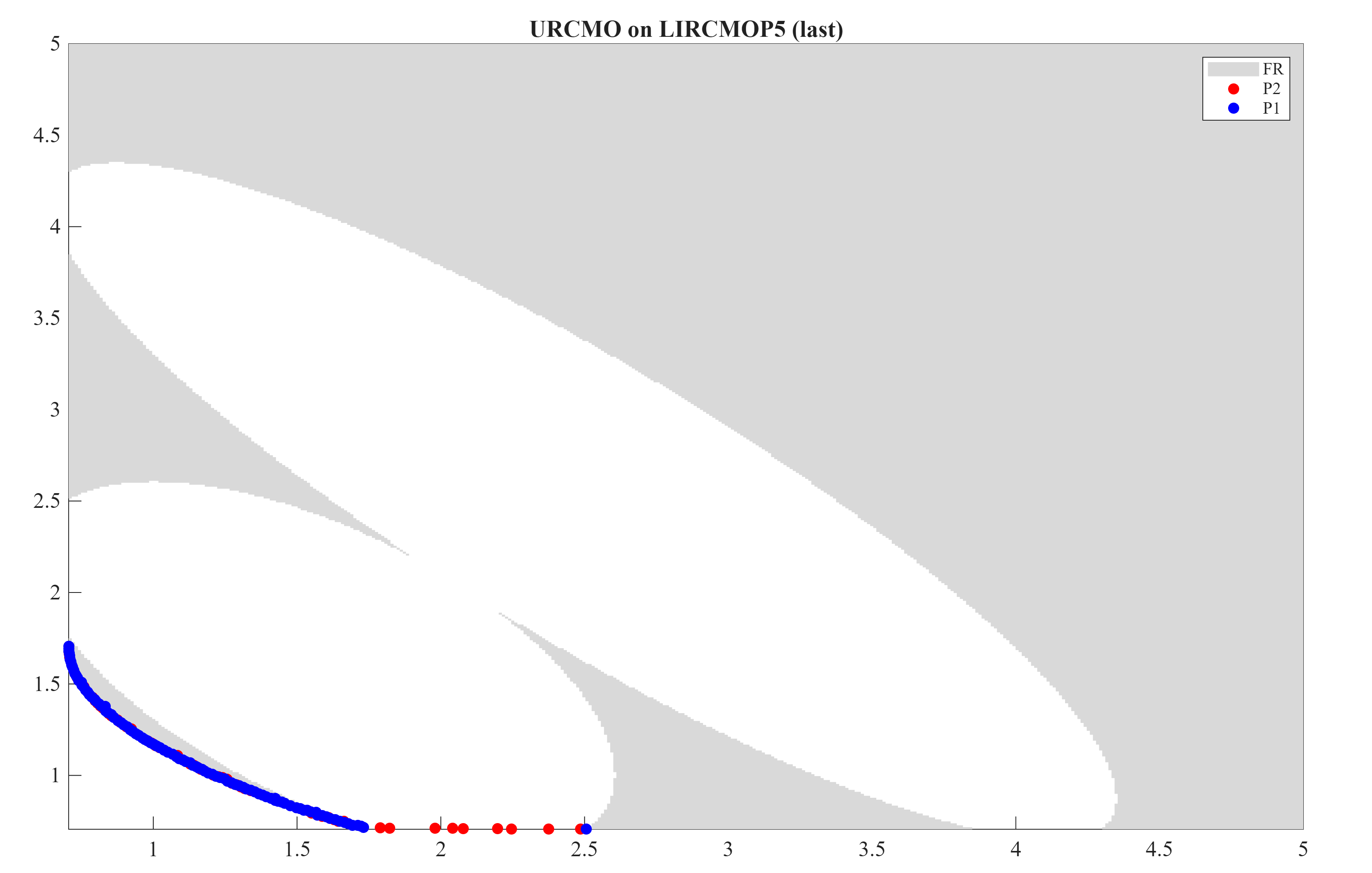}
    \end{minipage}
    
    
    \begin{minipage}{0.3\textwidth}
        \centering
        \includegraphics[width=\textwidth]{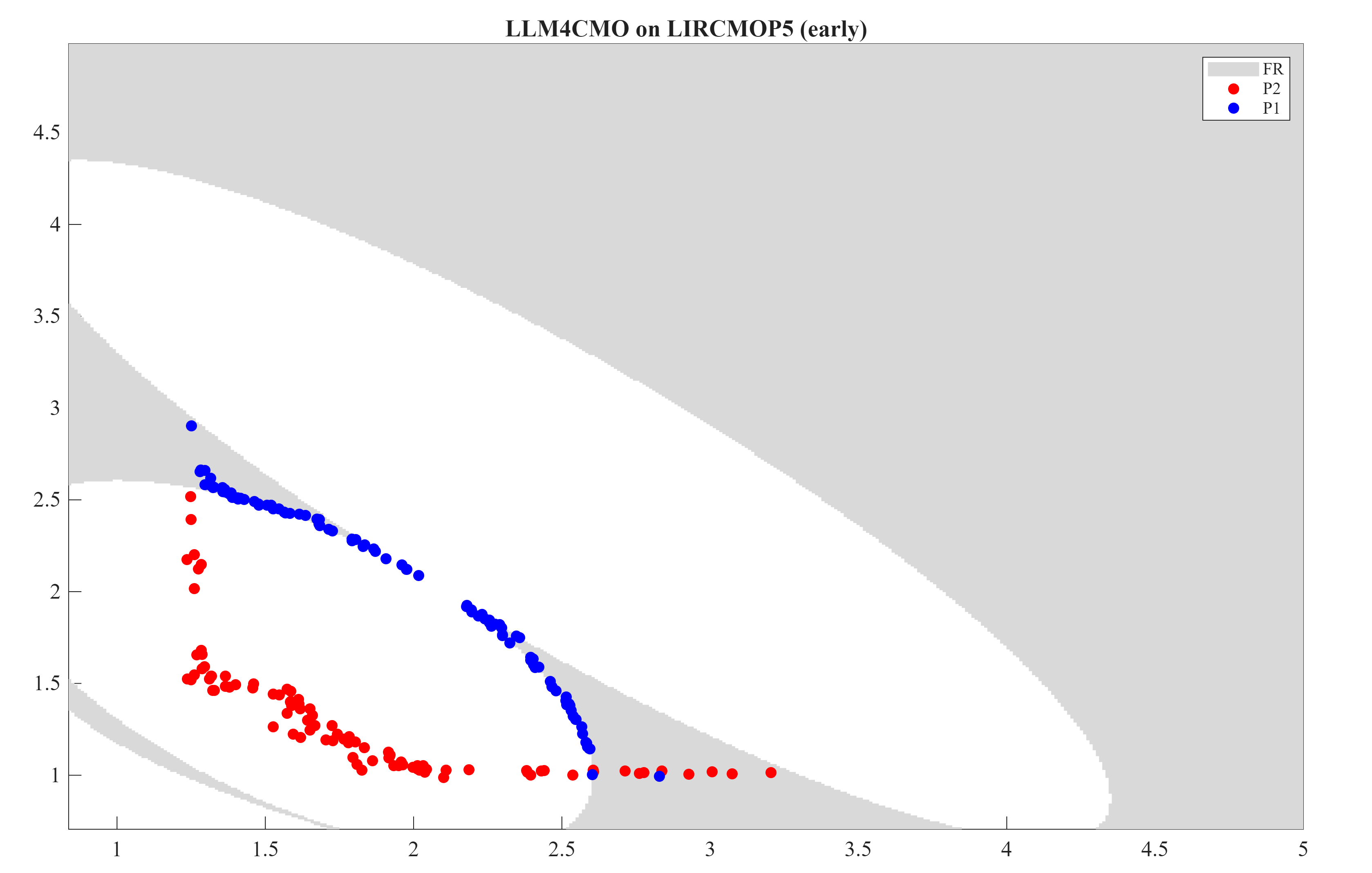}
    \end{minipage}
    \hfill
    \begin{minipage}{0.3\textwidth}
        \centering
        \includegraphics[width=\textwidth]{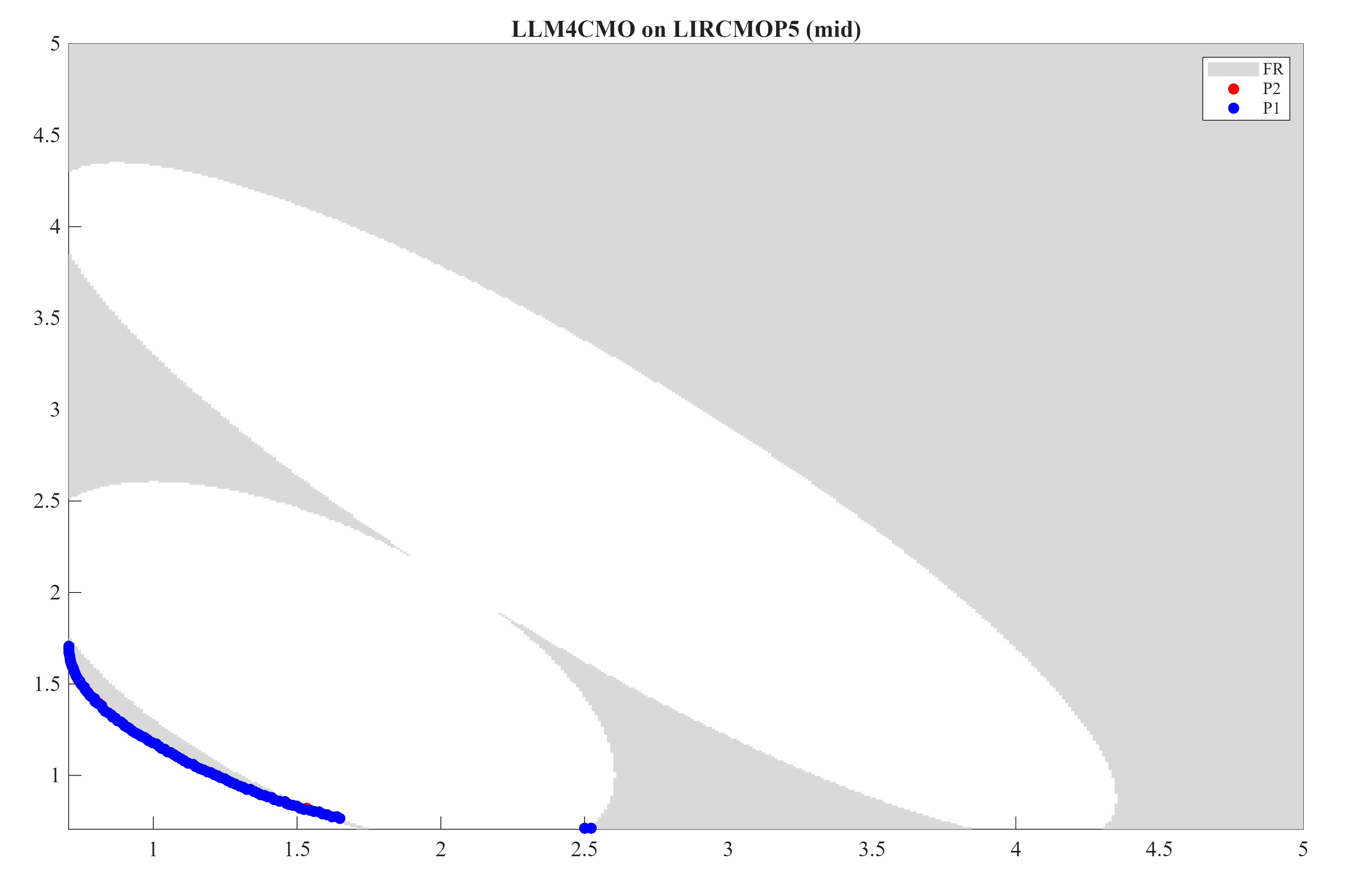}
    \end{minipage}
    \hfill
    \begin{minipage}{0.3\textwidth}
        \centering
        \includegraphics[width=\textwidth]{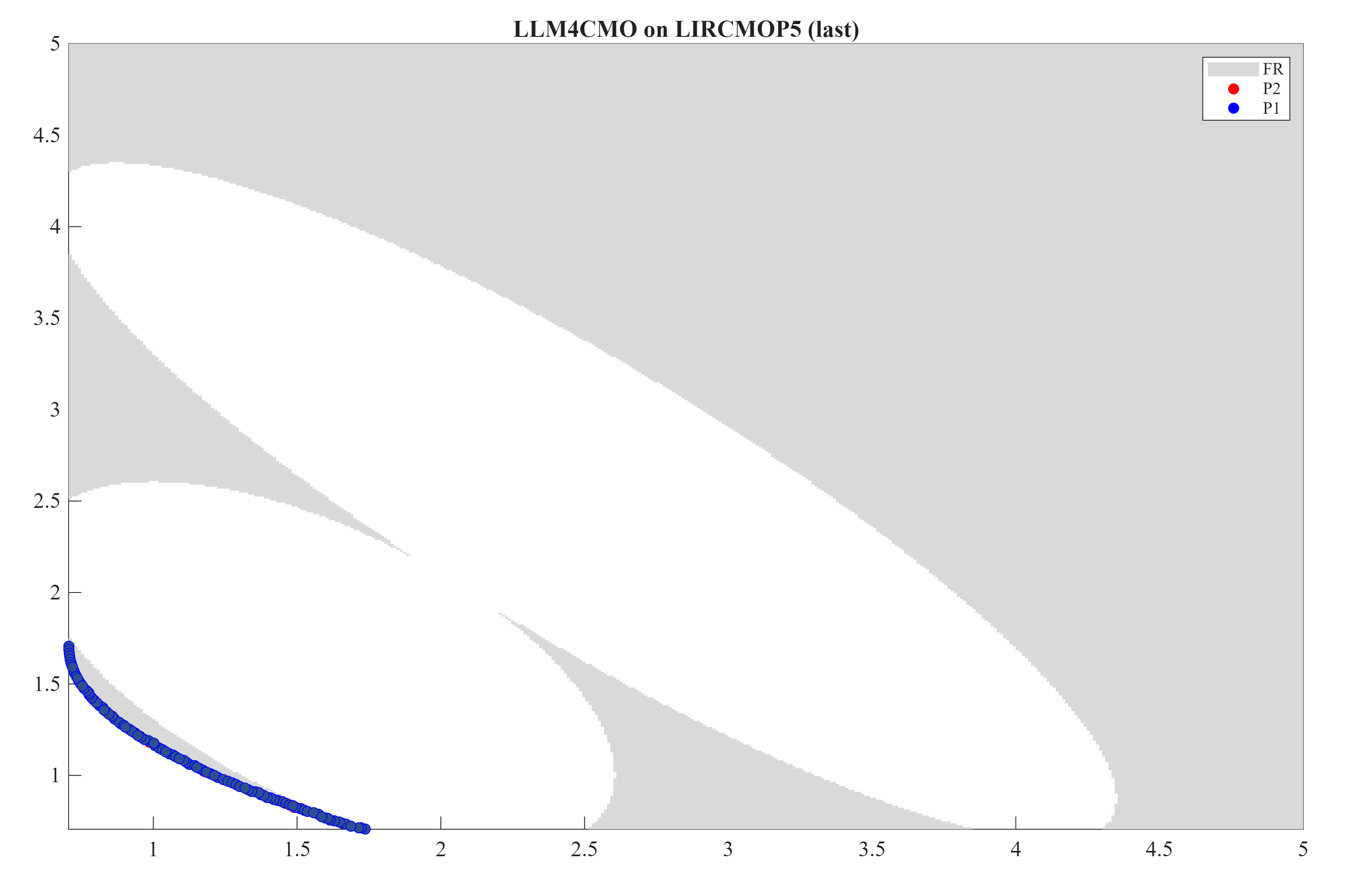}
    \end{minipage}
    
    \caption{The process charge of P1, P2 and of Bico, URCMO and LLM4CMOon LIRCMOP5(Type-1).}
    \label{fig:3ALIR5}
\end{figure}

\begin{figure}[htbp]
    \centering
    \begin{minipage}{0.3\textwidth}
        \centering
        \includegraphics[width=\textwidth]{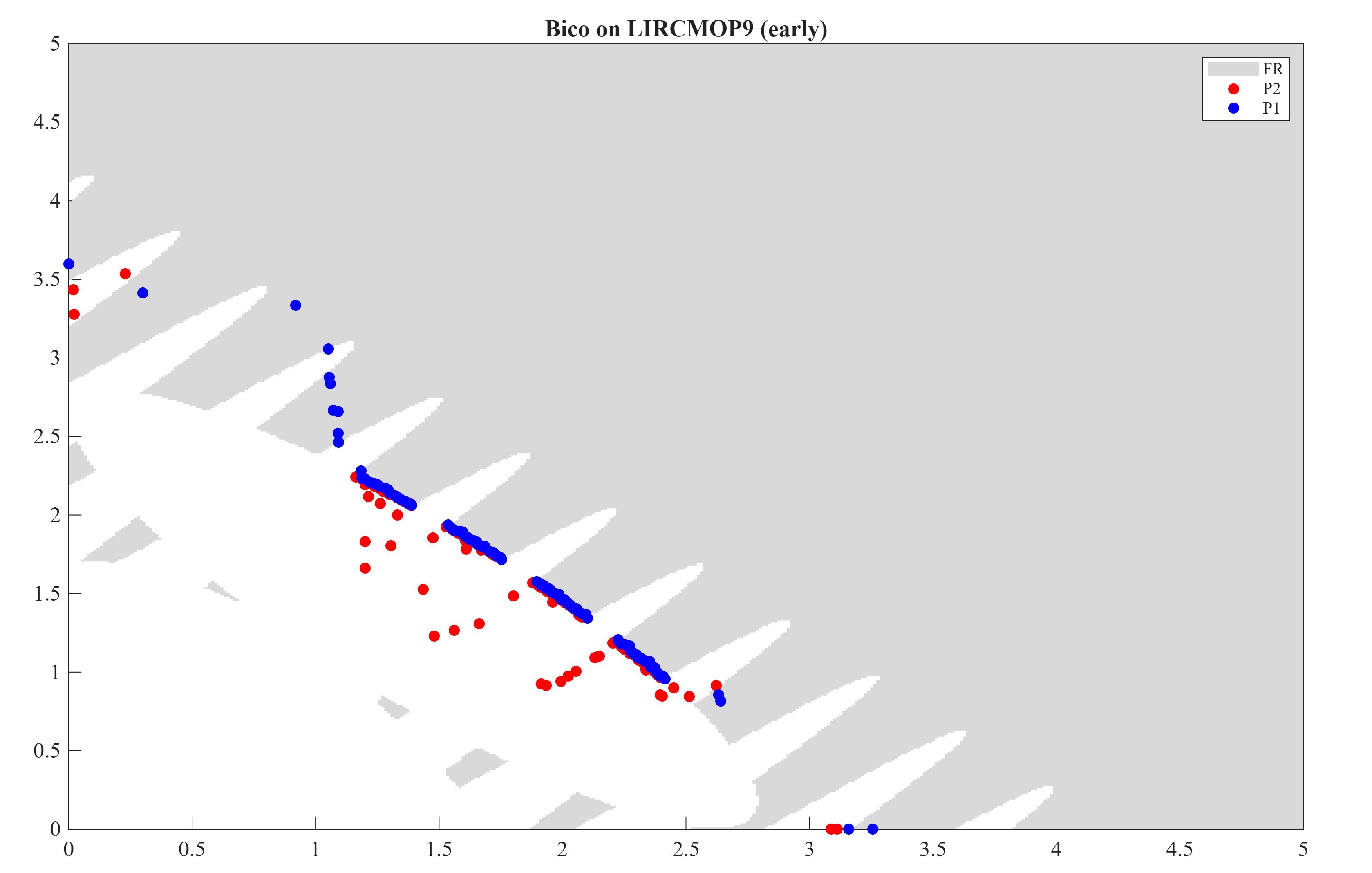}
    \end{minipage}
    \hfill
    \begin{minipage}{0.3\textwidth}
        \centering
        \includegraphics[width=\textwidth]{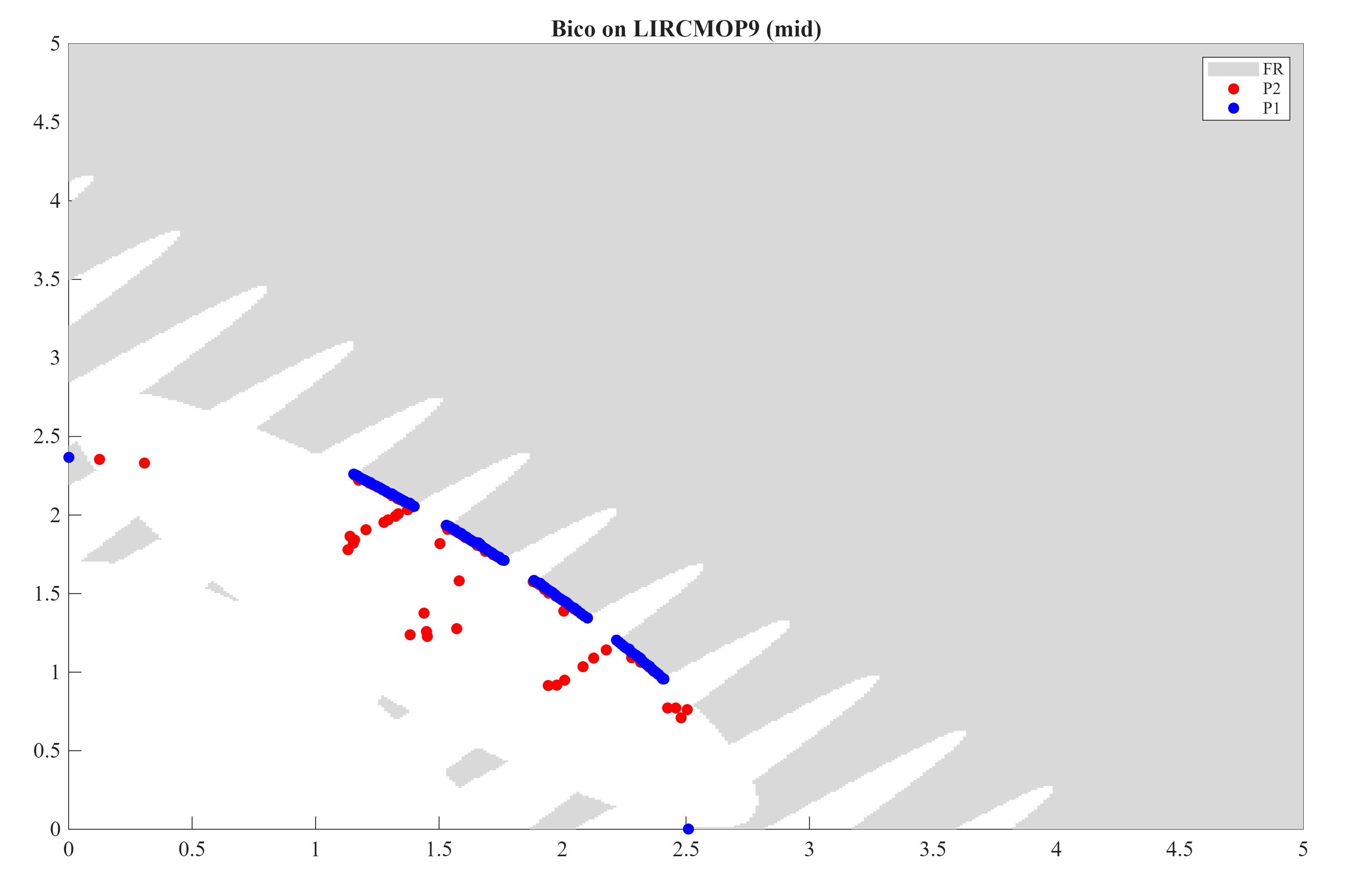}
    \end{minipage}
    \hfill
    \begin{minipage}{0.3\textwidth}
        \centering
        \includegraphics[width=\textwidth]{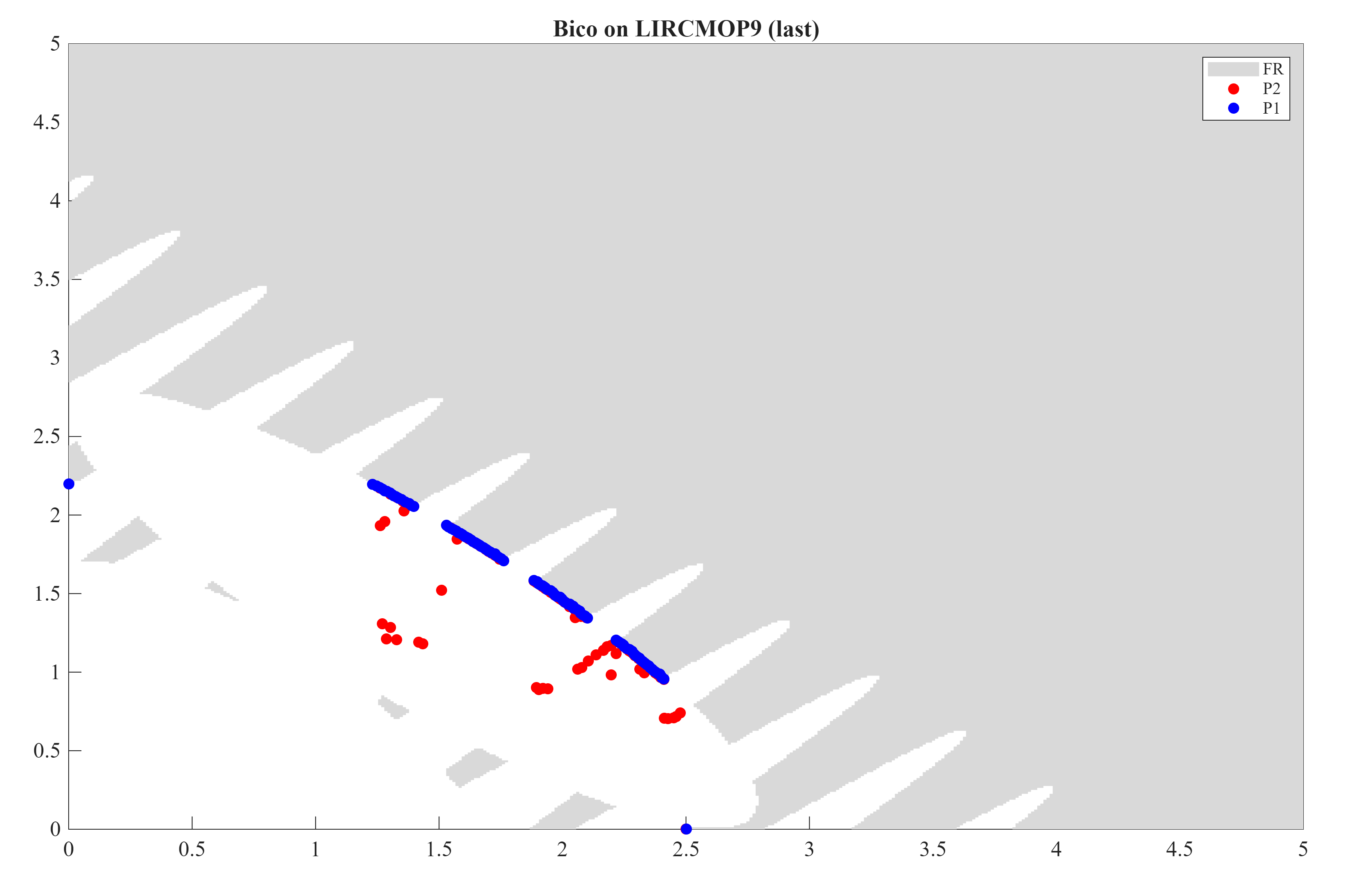}
    \end{minipage}
    
    
    \begin{minipage}{0.3\textwidth}
        \centering
        \includegraphics[width=\textwidth]{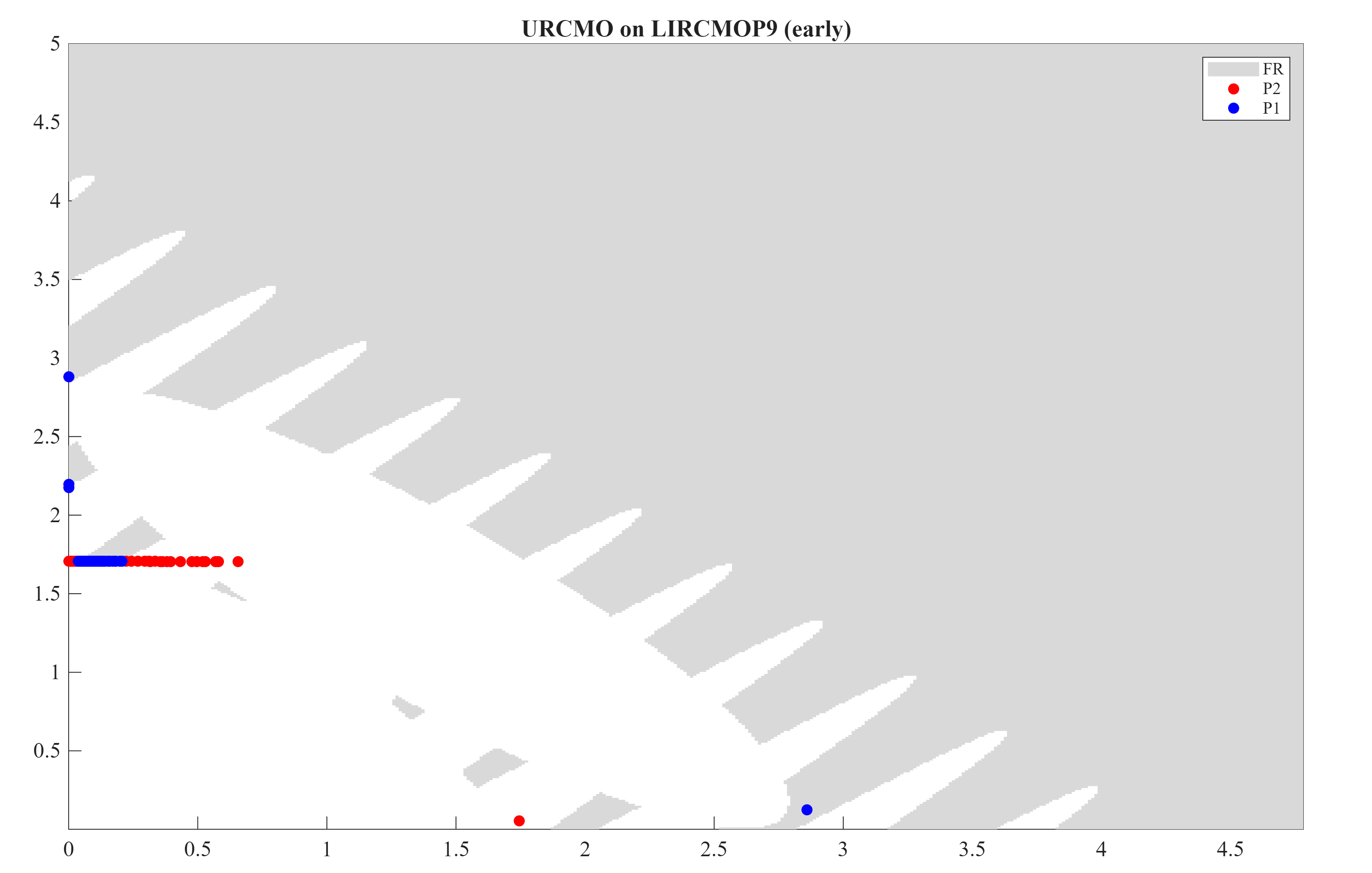}
    \end{minipage}
    \hfill
    \begin{minipage}{0.3\textwidth}
        \centering
        \includegraphics[width=\textwidth]{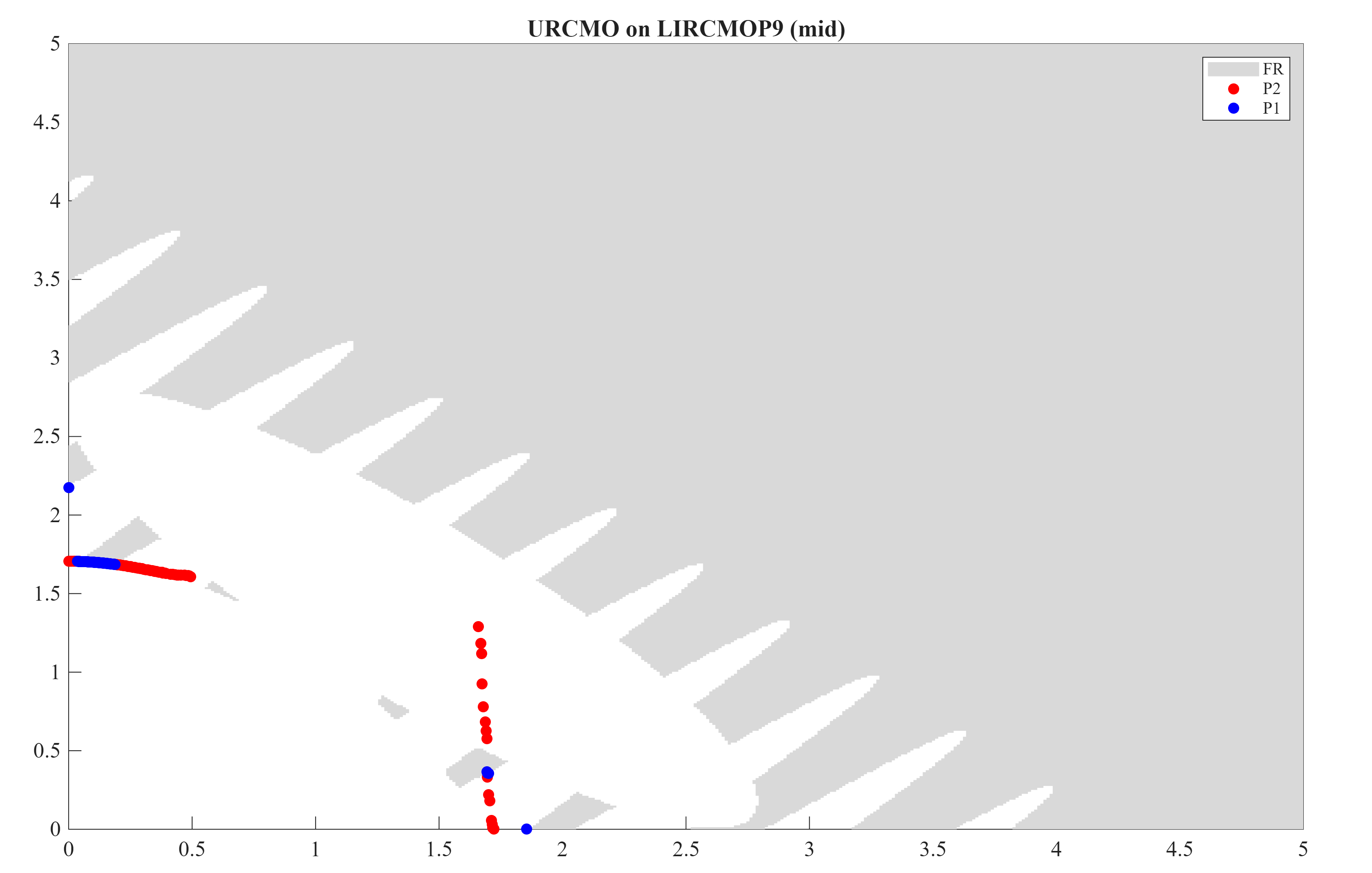}
    \end{minipage}
    \hfill
    \begin{minipage}{0.3\textwidth}
        \centering
        \includegraphics[width=\textwidth]{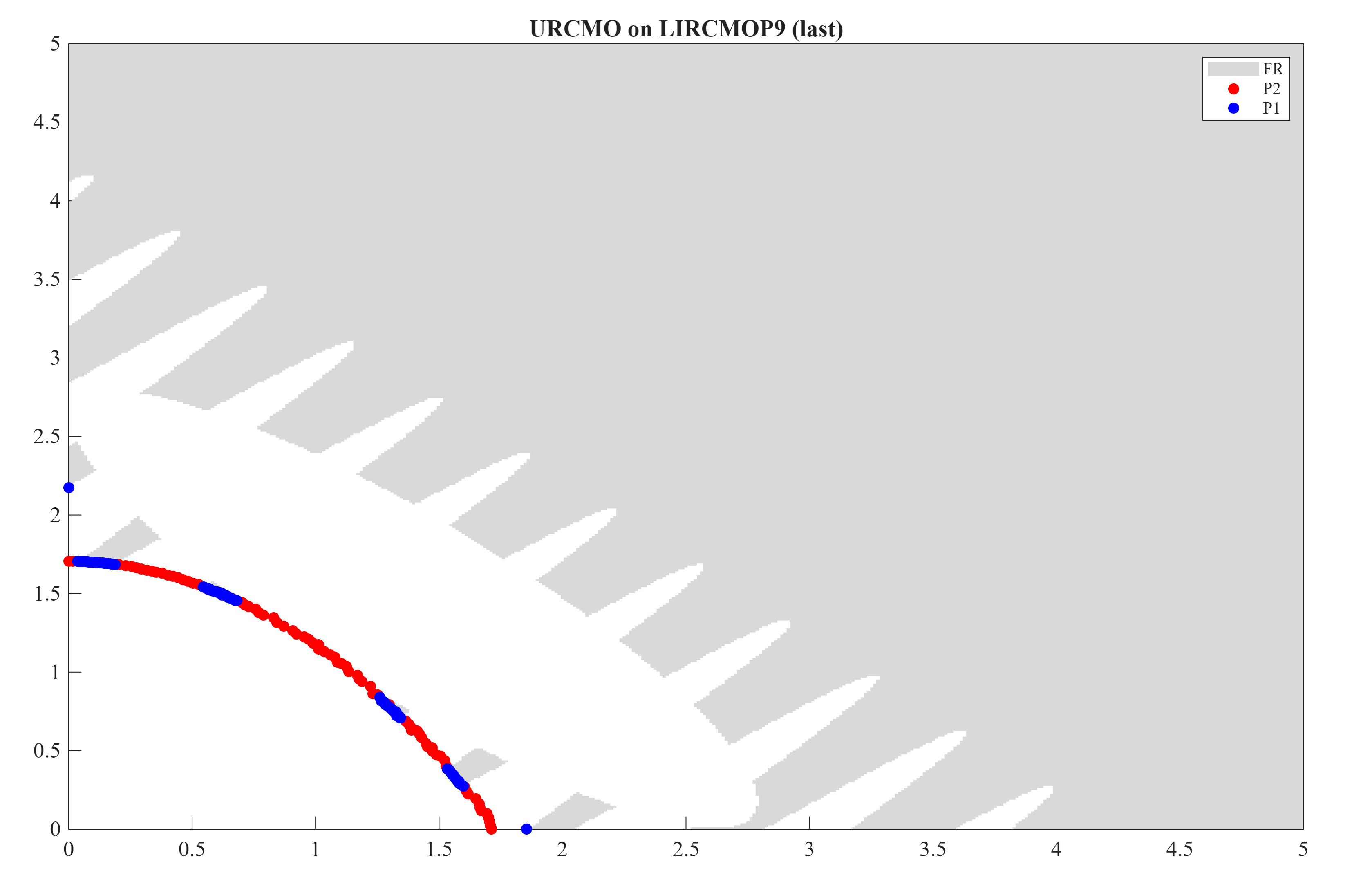}
    \end{minipage}
    
    
    \begin{minipage}{0.3\textwidth}
        \centering
        \includegraphics[width=\textwidth]{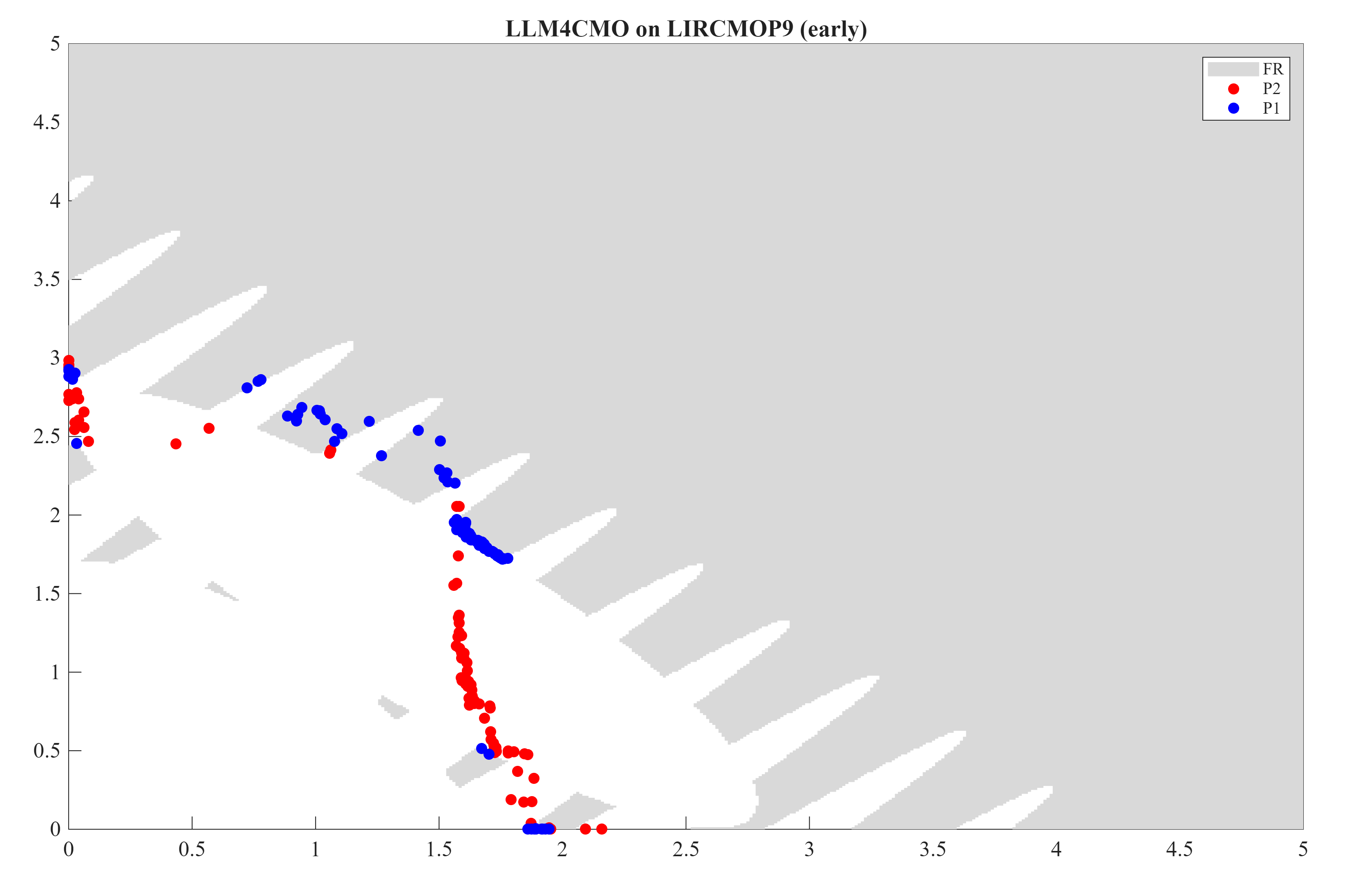}
    \end{minipage}
    \hfill
    \begin{minipage}{0.3\textwidth}
        \centering
        \includegraphics[width=\textwidth]{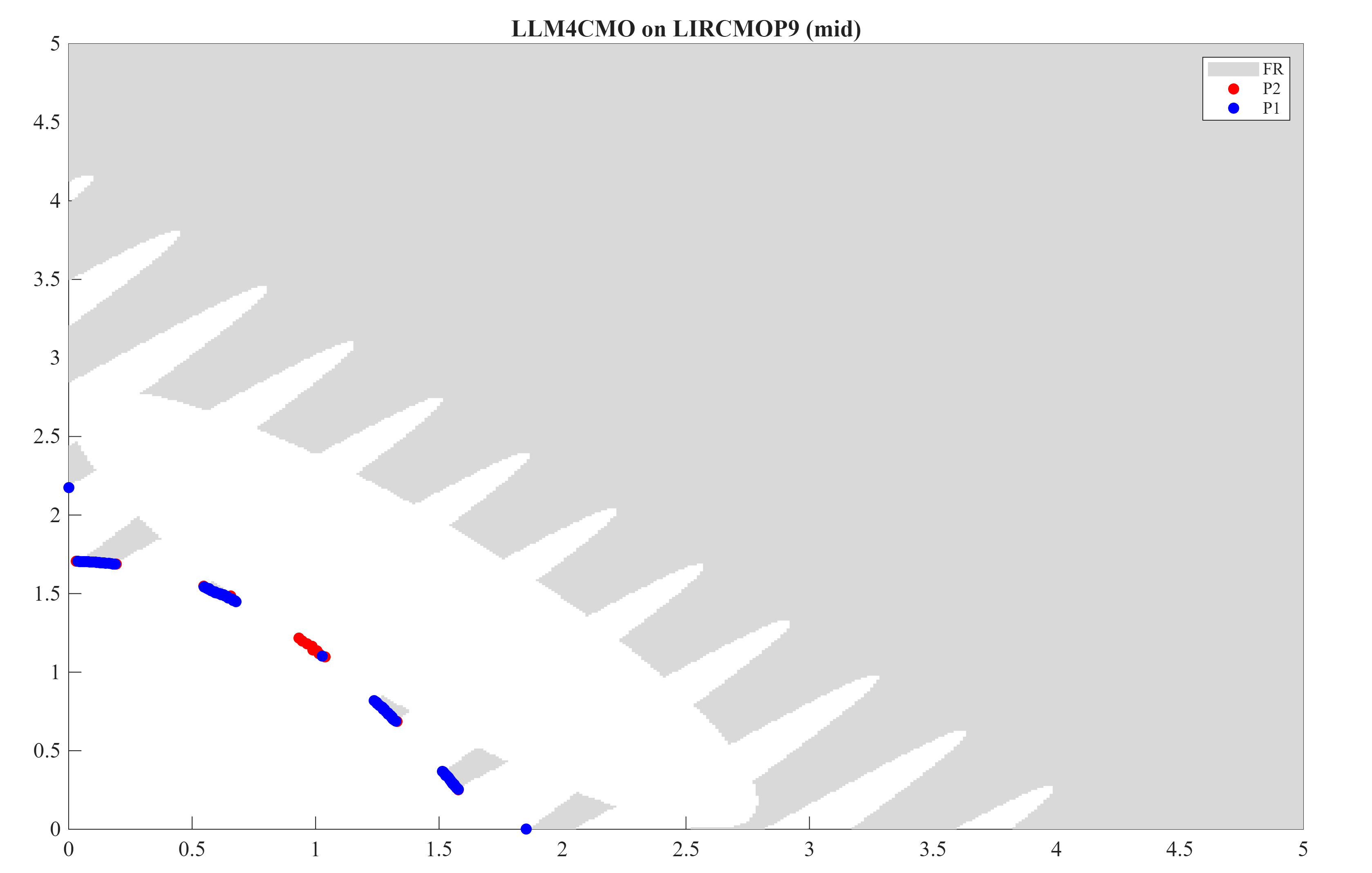}
    \end{minipage}
    \hfill
    \begin{minipage}{0.3\textwidth}
        \centering
        \includegraphics[width=\textwidth]{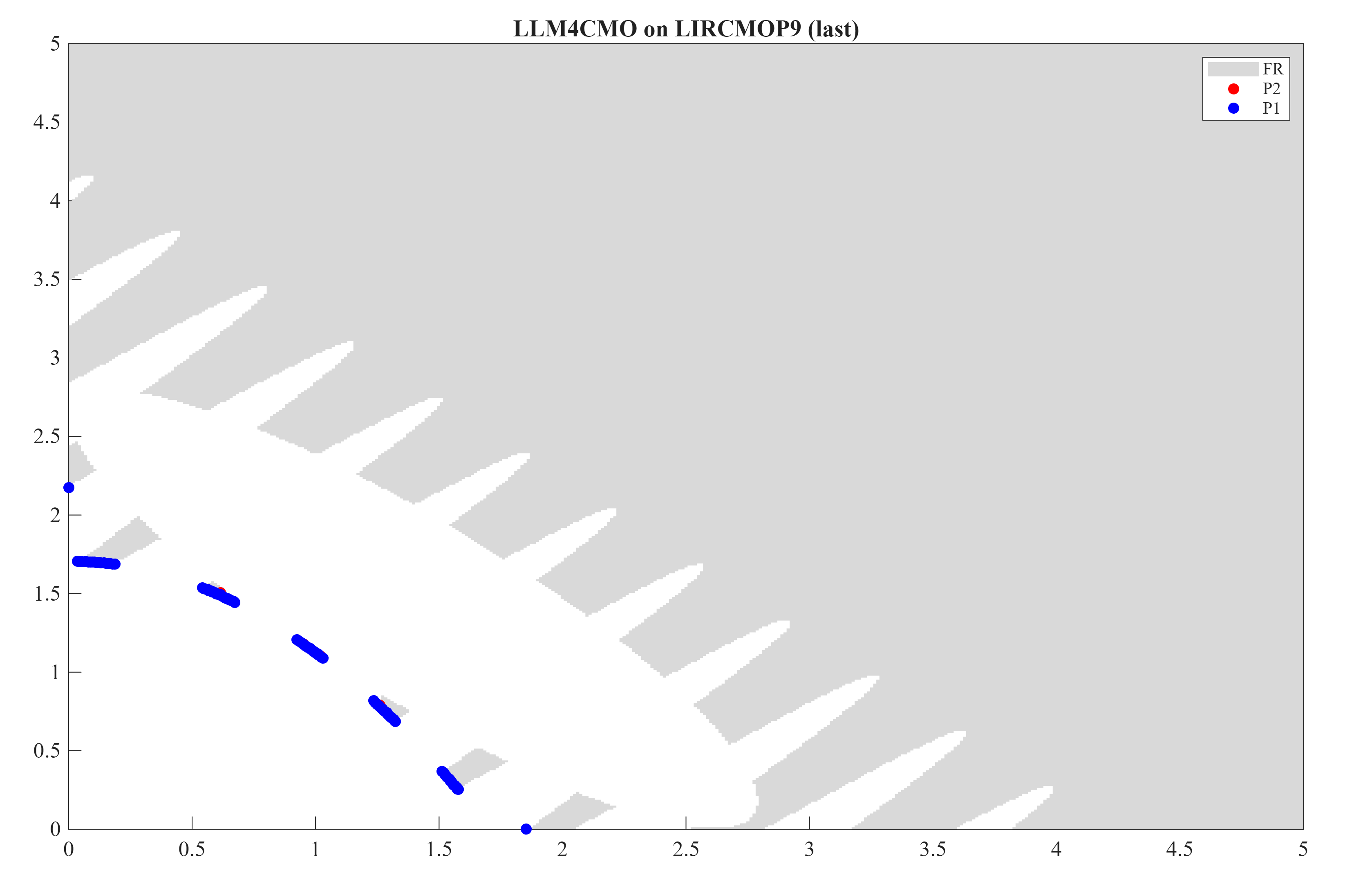}
    \end{minipage}
    
    \caption{The process charge of P1, P2 and  of Bico, URCMO and LLM4CMOon LIRCMOP9(Type-2).}
    \label{fig:3ALIR5}
\end{figure}

\begin{figure}[htbp]
    \centering
    \begin{minipage}{0.3\textwidth}
        \centering
        \includegraphics[width=\textwidth]{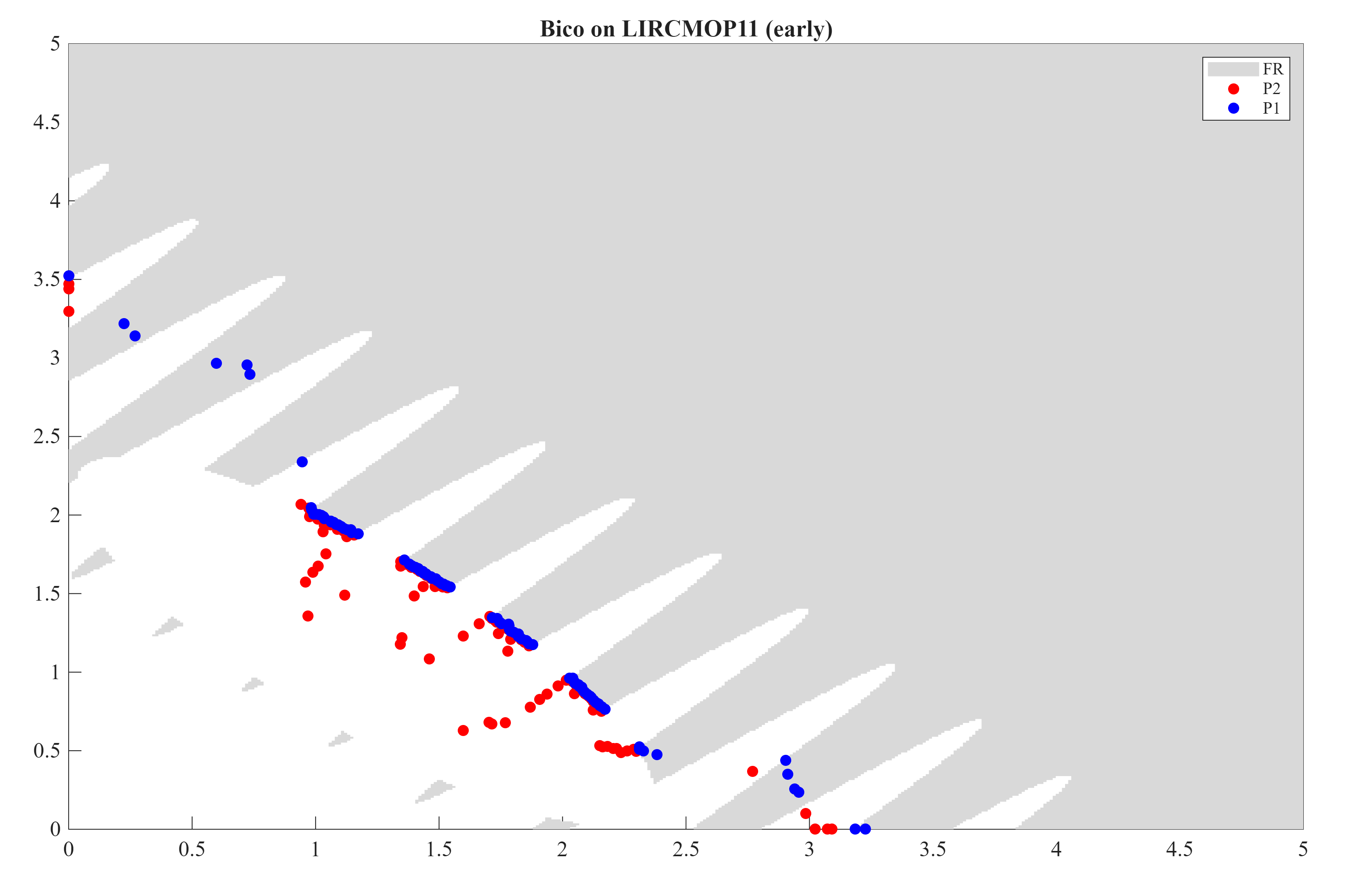}
    \end{minipage}
    \hfill
    \begin{minipage}{0.3\textwidth}
        \centering
        \includegraphics[width=\textwidth]{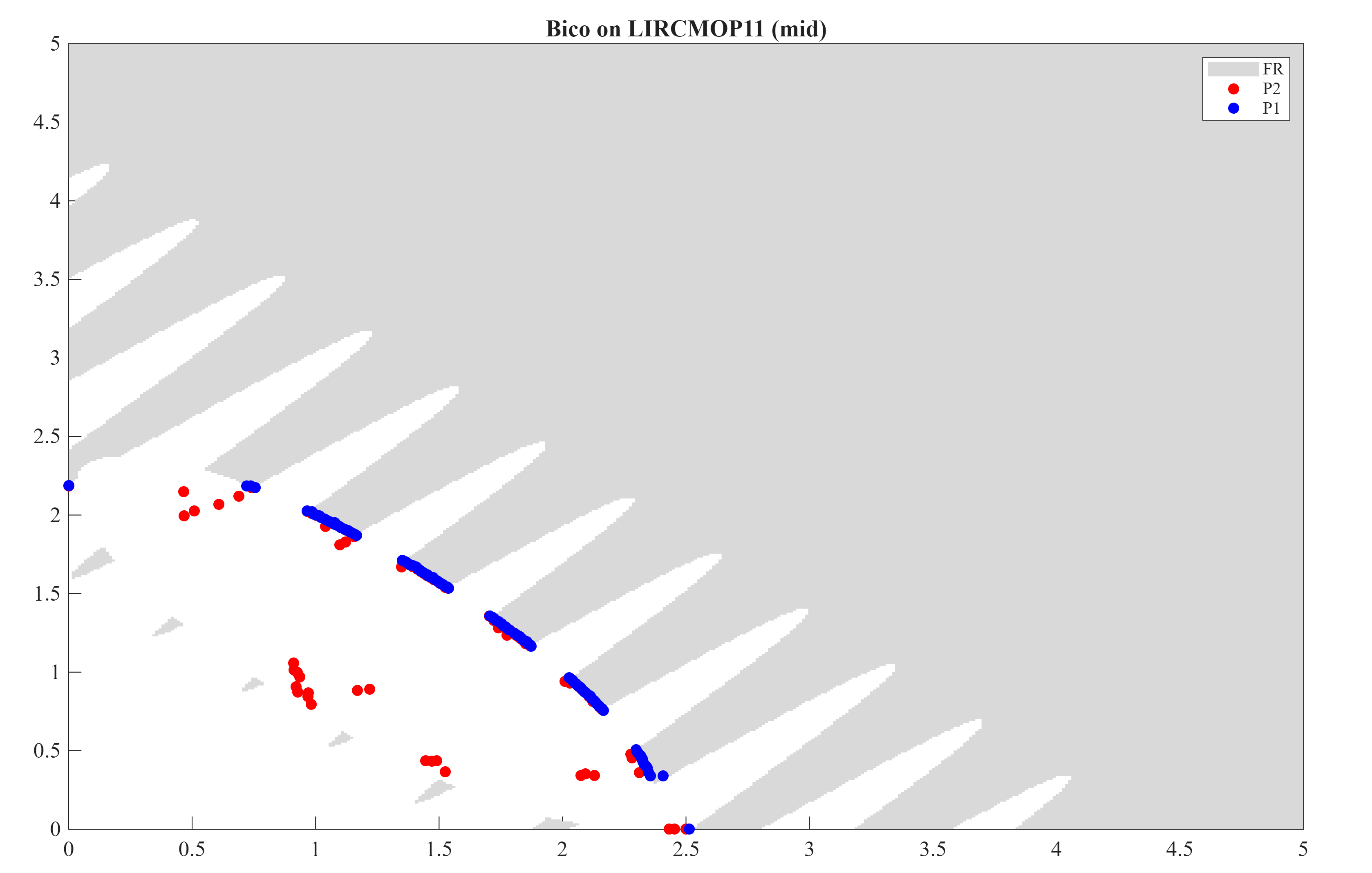}
    \end{minipage}
    \hfill
    \begin{minipage}{0.3\textwidth}
        \centering
        \includegraphics[width=\textwidth]{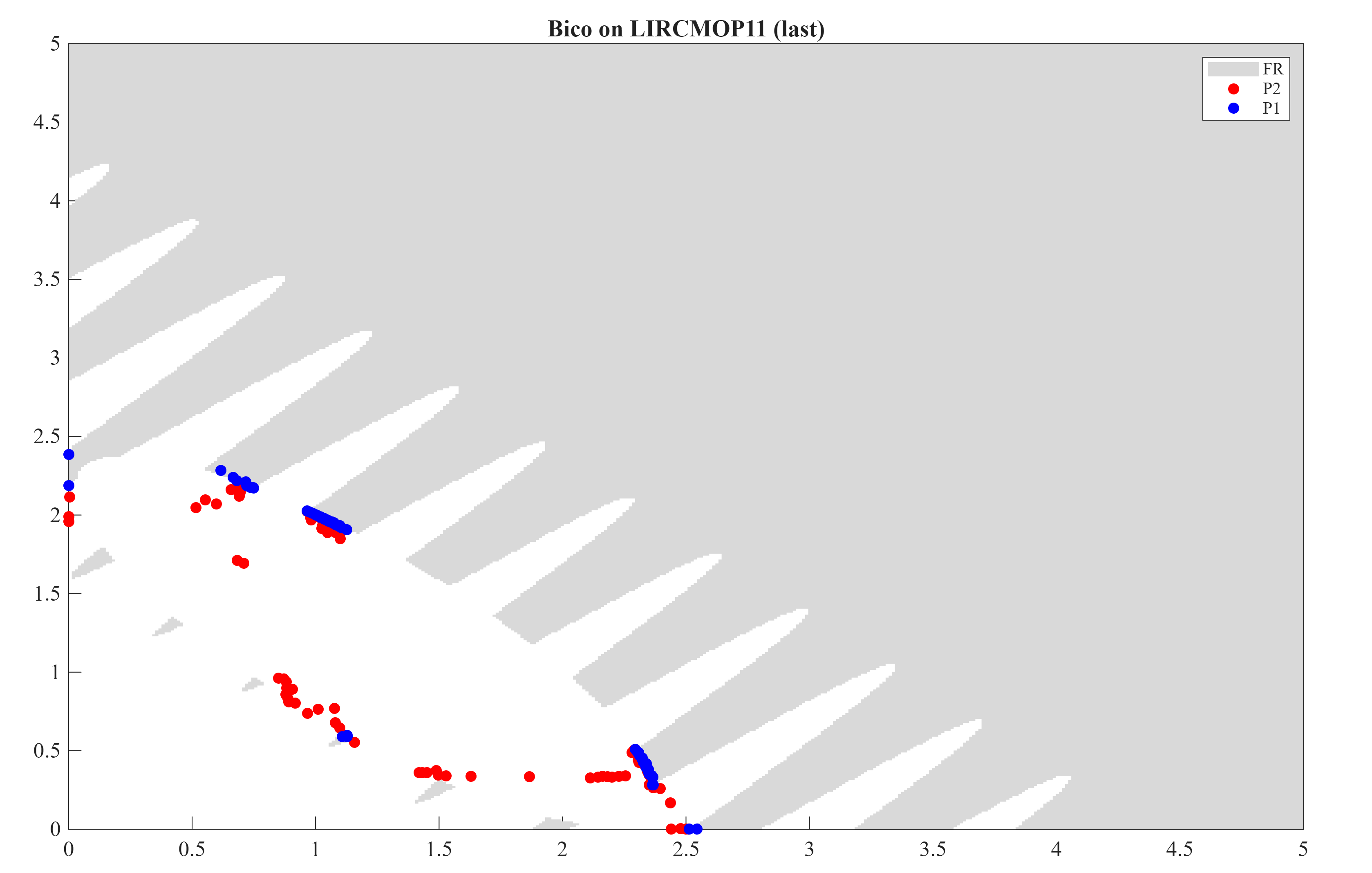}
    \end{minipage}
    
    
    \begin{minipage}{0.3\textwidth}
        \centering
        \includegraphics[width=\textwidth]{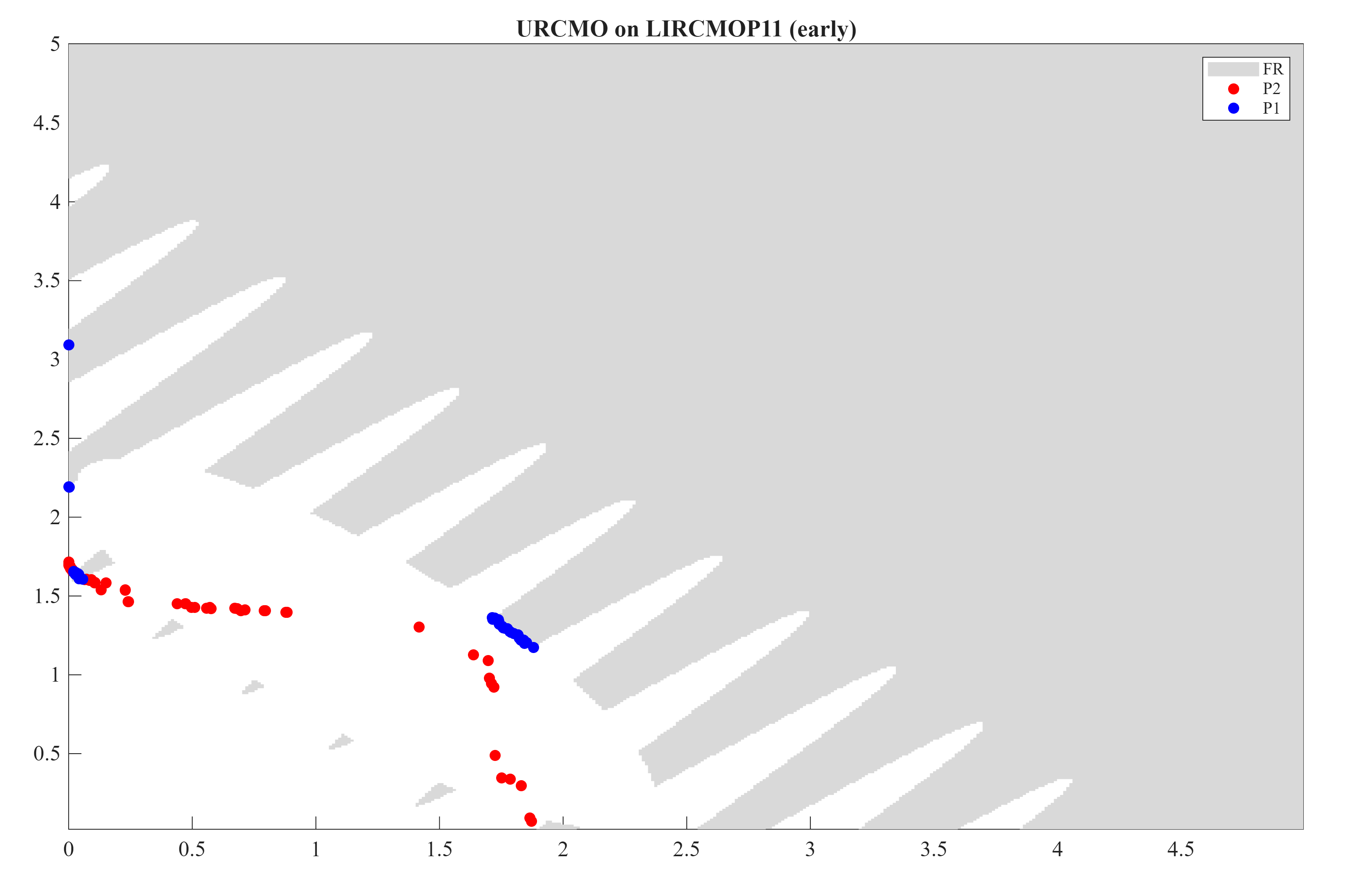}
    \end{minipage}
    \hfill
    \begin{minipage}{0.3\textwidth}
        \centering
        \includegraphics[width=\textwidth]{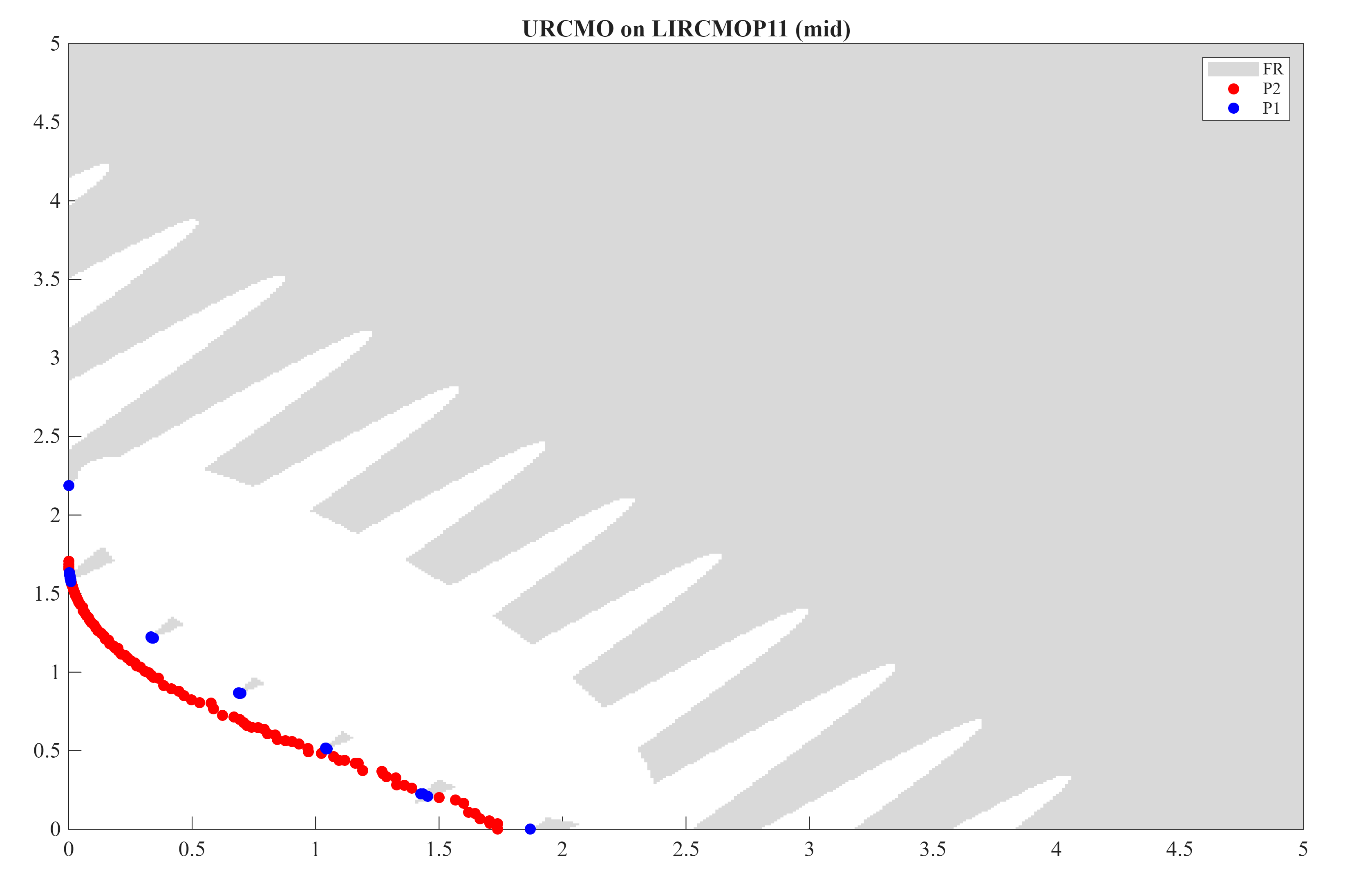}
    \end{minipage}
    \hfill
    \begin{minipage}{0.3\textwidth}
        \centering
        \includegraphics[width=\textwidth]{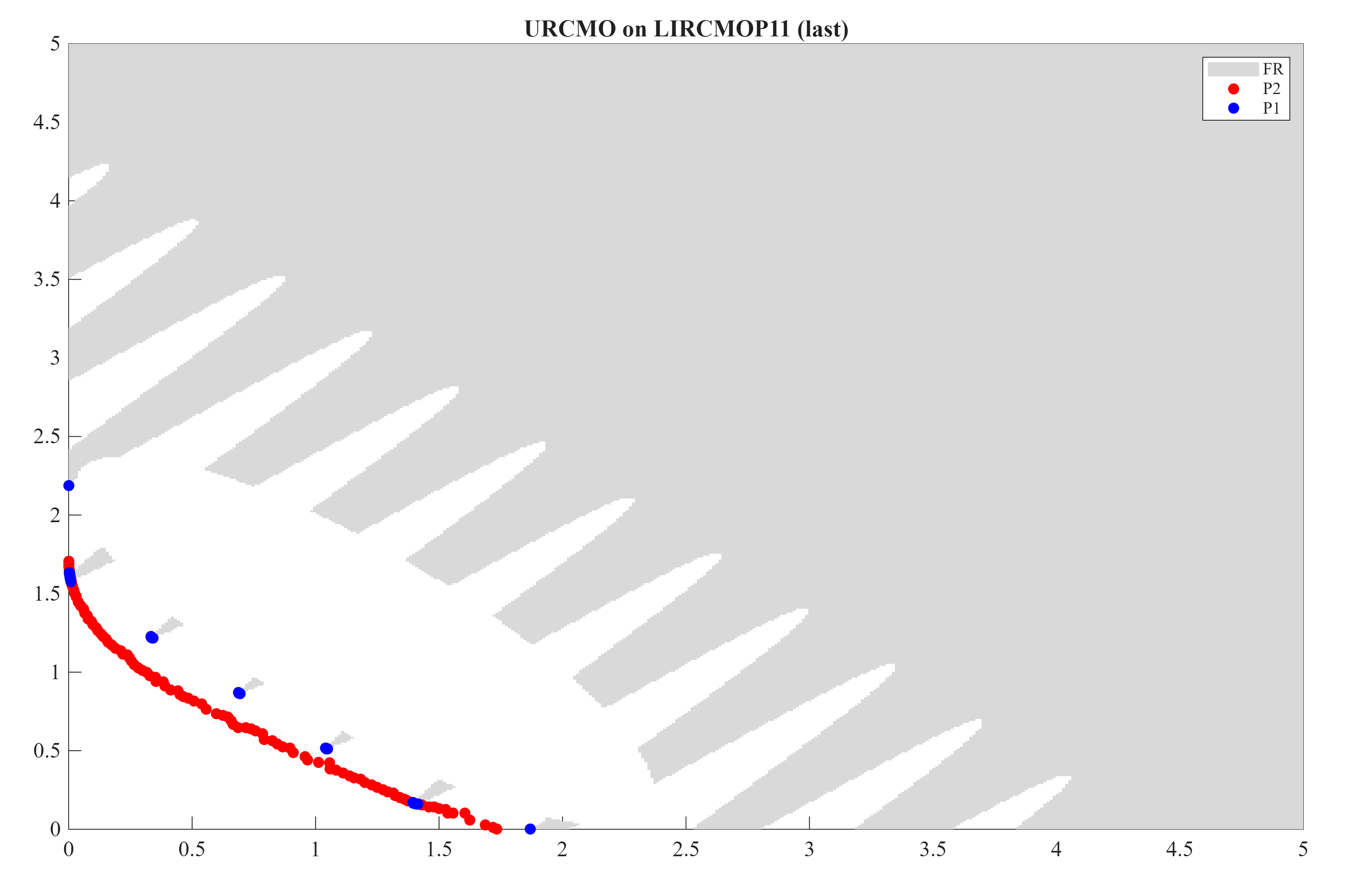}
    \end{minipage}
    
    
    \begin{minipage}{0.3\textwidth}
        \centering
        \includegraphics[width=\textwidth]{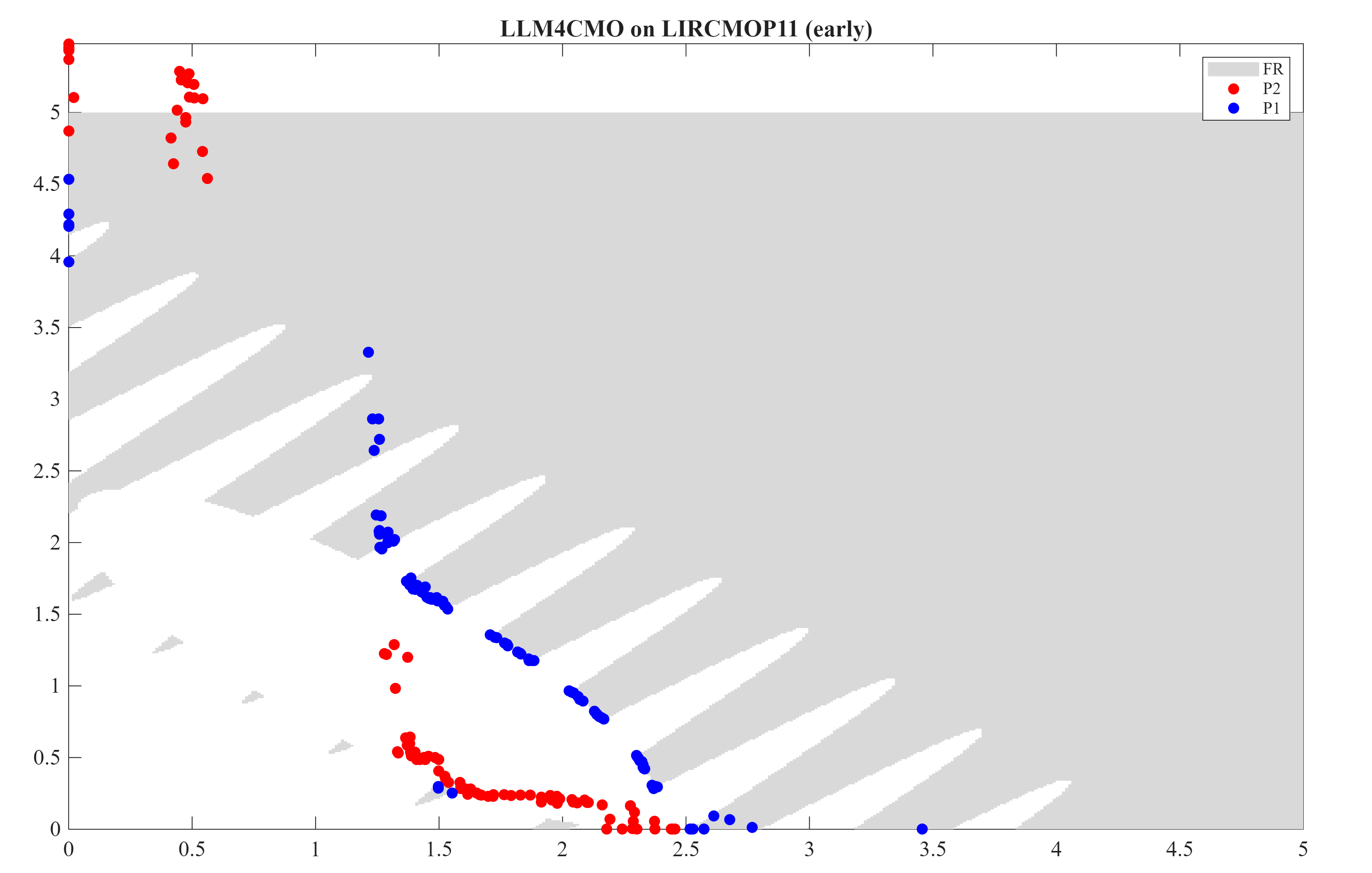}
    \end{minipage}
    \hfill
    \begin{minipage}{0.3\textwidth}
        \centering
        \includegraphics[width=\textwidth]{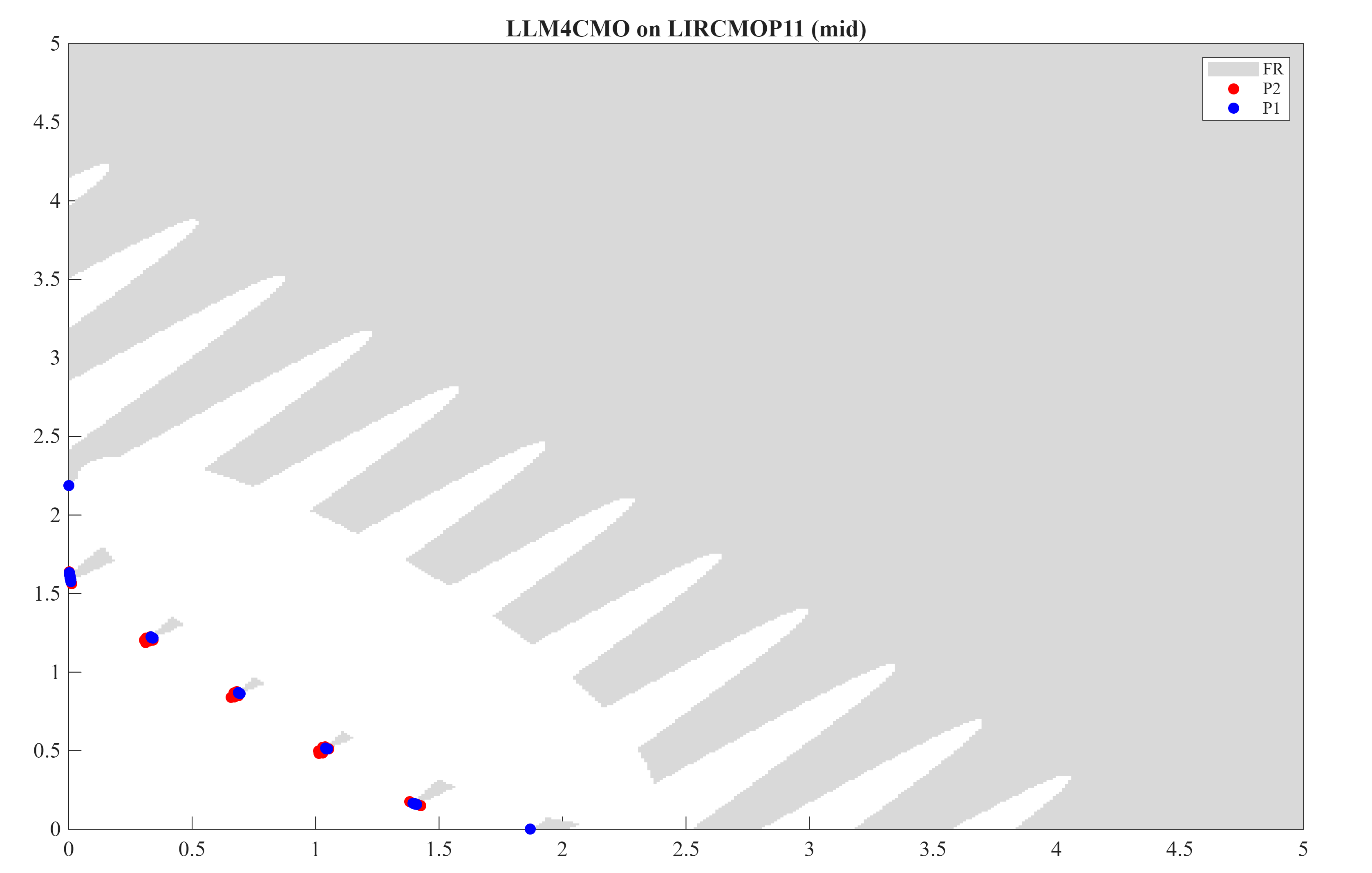}
    \end{minipage}
    \hfill
    \begin{minipage}{0.3\textwidth}
        \centering
        \includegraphics[width=\textwidth]{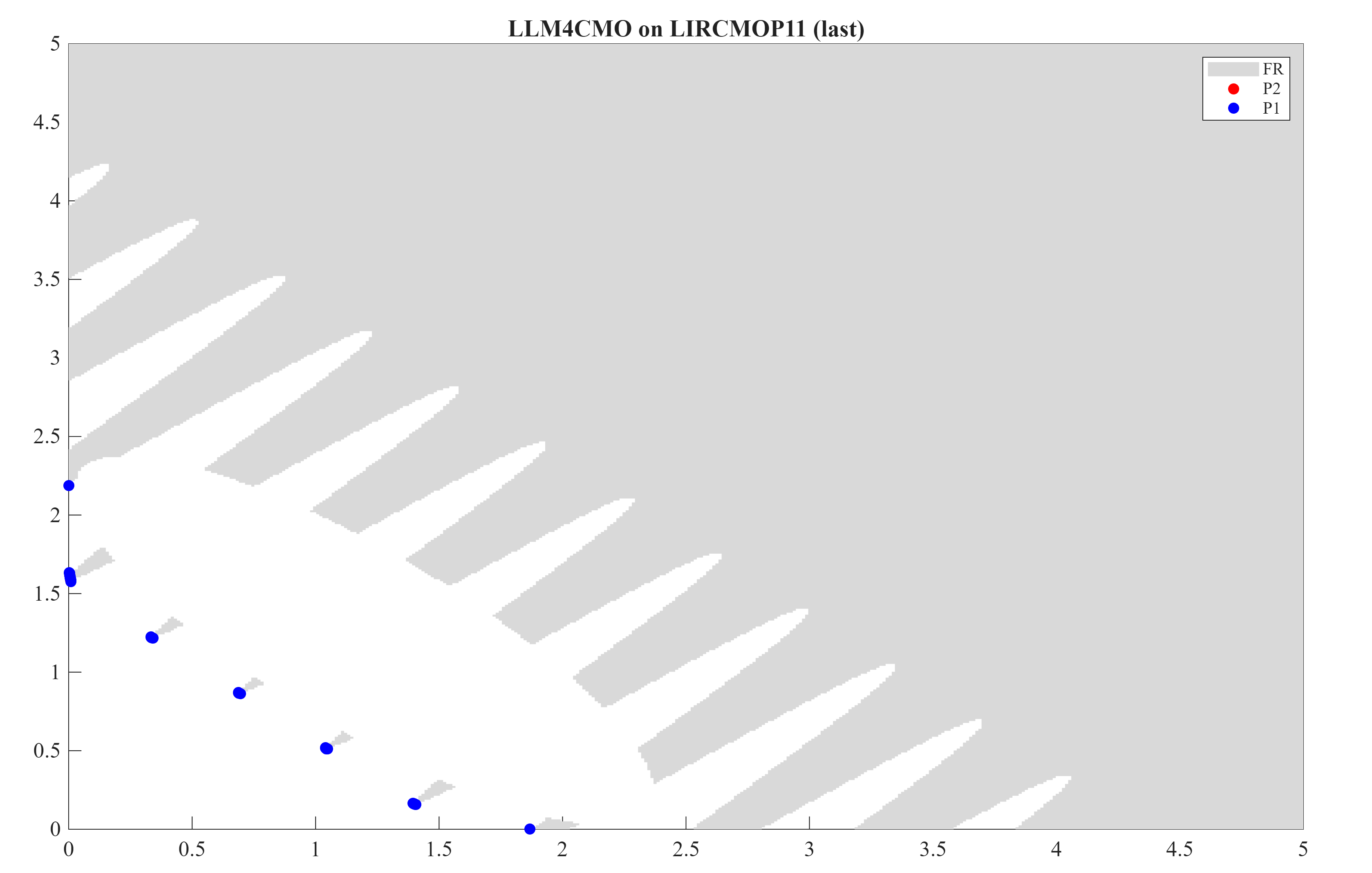}
    \end{minipage}
    
    \caption{The process charge of P1, P2 and of Bico, URCMO and LLM4CMOon LIRCMOP11(Type-3).}
    \label{fig:3ALIR11}
\end{figure}

\clearpage\label{sec:A7}
\section{Complete prompt template} \label{sec:A8}
\subsection{Complete prompt template for HOps task.} \label{sec:A8-1}
\begin{figure}[H]
    \centering
    \resizebox{0.87\textwidth}{!}{
        \includegraphics[]{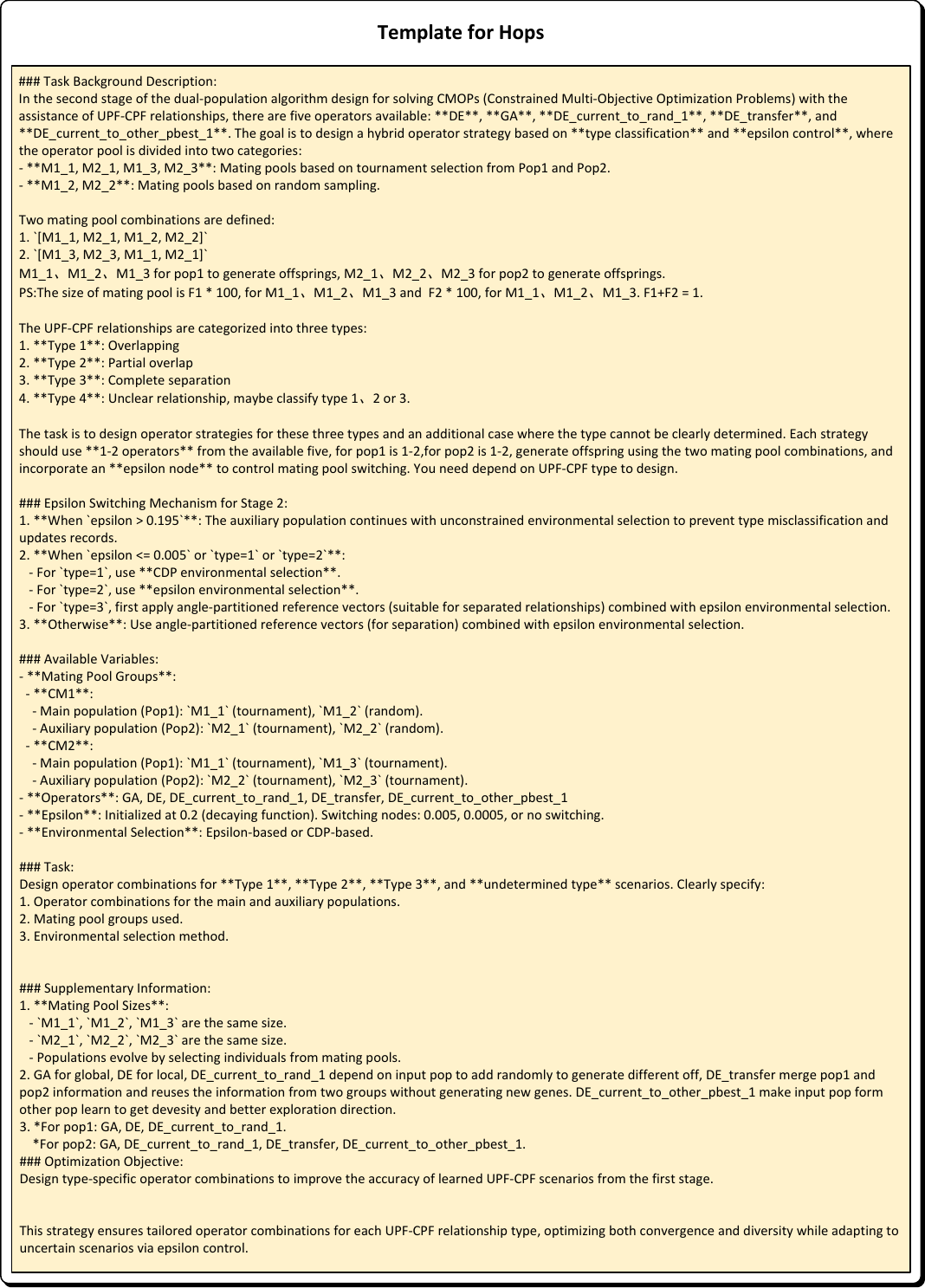}
    }
    \caption{The complete prompt template for HOps task.}
    \label{fig:PHOps}
\end{figure}

\subsection{Complete prompt template for Epsilon decay function design.} \label{sec:A8-2}
\begin{figure}[H]
    \centering
    \resizebox{0.87\textwidth}{!}{
        \includegraphics[]{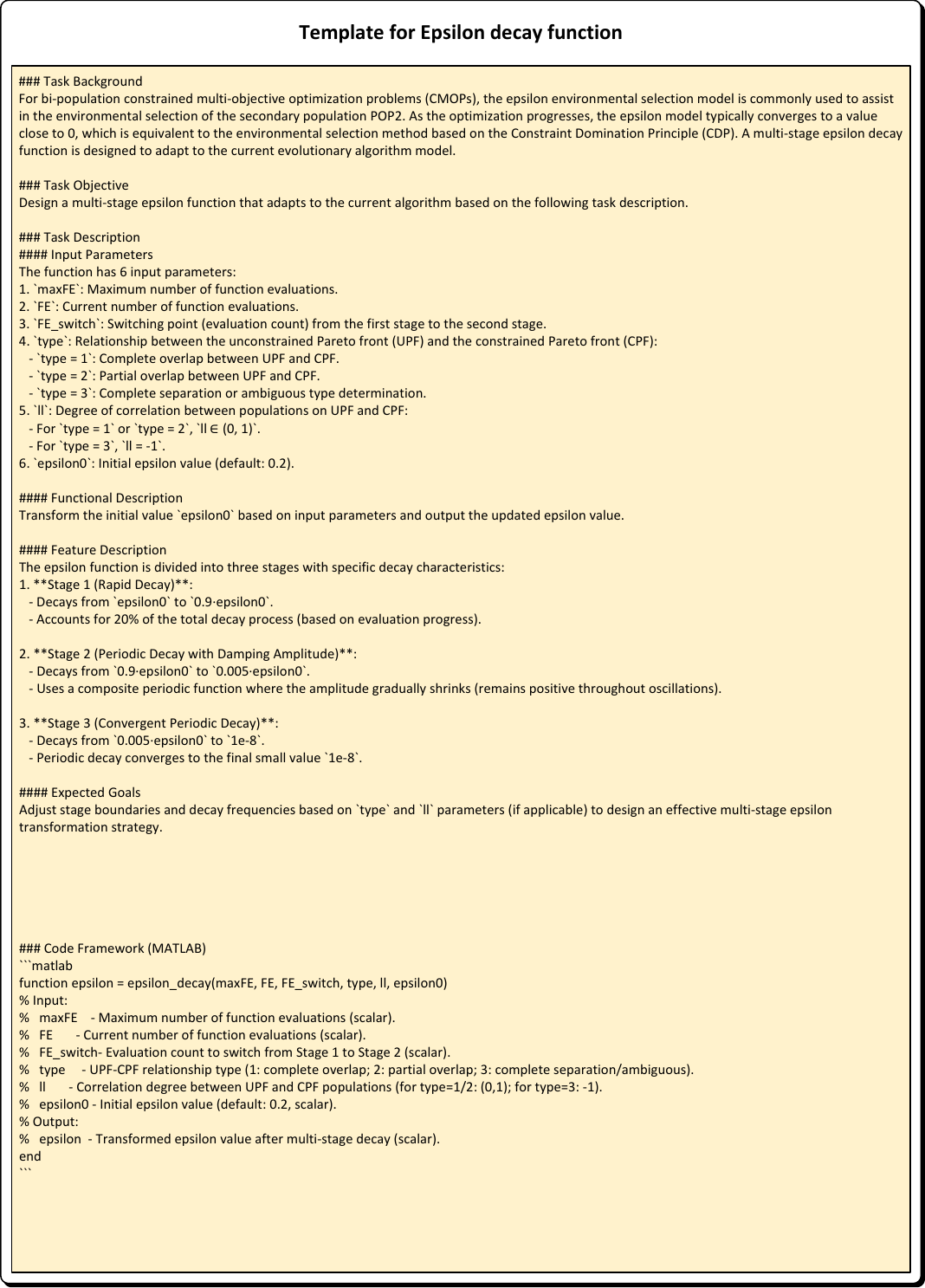}
    }
    \caption{The complete prompt template for Epsilon decay function design.}
    \label{fig:PEPS}
\end{figure}

\subsection{Complete prompt template for DRA.} \label{sec:A8-3}
\begin{figure}[H]
    \centering
    \resizebox{0.87\textwidth}{!}{
        \includegraphics[]{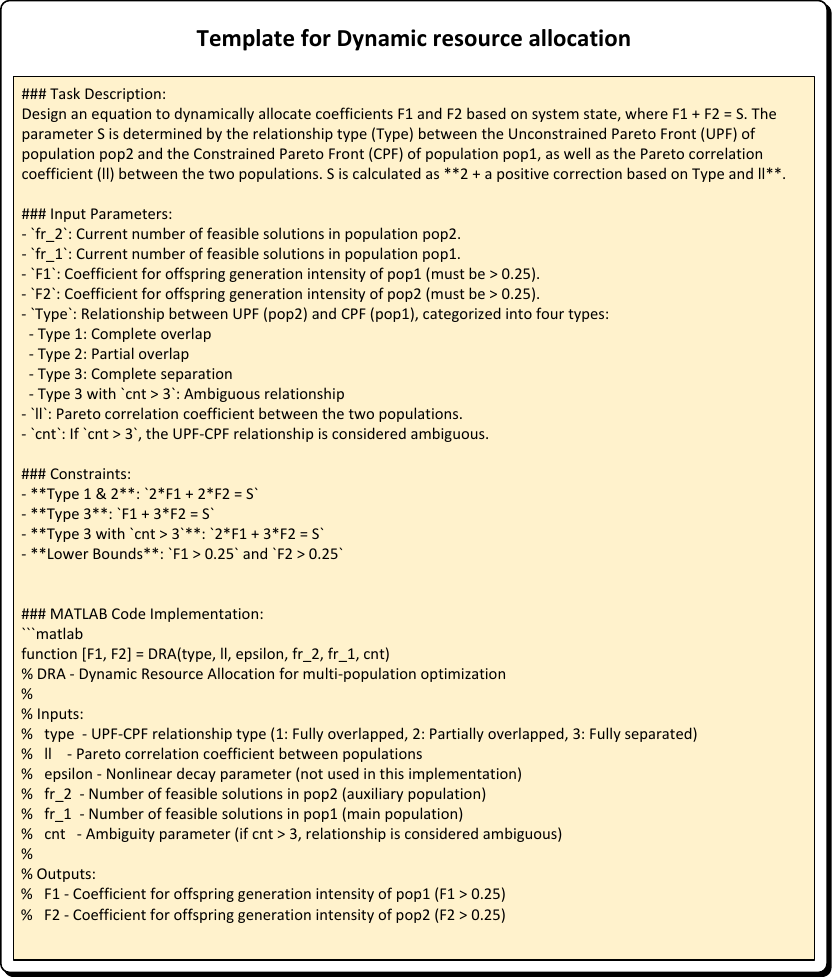}
    }
    \caption{The complete prompt template for dynastic resource allocation design.}
    \label{fig:PDRA}
\end{figure}

\section{Complete interaction process.} \label{sec:A9}
Our interaction approach, as mentioned in the text, involves testing problems based on different types and then communicating the optimization directions to the LLM. Given that the interaction process is complex, the text is relatively long, and it involves a strong subjective judgment process, for the complete interaction process of this part, please refer to our code repository~\url{https://anonymous.4open.science/r/LLM4CMO971}.


 \vfill

\end{document}